%% file: main.tex
\documentclass[10pt,twocolumn,letterpaper]{article}

 \usepackage[pagenumbers]{cvpr} %

\input{preamble}

\usepackage{graphicx}
\usepackage{amsmath}
\usepackage{amssymb}
\usepackage{booktabs}
\usepackage{makecell}
\usepackage{rotating}
\usepackage{multirow}
\usepackage{comment}
\usepackage{pifont}
\newcommand{\cmark}{\ding{51}}%
\newcommand{\xmark}{\ding{55}}%
\usepackage{tikz}

\definecolor{cvprblue}{rgb}{0.21,0.49,0.74}
\usepackage[pagebackref,breaklinks,colorlinks,citecolor=cvprblue]{hyperref}

\def\paperID{11} %
\def\confName{CVPR}
\def\confYear{2024}

\newcommand{\grafiqs}{\textsc{GraFIQs}\xspace}

\title{\grafiqs: Face Image Quality Assessment Using Gradient Magnitudes}

\author{Jan Niklas Kolf\textsuperscript{1,2} \hspace{5mm} Naser Damer\textsuperscript{1,2} \hspace{5mm} Fadi Boutros\textsuperscript{1}
\vspace{0.35cm}\\
\textsuperscript{1}Fraunhofer IGD \hspace{3mm} \textsuperscript{2}Technische Universität Darmstadt
\vspace{0.1cm}\\
{\tt\small \{jan.niklas.kolf, naser.damer, fadi.boutros\}@igd.fraunhofer.de}
}

\begin{document}
\maketitle

\input{00_abstract}

\input{01_introduction_02-relatedwork}
\input{03_methodology}
\input{04_experimental_setup}

\input{05_results}

\input{06_conclusion}
{\small
\bibliographystyle{ieee_fullname}
\bibliography{08_references}
}
\newpage
\input{09_appendix}

\end{document}

%% file: preamble.tex
\usepackage[dvipsnames]{xcolor}

%% file: 00_abstract.tex
\begin{abstract}
Face Image Quality Assessment (FIQA) estimates the utility of face images for automated face recognition (FR) systems.
We propose in this work a novel approach to assess the quality of face images based on inspecting the required changes in the pre-trained FR model weights to minimize differences between testing samples and the distribution of the FR training dataset. To achieve that, we propose quantifying the discrepancy in Batch Normalization statistics (BNS), including mean and variance, between those recorded during FR training and those obtained by processing testing samples through the pretrained FR model. We then generate gradient magnitudes of pretrained FR weights by backpropagating the BNS through the pretrained model. The cumulative absolute sum of these gradient magnitudes serves as the FIQ for our approach.
Through comprehensive experimentation, we demonstrate the effectiveness of our training-free and quality labeling-free approach, achieving competitive performance to recent state-of-the-art FIQA approaches without relying on quality labeling, the need to train regression networks, specialized architectures, or designing and optimizing specific loss functions.\footnote{\url{https://github.com/jankolf/GraFIQs}}
\end{abstract}

%% file: 01_introduction_02-relatedwork.tex
\vspace{-4mm}
\section{Introduction}
\vspace{-2mm}
Face image quality assessment (FIQA) estimates the utility of the captured sample for face recognition (FR) \cite{Quality_ISO}.
FIQA algorithms process an image to produce a scalar quality score. This score serves as a scalar measurement to quantify the quality of a face image in terms of its suitability for use in FR systems \cite{NISTQuaity}.  
Ensuring high FIQ can improve the performance of FR and enhance their effectiveness in applications such as automated border control \cite{NISTQuaity,DBLP:journals/csur/SchlettRHGFB22}. 
FIQA primarily targets assessing the utility of a face image for automated FR \cite{Quality_ISO,NISTQuaity,DBLP:journals/csur/SchlettRHGFB22} rather than assessing the perceived image quality \cite{elasticface}. Perceived image quality assessment has been addressed in the literature by general image quality assessment (IQA) methods \cite{BRISQE_IQA,nique,liu2017rankiqa}, which provide insights into image quality from human perception. IQA does not necessarily reflect the utility of face image for FR \cite{BiyingWACV}, e.g., a face image may have high perceived quality according to IQA metrics; however, it may still be relatively less suitable for FR due to factors such as occlusion.
On the other hand, FIQA algorithms provide a targeted assessment of the image's utility for FR tasks.
Thus, FIQA approaches in the literature demonstrate superiority over IQA in assessing the utility of face image FR, as presented in \cite{boutros_2023_crfiqa,MagFace,SDDFIQA,DBLP:conf/iwbf/BabnikDS23}. 

State-of-the-art (SOTA) FIQA can be categorized into two main categories. The approaches in the first category focus on labeling face images with quality labels and then training a regression network to predict the FIQ score of test samples \cite{faceqnetv1,SDDFIQA,RANKIQ_FIQA,best2018learning}. Best-Rowden et al. \cite{best2018learning} proposed learning FIQ based on target quality labels obtained from either human assessment of FIQ or quality score labels computed from genuine comparisons. FaceQNet \cite{faceqnetv1} proposed to label the images with quality scores based on the genuine comparison between a sample and an ICAO compliant sample and then used the labeled data to train a regression network. SDD-FIQ \cite{SDDFIQA} generated quality labels to train a regression network using Wasserstein distance between genuine and imposter similarity distributions. 
RankIQ \cite{RANKIQ_FIQA} proposed the learning-to-rank approach that predicts the sample quality as a rank using FRs performances on several datasets.  
The approaches in the second category \cite{boutros_2023_crfiqa,MagFace,SERFIQ,PFE_FIQA} assess the utility of face images by either learning to estimate the quality from face embedding properties during FR training or inspecting the robustness of embeddings. To assess the image quality, PFE \cite{PFE_FIQA} estimates an uncertainty of face embedding in the latent space and considers it as a reverse measure of face quality. MagFace \cite{MagFace} proposed
to learn a universal feature embedding and then utilize the magnitude of the embedding to measure the quality of a given face image.
SER-FIQ \cite{SERFIQ} inspected the robustness of face embedding and considered it as FIQ by passing the face images into an FR network multiple times, each with a different random dropout pattern, to output several embeddings of each sample. Then, the quality score is obtained by calculating the sigmoid of the negative mean of the Euclidean distances between the embeddings.
CR-FIQA \cite{boutros_2023_crfiqa} proposed to estimate the FIQ of a sample by learning to predict its relative classifiability, which is measured based on the allocation of the sample feature representation in the embedding space to its class center and the nearest negative class center. 
DifFIQA \cite{10449044} leverages a diffusion model to explore the stability of embeddings of face images through image perturbations caused by the processes of adding noise and denoising.
eDifFIQA \cite{babnikTBIOM2024} applied knowledge distillation to DifFIQA  \cite{10449044}, aiming at reducing the computational cost of DifFIQA  \cite{10449044}.

This paper presents a pioneering approach, namely \grafiqs, that leverages the gradient magnitude during the backpropagation step of pretrained FR model to assess the FIQ. Unlike recent high-performing FIQA approaches \cite{MagFace,SDDFIQA,boutros_2023_crfiqa,SERFIQ} that rely on face embeddings, our approach does not require quality labeling and training of regression networks \cite{faceqnetv1,SDDFIQA}, the need of specialized architectures \cite{SERFIQ}, or designing and optimizing specific loss functions \cite{MagFace}. In contrast, our approach inspects the necessary changes in the pretrained FR model parameters to minimize the difference between test samples and the FR model training dataset distribution. To achieve this, we propose to measure the shift in Batch Normalization statistics (BNS), mean and variance, between the ones recorded during the FR training and those obtained by passing the test samples into the FR model. Subsequently, after extracting the BNS, we backpropagate the difference between BNS into the pretrained model to generate gradient magnitudes, whose absolute sum serves as FIQ. Note that the BNS of the training dataset are integral parts of the pretrained model parameters and do not require retraining the model to extract these values. Through extensive experiments, we prove that our training-free and quality labeling-free approach can achieve competitive results with recent SOTA FIQA methods. This novel perspective on FIQA provides a new way of assessing the quality of face images based on the gradient magnitudes of the pretrained FR model.

%% file: 03_methodology.tex
\vspace{-3mm}
\section{Methodology}
\begin{figure}[ht!]
\begin{center}
\includegraphics[width=0.9\linewidth]{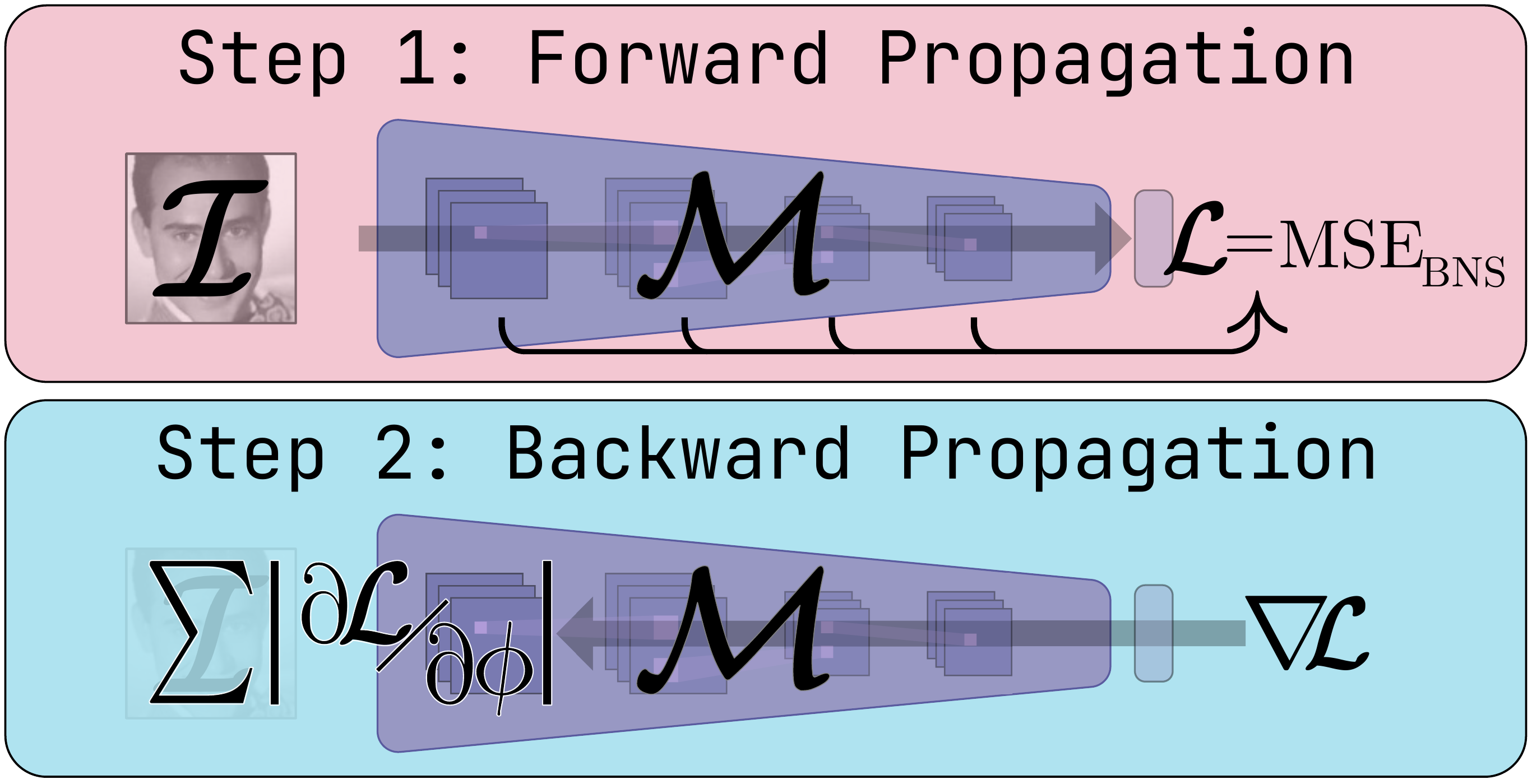}
\end{center}
\vspace{-6mm}
\caption{An overview of the proposed \grafiqs for assessing the quality of unseen testing samples. Sample $\mathcal{I}$ is passed into the pretrained FR model and BNS are extracted. Then, the MSE between the BNS obtained by processing the testing sample and the one recorded during the FR training is calculated. The MSE is backpropagated through the pretrained FR to extract the gradient magnitudes of parameter group $\phi$. Finally, the absolute sum of gradient magnitudes of $\phi$ is calculated and utilized as FIQ.}
\label{fig:overview}
\vspace{-5mm}
\end{figure}

\label{sec:methodology}
This section presents our novel training-free and label-free FIQA technique that leverages the gradient magnitudes obtained from forwarding and backpropagating any test sample in a given pretrained FR model. 

Conventionally, gradient optimization is only used during the training phase of a FR model to update the model parameters in the backpropagation step with respect to the training loss function \cite{bishop_2023_deeplearning}. The model parameters are iteratively updated until the model is converged, i.e., gradient descent converged to a local minimum. The parameters of the model are updated in a way that the loss function achieves the minimum value on the training dataset \cite{aggarwal_2020_ml_optim, prince_2023_understanding}.
Once the model is trained, the model weights are frozen, and no gradient optimizations are required or performed during the inference phase. In the case that the FR model is not designed and optimized to predict the FIQ, which is the case in our approach, the output of the FR can not be used directly to assess the utility of the test sample. To address the aforementioned challenge, we propose to assess the utility of any given test sample by calculating the required changes in the pretrained FR model weights to minimize the difference between the test sample and the model training data distribution. 
To achieve this objective, we calculate the difference between training data and test data distribution using mean squared error (MSE) between BNS \cite{ioffe_2015_batchnormalization,xu_2020_generativequant}. We then propose to backpropagate the MSE to generate gradient magnitudes from the pretrained model, in which their absolute sums are used to assess the sample quality, as detailed in Section \ref{subsec:grafiqs_approach}.  

Figure \ref{fig:overview} presents an overview of our proposed approach to estimate the utility of test samples using a pretrained FR model. 
In this approach, a test sample is forwarded into the pretrained FR model, and a Batch Normalization (BN) loss is calculated (details in Section \ref{subsec:batchnormalization}). Then, we calculate the gradient of the loss function with respect to the model parameters using backpropagation. The magnitude of the resulting gradients is used to measure the required changes in the model parameters to minimize the MSE between BNS of the training data and the BNS of the given sample.
If high magnitudes are present within the gradients of the FR model parameters, this indicates that the parameters require large changes to minimize the MSE based on the input \cite{prince_2023_understanding}.
On the contrary, low magnitudes indicate that the parameters do not have to be updated to a large degree to achieve minimal MSE loss of BNS.
We theorize that, given the BNS calculated on a training dataset and the BNS of an input image, high gradient magnitudes resulting from MSE loss indicate a low utility of the input image, and vice versa, further explained in Section \ref{subsec:gradient_based_optimization}.
Although calculating the BN loss would directly indicate the quality of the test samples, we show in this paper that the gradient magnitude (defined as how strong the required changes are) with respect to the model parameters is more informative in estimating the utility of the sample.
This concept and the required loss function will be introduced in this section, and the success of the underlying approach will be empirically proven later in this work.

This section first presents fundamentals on gradient-based optimization. Then, we present and formalize the basic principles of \grafiqs and the fundamental loss function introduced in this work.
\vspace{-1mm}
\subsection{Gradient-Based Optimization}
\label{subsec:gradient_based_optimization}
\vspace{-1mm}
In essence, the objective of a given optimization problem is to find an assignment for a given set of parameters $\theta \in \mathbb{R}^{d}$ that minimizes a given loss or cost function $\mathcal{L}(\theta)$ \cite{aggarwal_2020_ml_optim}:
\begin{align}
 \label{eq:thetaoptimization}
\theta^{*} &= \mathop{\mathrm{argmin}}_{\theta} \mathcal{L}(\theta)
\end{align}
A machine learning model $\mathcal{M}_{\theta}$, e.g. a FR model, is a single or a sequence of mathematical functions (also referred to as layers \cite{bishop_2023_deeplearning}) that maps an input, e.g. input image $\mathcal{I}$, to an output $y=\mathcal{M}_{\theta}(\mathcal{I})$, also called prediction, using the set of parameters $\theta$ \cite{aggarwal_2020_ml_optim}.

For a simpler explanation, consider the supervised learning approach.
In this approach, dataset $\mathcal{D}$ is given, which consists of input training samples $\mathcal{I}$ and their corresponding labels $\hat{y}$.
The objective of optimization is to learn a set of parameters $\theta$ such that the prediction $y=\mathcal{M}_{\theta}(\mathcal{I})$ matches $\hat{y}$ as closely as possible \cite{aggarwal_2020_ml_optim}.
The model $\mathcal{M}_{\theta}$ and therefore the parameters $\theta$ are fitted to $\mathcal{D}$ through this optimization process.
The scalar-valued loss function $\mathcal{L}(\theta)$ is used to quantify the mismatch between the prediction $\mathcal{M}_{\theta}(\mathcal{I})$ and the desired target output $\hat{y}$ \cite{prince_2023_understanding}.
Therefore, minimizing $\mathcal{L}(\theta)$ minimizes the mismatch between prediction $y=\mathcal{M}_{\theta}(\mathcal{I})$ and target $\hat{y}$.

Since analytical solutions that minimize $\mathcal{L}(\theta)$ and yield $\theta^{*}$ are often not feasible, a commonly used method in machine learning is gradient descent \cite{bishop_2023_deeplearning, prince_2023_understanding}, a variation of gradient-based optimization.
These optimization techniques require that the derivative of $\mathcal{L}(\theta)$ w.r.t. the model parameters $\theta$, $\nabla_{\theta} \mathcal{L}(\theta)$, is defined \cite{bishop_2023_deeplearning}.

The gradient $\nabla_{\theta} \mathcal{L}(\theta)$ is defined as the vector of partial derivatives of $\mathcal{L}(\theta)$ w.r.t. to each parameter $\theta_{i} \in \theta$:
\begin{align}
    \nabla_{\theta} \mathcal{L}(\theta) &= \begin{bmatrix}
                               \frac{\partial\mathcal{L}(\theta)}{\partial\theta_{1}}  &
                               \cdots &
                               \frac{\partial\mathcal{L}(\theta)}{\partial\theta_{m}}
                             \end{bmatrix}^{T}
\end{align}

The gradient $\nabla_{\theta} \mathcal{L}(\theta)$ specifies the direction of the fastest increase in $\mathcal{L}(\theta)$ \cite{prince_2023_understanding}.
The magnitude of $\nabla_{\theta} \mathcal{L}(\theta)$ specifies the rate of change in the direction of the gradient \cite{prince_2023_understanding}.
Therefore, changing the parameters in the direction of $-\nabla_{\theta} \mathcal{L}(\theta)$ reduces $\mathcal{L}(\theta)$.
As the gradient only gives a momentary and local rate of change that does not hold over larger distances, steps in the direction of the negative gradient have to be adjusted by a scaling parameter $\alpha>0$ \cite{aggarwal_2020_ml_optim}.
The parameter $\alpha$ is named the learning rate \cite{aggarwal_2020_ml_optim}, as it specifies the rate of change in the direction of the negative gradient.
To calculate $\theta^{*}$ that minimize $\mathcal{L}(\theta)$, an iterative update procedure with an initial set of parameters $\theta^{0}$ and learning rate $\alpha^{0}$ is used \cite{aggarwal_2020_ml_optim}:
\begin{align}
    \theta^{t+1} &= \theta^{t} - \alpha^{t} \nabla_{\theta^{t}} \mathcal{L}(\theta^{t})
\end{align}

With every parameter update, the overall mismatch between predictions $y=\mathcal{M}_{\theta}(\mathcal{I})$ and targets $\hat{y}$ of dataset $\mathcal{D}$, measured by $\mathcal{L}(\theta)$, is reduced.
After sufficient update steps are performed, it is assumed that $\theta^{t_{\text{max}}} \approx \theta^{*}$.

\vspace{-1mm}
\subsection{Batch Normalization}
\label{subsec:batchnormalization}
\vspace{-1mm}
The previous section presented preliminary on FR model training and calculating gradient magnitudes, which serve as a basis for our concept to assess the FIQ. 
Details on FR training is provided in Section \ref{sec:experimental_setup}.
Calculating gradient magnitudes, as we presented in Section \ref{subsec:gradient_based_optimization}, requires a loss function, e.g. multi-class classification loss, which is not feasible when training and test data are identity-disjoint (the case for the FR model). Alternatively, we propose to calculate MSE (as a loss function) between the BNS of the pretrained FR model and the BNS extracted by passing test samples into the pretrained FR. We then backpropagate MSE to calculate gradient magnitudes. Therefore, we briefly provide in this section insight into BN.

BN \cite{ioffe_2015_batchnormalization} is a technique applied in certain machine learning models to improve the speed, performance, and stability of the training phase \cite{bishop_2023_deeplearning, prince_2023_understanding, santurkar_2018_batchnormalization}.
It is included several times as a dedicated layer, the BN layer (BNL), within the sequence of layers \cite{prince_2023_understanding}.
The BNL receives a set (batch) of outputs (referred to as activations) from the previous layer as input \cite{bishop_2023_deeplearning}.
The activations of the input batch are normalized using the mean and standard deviation calculated throughout the batch.
During inference, mean $\mu$ and standard deviation $\sigma$ are used for normalization that were calculated during the model training phase using exponential moving averages \cite{bishop_2023_deeplearning, xu_2020_generativequant}, forming BNS.

To determine whether a particular input sample adheres to the same data distribution as the training data, the BNS, mean $\mu$ and standard deviation $\sigma$ of the training data, can be compared to the BNS obtained from that sample during inference ($\mu'$ and $\sigma'$)  \cite{xu_2020_generativequant}.
This technique serves as an effective method for domain adaptation, a process widely utilized in various applications \cite{kolf_2023_idnet, xu_2020_generativequant}. Note that BNS are part of the pretrained model parameters and do not require accessing the training data to extract these statistics \cite{bishop_2023_deeplearning}.

To quantify the divergence between the BNS of the training data and that of an input sample, a loss function based on the mean squared error is devised. This loss function evaluates the discrepancy in BNS across all BNL in the model and is defined as follows:
\begin{align}
    \label{eq:mse_bsn}
    \text{MSE}_{\text{BNS}} &= \frac{1}{|\text{BNL}|} \sum_{l\in \text{BNL}} \| \mu_{l} - \mu'_{l}\|_{2}^{2} + \| \sigma_{l} - \sigma'_{l}\|_{2}^{2},
\end{align}
where, $\mu_{l}$ and $\sigma_{l}$ denote the mean and standard deviation of the BNS for the $l$-th layer recorded during training, while $\mu'_{l}$ and $\sigma'^2_{l}$ represent the mean and variance for the same layer obtained from the input sample during inference. The norm $\| \cdot \|_{2}^{2}$ indicates the squared Euclidean (L2) norm.

A higher $\text{MSE}_\text{BNS}$ signifies a greater disparity between the input sample and the data distribution of the training dataset, compared to an input sample that yields a lower $\text{MSE}_\text{BNS}$.
\vspace{-1mm}
\subsection{Deriving \grafiqs Approach}
\label{subsec:grafiqs_approach}
\vspace{-1mm}
Although the $\text{MSE}_{\text{BNS}}$ value can be directly used to evaluate the FIQ of a given input sample (as presented in Section \ref{sec:ablation}), it does not reflect the required changes on the activation level to match the distribution between pretrained model and test sample.

Consider two different input samples, denoted as $\mathcal{I}_i$ and $\mathcal{I}_j$.
At a given BNL $l$, BNL $l$ receives activations as input that have respective mean and variance values $\mu'_{l,i}$, $\sigma'_{l,i}$ for $\mathcal{I}_i$ and $\mu'_{l,j}$, $\sigma'_{l,j}$ for $\mathcal{I}_j$, respectively.
The MSE between the BNS of the training dataset and the BNS of sample $\mathcal{I}_m$, $m \in \{i,j\}$, is defined as $\text{MSE}_{l,m} = \| \mu_{l} - \mu'_{l,m}\|_{2}^{2} + \| \sigma_{l} - \sigma'_{l,m}\|_{2}^{2}$.
Assume that $\mu'_{l,i} = \mu'_{l,j}$ and $\sigma'_{l,i} = \sigma'_{l,j}$ and therefore $\text{MSE}_{l,i} = \text{MSE}_{l,j}$.
This does not imply that the activations of the layer preceding the BNL, which result from the inputs $\mathcal{I}_i$ and $\mathcal{I}_j$, are identical.
Different activations of the samples, which are generated by the parameters in the pretrained FR model, can lead to the same mean and standard deviation.
However, the necessary changes within the parameters to minimize the difference between the BNS of the training set ($\mu'_{l}$ and $\sigma'^2_{l}$) and the BNS of the respective sample, $\mathcal{I}_i$ ($\mu'_{l,i}$, $\sigma'_{l,i}$) or $\mathcal{I}_j$ ($\mu'_{l,j}$, $\sigma'_{l,j}$), can differ significantly.
Using only the mean and standard deviation, either at a single BNL or over the entire network, is not enough to fully capture the required changes of the parameters to output activations that match the BNS of the training dataset.
To overcome the limitations that arise when the mean and standard deviations are used, we propose a new method that considers not only $\text{MSE}_\text{BNS}$, but also the necessary parameter changes that are described by gradient magnitudes.

This approach, gradient magnitude based FIQA (\grafiqs), is specified in the following.
An input image $\mathcal{I}$ is fed into the network and the loss $\mathcal{L}_{\text{BNS}} = \text{MSE}_\text{BNS}$ is calculated.
The gradients $\nabla_{\theta}\mathcal{L}_{\text{BNS}}$ are calculated by backpropagation through the network.
The gradient magnitudes, i.e. the absolute value of the gradient, from $\partial \mathcal{L}_{\text{BNS}} / \partial \phi$ are extracted and summed to give the total required changes in the parameters $\phi$.
The FIQ is therefore calculated as $\sum |\partial \mathcal{L} / \partial \phi|$.
Gradients can be extracted either at image level, $\phi = \mathcal{I}$, or for intermediate layers, $\phi = \text{B}i$, as detailed in Section \ref{sec:ablation}.

%% file: 04_experimental_setup.tex
\section{Experimental Setup}
\label{sec:experimental_setup}

\textbf{Pretrained model architecture}
The proposed approach \grafiqs is demonstrated using two pretrained FR models, ResNet100 and ResNet50, trained with ArcFace loss on MS1MV2 \cite{guo_2016_ms1m, deng2019arcface} and CASIA-WebFace \cite{yi_2014_casiawebface}, respectively. The pretrained models were released by \cite{deng2019arcface} and they are publically available. 
The results of the pretrained ResNet50 model are provided in supplementary materials.

\textbf{Evaluation Benchmarks} 
We reported the achieved results on the following benchmarks: Labeled Faces in the Wild (LFW) \cite{LFWTech}, AgeDB-30 \cite{agedb}, Celebrities in Frontal-Profile in the Wild (CFP-FP) \cite{cfp-fp}, Cross-age LFW (CALFW) \cite{CALFW}, Adience \cite{Adience}, Cross-Pose LFW (CPLFW)  \cite{CPLFWTech} and Cross-Quality LFW (XQLFW) \cite{XQLFW}. These benchmarks contain challenging pairs with large age variations (AgeDB-30 and CALFW), head-pose variations (CFP-FP and CPLFW), and face image quality variations (XQLFW). These benchmarks are chosen to be aligned with recent SOTA FIQ approaches \cite{boutros_2023_crfiqa}. 

\textbf{Evaluation Metric}
We evaluate the FIQA by plotting Error-versus-Discard Characteristic (EDC) Curves \cite{GT07,NISTQuaity}. It should be noted that several SOTA approaches in the literature referred to EDC as ERC (Error-versus-Reject Curves) \cite{10449044}.
The EDC is a widely used metric for evaluating FIQA performance \cite{GT07,NISTQuaity}. EDC curve demonstrates the effect of discarding a fraction of face images, of the lowest quality, on face verification performance in terms of False None Match Rate \cite{iso_metric} (FNMR) at a specific threshold calculated at fixed False Match Rate \cite{iso_metric} (FMR). Following SOTA FIQA approaches \cite{10449044, boutros_2023_crfiqa, MagFace}, the EDC curves for all benchmarks are plotted at two fixed FMRs, $1e-3$ and $1e-4$. 
We also report the Area under the Curve (AUC) of the EDC, to provide a quantitative aggregate measure of verification performance across all rejection ratios.

\textbf{FR Models}
To provide insight into the generalizability of our approach, we report the verification performance at different quality discard rates using four different FR models. The utilized models are:  ArcFace \cite{deng2019arcface}, ElasticFace (ElasticFace-Arc) \cite{elasticface}, MagFace \cite{MagFace}, and CurricularFace \cite{curricularFace}.
We utilized the pretrained models provided by the corresponding authors \cite{curricularFace,elasticface,MagFace,deng2019arcface}. All models were originally trained on MS1MV2 \cite{guo_2016_ms1m, deng2019arcface} and used ResNet100 \cite{he_2016_resnet} as network architecture.  All models process $112 \times 112$ aligned and cropped images to produce feature embedding of size 512-D. 

The results are reported under two protocols, same and cross model. The results reported using ArcFace \cite{deng2019arcface} follow the same model protocol, where ArcFace is used to calculate the quality of images and for reporting the verification accuracies at different quality discard rates. The results reported using   ElasticFace \cite{elasticface}, MagFace \cite{MagFace}, and CurricularFace \cite{curricularFace} are cross-model evaluation protocols, where ArcFace is used to calculate the quality of images and ElasticFace, MagFace, and CurricularFace were used to report the verification accuracies at different quality discard rates.

\textbf{Baseline and Comparisons with SOTA FIQ}
The achieved results by our \grafiqs are compared to the three general IQA methods, BRISQUE \cite{BRISQE_IQA}, RankIQA \cite{liu2017rankiqa}, and DeepIQA \cite{DEEPIQ_IQA} and to nine SOTA FIQA methods, RankIQ \cite{RANKIQ_FIQA}, PFE \cite{PFE_FIQA}, SER-FIQ \cite{SERFIQ}, FaceQnet (v1 \cite{faceqnetv1}) \cite{hernandez2019faceqnet,faceqnetv1}, MagFace \cite{MagFace}, SDD-FIQA \cite{SDDFIQA}, CR-FIQA \cite{boutros_2023_crfiqa}, DifFIQA \cite{10449044} and eDifFIQA \cite{babnikTBIOM2024}.

%% file: 05_results.tex
\begin{table}
\centering
\caption{Conceptual comparison on the design choices between our \grafiqs and recent FIQA approaches in the literature.}
\vspace{-3mm}
\label{tab:desinged_comparison}
\resizebox{0.95\linewidth}{!}{%
\begin{tabular}{ccccccccc}
\multicolumn{1}{l}{} & \multicolumn{1}{l}{} & \multicolumn{1}{l}{} & \multicolumn{1}{l}{} & \multicolumn{1}{l}{} & \multicolumn{4}{c}{Inference} \\ 
\cline{6-9}
FIQA & \multicolumn{1}{l}{\begin{sideways}\begin{tabular}[c]{@{}l@{}}Quality \\Labels\end{tabular}\end{sideways}} & \multicolumn{1}{l}{\begin{sideways}\begin{tabular}[c]{@{}l@{}}Specific\\Architecture\end{tabular}\end{sideways}} & \multicolumn{1}{l}{\begin{sideways}\begin{tabular}[c]{@{}l@{}}Requires \\Training\end{tabular}\end{sideways}} & \multicolumn{1}{l}{\begin{sideways}\begin{tabular}[c]{@{}l@{}}Custom\\~Loss\end{tabular}\end{sideways}} & \multicolumn{1}{l}{\begin{sideways}\begin{tabular}[c]{@{}l@{}}Feed-\\Forwards\end{tabular}\end{sideways}} & \multicolumn{1}{l}{\begin{sideways}Backwards\end{sideways}} & \multicolumn{1}{l}{\begin{sideways}\begin{tabular}[c]{@{}l@{}}Embedding-\\Level\end{tabular}\end{sideways}} & \multicolumn{1}{l}{\begin{sideways}\begin{tabular}[c]{@{}l@{}}Gradient-\\Level\end{tabular}\end{sideways}} \\ 
\hline
CR-FIQA \cite{boutros_2023_crfiqa} & \xmark & \xmark & \cmark & \cmark & 1 & 0 & \cmark & \xmark \\
DifFIQA \cite{10449044} & \xmark & \xmark & \cmark & \cmark & 1 & 0 & \cmark & \xmark \\
eDifFIQA \cite{babnikTBIOM2024} & \cmark & \xmark & \cmark & \cmark & 1 & 0 & \cmark & \xmark \\
MagFace \cite{MagFace} & \xmark & \xmark & \cmark & \cmark & 1 & 0 & \cmark & \xmark \\
FaceQnet \cite{faceqnetv1} & \cmark & \xmark & \cmark & \xmark & 1 & 0 & \cmark & \xmark \\
SDD-FIQA \cite{SDDFIQA} & \cmark & \xmark & \cmark & \xmark & 1 & 0 & \cmark & \xmark \\
SER-FIQ \cite{SERFIQ} & \xmark & \cmark & \xmark & \xmark & 100 & 0 & \cmark & \xmark \\
PFE \cite{PFE_FIQA} & \xmark & \xmark & \cmark & \cmark & 1 & 0 & \cmark & \xmark \\ 
\hline
\grafiqs (Our) & \xmark & \xmark & \xmark & \xmark & 1 & 1 & \xmark & \cmark \\
\hline
\end{tabular}
}
\vspace{-5mm}
\end{table}

\begin{table*}[ht!]
\begin{center}
\caption{The achieved AUC of EDC by using two approaches presented in this paper, MSE of BNS ($\text{MSE}_{\text{BNS}}$) and gradient magnitudes ($\mathcal{L}_{\text{BNS}}$), and under different settings. The gradient magnitudes are extracted during the backpropagation step from different intermediate layers, B1, B2, B3 and B4 ($\phi = \text{B}1 - \phi = \text{B}4$) as well as on the pixel level  ($\phi = \mathcal{I}$).  The results are reported under two operation threshold FMR$=1e-3$ and FMR$=1e-4$ and under two protocols, same model (ArcFace) and cross-model (ElasticFace, MagFace and CurricularFace). Utilizing gradient magnitudes from B$2$ achieved the best overall performance.}
\label{tbl:ablation}
\vspace{-3mm}
\resizebox{0.98\textwidth}{!}{
\begin{tabular}{|c|c|c|c|c|c|c|c|c|c|c|c|c|c|c|c|c|c|c|c|}
\hline
\multirow{2}{*}{FR} & \multirow{2}{*}{Loss $\mathcal{L}$} & \multirow{2}{*}{FIQ} & \multirow{2}{*}{Gradient} & \multicolumn{2}{c|}{Adience \cite{Adience}} & \multicolumn{2}{c|}{AgeDB30 \cite{agedb}} & \multicolumn{2}{c|}{CFP-FP \cite{cfp-fp}} & \multicolumn{2}{c|}{LFW \cite{LFWTech}} & \multicolumn{2}{c|}{CALFW \cite{CALFW}} & \multicolumn{2}{c|}{CPLFW \cite{CPLFWTech}} & \multicolumn{2}{c|}{XQLFW \cite{XQLFW}} & \multicolumn{2}{c|}{Mean AUC}\\ 
 \cline{5-20}
 & & & & $1e{-3}$ & $1e{-4}$ & $1e{-3}$ & $1e{-4}$ & $1e{-3}$ & $1e{-4}$ & $1e{-3}$ & $1e{-4}$ & $1e{-3}$ & $1e{-4}$ & $1e{-3}$ & $1e{-4}$ & $1e{-3}$ & $1e{-4}$ & $1e{-3}$ & $1e{-4}$\\ 
 \cline{2-20}
\hline 
\hline
\multirow{6}{*}{\rotatebox[origin=c]{90}{ArcFace \cite{deng2019arcface}}} & - & $\text{MSE}_{\text{BNS}}$ & - & 0.0262 & 0.0578 & 0.0223 & 0.0287 & 0.0151 & 0.0213 & 0.0029 & 0.0035 & 0.0631 & 0.0669 & 0.0475 & 0.0683 & 0.2207 & 0.2612 & 0.0568 & 0.0725\\ \cline{2-20}
 & $\mathcal{L}_\text{BNS}$ & $\sum |\partial \mathcal{L} / \partial \phi|$ & $\phi =  \mathcal{I}$ & 0.0245 & 0.0502 & 0.0196 & 0.0252 & 0.0117 & 0.0168 & 0.0030 & 0.0039 & 0.0609 & 0.0658 & 0.0489 & 0.0665 & 0.2664 & 0.3200 & 0.0621 & 0.0783\\ 
 & $\mathcal{L}_\text{BNS}$ & $\sum |\partial \mathcal{L} / \partial \phi|$ & $\phi = \text{B}1$ & 0.0239 & 0.0464 & 0.0202 & 0.0269 & 0.0117 & 0.0170 & 0.0027 & 0.0034 & 0.0615 & 0.0656 & 0.0481 & 0.0677 & 0.2627 & 0.3130 & 0.0615 & 0.0771\\ 
 & $\mathcal{L}_\text{BNS}$ & $\sum |\partial \mathcal{L} / \partial \phi|$ & $\phi = \text{B}2$ & 0.0225 & 0.0403 & 0.0176 & 0.0219 & 0.0070 & 0.0111 & 0.0032 & 0.0038 & 0.0644 & 0.0692 & 0.0415 & 0.0612 & 0.2058 & 0.2447 & \textbf{0.0517} & \textbf{0.0646}\\ 
 & $\mathcal{L}_\text{BNS}$ & $\sum |\partial \mathcal{L} / \partial \phi|$ & $\phi = \text{B}3$ & 0.0222 & 0.0406 & 0.0194 & 0.0276 & 0.0081 & 0.0125 & 0.0037 & 0.0041 & 0.0595 & 0.0636 & 0.0457 & 0.0675 & 0.2105 & 0.2514 & 0.0527 & 0.0668\\ 
 & $\mathcal{L}_\text{BNS}$ & $\sum |\partial \mathcal{L} / \partial \phi|$ & $\phi = \text{B}4$ & 0.0239 & 0.0458 & 0.0155 & 0.0246 & 0.0186 & 0.0277 & 0.0031 & 0.0037 & 0.0562 & 0.0598 & 0.0598 & 0.0853 & 0.2284 & 0.2721 & 0.0579 & 0.0741\\ \cline{2-20}
\hline 
\hline 
\multirow{6}{*}{\rotatebox[origin=c]{90}{ElasticFace \cite{elasticface}}} & - & $\text{MSE}_{\text{BNS}}$ & - & 0.0284 & 0.0540 & 0.0220 & 0.0236 & 0.0134 & 0.0168 & 0.0027 & 0.0035 & 0.0613 & 0.0630 & 0.0455 & 0.0595 & 0.1983 & 0.2335 & 0.0531 & 0.0648\\ \cline{2-20}
 & $\mathcal{L}_\text{BNS}$ & $\sum |\partial \mathcal{L} / \partial \phi|$ & $\phi =  \mathcal{I}$ & 0.0262 & 0.0474 & 0.0202 & 0.0217 & 0.0116 & 0.0142 & 0.0032 & 0.0039 & 0.0589 & 0.0604 & 0.0455 & 0.0573 & 0.2515 & 0.2838 & 0.0596 & 0.0698\\ 
 & $\mathcal{L}_\text{BNS}$ & $\sum |\partial \mathcal{L} / \partial \phi|$ & $\phi = \text{B}1$ & 0.0254 & 0.0448 & 0.0202 & 0.0217 & 0.0114 & 0.0140 & 0.0026 & 0.0034 & 0.0595 & 0.0608 & 0.0452 & 0.0695 & 0.2501 & 0.2802 & 0.0592 & 0.0706\\ 
 & $\mathcal{L}_\text{BNS}$ & $\sum |\partial \mathcal{L} / \partial \phi|$ & $\phi = \text{B}2$ & 0.0233 & 0.0394 & 0.0182 & 0.0200 & 0.0070 & 0.0091 & 0.0029 & 0.0037 & 0.0614 & 0.0632 & 0.0393 & 0.0633 & 0.1930 & 0.2319 & \textbf{0.0493} & \textbf{0.0615}\\ 
 & $\mathcal{L}_\text{BNS}$ & $\sum |\partial \mathcal{L} / \partial \phi|$ & $\phi = \text{B}3$ & 0.0232 & 0.0396 & 0.0193 & 0.0207 & 0.0074 & 0.0100 & 0.0035 & 0.0041 & 0.0574 & 0.0591 & 0.0422 & 0.0668 & 0.1947 & 0.2384 & 0.0497 & 0.0627\\ 
 & $\mathcal{L}_\text{BNS}$ & $\sum |\partial \mathcal{L} / \partial \phi|$ & $\phi = \text{B}4$ & 0.0250 & 0.0431 & 0.0158 & 0.0169 & 0.0133 & 0.0179 & 0.0030 & 0.0035 & 0.0539 & 0.0554 & 0.0498 & 0.0769 & 0.2058 & 0.2615 & 0.0524 & 0.0679\\ \cline{2-20}
\hline 
\hline
\multirow{6}{*}{\rotatebox[origin=c]{90}{MagFace \cite{MagFace}}} & - & $\text{MSE}_{\text{BNS}}$ & - & 0.0272 & 0.0587 & 0.0222 & 0.0366 & 0.0178 & 0.0328 & 0.0033 & 0.0041 & 0.0635 & 0.0650 & 0.0496 & 0.1000 & 0.2568 & 0.3064 & 0.0629 & 0.0862\\ \cline{2-20}
 & $\mathcal{L}_\text{BNS}$ & $\sum |\partial \mathcal{L} / \partial \phi|$ & $\phi =  \mathcal{I}$ & 0.0254 & 0.0508 & 0.0204 & 0.0355 & 0.0156 & 0.0290 & 0.0033 & 0.0043 & 0.0610 & 0.0630 & 0.0502 & 0.0984 & 0.2953 & 0.3355 & 0.0673 & 0.0881\\ 
 & $\mathcal{L}_\text{BNS}$ & $\sum |\partial \mathcal{L} / \partial \phi|$ & $\phi = \text{B}1$ & 0.0246 & 0.0472 & 0.0209 & 0.0395 & 0.0155 & 0.0287 & 0.0030 & 0.0037 & 0.0616 & 0.0634 & 0.0499 & 0.1355 & 0.2913 & 0.3384 & 0.0667 & 0.0938\\ 
 & $\mathcal{L}_\text{BNS}$ & $\sum |\partial \mathcal{L} / \partial \phi|$ & $\phi = \text{B}2$ & 0.0233 & 0.0419 & 0.0182 & 0.0253 & 0.0087 & 0.0186 & 0.0033 & 0.0041 & 0.0640 & 0.0652 & 0.0428 & 0.0987 & 0.2524 & 0.3018 & 0.0590 & \textbf{0.0794}\\ 
 & $\mathcal{L}_\text{BNS}$ & $\sum |\partial \mathcal{L} / \partial \phi|$ & $\phi = \text{B}3$ & 0.0229 & 0.0420 & 0.0203 & 0.0409 & 0.0098 & 0.0187 & 0.0038 & 0.0045 & 0.0590 & 0.0603 & 0.0462 & 0.1073 & 0.2479 & 0.2905 & \textbf{0.0586} & 0.0806\\ 
 & $\mathcal{L}_\text{BNS}$ & $\sum |\partial \mathcal{L} / \partial \phi|$ & $\phi = \text{B}4$ & 0.0247 & 0.0459 & 0.0169 & 0.0391 & 0.0208 & 0.0380 & 0.0033 & 0.0039 & 0.0557 & 0.0569 & 0.0577 & 0.1266 & 0.2654 & 0.3187 & 0.0635 & 0.0899\\ \cline{2-20}
\hline 
\hline
\multirow{6}{*}{\rotatebox[origin=c]{90}{\makecell{Curricular- \\ Face \cite{curricularFace}}}} & - & $\text{MSE}_{\text{BNS}}$ & - & 0.0245 & 0.0496 & 0.0212 & 0.0250 & 0.0145 & 0.0191 & 0.0029 & 0.0035 & 0.0621 & 0.0655 & 0.0426 & 0.0618 & 0.1863 & 0.2163 & 0.0506 & 0.0630\\ \cline{2-20}
 & $\mathcal{L}_\text{BNS}$ & $\sum |\partial \mathcal{L} / \partial \phi|$ & $\phi =  \mathcal{I}$ & 0.0233 & 0.0429 & 0.0211 & 0.0238 & 0.0120 & 0.0161 & 0.0034 & 0.0039 & 0.0591 & 0.0623 & 0.0422 & 0.0578 & 0.2233 & 0.2652 & 0.0549 & 0.0674\\ 
 & $\mathcal{L}_\text{BNS}$ & $\sum |\partial \mathcal{L} / \partial \phi|$ & $\phi = \text{B}1$ & 0.0229 & 0.0410 & 0.0208 & 0.0237 & 0.0117 & 0.0160 & 0.0028 & 0.0034 & 0.0597 & 0.0624 & 0.0424 & 0.0728 & 0.2263 & 0.2639 & 0.0552 & 0.0690\\ 
 & $\mathcal{L}_\text{BNS}$ & $\sum |\partial \mathcal{L} / \partial \phi|$ & $\phi = \text{B}2$ & 0.0220 & 0.0365 & 0.0167 & 0.0200 & 0.0068 & 0.0099 & 0.0033 & 0.0038 & 0.0610 & 0.0641 & 0.0369 & 0.0663 & 0.1713 & 0.1959 & \textbf{0.0454} & \textbf{0.0566}\\ 
 & $\mathcal{L}_\text{BNS}$ & $\sum |\partial \mathcal{L} / \partial \phi|$ & $\phi = \text{B}3$ & 0.0214 & 0.0357 & 0.0179 & 0.0211 & 0.0079 & 0.0117 & 0.0038 & 0.0043 & 0.0591 & 0.0616 & 0.0387 & 0.0694 & 0.1860 & 0.2143 & 0.0478 & 0.0597\\ 
 & $\mathcal{L}_\text{BNS}$ & $\sum |\partial \mathcal{L} / \partial \phi|$ & $\phi = \text{B}4$ & 0.0225 & 0.0390 & 0.0151 & 0.0190 & 0.0159 & 0.0240 & 0.0033 & 0.0037 & 0.0550 & 0.0575 & 0.0448 & 0.0742 & 0.2025 & 0.2300 & 0.0513 & 0.0639\\ \cline{2-20}
\hline 
\end{tabular}
}
\end{center}
\vspace{-3mm}
\end{table*}

\vspace{-1mm}
\section{Results}
\label{sec:ablation}
\vspace{-1mm}
This section first provides empirical proof of the use of gradient magnitude to assess the utility of face images. 
Table \ref{tbl:ablation} presents the verification performance as AUC at FMR$=1e-3$ and FMR$=1e-4$ reported on seven benchmarks.
The results are reported using two protocols, same model and cross-model protocol.
All quality values are calculated using the ArcFace model and the verification performances at different discard rates are provided using ArcFace (same model) as well as ElasticFace, MagFace and CurricularFace  (cross-model).
The corresponding EDCs are provided in Figure \ref{fig:iresnet100_short_overview}.

For each protocol, we first evaluate the use of the BNS based $\text{MSE}_{\text{BSN}}$ function (Equation \ref{eq:mse_bsn}) as FIQ.
The function $\text{MSE}_{\text{BSN}}$ measures the change in the distribution between the model training data and test samples.
In this experiment, we use $\text{MSE}_{\text{BSN}}$ directly to assess the quality and we do not backpropagate $\text{MSE}_{\text{BSN}}$ through the network.
Higher values resulting from $\text{MSE}_{\text{BSN}}$ indicate low quality and vice versa. 
It can be seen in Figure \ref{fig:iresnet100_short_overview} (black, dashed line) that discarding samples with high $\text{MSE}_{\text{BSN}}$ (low quality) improves the verification accuracies where EDC (in both same and cross model evaluation protocols) is dropped when discarding a fraction of low-quality samples. 

In the next study, we evaluate the use of gradient magnitude to assess the quality of face images.
As we discussed in Section \ref{sec:methodology}, the $\text{MSE}_{\text{BSN}}$ does not fully describe the strength and level of difference between the sample and the data distribution learned from the model.
For each of the evaluation protocols considered, we provide detailed evaluations of the sums of absolute gradient magnitudes (as FIQ) extracted during the backpropagation step from four intermediate layers and on the image level (noted as as $\mathcal{I}$). 
The network architecture of utilized ArcFace models is ResNet100 which consists mainly of 4 stages, each consisting of several stacked residual blocks equal to 3, 4, 23, and 3, respectively. 
We consider the output of each of the stages (during the backpropagation step) as intermediate gradient magnitude, noted as B$1$, B$2$, B$3$, and B$4$, respectively.
It can be clearly noticed that utilizing gradient magnitudes that result from $\mathcal{L}_{\text{BSN}}$ and are extracted from different intermediate layers to assess face image quality achieved higher verification than directly using $\text{MSE}_{\text{BSN}}$ in most of the considered settings.
Specifically, utilizing gradient magnitude from B$2$ achieved the best overall performance. The results are also supported by EDC curves in Figure \ref{fig:iresnet100_short_overview}, where EDC is dropped when discarding a fraction of low-quality (high gradient magnitude) samples. 

\input{figures/fig_iresnet100_short-overview-paper}

\begin{table*}[ht!]
    \begin{center}%
    \caption{The AUCs of EDC achieved by our \grafiqs and the SOTA methods under different experimental settings. The notions of $1e-3$ and $1e-4$ indicate the value of the fixed FMR at which the EDC curves (FNMR vs.~reject) were calculated. The results are compared to three IQA and nine ten FIQA approaches. The mean AUC overall evaluation datasets (except XQLFW as it was labeled by SER-FIQ \cite{SERFIQ}) at FMR$=1e-3$ and FMR$=1e-4$ per method are shown in the last column. The XQLFW dataset uses SER-FIQ (marked with *) as FIQ labeling method.}
    \label{tab:erc_sota_comparison}
    \vspace{-3mm}
    \resizebox{0.99\textwidth}{!}{
\begin{tabular}{|c|cl|ll|ll|ll|ll|ll|ll|ll|ll|}
    \hline
     \multirow{2}{*}{FR}                                                            & \multicolumn{2}{c|}{\multirow{2}{*}{Method}}      & \multicolumn{2}{c|}{Adience\cite{Adience}} & \multicolumn{2}{c|}{AgeDB-30\cite{agedb}} & \multicolumn{2}{c|}{CFP-FP\cite{cfp-fp}} & \multicolumn{2}{c|}{LFW\cite{LFWTech}} & \multicolumn{2}{c|}{CALFW\cite{CALFW}} & \multicolumn{2}{c|}{CPLFW\cite{CPLFWTech}} & \multicolumn{2}{c|}{XQLFW\cite{XQLFW}} & \multicolumn{2}{c|}{Mean AUC}\\ 
                                                                                    &                                                   &  & $1e{-3}$ & $1e{-4}$ & $1e{-3}$ & $1e{-4}$  & $1e{-3}$ & $1e{-4}$  & $1e{-3}$ & $1e{-4}$  & $1e{-3}$ & $1e{-4}$ & $1e^{-3}$ & $1e{-4}$ & $1e{-3}$ & $1e{-4}$ & $1e{-3}$ & $1e{-4}$\\ \hline
     \hline 
     \multirow{14}{*}{\rotatebox[origin=c]{90}{ArcFace\cite{deng2019arcface}}}      & \multirow{3}{*}{\rotatebox[origin=c]{90}{IQA}}    & \multicolumn{1}{|l|}{BRISQUE\cite{BRISQE_IQA}} &0.0565 & 0.1285 &0.0400 & 0.0585 &0.0343 & 0.0433 &0.0043 & 0.0049 &0.0755 & 0.0813 &0.2558 & 0.3037 &0.6680 & 0.7122                                             &	0.0777	&	0.1034     \\ 
                                                                                    &                                                   & \multicolumn{1}{|l|}{RankIQA\cite{liu2017rankiqa}} &0.0400 & 0.0933 &0.0372 & 0.0523 &0.0301 & 0.0384 &0.0039 & 0.0045 &0.0846 & 0.0915 &0.2437 & 0.2969 &0.6584 & 0.7039                                         &	0.0733	&	0.0962\\ 
                                                                                    &                                                   & \multicolumn{1}{|l|}{DeepIQA\cite{DEEPIQ_IQA}} &0.0568 & 0.1372 &0.0403 & 0.0523 &0.0238 & 0.0292 &0.0049 & 0.0056 &0.0793 & 0.0850 &0.2309 & 0.2856 &0.5958 & 0.6458                                             &	0.0727	&	0.0992\\ \cline{2-19}
                                                                                    & \multirow{11}{*}{\rotatebox[origin=c]{90}{FIQA}}   & \multicolumn{1}{|l|}{RankIQ\cite{RANKIQ_FIQA} }&0.0353 & 0.0873 &0.0322 & 0.0420 &0.0152 & 0.0260 &0.0018 & 0.0024 &0.0608 & 0.0672 &0.0633 & 0.0848 &0.2789 & 0.3332                                             &	0.0348	&	0.0516\\ 
                                                                                    &                                                   & \multicolumn{1}{|l|}{PFE\cite{PFE_FIQA}} &0.0212 & 0.0428 &0.0172 & 0.0226 &0.0092 & 0.0129 &0.0023 & 0.0028 &0.0647 & 0.0681 & 0.0450 & 0.0638 &0.2302 & 0.2710                                                  &	0.0266	&	0.0355\\ 
                                                                                    &                                                   & \multicolumn{1}{|l|}{SER-FIQ\cite{SERFIQ}} &0.0223 & 0.0434 &0.0167 & 0.0223 & 0.0065 & 0.0103 & 0.0023 & 0.0028 & 0.0595 & 0.0627 &0.0389 & 0.0584 & 0.1812$^{*}$ & 0.2295$^{*}$                                 &	0.0244	&	0.0333\\ 
                                                                                    &                                                   & \multicolumn{1}{|l|}{FaceQnet\cite{hernandez2019faceqnet,faceqnetv1}} & 0.0346 & 0.0734 &0.0197 & 0.0245 &0.0240 & 0.0273 &0.0022 & 0.0027 &0.0774 & 0.0822 &0.1504 & 0.1751 &0.5829  & 0.6136                    &	0.0514	&	0.0642 \\ 
                                                                                    &                                                   & \multicolumn{1}{|l|}{MagFace\cite{MagFace}} & 0.0207 & 0.0425 & 0.0156 & 0.0198 &0.0073 & 0.0105 & 0.0016 & 0.0021 & 0.0568 & 0.0602 &0.0492 & 0.0642 &0.4022 & 0.4636                                            &	0.0252	&	0.0332\\ 
                                                                                    &                                                   & \multicolumn{1}{|l|}{SDD-FIQA\cite{SDDFIQA}} &0.0248 & 0.0562 &0.0186 & 0.0206 &0.0122 & 0.0193 &0.0021 & 0.0027 &0.0641 & 0.0698 &0.0517 & 0.0670 &0.3090 & 0.3561                                               &	0.0289	&	0.0393\\ 
                                                                                    &                                                   & \multicolumn{1}{|l|}{CR-FIQA(S) \cite{boutros_2023_crfiqa}} & 0.0241 & 0.0517 & 0.0144 & 0.0187 & 0.0090 & 0.0145 & 0.0020 & 0.0025 & 0.0521 & 0.0554 & 0.0391 & 0.0567 & 0.2377 & 0.2740                         & \textit{0.0234}    & 0.0333 \\
                                                                                    &                                                   & \multicolumn{1}{|l|}{CR-FIQA(L) \cite{boutros_2023_crfiqa}} & 0.0204 & 0.0353 & 0.0159 & 0.0189 & 0.0050 & 0.0082 & 0.0023 & 0.0029 & 0.0616 & 0.0632 & 0.0360 & 0.0515 & 0.2084 & 0.2441                         &	0.0235	&	\textbf{0.0300}\\ 
                                                                                    &                                                   & \multicolumn{1}{|l|}{DifFIQA(R) \cite{10449044}} & 0.0251 & 0.0619 & 0.0194 & 0.0262 & 0.0053 & 0.0091 & 0.0020 & 0.0025 & 0.0629 & 0.0688 & 0.0365 & 0.0531 & 0.1847 & 0.2397                                                      &   0.0252  & 0.0369\\
                                                                                    &                                                   & \multicolumn{1}{|l|}{eDifFIQA(L) \cite{babnikTBIOM2024}} & 0.0210 & 0.0402 & 0.0148 & 0.0176 & 0.0049 & 0.0083 & 0.0014 & 0.0019 & 0.0574 & 0.0627 & 0.0342 & 0.0500 & 0.1917 & 0.2469                                                     &   \textbf{0.0223}  & \textit{0.0301} \\  \cline{3-19}
                                                                                    &                                                   & \multicolumn{1}{|l|}{\grafiqs $\mathcal{L}_{\text{BNS}},\phi_{2}$ (Our)} & 0.0225 & 0.0403 & 0.0176 & 0.0219 & 0.0070 & 0.0111 & 0.0032 & 0.0038 & 0.0644 & 0.0692 & 0.0415 & 0.0612 & 0.2058 & 0.2447            &	0.0260	&	0.0346\\ 
     \hline \hline 
     \multirow{14}{*}{\rotatebox[origin=c]{90}{ElasticFace\cite{elasticface}}}      & \multirow{3}{*}{\rotatebox[origin=c]{90}{IQA}}    & \multicolumn{1}{|l|}{BRISQUE\cite{BRISQE_IQA}} &0.0644 & 0.1184 &0.0375 & 0.0403 &0.0281 & 0.0372 &0.0034 & 0.0047 &0.0726 & 0.0747 &0.2641 & 0.4688 &0.6343 & 0.6964                                             &	0.0784	&	0.1240\\ 
                                                                                    &                                                   & \multicolumn{1}{|l|}{RankIQA\cite{liu2017rankiqa}} &0.0433 & 0.0862 &0.0374 & 0.0436 &0.0269 & 0.0318 &0.0033 & 0.0045 &0.0810 & 0.0835 &0.2325 & 0.4306 &0.6189 & 0.6856                                         &	0.0707	&	0.1134\\ 
                                                                                    &                                                   & \multicolumn{1}{|l|}{DeepIQA\cite{DEEPIQ_IQA}} &0.0645 & 0.1203 &0.0384 & 0.0411 &0.0191 & 0.0256 &0.0043 & 0.0056 &0.0756 & 0.0772 &0.2401 & 0.4541 &0.5400 & 0.5832                                             &	0.0737	&	0.1207\\ \cline{2-19}
                                                                                    &\multirow{11}{*}{\rotatebox[origin=c]{90}{FIQA}}    & \multicolumn{1}{|l|}{RankIQ\cite{RANKIQ_FIQA} }& 0.0400 & 0.0777 &0.0309 & 0.0337 &0.0149 & 0.0180 & 0.0013 & 0.0020 &0.0598 & 0.0614 &0.0581 & 0.0727 &0.2468 & 0.2776                                           &	0.0342	&	0.0443\\ 
                                                                                    &                                                   & \multicolumn{1}{|l|}{PFE\cite{PFE_FIQA}} & 0.0222 & 0.0381 &0.0163 & 0.0172 &0.0088 & 0.0113 &0.0018 & 0.0025 &0.0628 & 0.0643 &0.0419 & 0.0895 &0.2112 & 0.2436                                                  &	0.0256	&	0.0372\\ 
                                                                                    &                                                   & \multicolumn{1}{|l|}{SER-FIQ\cite{SERFIQ} } &0.0240 & 0.0417 &0.0163 & 0.0179 &0.0061 & 0.0085 &0.0021 & 0.0028 &0.0574 & 0.0590 &0.0387 & 0.0513 & 0.1576$^{*}$ & 0.1868$^{*}$                                   &	0.0241	&	0.0302\\ 
                                                                                    &                                                   & \multicolumn{1}{|l|}{FaceQnet\cite{hernandez2019faceqnet,faceqnetv1}} &0.0369 & 0.0667 &0.0194 & 0.0207 &0.0227 & 0.0247 &0.0021 & 0.0026 &0.0763 & 0.0777 &0.1420 & 0.2880 &0.5549 & 0.5844                      &	0.0499	&	0.0801\\ 
                                                                                    &                                                   & \multicolumn{1}{|l|}{MagFace\cite{MagFace}} & 0.0225 & 0.0385 &0.0150 & 0.0158 &0.0069 & 0.0095 & 0.0014 &0.0021 & 0.0553 & 0.0563 & 0.0474 & 0.0597 &0.3973 & 0.4282                                             &	0.0248	&	0.0303\\ 
                                                                                    &                                                   & \multicolumn{1}{|l|}{SDD-FIQA\cite{SDDFIQA}} &0.0277 & 0.0512 &0.0187 & 0.0200 &0.0098 & 0.0118 &0.0019 & 0.0027 &0.0624 & 0.0638 &0.0493 & 0.0634 &0.3052 & 0.3562                                               &	0.0283	&	0.0355\\
                                                                                    &                                                   & \multicolumn{1}{|l|}{CR-FIQA(S) \cite{boutros_2023_crfiqa}} & 0.0257 & 0.0465 & 0.0146 & 0.0160 & 0.0070 & 0.0096 & 0.0015 & 0.0022 & 0.0509 & 0.0522 & 0.0383 & 0.0502 & 0.2093 & 0.2835 & 0.0230 & 0.0295 \\
                                                                                    &                                                   & \multicolumn{1}{|l|}{CR-FIQA(L) \cite{boutros_2023_crfiqa}} & 0.0214 & 0.0357 &0.0149 & 0.0159 & 0.0045 & 0.0065 &0.0018 & 0.0025 &0.0594 & 0.0608 & 0.0350 & 0.0462 &0.1798 & 0.2060                             &	\textit{0.0228}	&	\textit{0.0279}\\
                                                                                    &                                                   & \multicolumn{1}{|l|}{DifFIQA(R) \cite{10449044}} & 0.0278 & 0.0536 & 0.0194 & 0.0207 & 0.0050 & 0.0073 & 0.0019 & 0.0025 & 0.0616 & 0.0634 & 0.0330 & 0.0445 & 0.1599 & 0.1890                                                      & 0.0248            & 0.0320\\
                                                                                    &                                                   & \multicolumn{1}{|l|}{eDifFIQA(L) \cite{babnikTBIOM2024}} & 0.0222 & 0.0374 & 0.0139 & 0.0148 & 0.0043 & 0.0066 & 0.0014 & 0.0019 & 0.0564 & 0.0576 & 0.0323 & 0.0440 & 0.1688 & 0.1996                                                     & \textbf{0.0218}            & \textbf{0.0271}\\  \cline{3-19}
                                                                                    &                                                   & \multicolumn{1}{|l|}{\grafiqs $\mathcal{L}_{\text{BNS}},\phi_{2}$ (Our)} & 0.0233 & 0.0394 & 0.0182 & 0.0200 & 0.0070 & 0.0091 & 0.0029 & 0.0037 & 0.0614 & 0.0632 & 0.0393 & 0.0633 & 0.1930 & 0.2319            &	0.0254	&	0.0331\\ 
     \hline \hline 
     \multirow{14}{*}{\rotatebox[origin=c]{90}{MagFace\cite{MagFace}}}              & \multirow{3}{*}{\rotatebox[origin=c]{90}{IQA}} & \multicolumn{1}{|l|}{BRISQUE\cite{BRISQE_IQA}} &0.0594 & 0.1308 &0.0442 & 0.0799 &0.0422 & 0.0589 &0.0043 & 0.0058 &0.0758 & 0.0788 &0.4649 & 0.6809 &0.6911 & 0.7229                                                 &	0.1151	&	0.1725\\ 
                                                                                    &                                                   & \multicolumn{1}{|l|}{RankIQA\cite{liu2017rankiqa}} &0.0407 & 0.0889 &0.0370 & 0.0681 &0.0369 & 0.0543 &0.0041 & 0.0056 &0.0829 & 0.0857 &0.3251 & 0.6475 &0.6706 & 0.7046                                         &	0.0878	&	0.1584\\ 
                                                                                    &                                                   & \multicolumn{1}{|l|}{DeepIQA\cite{DEEPIQ_IQA}} &0.0571 & 0.1302 &0.0417 & 0.0721 &0.0322 & 0.0545 &0.0048 & 0.0059 &0.0787 & 0.0809 &0.3672 & 0.6632 &0.6162 & 0.6519                                             &	0.0970	&	0.1678\\  \cline{2-19}
                                                                                    & \multirow{11}{*}{\rotatebox[origin=c]{90}{FIQA}} & \multicolumn{1}{|l|}{RankIQ\cite{RANKIQ_FIQA}}&0.0359 & 0.0837 &0.0361 & 0.0531 &0.0213 & 0.0332 & 0.0019 & 0.0027 &0.0602 & 0.0629 &0.0659 & 0.1642 &0.3076 & 0.3475                                                &	0.0369	&	0.0666\\ 
                                                                                    &                                                   & \multicolumn{1}{|l|}{PFE\cite{PFE_FIQA}} &0.0215 & 0.0423 &0.0192 & 0.0317 &0.0107 & 0.0138 &0.0023 & 0.0029 &0.0640 & 0.0652 &0.0449 & 0.1435 &0.2615 & 0.2926                                                    &	0.0271	&	0.0499\\ 
                                                                                    &                                                   & \multicolumn{1}{|l|}{SER-FIQ\cite{SERFIQ} } &0.0233 & 0.0451 &0.0185 & 0.0293 & 0.0080 & 0.0139 &0.0025 & 0.0033 &0.0590 & 0.0607 & 0.0397 & 0.0821 & 0.2139$^*$ & 0.2562$^*$                                         &	0.0252	&	0.0391\\ 
                                                                                    &                                                   & \multicolumn{1}{|l|}{FaceQnet\cite{hernandez2019faceqnet,faceqnetv1}} &0.0365 & 0.0720 &0.0217 & 0.0314 &0.0271 & 0.0351 &0.0022 & 0.0027 &0.0763 & 0.0773 &0.2988 & 0.5218 &0.6016 & 0.6210                      &	0.0771	&	0.1234\\ 
                                                                                    &                                                   & \multicolumn{1}{|l|}{MagFace\cite{MagFace}} & 0.0212 & 0.0417 & 0.0159 & 0.0247 &0.0085 & 0.0129 & 0.0017 & 0.0022 & 0.0562 & 0.0578 &0.0506 & 0.0887 &0.4478 & 0.4900                                            &	 0.0257	&	0.0380 \\ 
                                                                                    &                                                   & \multicolumn{1}{|l|}{SDD-FIQA\cite{SDDFIQA}} &0.0253 & 0.0562 &0.0216 & 0.0305 &0.0146 & 0.0201 &0.0021 & 0.0027 &0.0643 & 0.0657 &0.0525 & 0.1188 &0.3404 & 0.3928                                                &	0.0301	&	0.0490\\ 
                                                                                    &                                                   & \multicolumn{1}{|l|}{CR-FIQA(S) \cite{boutros_2023_crfiqa}} & 0.0244 & 0.0507 & 0.0165 & 0.0234 & 0.0102 & 0.0121 & 0.0020 & 0.0028 & 0.0516 & 0.0528 & 0.0409 & 0.0840 & 0.2670 & 0.3336 & \textit{0.0243} & 0.0376 \\
                                                                                    &                                                   & \multicolumn{1}{|l|}{CR-FIQA(L) \cite{boutros_2023_crfiqa}} & 0.0211 & 0.0372 &0.0174 & 0.0235 &0.0062 & 0.0080 &0.0023 & 0.0028 &0.0614 & 0.0628 & 0.0374 &0.0679 & 0.2369 & 0.2839                              &	\textit{0.0243}	& \textbf{0.0337}\\
                                                                                    &                                                   & \multicolumn{1}{|l|}{DifFIQA(R) \cite{10449044}} & 0.0256 & 0.0585 & 0.0223 & 0.0363 & 0.0066 & 0.0150 & 0.0020 & 0.0025 & 0.0638 & 0.0660 & 0.0371 & 0.0851 & 0.2177 & 0.2642                                                      & 0.0262 & 0.0439\\
                                                                                    &                                                   & \multicolumn{1}{|l|}{eDifFIQA(L) \cite{babnikTBIOM2024}} & 0.0216 & 0.0403 & 0.0168 & 0.0246 & 0.0058 & 0.0121 & 0.0014 & 0.0019 & 0.0580 & 0.0595 & 0.0357 & 0.0810 & 0.2278 & 0.2792                                                     & \textbf{0.0232} & \textit{0.0366}\\  \cline{3-19}
                                                                                    &                                                   & \multicolumn{1}{|l|}{\grafiqs $\mathcal{L}_{\text{BNS}},\phi_{2}$ (Our)} & 0.0233 & 0.0419 & 0.0182 & 0.0253 & 0.0087 & 0.0186 & 0.0033 & 0.0041 & 0.0640 & 0.0652 & 0.0428 & 0.0987 & 0.2524 & 0.3018             &	0.0267	&	0.0423\\ 
     \hline \hline 
     \multirow{14}{*}{\rotatebox[origin=c]{90}{CurricularFace\cite{curricularFace}}}& \multirow{3}{*}{\rotatebox[origin=c]{90}{IQA}}    & \multicolumn{1}{|l|}{BRISQUE\cite{BRISQE_IQA}} &0.0502 & 0.1095 &0.0433 & 0.0491 &0.0323 & 0.0357 &0.0041 & 0.0047 &0.0755 & 0.0784 &0.2709 & 0.5057 &0.6146 & 0.6336                                             &	0.0794	&	0.1305 \\ 
                                                                                    &                                                   & \multicolumn{1}{|l|}{RankIQA\cite{liu2017rankiqa}} &0.0359 & 0.0752 &0.0394 & 0.0510 &0.0298 & 0.0356 &0.0039 & 0.0045 &0.0806 & 0.0865 &0.2346 & 0.4654 &0.5900 & 0.6212                                          &	0.0707	&	0.1197 \\ 
                                                                                    &                                                   & \multicolumn{1}{|l|}{DeepIQA\cite{DEEPIQ_IQA}} &0.0492 & 0.1070 &0.0407 & 0.0476 &0.0227 & 0.0278 &0.0050 & 0.0056 &0.0764 & 0.0786 &0.2488 & 0.4961 &0.5165 & 0.5526                                             &	0.0738	&	0.1271 \\  \cline{2-19}
                                                                                    & \multirow{11}{*}{\rotatebox[origin=c]{90}{FIQA}}   & \multicolumn{1}{|l|}{RankIQ\cite{RANKIQ_FIQA} }&0.0314 & 0.0715 &0.0365 & 0.0417 &0.0186 & 0.0249 & 0.0018 & 0.0024 &0.0590 & 0.0640 &0.0541 & 0.0730 &0.2449 & 0.2880                                                & 0.0336	&	0.0463	\\ 
                                                                                    &                                                   & \multicolumn{1}{|l|}{PFE\cite{PFE_FIQA}} & 0.0198 & 0.0365 &0.0197 & 0.0227 &0.0100 & 0.0134 &0.0024 & 0.0028 &0.0630 & 0.0657 &0.0402 & 0.0983 &0.1982 & 0.2220                                                      & 0.0259	&	0.0399	\\ 
                                                                                    &                                                   & \multicolumn{1}{|l|}{SER-FIQ\cite{SERFIQ}} &0.0211 & 0.0381 &0.0167 & 0.0193 & 0.0074 & 0.0111 &0.0025 & 0.0030 &0.0587 & 0.0610 &0.0356 & 0.0520 & 0.1558$^*$ & 0.1866$^*$                                           & 0.0237	&	0.0308	\\ 
                                                                                    &                                                   & \multicolumn{1}{|l|}{FaceQNet\cite{hernandez2019faceqnet,faceqnetv1}} &0.0326 & 0.0626 &0.0221 & 0.0267 &0.0226 & 0.0274 &0.0022 & 0.0027 &0.0767 & 0.0799 &0.1384 & 0.3229 &0.5035 & 0.5411                          & 0.0491	&	0.0870	\\ 
                                                                                    &                                                   & \multicolumn{1}{|l|}{MagFace\cite{MagFace}} & 0.0200 & 0.0364 & 0.0167 & 0.0195 &0.0078 & 0.0111 & 0.0016 & 0.0021 & 0.0563 & 0.0590 &0.0449 & 0.0607 &0.3758 & 0.4178                                                & 0.0246	&	0.0315	\\ 
                                                                                    &                                                   & \multicolumn{1}{|l|}{SDD-FIQA\cite{SDDFIQA}} &0.0230 & 0.0462 &0.0219 & 0.0254 &0.0138 & 0.0185 &0.0021 & 0.0027 &0.0637 & 0.0675 &0.0465 & 0.0671 &0.2649 & 0.3053                                                   & 0.0285	&	0.0379	\\ 
                                                                                    &                                                   & \multicolumn{1}{|l|}{CR-FIQA(S) \cite{boutros_2023_crfiqa}} & 0.0227 & 0.0446 & 0.0156 & 0.0198 & 0.0097 & 0.0148 & 0.0020 & 0.0025 & 0.0513 & 0.0534 & 0.0340 & 0.0501 & 0.2101 & 0.2470 & \textit{0.0226} & 0.0309 \\
                                                                                    &                                                   & \multicolumn{1}{|l|}{CR-FIQA(L) \cite{boutros_2023_crfiqa}} & 0.0198 & 0.0336 & 0.0162 & 0.0200 & 0.0054 & 0.0080 &0.0023 & 0.0029 &0.0605 & 0.0618 & 0.0324 & 0.0462 & 0.1716 & 0.2318                               & 0.0228	&	\textit{0.0288}	\\
                                                                                    &                                                   & \multicolumn{1}{|l|}{DifFIQA(R) \cite{10449044}} & 0.0230 & 0.0475 & 0.0227 & 0.0260 & 0.0055 & 0.0092 & 0.0020 & 0.0025 & 0.0608 & 0.0657 & 0.0305 & 0.0441 & 0.1600 & 0.1871                                                          & 0.0241 & 0.0325\\
                                                                                    &                                                   & \multicolumn{1}{|l|}{eDifFIQA(L) \cite{babnikTBIOM2024}} & 0.0199 & 0.0338 & 0.0170 & 0.0195 & 0.0048 & 0.0084 & 0.0014 & 0.0019 & 0.0566 & 0.0601 & 0.0303 & 0.0455 & 0.1717 & 0.2080                                                         & \textbf{0.0217} & \textbf{0.0282}\\  \cline{3-19}
                                                                                    &                                                   & \multicolumn{1}{|l|}{\grafiqs $\mathcal{L}_{\text{BNS}},\phi_{2}$ (Our)} & 0.0220 & 0.0365 & 0.0167 & 0.0200 & 0.0068 & 0.0099 & 0.0033 & 0.0038 & 0.0610 & 0.0641 & 0.0369 & 0.0663 & 0.1713 & 0.1959                & 0.0245	&	0.0334	\\ 
     \hline 
    \end{tabular}}
    \end{center}
    \vspace{-5mm}
    \end{table*}

\subsection{Comparison with SOTA FIQ}
\begin{figure*}[h!]
\centering
	\begin{subfigure}[b]{0.92\textwidth}
		\centering
		\includegraphics[width=\textwidth]{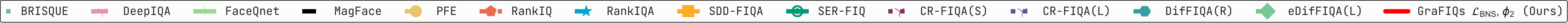}
	\end{subfigure}
\\
	\begin{subfigure}[b]{0.32\textwidth}
		 \centering
		 \includegraphics[width=0.8\textwidth]{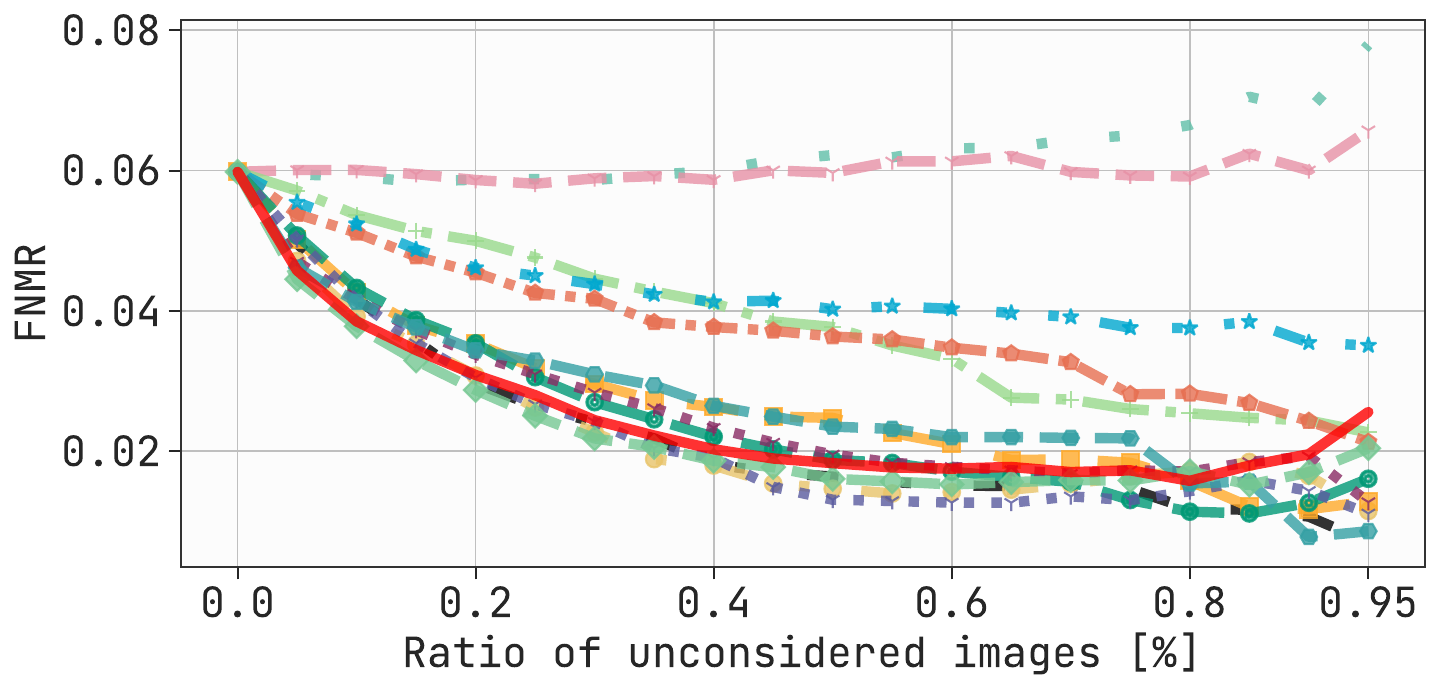}
		 \caption{MagFace Model, Adience Dataset}
	\end{subfigure}
\hfill
	\begin{subfigure}[b]{0.32\textwidth}
		 \centering
		 \includegraphics[width=0.8\textwidth]{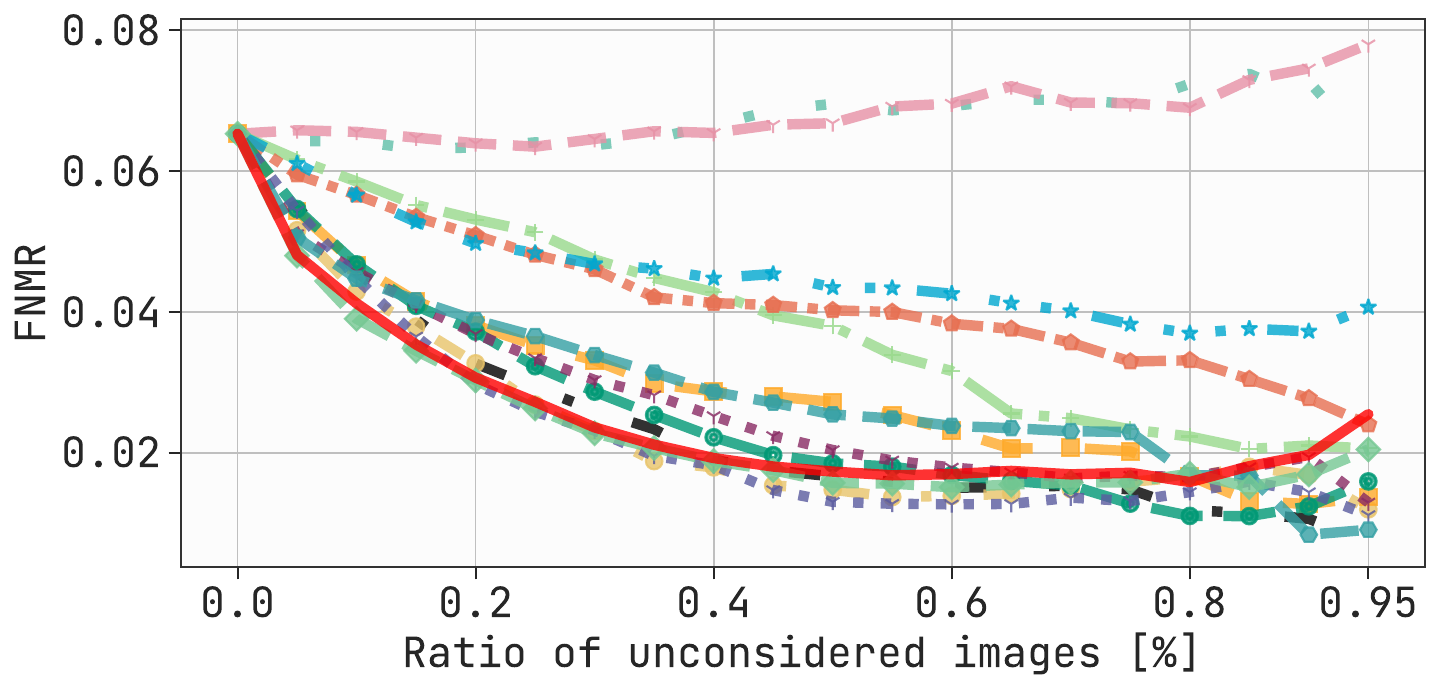}
		 \caption{ElasticFace Model, Adience Dataset}
	\end{subfigure}
\hfill
	\begin{subfigure}[b]{0.32\textwidth}
		 \centering
		 \includegraphics[width=0.8\textwidth]{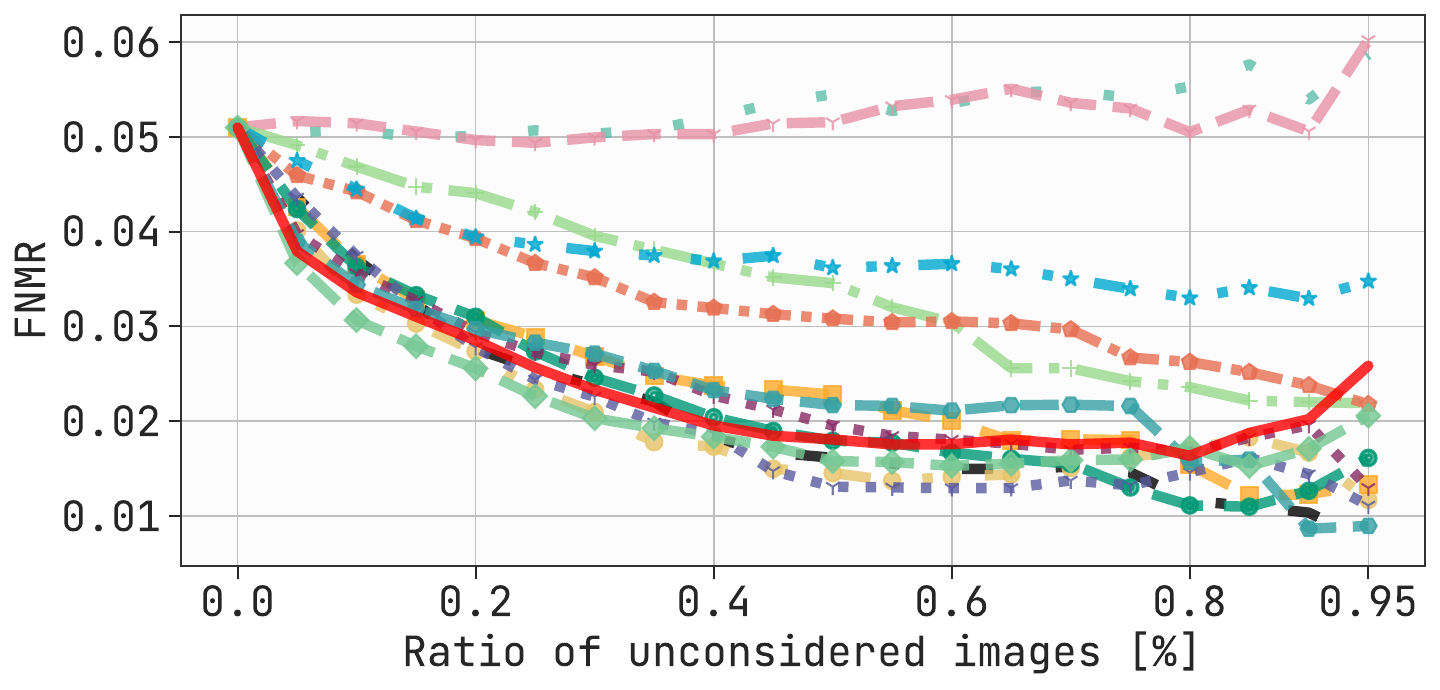}
		 \caption{CurricularFace Model, Adience Dataset}
	\end{subfigure}
\\
	\begin{subfigure}[b]{0.32\textwidth}
		 \centering
		 \includegraphics[width=0.8\textwidth]{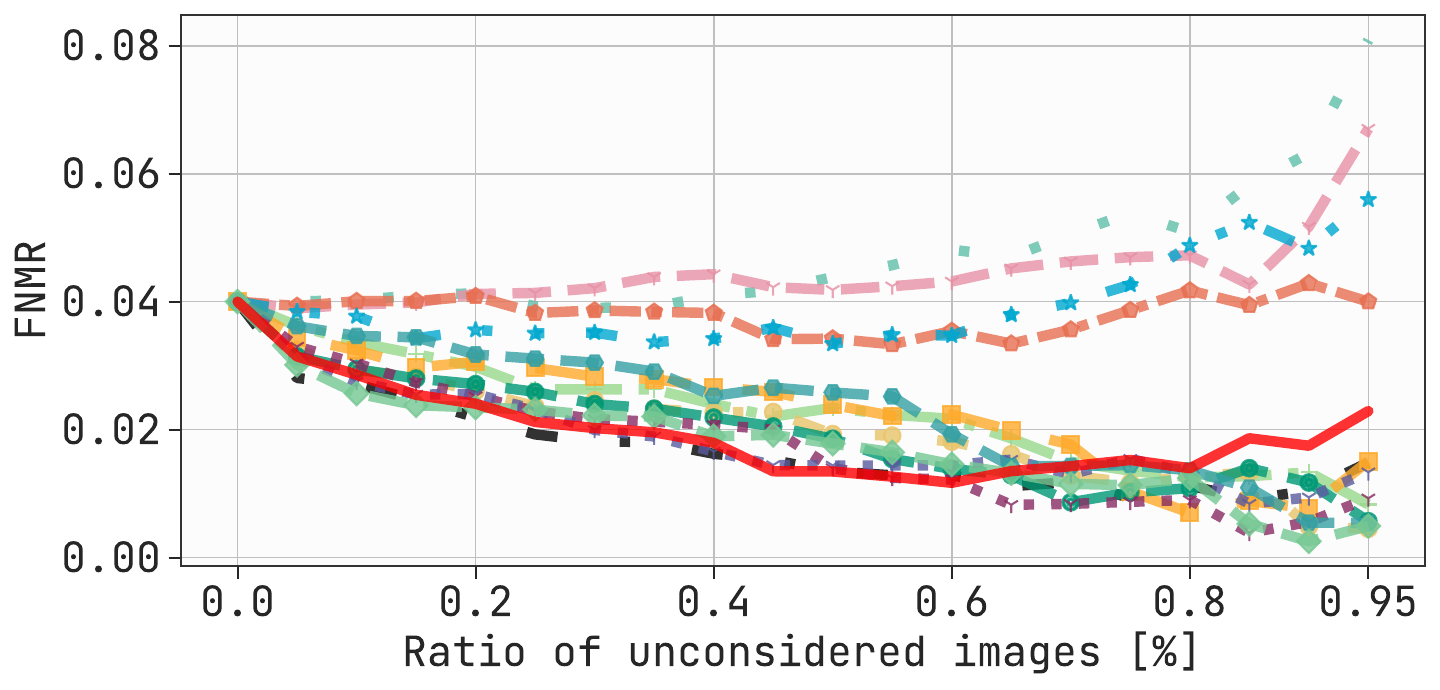}
		 \caption{MagFace Model, AgeDB30 Dataset}
	\end{subfigure}
\hfill
	\begin{subfigure}[b]{0.32\textwidth}
		 \centering
		 \includegraphics[width=0.8\textwidth]{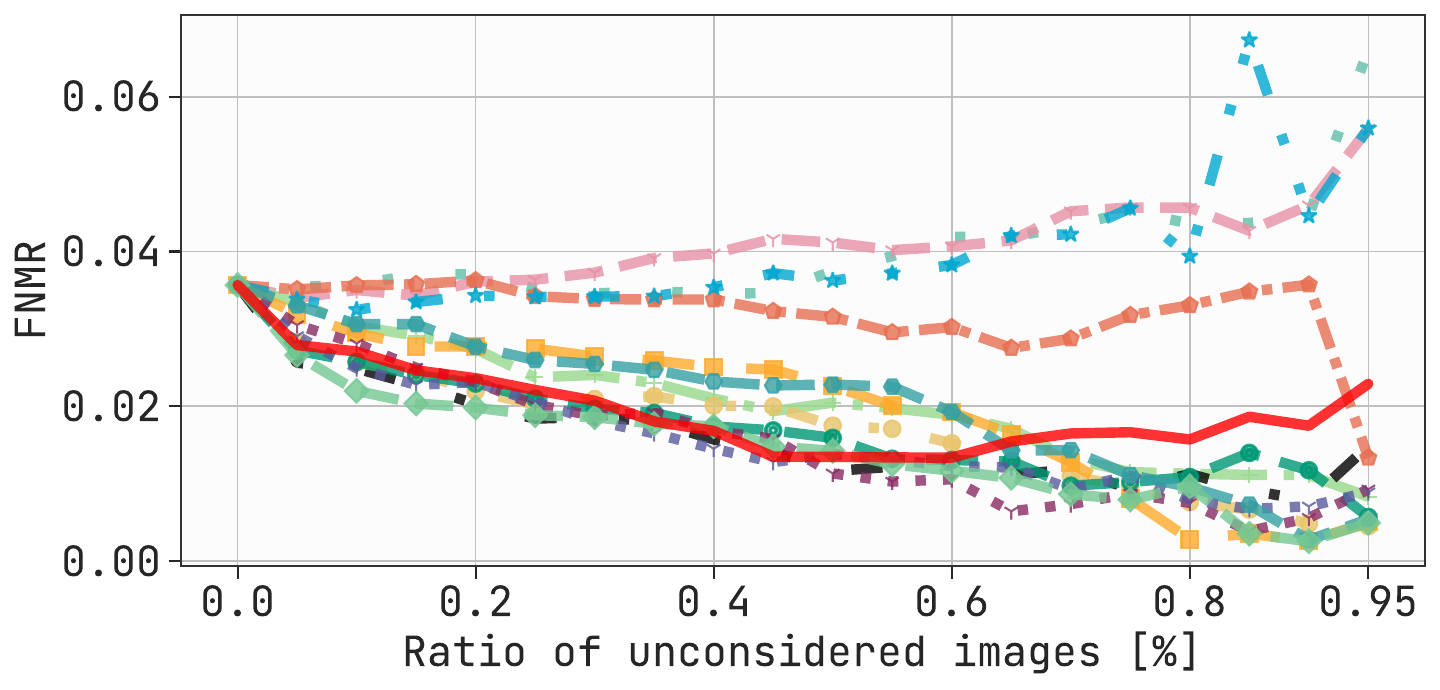}
		 \caption{ElasticFace Model, AgeDB30 Dataset}
	\end{subfigure}
\hfill
	\begin{subfigure}[b]{0.32\textwidth}
		 \centering
		 \includegraphics[width=0.8\textwidth]{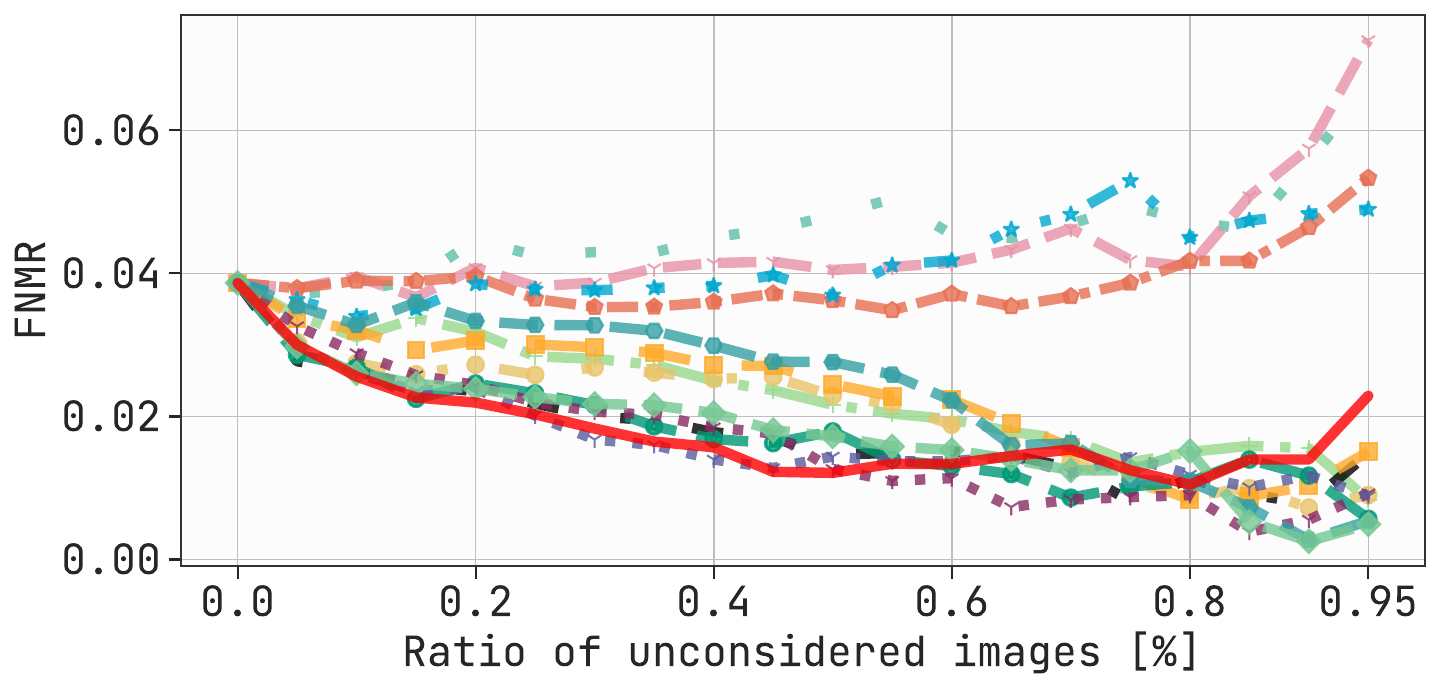}
		 \caption{CurricularFace Model, AgeDB30 Dataset}
	\end{subfigure}
\\
	\begin{subfigure}[b]{0.32\textwidth}
		 \centering
		 \includegraphics[width=0.8\textwidth]{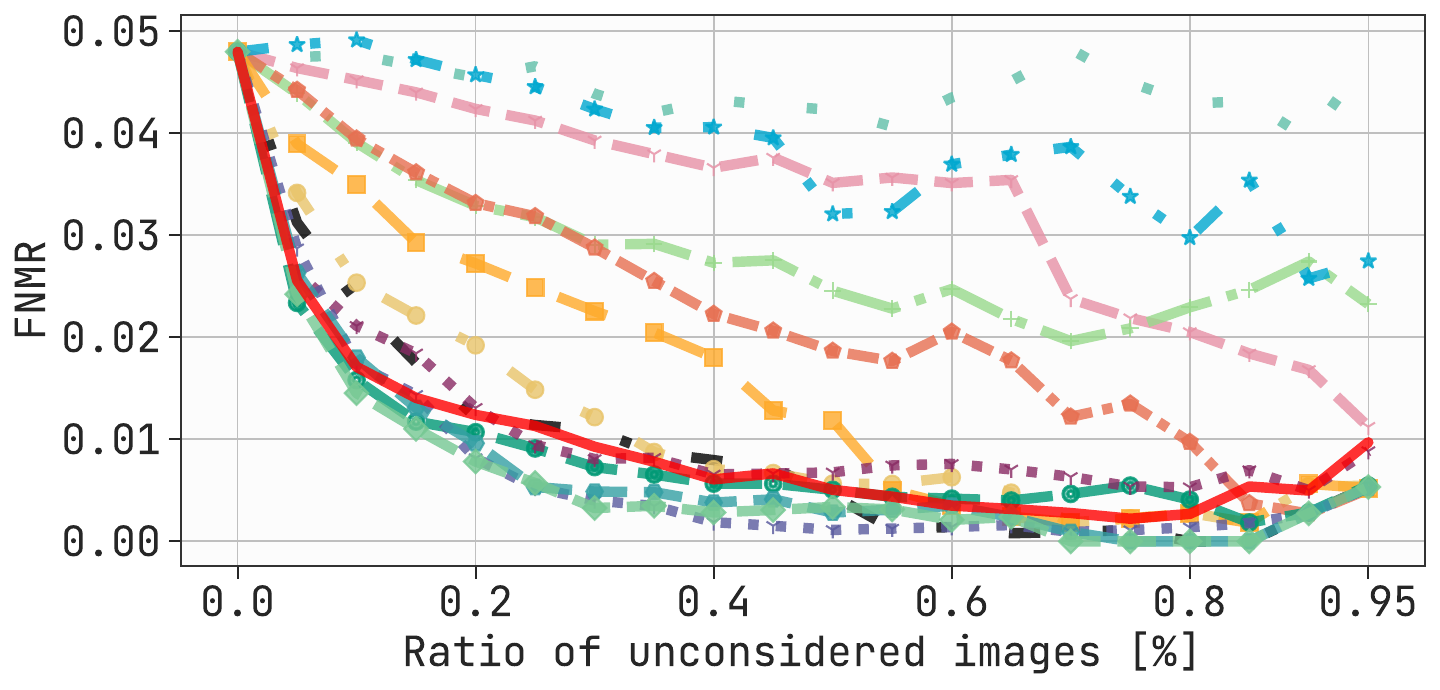}
		 \caption{MagFace Model, CFP-FP Dataset}
	\end{subfigure}
\hfill
	\begin{subfigure}[b]{0.32\textwidth}
		 \centering
		 \includegraphics[width=0.8\textwidth]{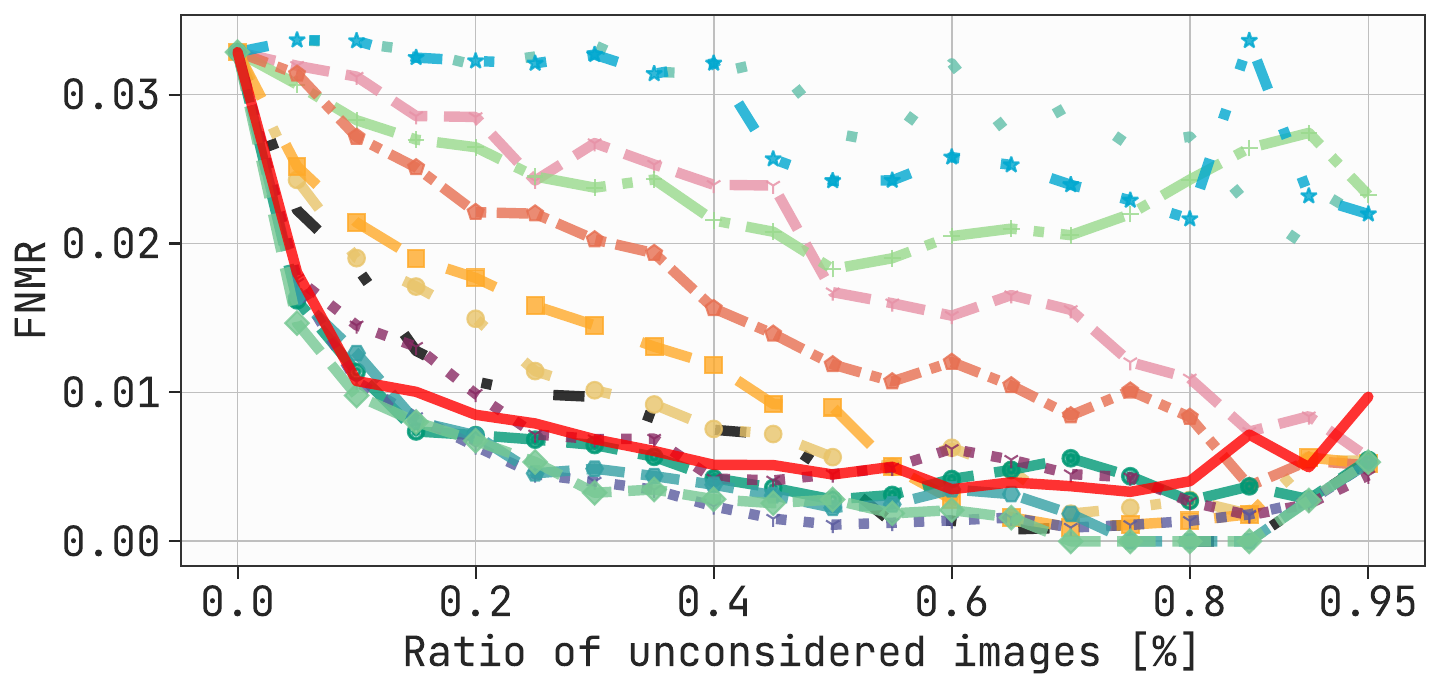}
		 \caption{ElasticFace Model, CFP-FP Dataset}
	\end{subfigure}
\hfill
	\begin{subfigure}[b]{0.32\textwidth}
		 \centering
		 \includegraphics[width=0.8\textwidth]{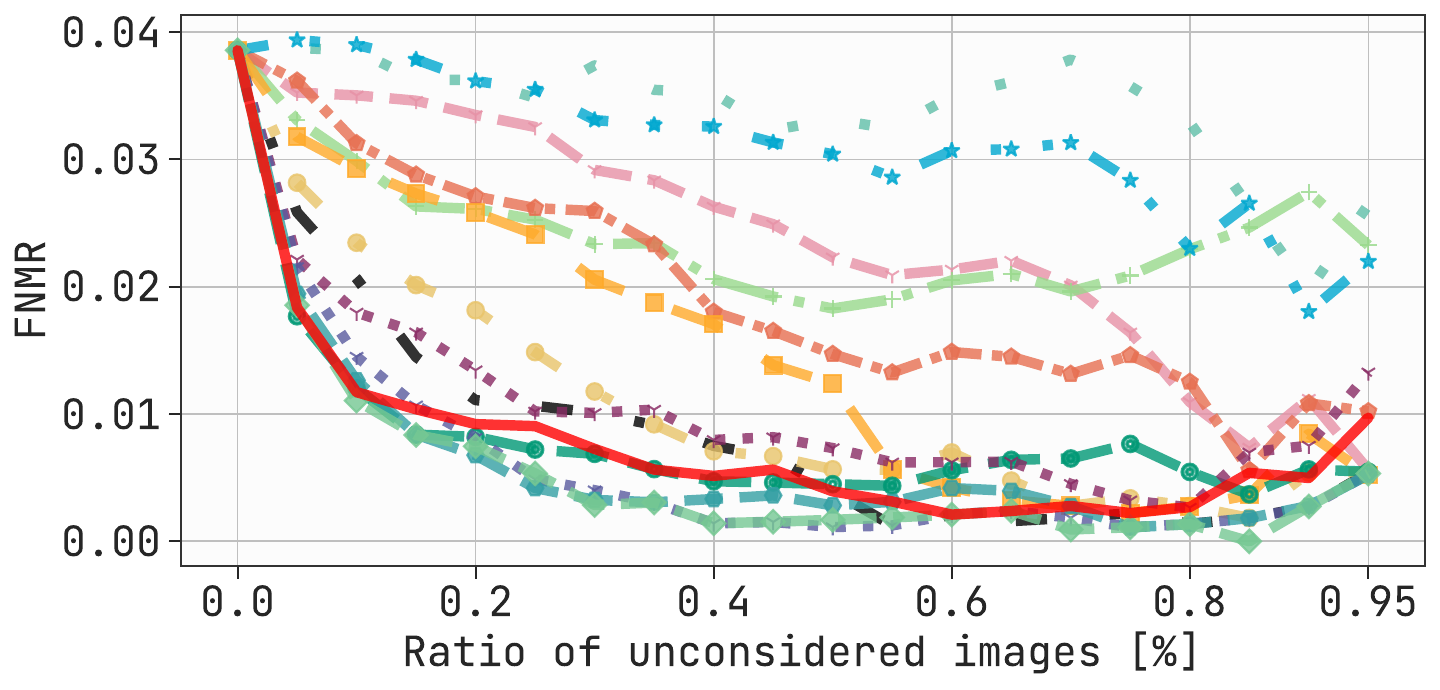}
		 \caption{CurricularFace Model, CFP-FP Dataset}
	\end{subfigure}
\\
	\begin{subfigure}[b]{0.32\textwidth}
		 \centering
		 \includegraphics[width=0.8\textwidth]{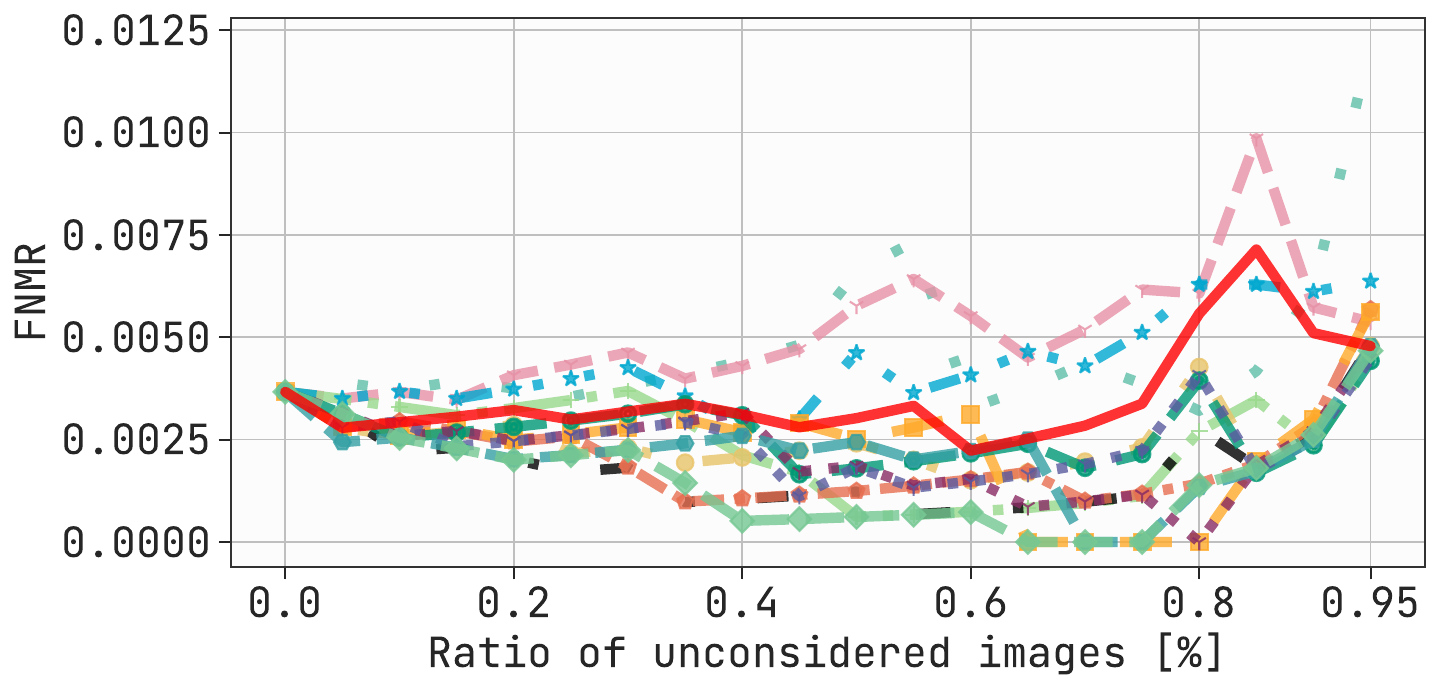}
		 \caption{MagFace Model, LFW Dataset}
	\end{subfigure}
\hfill
	\begin{subfigure}[b]{0.32\textwidth}
		 \centering
		 \includegraphics[width=0.8\textwidth]{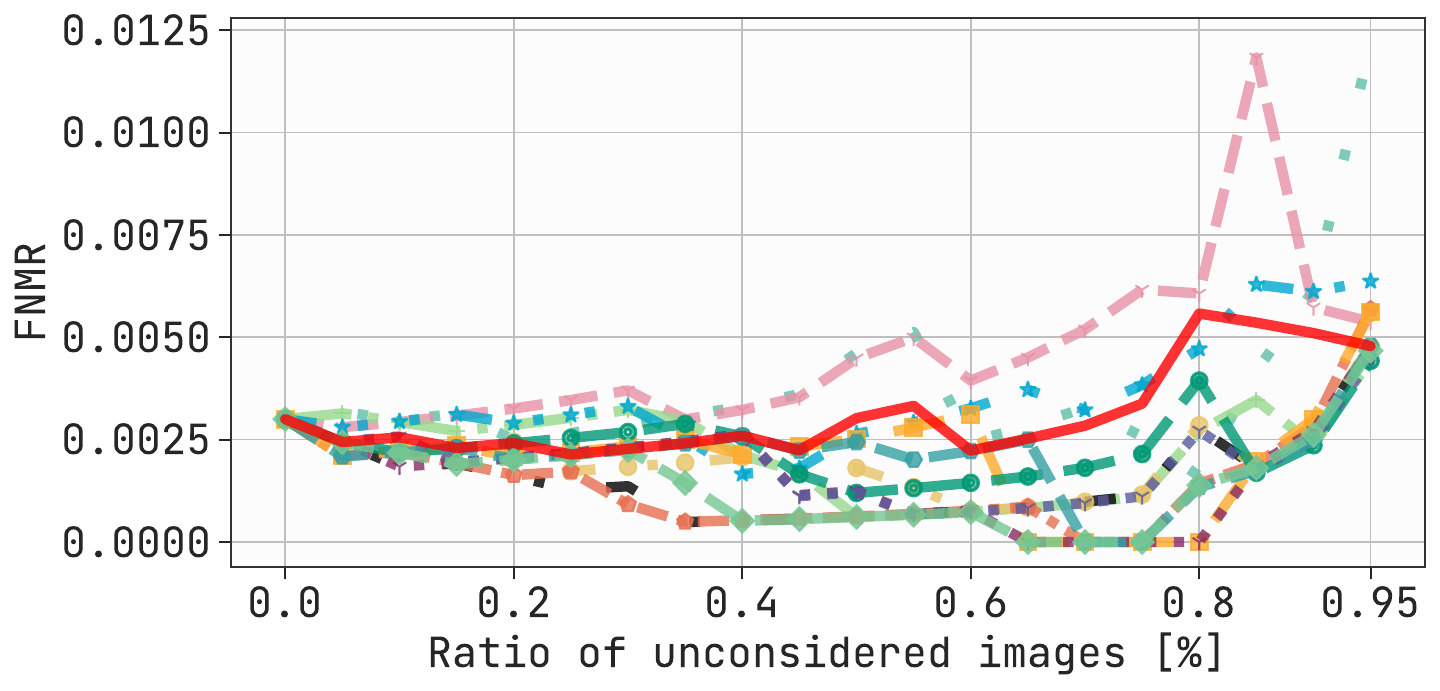}
		 \caption{ElasticFace Model, LFW Dataset}
	\end{subfigure}
\hfill
	\begin{subfigure}[b]{0.32\textwidth}
		 \centering
		 \includegraphics[width=0.8\textwidth]{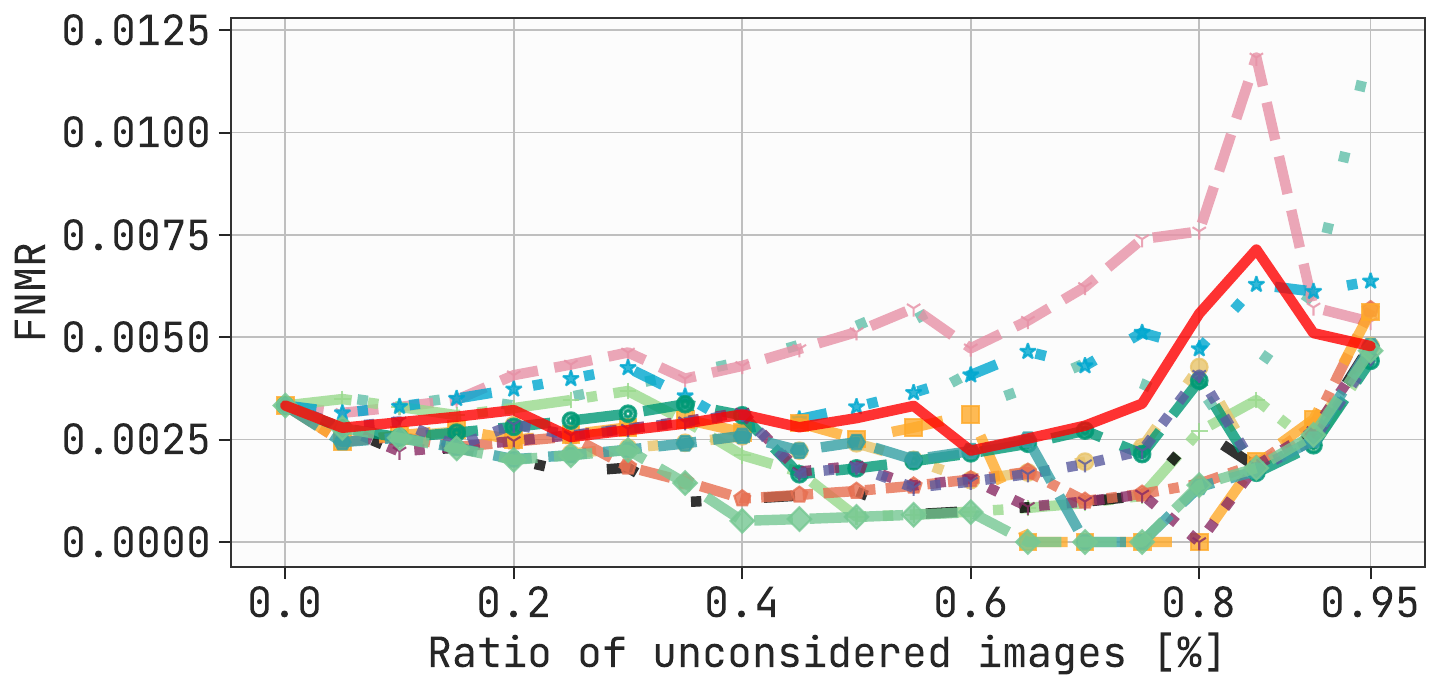}
		 \caption{CurricularFace Model, LFW Dataset}
	\end{subfigure}
\\
	\begin{subfigure}[b]{0.32\textwidth}
		 \centering
		 \includegraphics[width=0.8\textwidth]{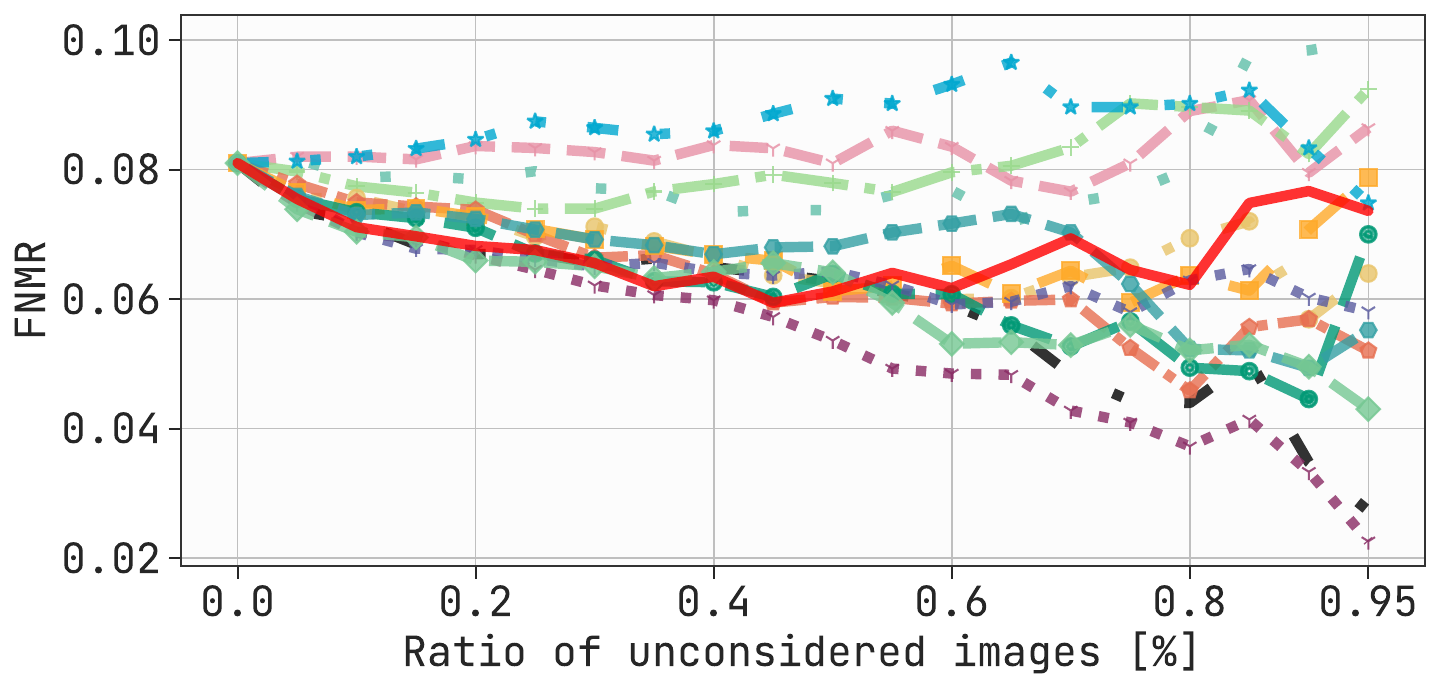}
		 \caption{MagFace Model, CALFW Dataset}
	\end{subfigure}
\hfill
	\begin{subfigure}[b]{0.32\textwidth}
		 \centering
		 \includegraphics[width=0.8\textwidth]{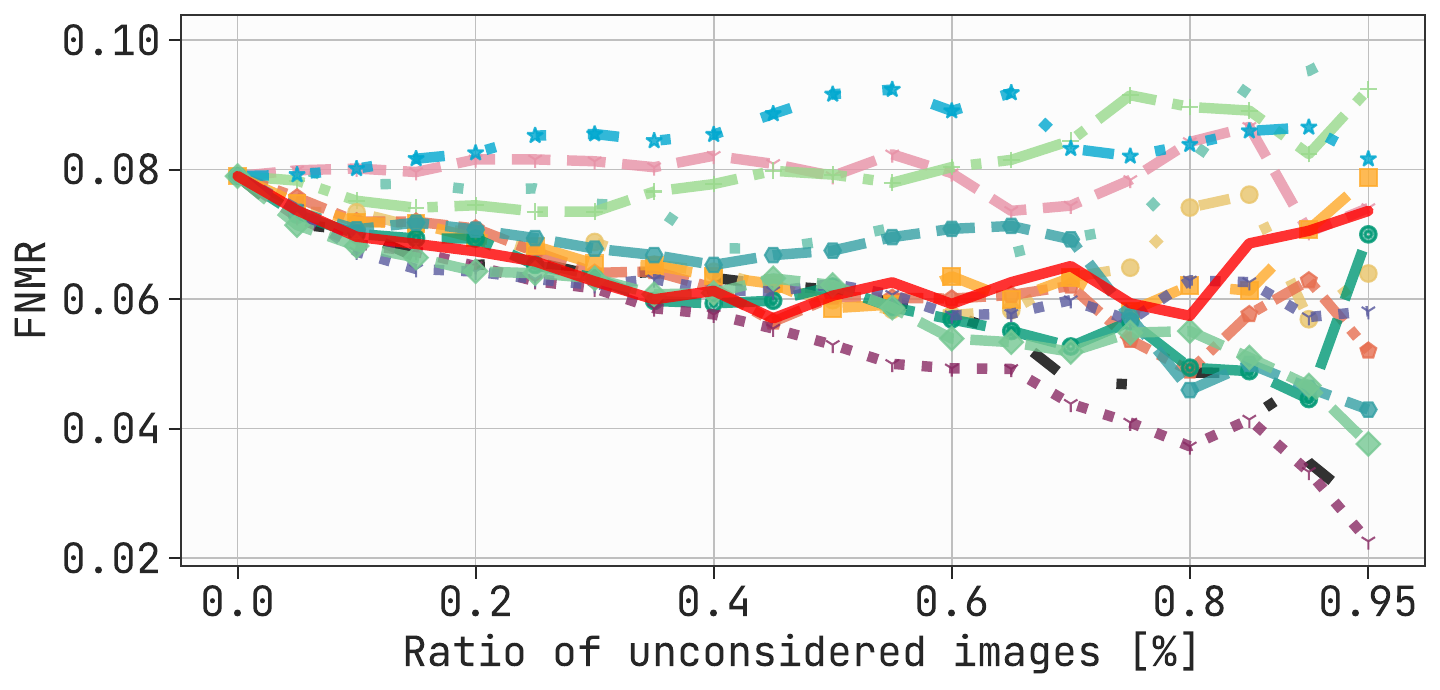}
		 \caption{ElasticFace Model, CALFW Dataset}
	\end{subfigure}
\hfill
	\begin{subfigure}[b]{0.32\textwidth}
		 \centering
		 \includegraphics[width=0.8\textwidth]{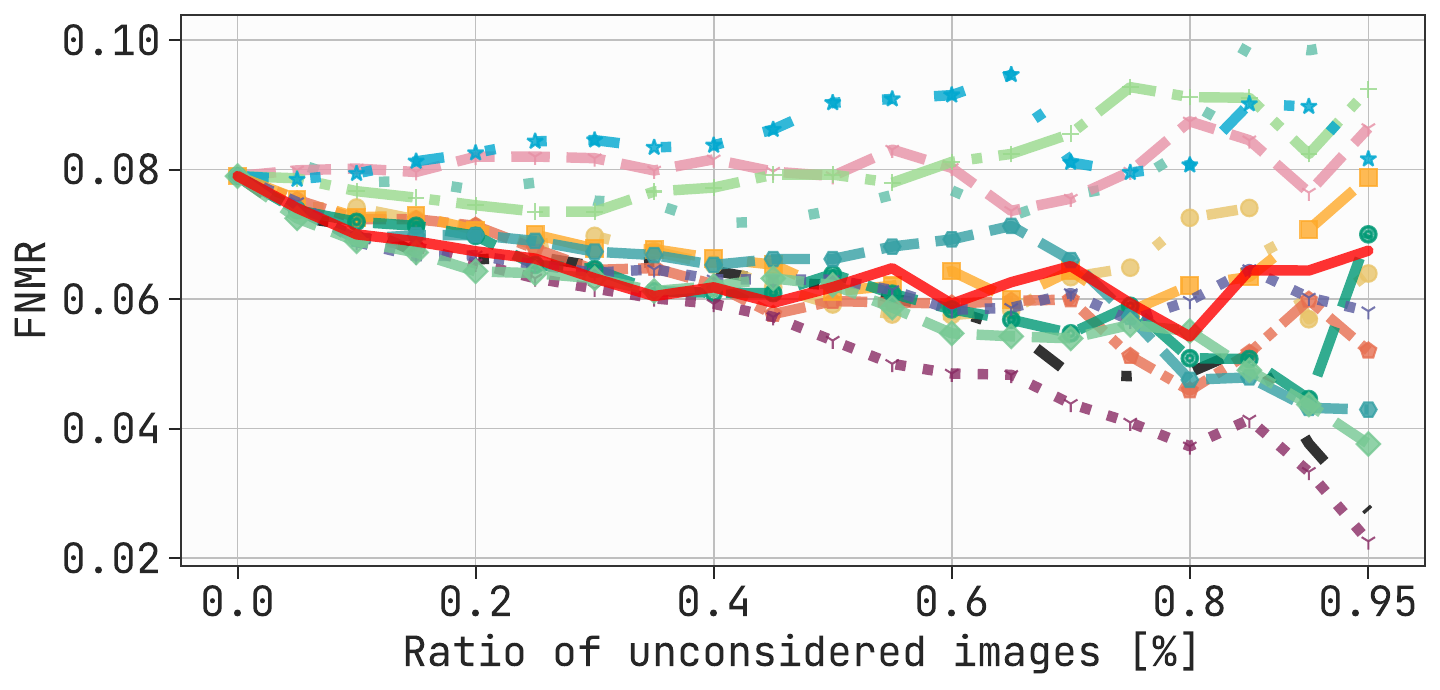}
		 \caption{CurricularFace Model, CALFW Dataset}
	\end{subfigure}
\\
	\begin{subfigure}[b]{0.32\textwidth}
		 \centering
		 \includegraphics[width=0.8\textwidth]{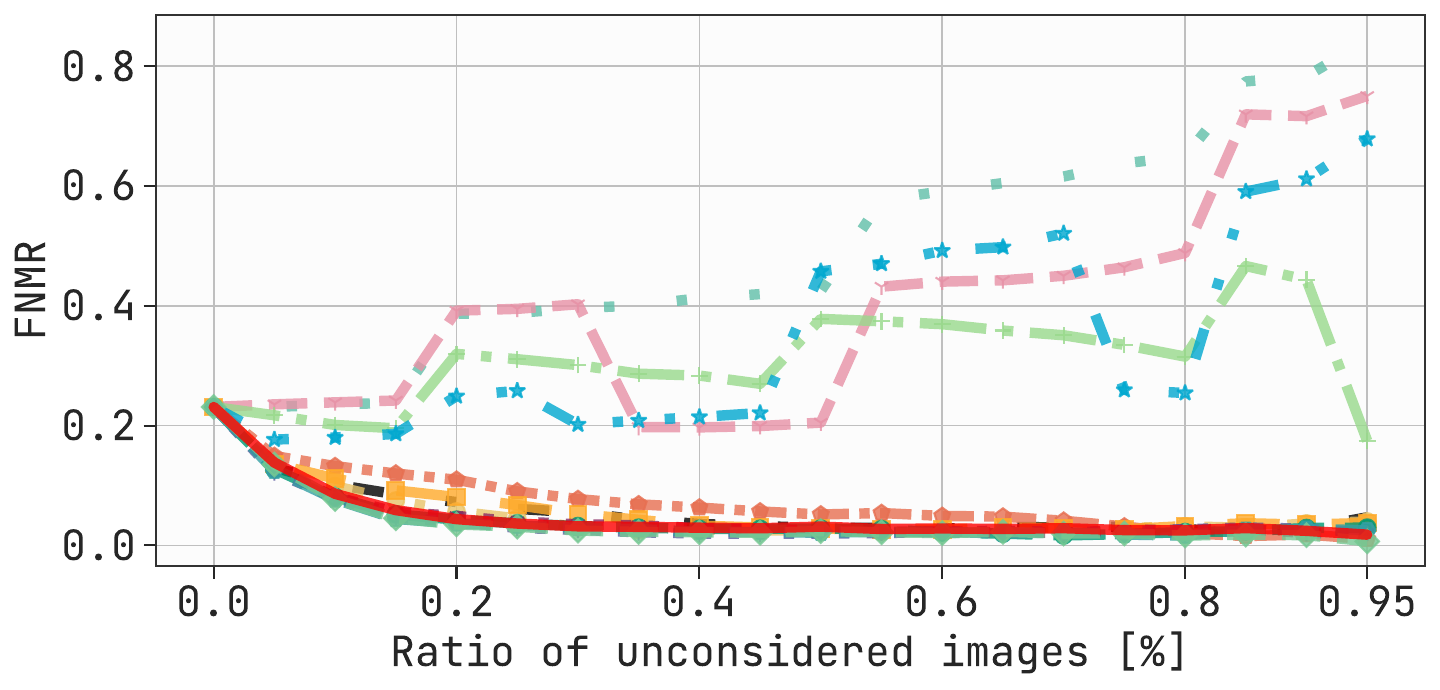}
		 \caption{MagFace Model, CPLFW Dataset}
	\end{subfigure}
\hfill
	\begin{subfigure}[b]{0.32\textwidth}
		 \centering
		 \includegraphics[width=0.8\textwidth]{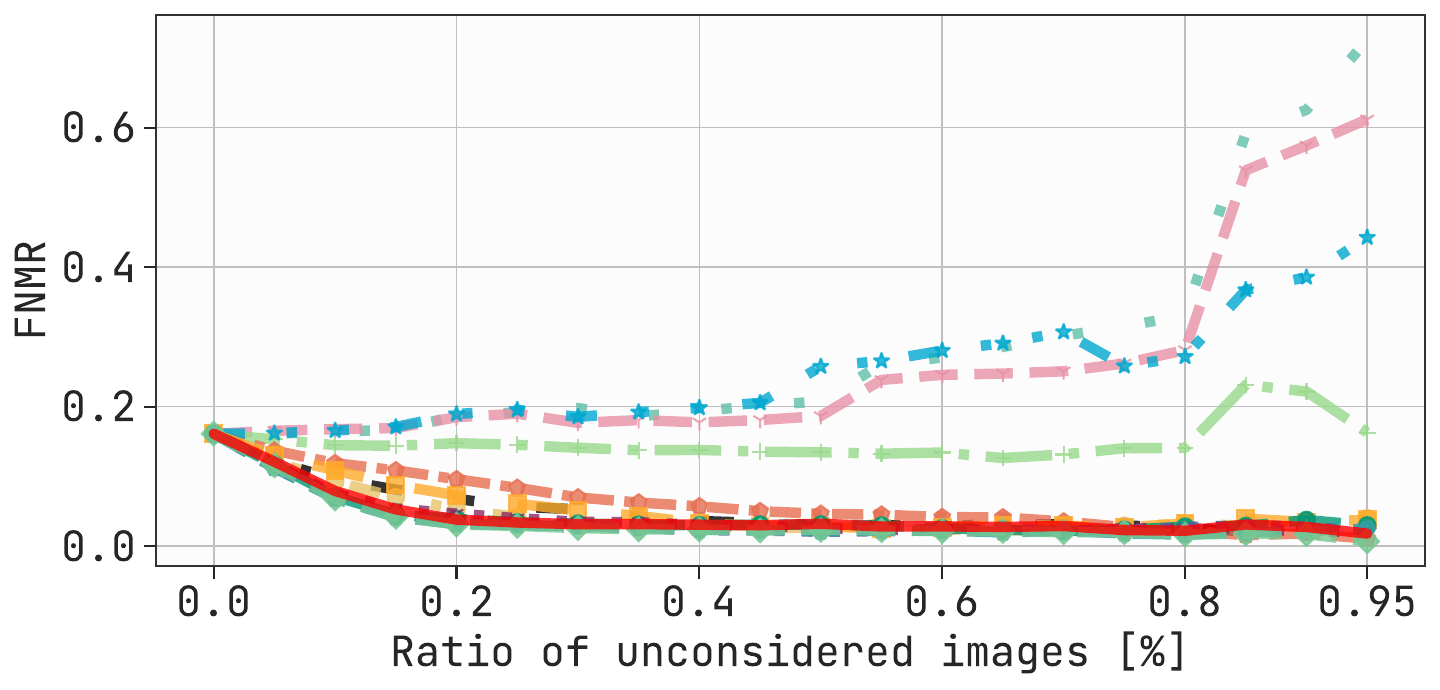}
		 \caption{ElasticFace Model, CPLFW Dataset}
	\end{subfigure}
\hfill
	\begin{subfigure}[b]{0.32\textwidth}
		 \centering
		 \includegraphics[width=0.8\textwidth]{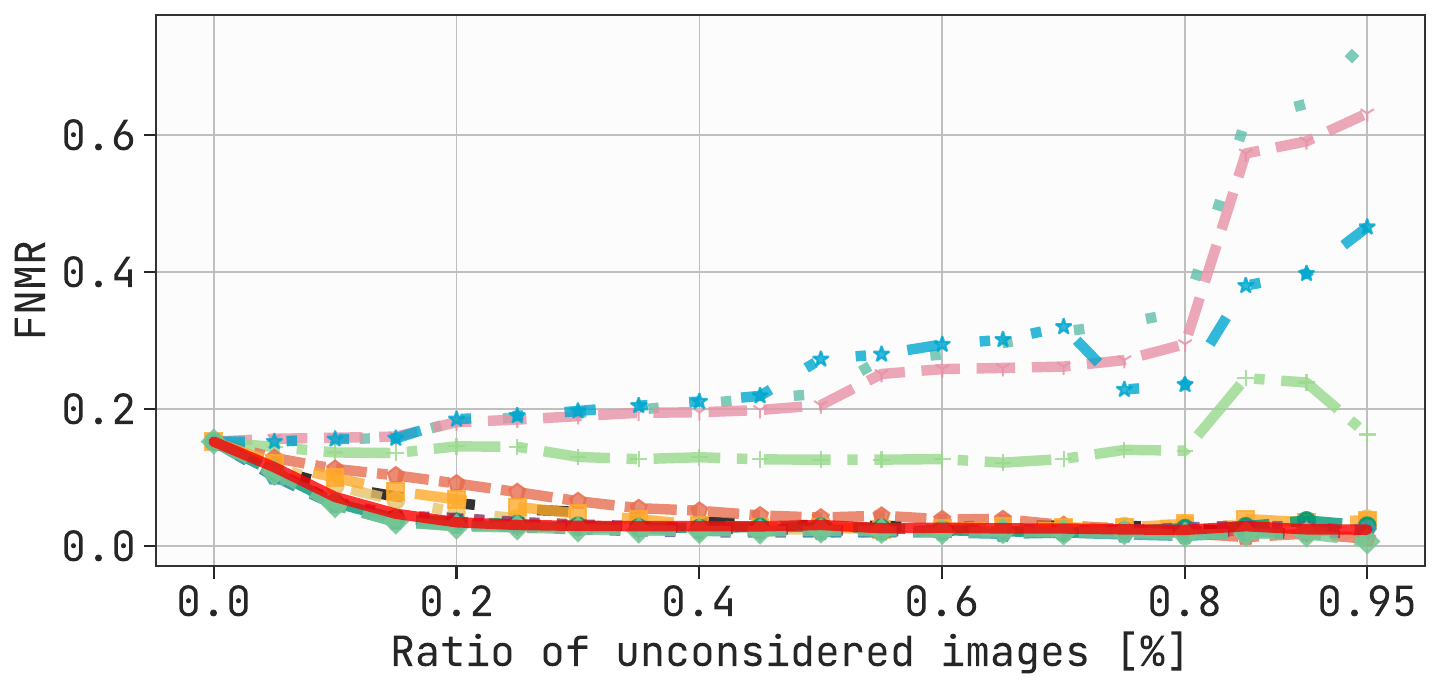}
		 \caption{CurricularFace Model, CPLFW Dataset}
	\end{subfigure}
\\
	\begin{subfigure}[b]{0.32\textwidth}
		 \centering
		 \includegraphics[width=0.8\textwidth]{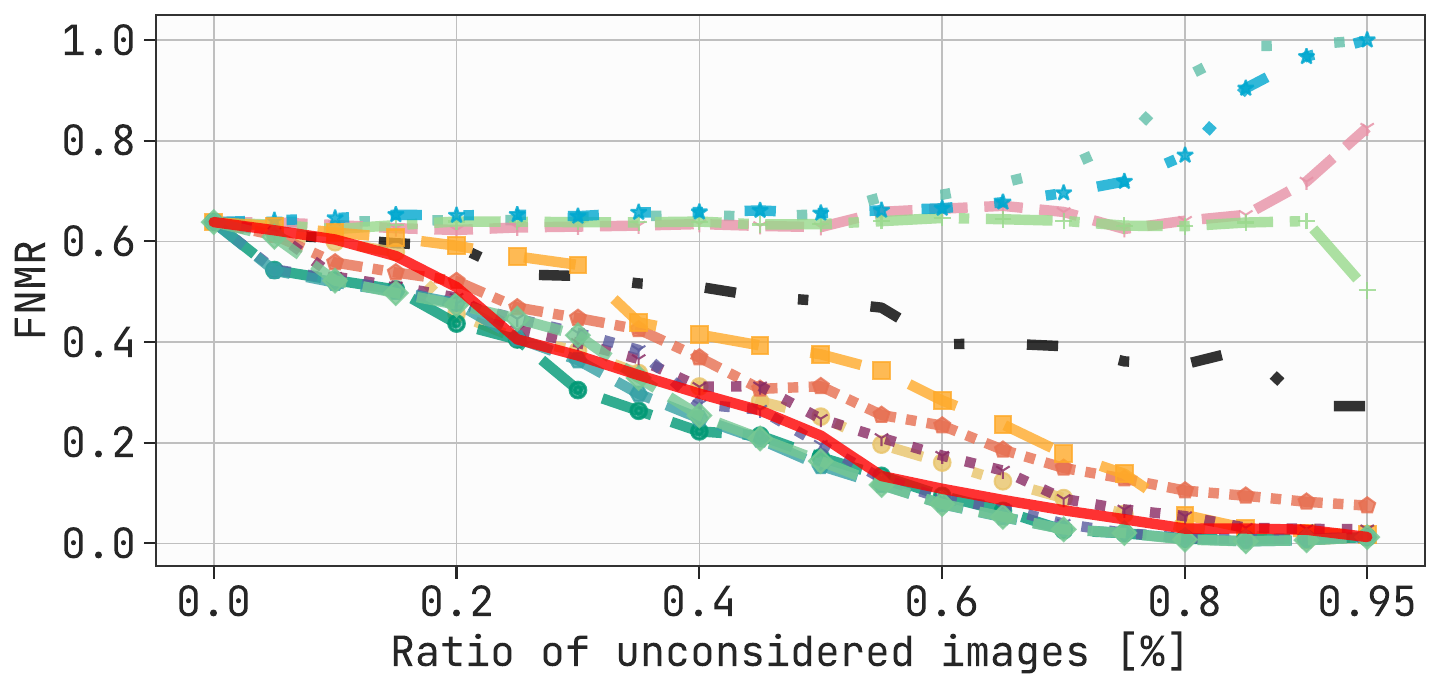}
		 \caption{MagFace Model, XQLFW Dataset}
	\end{subfigure}
\hfill
	\begin{subfigure}[b]{0.32\textwidth}
		 \centering
		 \includegraphics[width=0.8\textwidth]{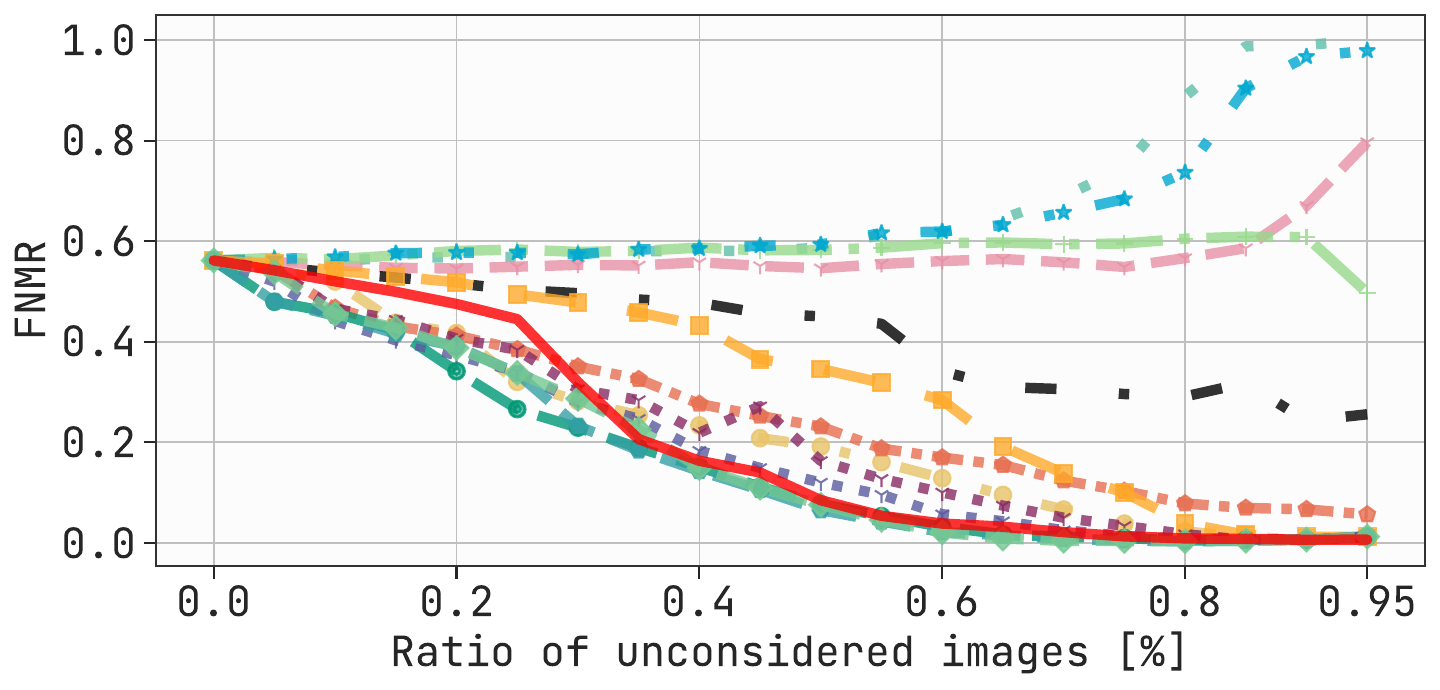}
		 \caption{ElasticFace Model, XQLFW Dataset}
	\end{subfigure}
\hfill
	\begin{subfigure}[b]{0.32\textwidth}
		 \centering
		 \includegraphics[width=0.8\textwidth]{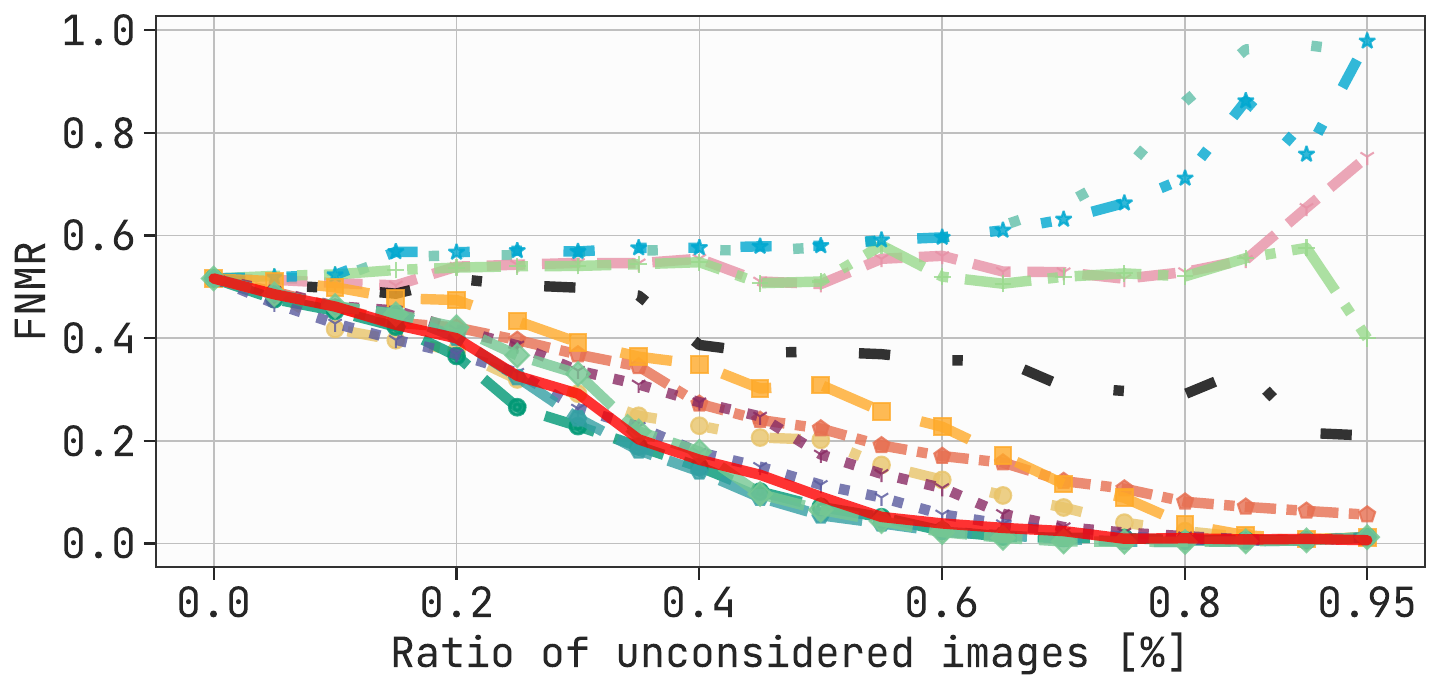}
		 \caption{CurricularFace Model, XQLFW Dataset}
	\end{subfigure}
 \vspace{-3mm}
\caption{EDC curves for FNMR@FMR=$1e-3$ for all evaluated benchmarks using MagFace, ElasticFace, and CurricularFace FR models. AUC are shown in Table \ref{tab:erc_sota_comparison}. EDC curves for ArcFace are provided in the supplementary material. The proposed \grafiqs method, shown in \textcolor{red}{solid red}, utilizes gradient magnitudes and it is reported using the best setting from Table \ref{tbl:ablation}.
}
\vspace{-4mm}
\label{fig:sota_comparison_cross_model}
\end{figure*}

We compared our achieved results with three IQA approaches, BRISQUE, RankIQA, and DeepIQA, as well as with seven SOTA FIQA approaches, RankIQ, PFE, SER-FIQ, FaceQnet, MagFace, SDD-FIQA, and CR-FIQA. 
Table \ref{tab:erc_sota_comparison} presents the achieved results as AUC of EDR calculated at two FMR thresholds, FRM$1e-3$ and FMR$1e-4$, using four different FR models. The corresponds EDC curves are presented in Figure \ref{fig:sota_comparison_cross_model}.
The results are reported on seven benchmarks described in Section \ref{sec:experimental_setup}.
The results of our \grafiqs approach are reported based on the best achieved settings from Table \ref{tbl:ablation} (i.e. B$2$) which was previously discussed in Section \ref{sec:ablation}.

We made the following observations from the results reported in Table \ref{tab:erc_sota_comparison}:
\begin{itemize}

    \item In comparison to IQA approaches, our \grafiqs outperformed all IQA approaches, BRISQUE \cite{BRISQE_IQA}, RankIQ \cite{liu2017rankiqa} and DeepIQA \cite{DEEPIQ_IQA}, in all settings, as shown in Table \ref{tab:erc_sota_comparison} and Figure \ref{fig:sota_comparison_cross_model}. 
    \item  On the benchmarks that include large age gaps (Adience, AgeDB-30 and CALFW) and in comparison to FIQA approaches, our \grafiqs achieved competitive results to the SOTA approaches. For example, our \grafiqs achieved competitive results, even outperformed them in many settings, to SDD-FIQA \cite{SDDFIQA}, FaceQNet \cite{faceqnetv1}, PFE \cite{PFE_FIQA}, DifFIQA \cite{10449044} and SER-FIQ \cite{SERFIQ}. Also, our \grafiqs scored slightly behind SOTA approaches, eDifFIQA \cite{babnikTBIOM2024}, CR-FIQA \cite{boutros_2023_crfiqa} and MagFace \cite{boutros_2023_crfiqa} on benchmarks with large age gaps, Table \ref{tab:erc_sota_comparison} and Figure \ref{fig:sota_comparison_cross_model}. 

    \item On the benchmarks that include large pose variations (CFP-FP and CPLFW) and compared to FIQA approaches, our \grafiqs scored slightly behind CR-FIQA \cite{boutros_2023_crfiqa}, DifFIQA \cite{10449044}, and eDifFIQA \cite{babnikTBIOM2024}, and outperformed all other approaches in most settings. 

    \item  On LFW and in comparison to FIQA approaches, our \grafiqs achieved slightly lower results than the SOTA FIQA approaches. 

    \item The XQLFW benchmark, derived from LFW, includes pairs of images with maximal disparities in quality, which are selected based on quality scores from BRISQUE \cite{BRISQE_IQA} and SER-FIQ \cite{SERFIQ}, ensuring that the images are categorized as either exceptionally high or low quality. On XQLFW and in comparison to FIQA approaches, our \grafiqs outperformed well-performing SOTA approaches, including MagFace \cite{MagFace}, SDD-FIQA \cite{SDDFIQA} and CR-FIQA \cite{boutros_2023_crfiqa} and achieved very close results to DifFIQA \cite{10449044}, eDifFIQA \cite{babnikTBIOM2024}, and the labeling approach SER-FIQ \cite{SERFIQ}.

    \item  Under same-model (ArcFace) and cross-model (ElasticFace, MagFace, and CurricularFace) experimental settings, our \grafiqs achieved consistent results, proving the generalizability of our concept for FIQA, as shown in Table \ref{tab:erc_sota_comparison}.

\end{itemize}

To conclude, the presented novel concept proved to be highly effective. Our gradient magnitude-based approach \grafiqs achieved very competitive results to the SOTA approaches without the need for quality labeling and regression network training \cite{faceqnetv1,SDDFIQA, babnikTBIOM2024, 10449044} or designing and optimizing special losses \cite{MagFace,boutros_2023_crfiqa}, as summarized in Table \ref{tab:desinged_comparison}. Unlike SER-FIQ which requires passing the sample 100 times into the pretrained FR model designed and trained with dropout layers, our \grafiqs requires a single forward and backward pass. This work is the pioneer FIQA approach that assesses the quality of face images by inspecting the required changes in pretrained FR model weights to match pretrained FR model data distributions. 

%% file: figures/fig_iresnet100_short-overview-paper.tex
\begin{figure*}[h!]
    \vspace{-3mm}
    \caption{EDC for FNMR@FMR=$1e-3$ and FNMR@FMR=$1e-4$ of our proposed method using $\mathcal{L}_{\text{BNS}}$ as backpropagation loss and absolute sum as FIQ. The gradients at image level ($\phi=\mathcal{I}$), and block levels ($\phi=\text{B}1$ $-$ $\phi=\text{B}4$) are used to calculate FIQ. $\text{MSE}_{\text{BNS}}$ as FIQ is shown in black. Results are shown on benchmarks Adience, AgeDB30 and XQLFW datasets using ArcFace model. The proposed \grafiqs method leads to lower verification error when images with the lowest utility score estimated from gradient magnitudes are rejected. Furthermore, estimating FIQ by backpropagating $\mathcal{L}_{\text{BNS}}$ yields significantly better results than using $\text{MSE}_{\text{BNS}}$ directly.}
      \vspace{-3mm}
      \label{fig:iresnet100_short_overview}
     \centering
        \begin{subfigure}[b]{0.6\textwidth}
            \centering
            \includegraphics[width=\textwidth]{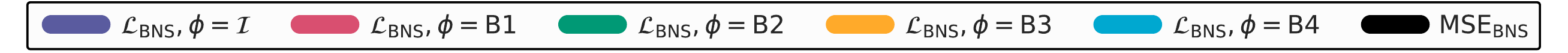}
        \end{subfigure}
        \\
        \begin{subfigure}[b]{0.28\textwidth}
            \centering
            \includegraphics[width=0.90\textwidth]{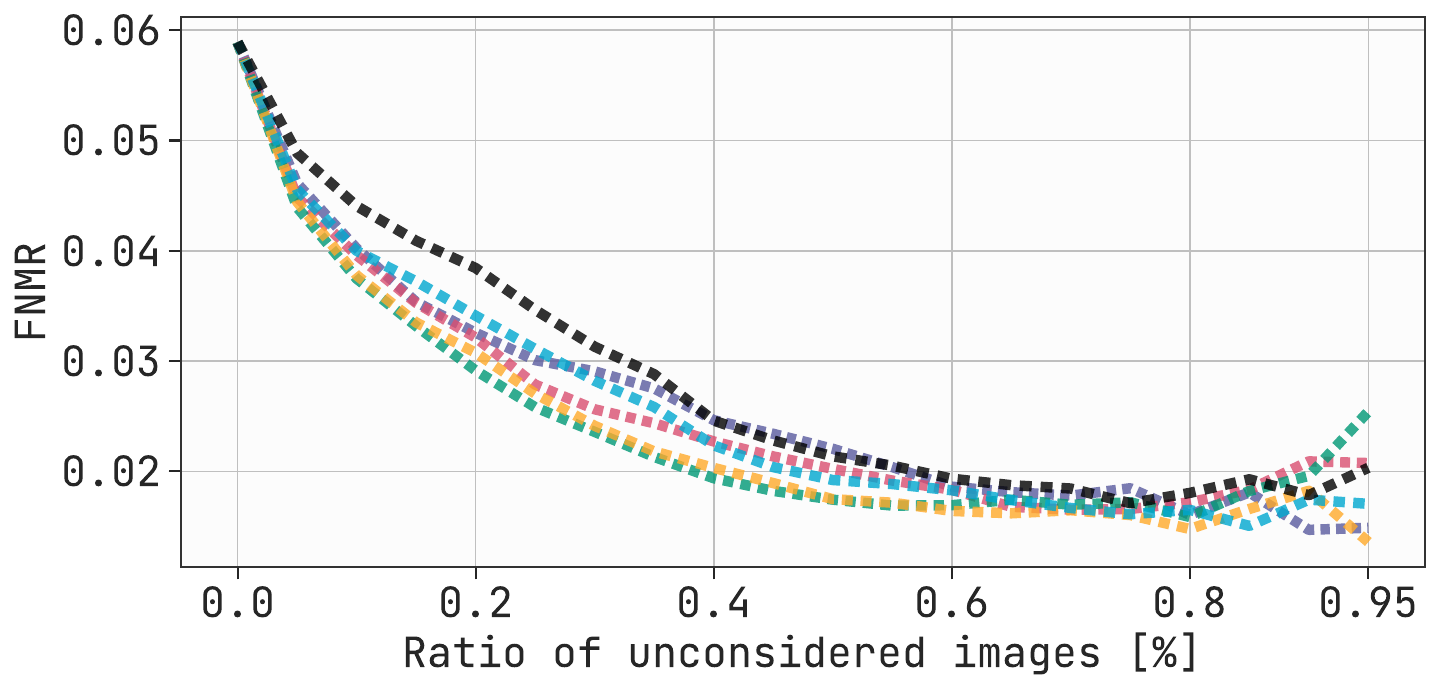}
            \vspace{-2mm}
            \caption{Adience, FMR$=1e-3$}
        \end{subfigure}
        \hfill
        \begin{subfigure}[b]{0.28\textwidth}
            \centering
            \includegraphics[width=0.90\textwidth]{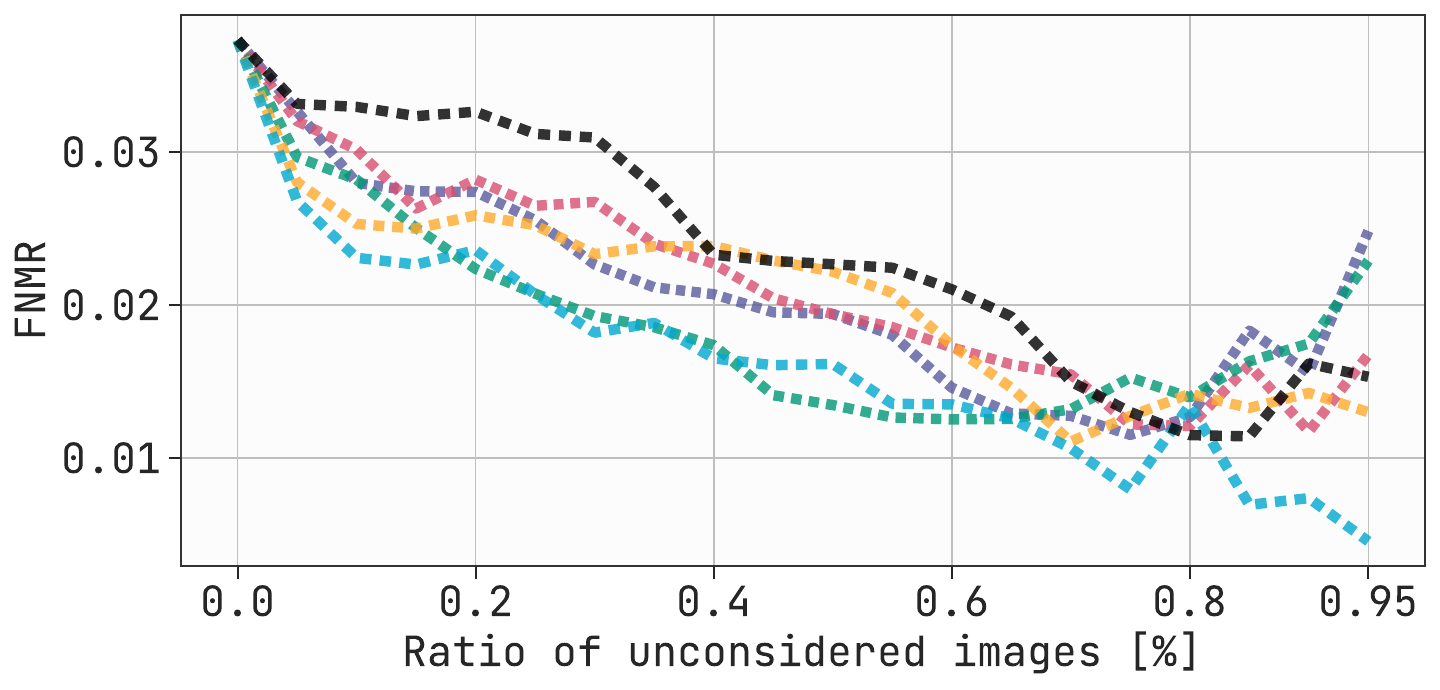}
            \vspace{-2mm}
            \caption{AgeDB30, FMR$=1e-3$}
        \end{subfigure}
        \hfill
        \begin{subfigure}[b]{0.28\textwidth}
            \centering
            \includegraphics[width=0.90\textwidth]{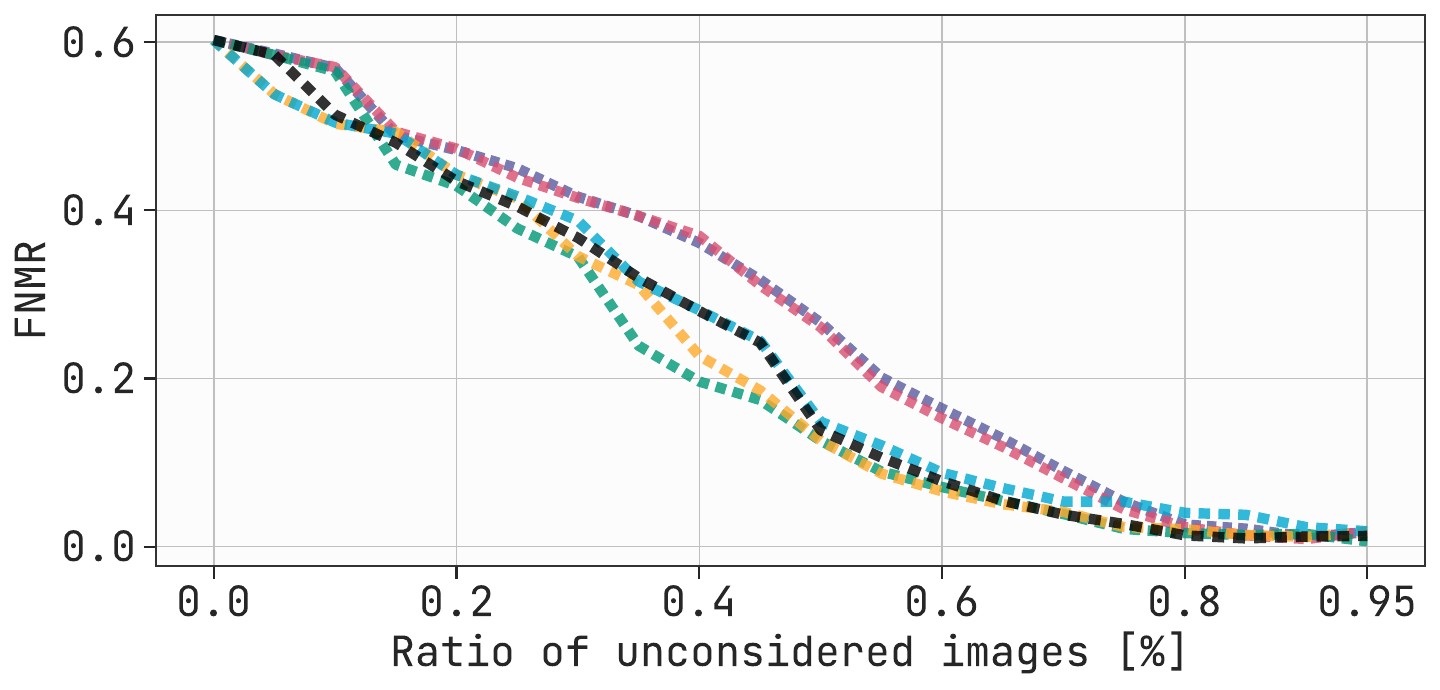}
            \vspace{-2mm}
            \caption{XQLFW, FMR$=1e-3$}
        \end{subfigure}
        \hfill
        \\
        \begin{subfigure}[b]{0.28\textwidth}
            \centering
            \includegraphics[width=0.90\textwidth]{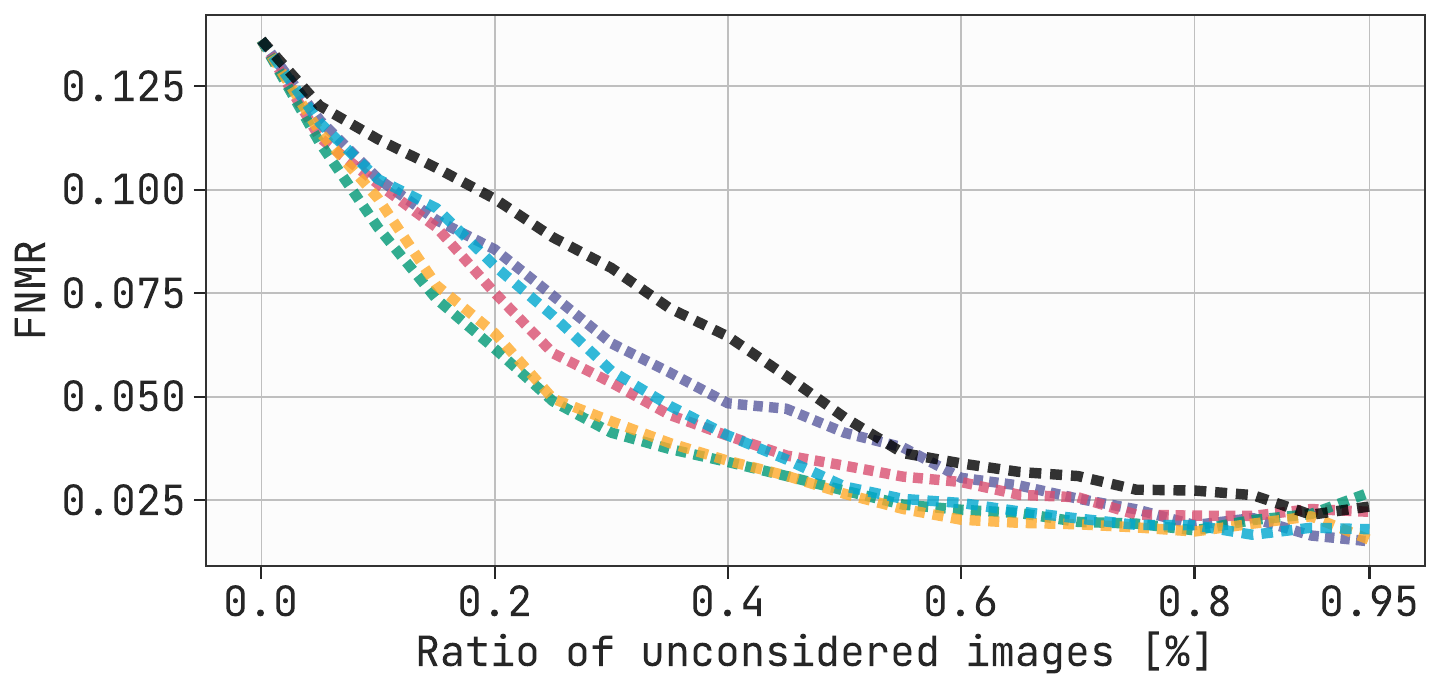}
            \vspace{-2mm}
            \caption{Adience, FMR$=1e-4$}
        \end{subfigure}
        \hfill
        \begin{subfigure}[b]{0.28\textwidth}
            \centering
            \includegraphics[width=0.90\textwidth]{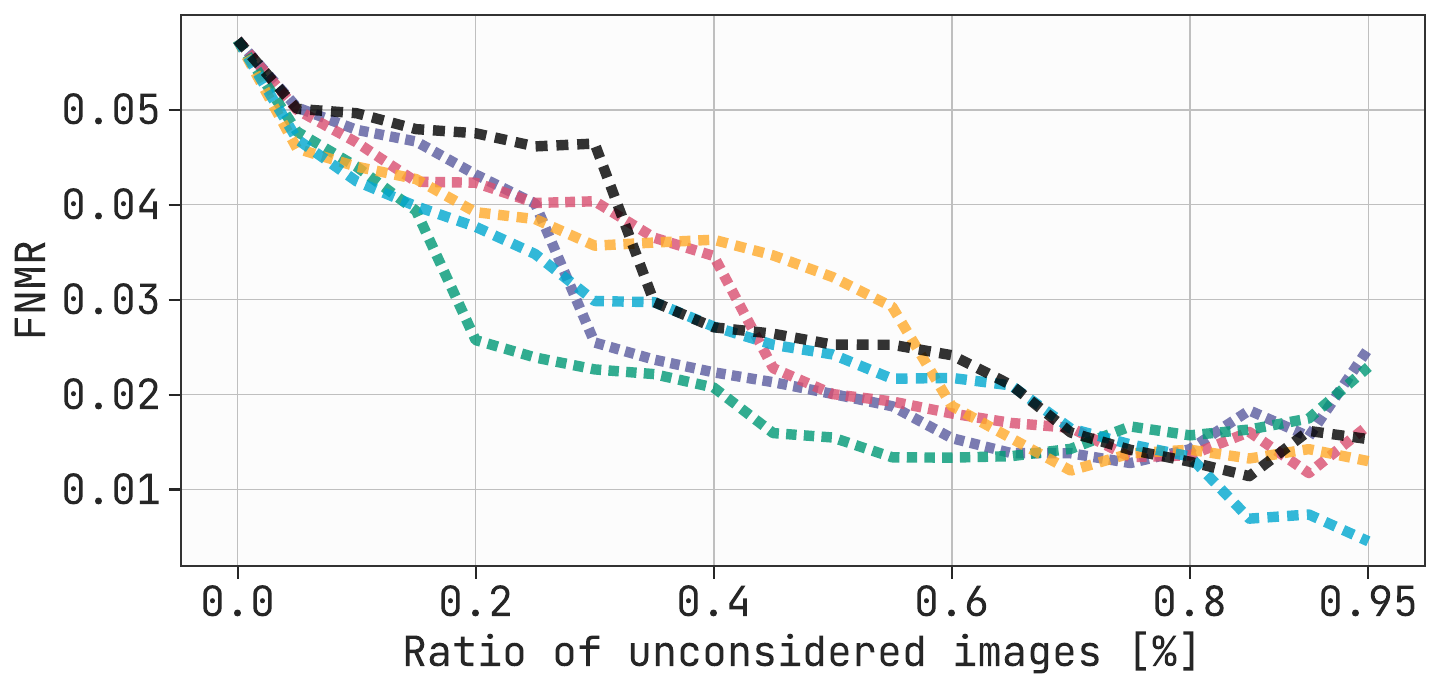}
            \vspace{-2mm}
            \caption{AgeDB30, FMR$=1e-4$}  
        \end{subfigure}
        \hfill
        \begin{subfigure}[b]{0.28\textwidth}
            \centering
            \includegraphics[width=0.90\textwidth]{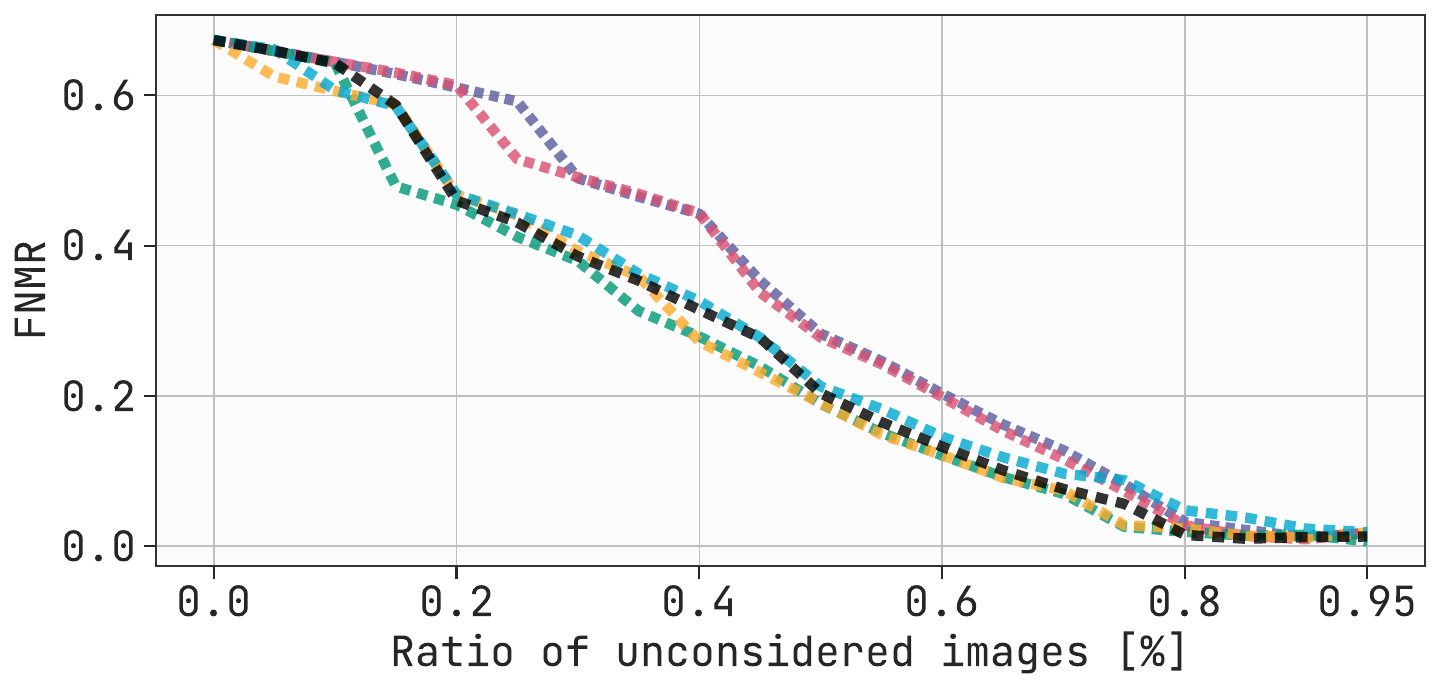}
            \vspace{-2mm}
            \caption{XQLFW, FMR$=1e-4$}
        \end{subfigure}
      \vspace{-5mm}
      
\end{figure*}

%% file: 06_conclusion.tex
\vspace{-2mm}
\section{Conclusion}
\vspace{-2mm}
In this paper, we are the first to introduce a method that assesses the quality of face images by measuring the required changes in the pre-trained FR model weights to minimize dissimilarities between testing samples and the distribution of the FR training dataset.
To accomplish this, the differences in BNS between those recorded during the training phase of FR and those acquired through the processing of testing samples using the pre-trained FR model were calculated. Then, the cumulative absolute sum of gradient magnitudes of the pre-trained FR weights was calculated by backpropagating the BNS through the pre-trained FR model.  Unlike previous FIQA approaches, our approach did not rely on quality labeling, the necessity to train regression networks, specialized architectures, or the design and optimization of specific training loss functions.
These findings suggest potential directions for future research aimed at assessing FIQ from different perspectives.

\vspace{-2mm}
\section*{Acknowledgments}
\vspace{-2mm}
This research work has been funded by the German Federal Ministry of Education and Research and the Hessen State Ministry for Higher Education, Research and the Arts within their joint support of the National Research Center for Applied Cybersecurity ATHENE.

%% file: 09_appendix.tex
\section{Supplementary Material}

This supplementary material contains the following supporting content:

\begin{itemize}
    \item An ablation on a different backbone (ResNet50 \cite{he_2016_resnet}): To validate the generalizability over different backbones, we present in Table \ref{tbl:resnet50_full_overview} the achieved AUC over the utilized benchmarks and FR models, similar to Table 2 in the main paper, that shows the AUC for the ResNet100 backbone. Similar to Figure 2 in the main paper, Figures \ref{fig:iresnet50_supplementary_adience}, \ref{fig:iresnet50_supplementary_agedb_30}, \ref{fig:iresnet50_supplementary_cfp_fp}, \ref{fig:iresnet50_supplementary_lfw}, \ref{fig:iresnet50_supplementary_calfw}, \ref{fig:iresnet50_supplementary_cplfw}, \ref{fig:iresnet50_supplementary_XQLFW} show the EDC curves achieved by our $\text{MSE}_{\text{BNS}}$ and $\mathcal{L}_{\text{BNS}}$ approaches on datasets Adience, AgeDB30, CFP-FP, LFW, CALFW, CPLFW, and XQLFW for all utilized FR models, respectively.

    \item Due to limited space in the main paper, we present here all EDC figures of ResNet100 backbone over all utilized datasets and FR models. Figures \ref{fig:iresnet100_supplementary_adience}, \ref{fig:iresnet100_supplementary_agedb_30}, \ref{fig:iresnet100_supplementary_cfp_fp}, \ref{fig:iresnet100_supplementary_lfw}, \ref{fig:iresnet100_supplementary_calfw}, \ref{fig:iresnet100_supplementary_cplfw}, \ref{fig:iresnet100_supplementary_XQLFW} show the EDC curves achieved by our $\text{MSE}_{\text{BNS}}$ and $\mathcal{L}_{\text{BNS}}$ approaches on datasets Adience, AgeDB30, CFP-FP, LFW, CALFW, CPLFW, and XQLFW for all utilized FR models, respectively.

    \item Furthermore we provide all EDC figures that show a comparison between \grafiqs using ResNet100, reported using the best setting from Table 2 in the main paper, and SOTA methods. Figures \ref{fig:iresnet100_supplementary_sota_adience}, \ref{fig:iresnet100_supplementary_sota_agedb_30}, \ref{fig:iresnet100_supplementary_sota_cfp_fp}, \ref{fig:iresnet100_supplementary_sota_lfw}, \ref{fig:iresnet100_supplementary_sota_calfw}, \ref{fig:iresnet100_supplementary_sota_cplfw}, \ref{fig:iresnet100_supplementary_sota_XQLFW} show the EDC curves on datasets Adience, AgeDB30, CFP-FP, LFW, CALFW, CPLFW, and XQLFW for all utilized FR models, respectively.  
\end{itemize}

\include{tables/table_resnet50_full_overview}
\include{figures/fig_iresnet50_supplementary_adience}
\include{figures/fig_iresnet50_supplementary_agedb_30}
\include{figures/fig_iresnet50_supplementary_cfp_fp}
\include{figures/fig_iresnet50_supplementary_lfw}
\include{figures/fig_iresnet50_supplementary_calfw}
\include{figures/fig_iresnet50_supplementary_cplfw}
\include{figures/fig_iresnet50_supplementary_XQLFW}

\include{figures/fig_iresnet100_supplementary_adience}
\include{figures/fig_iresnet100_supplementary_agedb_30}
\include{figures/fig_iresnet100_supplementary_cfp_fp}
\include{figures/fig_iresnet100_supplementary_lfw}
\include{figures/fig_iresnet100_supplementary_calfw}
\include{figures/fig_iresnet100_supplementary_cplfw}
\include{figures/fig_iresnet100_supplementary_XQLFW}
\include{figures/fig_iresnet100_supplementary_sota_adience}
\include{figures/fig_iresnet100_supplementary_sota_agedb_30}
\include{figures/fig_iresnet100_supplementary_sota_cfp_fp}
\include{figures/fig_iresnet100_supplementary_sota_lfw}
\include{figures/fig_iresnet100_supplementary_sota_calfw}
\include{figures/fig_iresnet100_supplementary_sota_cplfw}
\include{figures/fig_iresnet100_supplementary_sota_XQLFW}

%% file: tables/table_resnet50_full_overview.tex
\begin{table*}[]
\begin{center}
\resizebox{0.93\textwidth}{!}{
\begin{tabular}{|c|c|c|c|c|c|c|c|c|c|c|c|c|c|c|c|c|c|c|c|}
\hline
\multirow{2}{*}{FR} & \multirow{2}{*}{Loss $\mathcal{L}$} & \multirow{2}{*}{FIQ} & \multirow{2}{*}{Gradient} & \multicolumn{2}{c|}{Adience \cite{Adience}} & \multicolumn{2}{c|}{AgeDB30 \cite{agedb}} & \multicolumn{2}{c|}{CFP-FP \cite{cfp-fp}} & \multicolumn{2}{c|}{LFW \cite{LFWTech}} & \multicolumn{2}{c|}{CALFW \cite{CALFW}} & \multicolumn{2}{c|}{CPLFW \cite{CPLFWTech}} & \multicolumn{2}{c|}{XQLFW \cite{XQLFW}} & \multicolumn{2}{c|}{Mean AUC}\\ 
 \cline{5-20}
 & & & & $1e{-3}$ & $1e{-4}$ & $1e{-3}$ & $1e{-4}$ & $1e{-3}$ & $1e{-4}$ & $1e{-3}$ & $1e{-4}$ & $1e{-3}$ & $1e{-4}$ & $1e{-3}$ & $1e{-4}$ & $1e{-3}$ & $1e{-4}$ & $1e{-3}$ & $1e{-4}$\\ 
 \cline{2-20}
\hline 
\multirow{6}{*}{\rotatebox[origin=c]{90}{ArcFace \cite{deng2019arcface}}} & - & $\text{MSE}(\text{BNS}_{\mathcal{M}}, \text{BNS}_{\mathcal{I}})$ & - & 0.0311 & 0.0668 & 0.0337 & 0.0470 & 0.0204 & 0.0266 & 0.0034 & 0.0041 & 0.0665 & 0.0700 & 0.0630 & 0.0903 & 0.2436 & 0.2946 & 0.0660 & 0.0856\\ \cline{2-20}
 & $\mathcal{L}_\text{BNS}$ & $\sum |\partial \mathcal{L} / \partial \phi|$ & $\phi =  \mathcal{I}$ & 0.0262 & 0.0527 & 0.0312 & 0.0412 & 0.0126 & 0.0199 & 0.0034 & 0.0041 & 0.0637 & 0.0679 & 0.0552 & 0.0867 & 0.2382 & 0.2896 & 0.0615 & 0.0803\\ 
 & $\mathcal{L}_\text{BNS}$ & $\sum |\partial \mathcal{L} / \partial \phi|$ & $\phi = \text{B}1$ & 0.0261 & 0.0523 & 0.0288 & 0.0385 & 0.0137 & 0.0236 & 0.0033 & 0.0040 & 0.0673 & 0.0725 & 0.0588 & 0.0898 & 0.2370 & 0.2925 & 0.0621 & 0.0819\\ 
 & $\mathcal{L}_\text{BNS}$ & $\sum |\partial \mathcal{L} / \partial \phi|$ & $\phi = \text{B}2$ & 0.0258 & 0.0522 & 0.0286 & 0.0377 & 0.0142 & 0.0235 & 0.0028 & 0.0036 & 0.0732 & 0.0786 & 0.0607 & 0.0927 & 0.2409 & 0.2905 & 0.0637 & 0.0827\\ 
 & $\mathcal{L}_\text{BNS}$ & $\sum |\partial \mathcal{L} / \partial \phi|$ & $\phi = \text{B}3$ & 0.0289 & 0.0606 & 0.0308 & 0.0417 & 0.0279 & 0.0377 & 0.0028 & 0.0035 & 0.0771 & 0.0833 & 0.1125 & 0.1420 & 0.3638 & 0.4054 & 0.0920 & 0.1106\\ 
 & $\mathcal{L}_\text{BNS}$ & $\sum |\partial \mathcal{L} / \partial \phi|$ & $\phi = \text{B}4$ & 0.0245 & 0.0510 & 0.0248 & 0.0368 & 0.0181 & 0.0285 & 0.0041 & 0.0046 & 0.0582 & 0.0650 & 0.0595 & 0.0893 & 0.2441 & 0.2761 & 0.0619 & 0.0788\\ \cline{2-20}
\hline 
\hline 
\multirow{6}{*}{\rotatebox[origin=c]{90}{ElasticFace \cite{elasticface}}} & - & $\text{MSE}(\text{BNS}_{\mathcal{M}}, \text{BNS}_{\mathcal{I}})$ & - & 0.0335 & 0.0631 & 0.0314 & 0.0329 & 0.0194 & 0.0243 & 0.0031 & 0.0041 & 0.0628 & 0.0651 & 0.0572 & 0.0754 & 0.2143 & 0.2583 & 0.0602 & 0.0747\\ \cline{2-20}
 & $\mathcal{L}_\text{BNS}$ & $\sum |\partial \mathcal{L} / \partial \phi|$ & $\phi =  \mathcal{I}$ & 0.0280 & 0.0505 & 0.0328 & 0.0360 & 0.0118 & 0.0152 & 0.0032 & 0.0041 & 0.0620 & 0.0638 & 0.0549 & 0.0705 & 0.2160 & 0.2445 & 0.0584 & 0.0692\\ 
 & $\mathcal{L}_\text{BNS}$ & $\sum |\partial \mathcal{L} / \partial \phi|$ & $\phi = \text{B}1$ & 0.0279 & 0.0503 & 0.0313 & 0.0347 & 0.0127 & 0.0160 & 0.0031 & 0.0040 & 0.0647 & 0.0667 & 0.0556 & 0.0837 & 0.2168 & 0.2468 & 0.0589 & 0.0717\\ 
 & $\mathcal{L}_\text{BNS}$ & $\sum |\partial \mathcal{L} / \partial \phi|$ & $\phi = \text{B}2$ & 0.0278 & 0.0500 & 0.0317 & 0.0353 & 0.0132 & 0.0165 & 0.0027 & 0.0036 & 0.0711 & 0.0732 & 0.0577 & 0.0852 & 0.2183 & 0.2555 & 0.0604 & 0.0742\\ 
 & $\mathcal{L}_\text{BNS}$ & $\sum |\partial \mathcal{L} / \partial \phi|$ & $\phi = \text{B}3$ & 0.0311 & 0.0571 & 0.0358 & 0.0395 & 0.0244 & 0.0320 & 0.0026 & 0.0035 & 0.0748 & 0.0763 & 0.1044 & 0.1536 & 0.3376 & 0.3845 & 0.0872 & 0.1066\\ 
 & $\mathcal{L}_\text{BNS}$ & $\sum |\partial \mathcal{L} / \partial \phi|$ & $\phi = \text{B}4$ & 0.0262 & 0.0488 & 0.0258 & 0.0280 & 0.0139 & 0.0183 & 0.0039 & 0.0046 & 0.0556 & 0.0578 & 0.0546 & 0.0693 & 0.2012 & 0.2416 & 0.0545 & 0.0669\\ \cline{2-20}
\hline 
\hline 
\multirow{6}{*}{\rotatebox[origin=c]{90}{MagFace \cite{meng_2021_magface}}} & - & $\text{MSE}(\text{BNS}_{\mathcal{M}}, \text{BNS}_{\mathcal{I}})$ & - & 0.0326 & 0.0663 & 0.0329 & 0.0699 & 0.0305 & 0.0486 & 0.0041 & 0.0049 & 0.0664 & 0.0682 & 0.0627 & 0.1414 & 0.2727 & 0.3274 & 0.0717 & 0.1038\\ \cline{2-20}
 & $\mathcal{L}_\text{BNS}$ & $\sum |\partial \mathcal{L} / \partial \phi|$ & $\phi =  \mathcal{I}$ & 0.0276 & 0.0546 & 0.0339 & 0.0524 & 0.0188 & 0.0320 & 0.0035 & 0.0046 & 0.0648 & 0.0671 & 0.0601 & 0.1230 & 0.2654 & 0.3009 & 0.0677 & 0.0907\\ 
 & $\mathcal{L}_\text{BNS}$ & $\sum |\partial \mathcal{L} / \partial \phi|$ & $\phi = \text{B}1$ & 0.0275 & 0.0550 & 0.0318 & 0.0504 & 0.0197 & 0.0318 & 0.0034 & 0.0045 & 0.0676 & 0.0700 & 0.0624 & 0.1460 & 0.2698 & 0.3119 & 0.0689 & 0.0957\\ 
 & $\mathcal{L}_\text{BNS}$ & $\sum |\partial \mathcal{L} / \partial \phi|$ & $\phi = \text{B}2$ & 0.0274 & 0.0552 & 0.0319 & 0.0486 & 0.0206 & 0.0328 & 0.0029 & 0.0041 & 0.0736 & 0.0759 & 0.0673 & 0.1461 & 0.2728 & 0.3099 & 0.0709 & 0.0961\\ 
 & $\mathcal{L}_\text{BNS}$ & $\sum |\partial \mathcal{L} / \partial \phi|$ & $\phi = \text{B}3$ & 0.0310 & 0.0631 & 0.0353 & 0.0529 & 0.0393 & 0.0584 & 0.0028 & 0.0040 & 0.0768 & 0.0798 & 0.1239 & 0.2893 & 0.4115 & 0.4721 & 0.1029 & 0.1457\\ 
 & $\mathcal{L}_\text{BNS}$ & $\sum |\partial \mathcal{L} / \partial \phi|$ & $\phi = \text{B}4$ & 0.0257 & 0.0525 & 0.0266 & 0.0540 & 0.0214 & 0.0376 & 0.0045 & 0.0060 & 0.0576 & 0.0597 & 0.0625 & 0.1123 & 0.2679 & 0.3092 & 0.0666 & 0.0902\\ \cline{2-20}
\hline 
\hline 
\multirow{6}{*}{\rotatebox[origin=c]{90}{\makecell{Curricular- \\ Face \cite{curricularFace}}}} & - & $\text{MSE}(\text{BNS}_{\mathcal{M}}, \text{BNS}_{\mathcal{I}})$ & - & 0.0284 & 0.0583 & 0.0313 & 0.0388 & 0.0215 & 0.0280 & 0.0035 & 0.0041 & 0.0659 & 0.0698 & 0.0529 & 0.0798 & 0.1916 & 0.2380 & 0.0564 & 0.0738\\ \cline{2-20}
 & $\mathcal{L}_\text{BNS}$ & $\sum |\partial \mathcal{L} / \partial \phi|$ & $\phi =  \mathcal{I}$ & 0.0249 & 0.0459 & 0.0330 & 0.0387 & 0.0140 & 0.0194 & 0.0034 & 0.0041 & 0.0628 & 0.0659 & 0.0499 & 0.0752 & 0.1988 & 0.2275 & 0.0553 & 0.0681\\ 
 & $\mathcal{L}_\text{BNS}$ & $\sum |\partial \mathcal{L} / \partial \phi|$ & $\phi = \text{B}1$ & 0.0248 & 0.0457 & 0.0305 & 0.0358 & 0.0149 & 0.0197 & 0.0033 & 0.0040 & 0.0646 & 0.0675 & 0.0509 & 0.0909 & 0.2036 & 0.2288 & 0.0561 & 0.0703\\ 
 & $\mathcal{L}_\text{BNS}$ & $\sum |\partial \mathcal{L} / \partial \phi|$ & $\phi = \text{B}2$ & 0.0246 & 0.0455 & 0.0306 & 0.0361 & 0.0153 & 0.0198 & 0.0029 & 0.0036 & 0.0703 & 0.0731 & 0.0516 & 0.0912 & 0.2029 & 0.2320 & 0.0569 & 0.0716\\ 
 & $\mathcal{L}_\text{BNS}$ & $\sum |\partial \mathcal{L} / \partial \phi|$ & $\phi = \text{B}3$ & 0.0274 & 0.0512 & 0.0333 & 0.0401 & 0.0294 & 0.0335 & 0.0030 & 0.0037 & 0.0735 & 0.0765 & 0.0981 & 0.1657 & 0.3322 & 0.3716 & 0.0853 & 0.1060\\ 
 & $\mathcal{L}_\text{BNS}$ & $\sum |\partial \mathcal{L} / \partial \phi|$ & $\phi = \text{B}4$ & 0.0230 & 0.0438 & 0.0253 & 0.0299 & 0.0170 & 0.0195 & 0.0041 & 0.0046 & 0.0572 & 0.0600 & 0.0495 & 0.0710 & 0.2204 & 0.2639 & 0.0566 & 0.0704\\ \cline{2-20}
\hline 
\end{tabular}}
\caption{The achieved AUC of EDC by using two approaches presented in this paper, MSE of BNS ($\text{MSE}_{\text{BNS}}$) and gradient magnitudes ($\mathcal{L}_{\text{BNS}}$), and under different settings. The ResNet50 model is used. The gradient magnitudes are extracted during the backpropagation step from different intermediate layers, B1, B2, B3 and B4 ($\phi = \text{B}1 - \phi = \text{B}4$) as well as on the pixel level  ($\phi = \mathcal{I}$).  The results are reported under two operation threshold FMR$=1e-3$ and FMR$=1e-4$ and under two protocols, same model (ArcFace) and cross-model (ElasticFace, MagFace and CurricularFace).}
\label{tbl:resnet50_full_overview}
\end{center}
\end{table*}

%% file: figures/fig_iresnet50_supplementary_adience.tex
\begin{figure*}[h!]
\centering
	\begin{subfigure}[b]{0.9\textwidth}
		\centering
		\includegraphics[width=\textwidth]{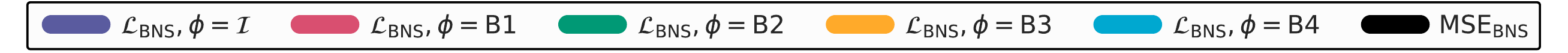}
	\end{subfigure}
\\
	\begin{subfigure}[b]{0.48\textwidth}
		 \centering
		 \includegraphics[width=0.95\textwidth]{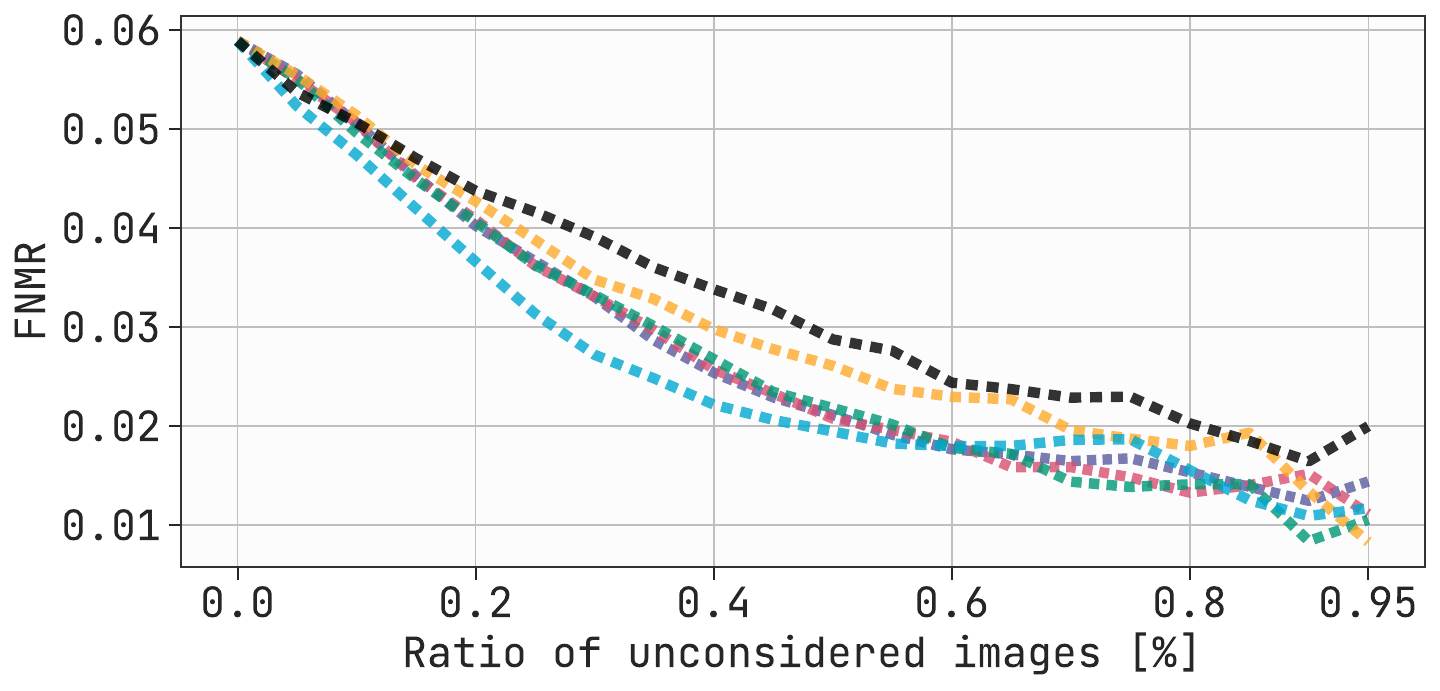}
		 \caption{ArcFace \cite{deng2019arcface} Model, Adience \cite{Adience} Dataset \\ \grafiqs with ResNet50, FMR$=1e-3$}
	\end{subfigure}
\hfill
	\begin{subfigure}[b]{0.48\textwidth}
		 \centering
		 \includegraphics[width=0.95\textwidth]{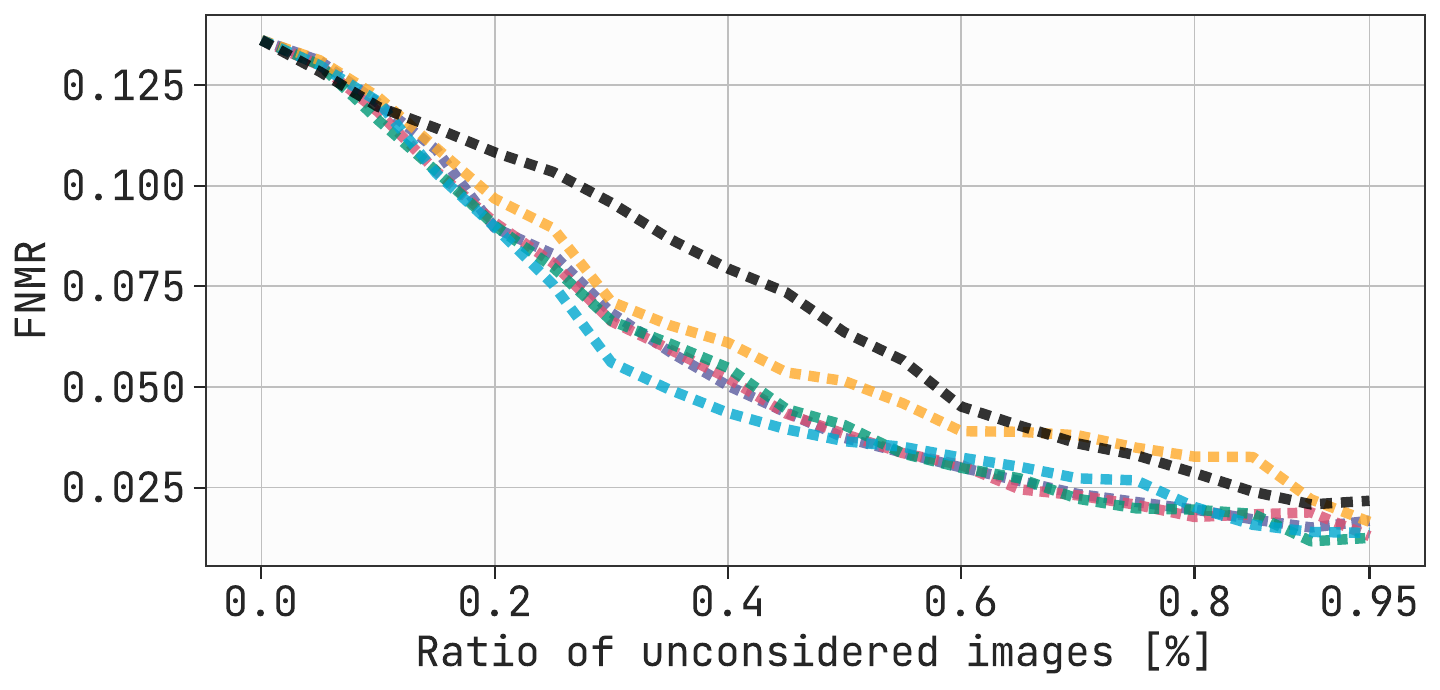}
		 \caption{ArcFace \cite{deng2019arcface} Model, Adience \cite{Adience} Dataset \\ \grafiqs with ResNet50, FMR$=1e-4$}
	\end{subfigure}
\\
	\begin{subfigure}[b]{0.48\textwidth}
		 \centering
		 \includegraphics[width=0.95\textwidth]{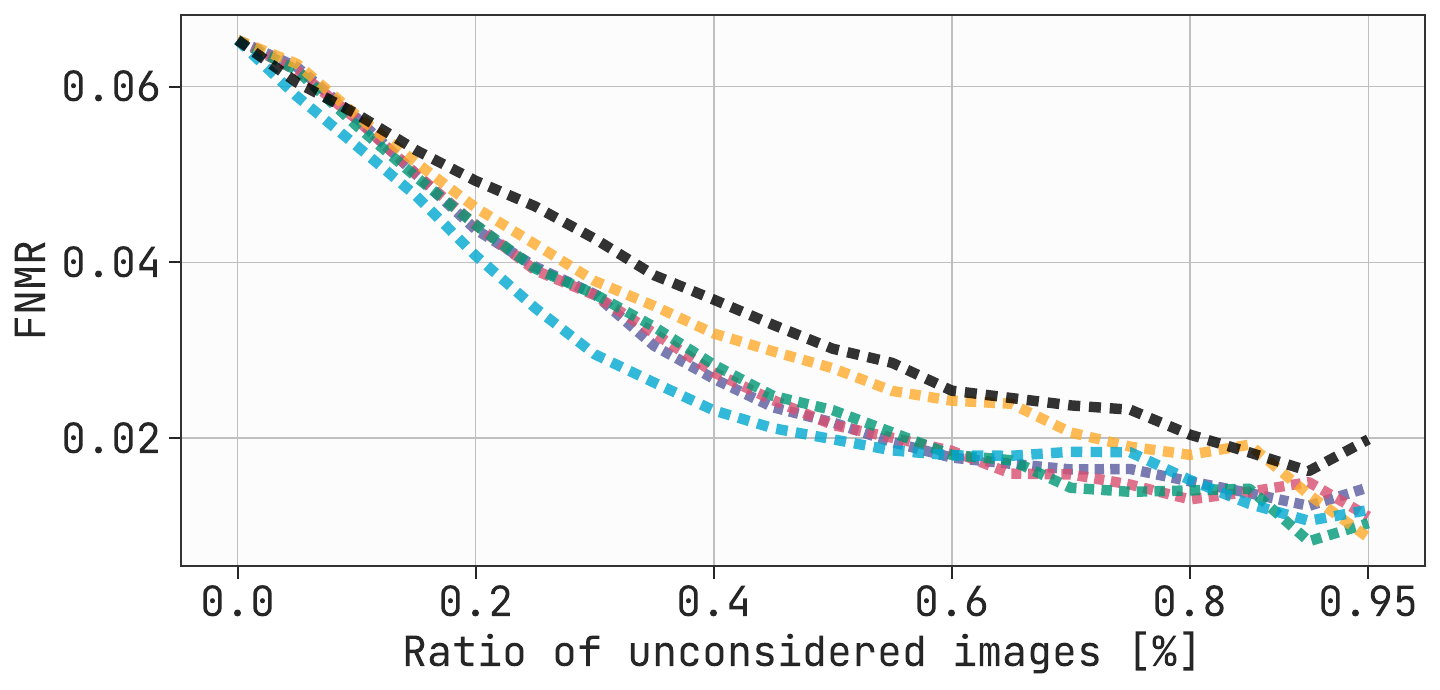}
		 \caption{ElasticFace \cite{elasticface} Model, Adience \cite{Adience} Dataset \\ \grafiqs with ResNet50, FMR$=1e-3$}
	\end{subfigure}
\hfill
	\begin{subfigure}[b]{0.48\textwidth}
		 \centering
		 \includegraphics[width=0.95\textwidth]{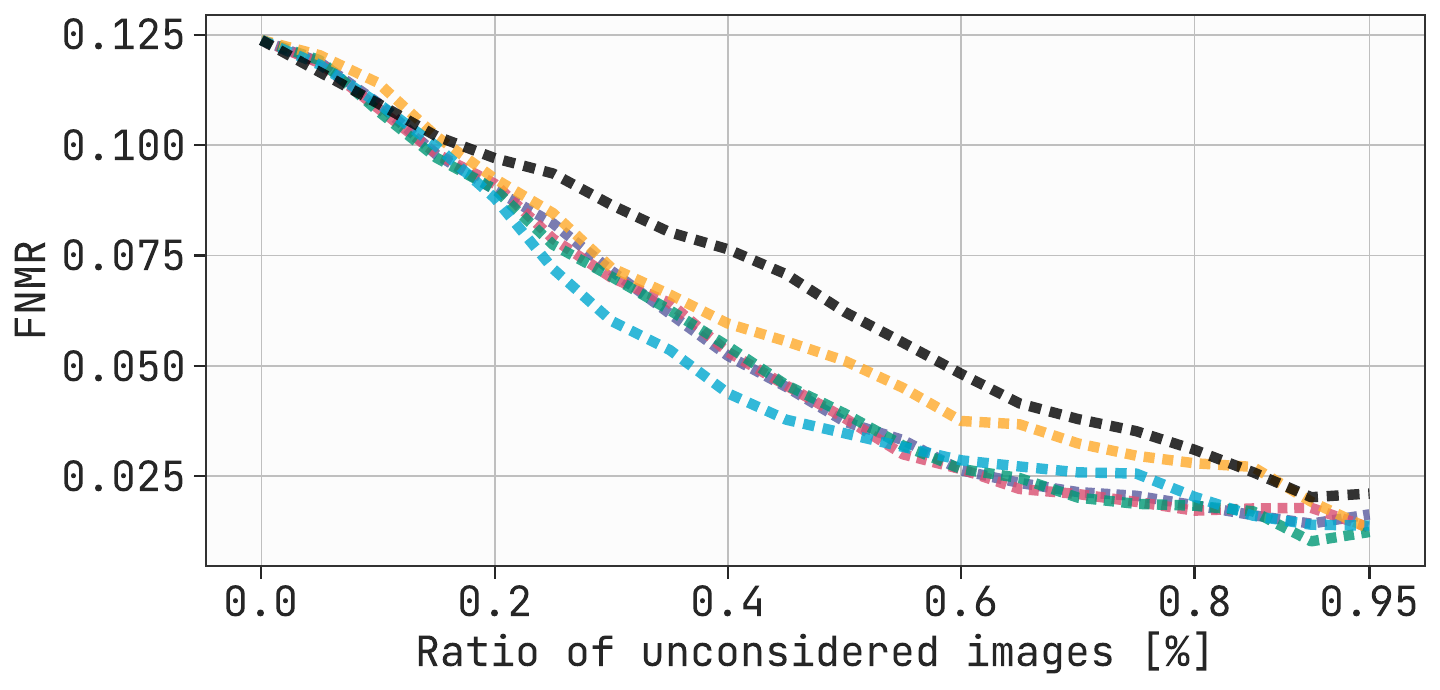}
		 \caption{ElasticFace \cite{elasticface} Model, Adience \cite{Adience} Dataset \\ \grafiqs with ResNet50, FMR$=1e-4$}
	\end{subfigure}
\\
	\begin{subfigure}[b]{0.48\textwidth}
		 \centering
		 \includegraphics[width=0.95\textwidth]{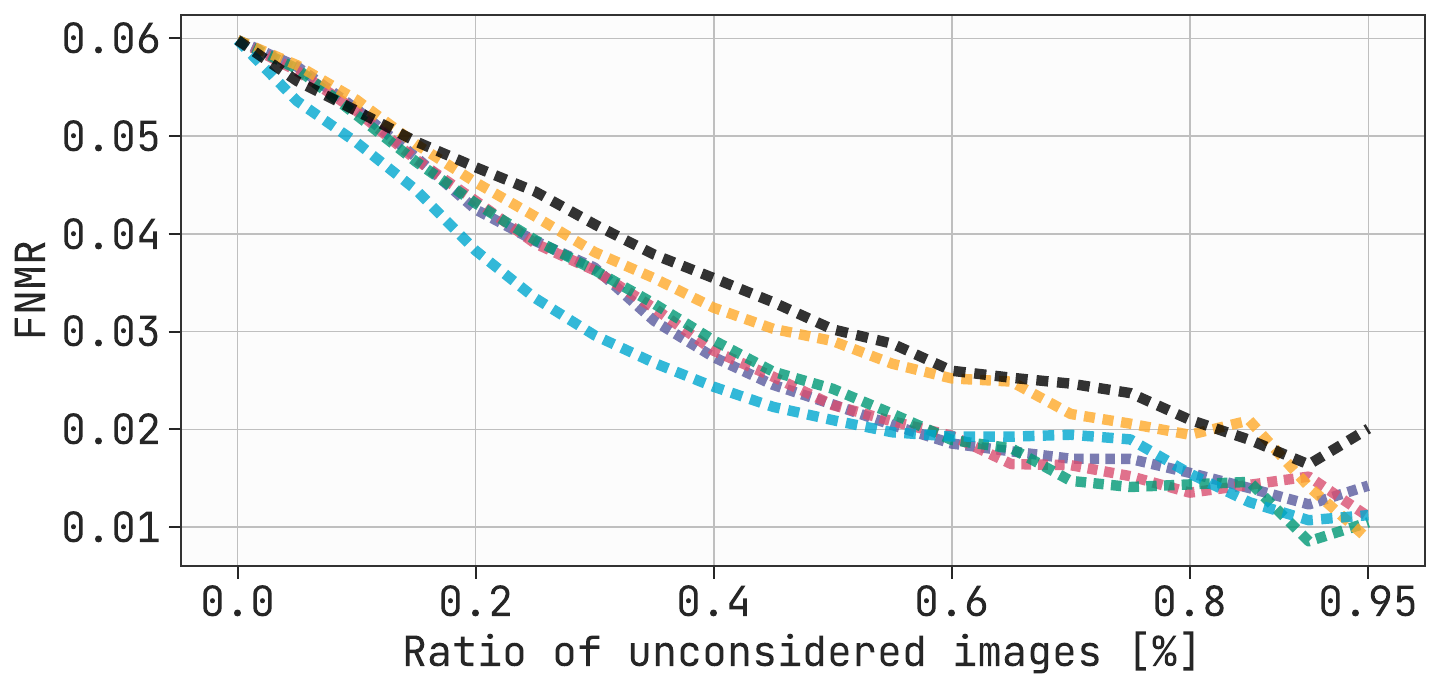}
		 \caption{MagFace \cite{meng_2021_magface} Model, Adience \cite{Adience} Dataset \\ \grafiqs with ResNet50, FMR$=1e-3$}
	\end{subfigure}
\hfill
	\begin{subfigure}[b]{0.48\textwidth}
		 \centering
		 \includegraphics[width=0.95\textwidth]{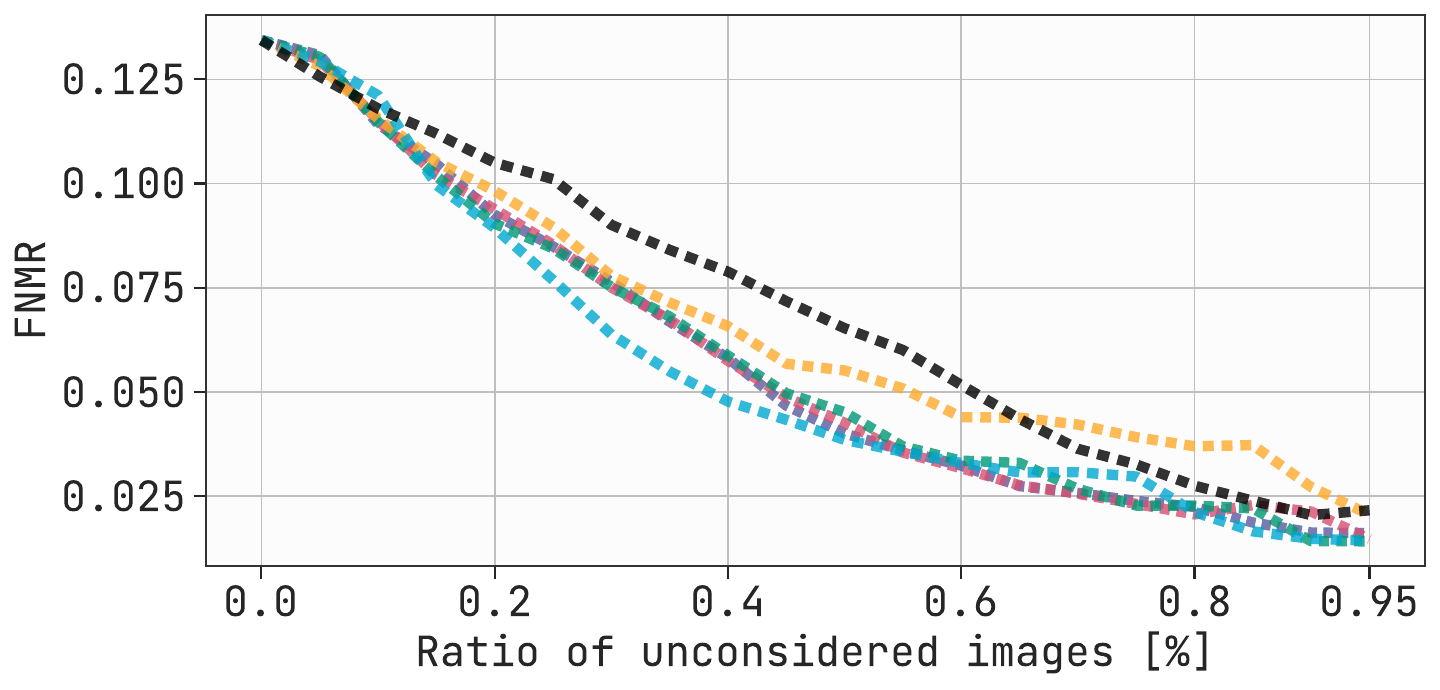}
		 \caption{MagFace \cite{meng_2021_magface} Model, Adience \cite{Adience} Dataset \\ \grafiqs with ResNet50, FMR$=1e-4$}
	\end{subfigure}
\\
	\begin{subfigure}[b]{0.48\textwidth}
		 \centering
		 \includegraphics[width=0.95\textwidth]{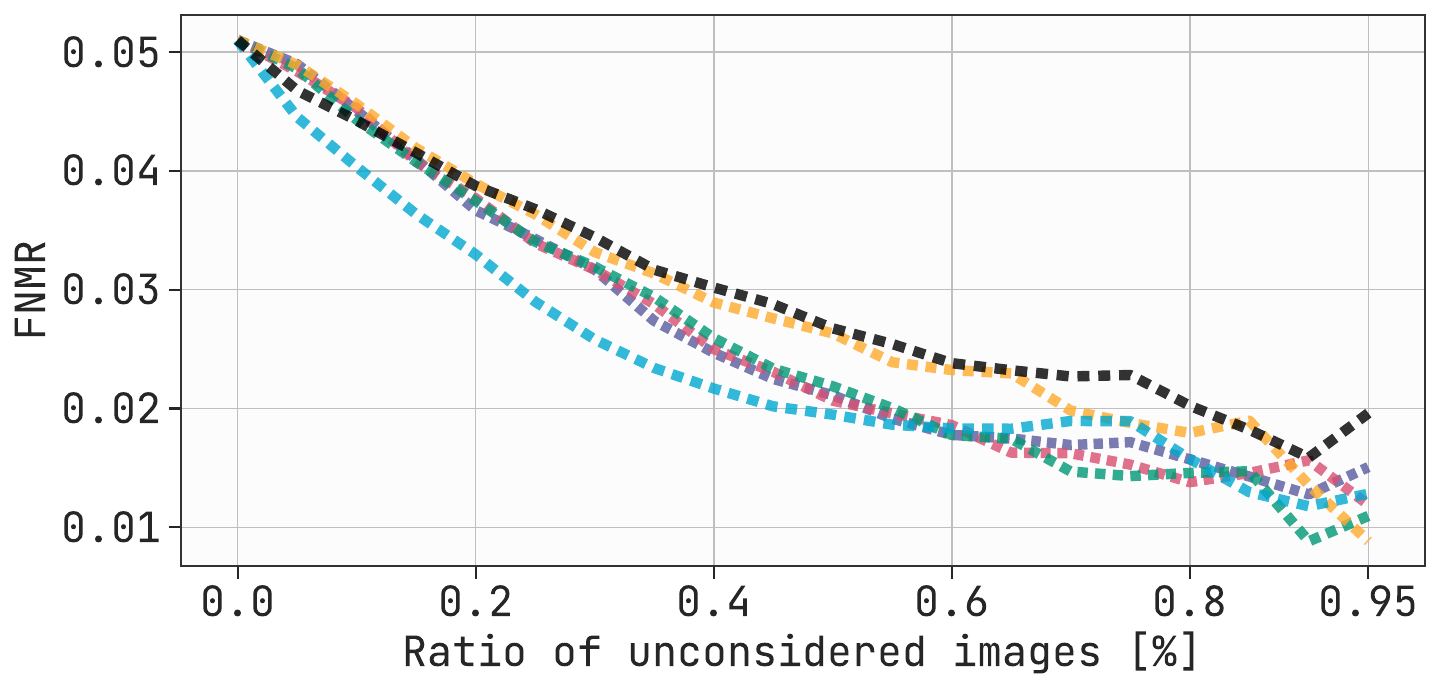}
		 \caption{CurricularFace \cite{curricularFace} Model, Adience \cite{Adience} Dataset \\ \grafiqs with ResNet50, FMR$=1e-3$}
	\end{subfigure}
\hfill
	\begin{subfigure}[b]{0.48\textwidth}
		 \centering
		 \includegraphics[width=0.95\textwidth]{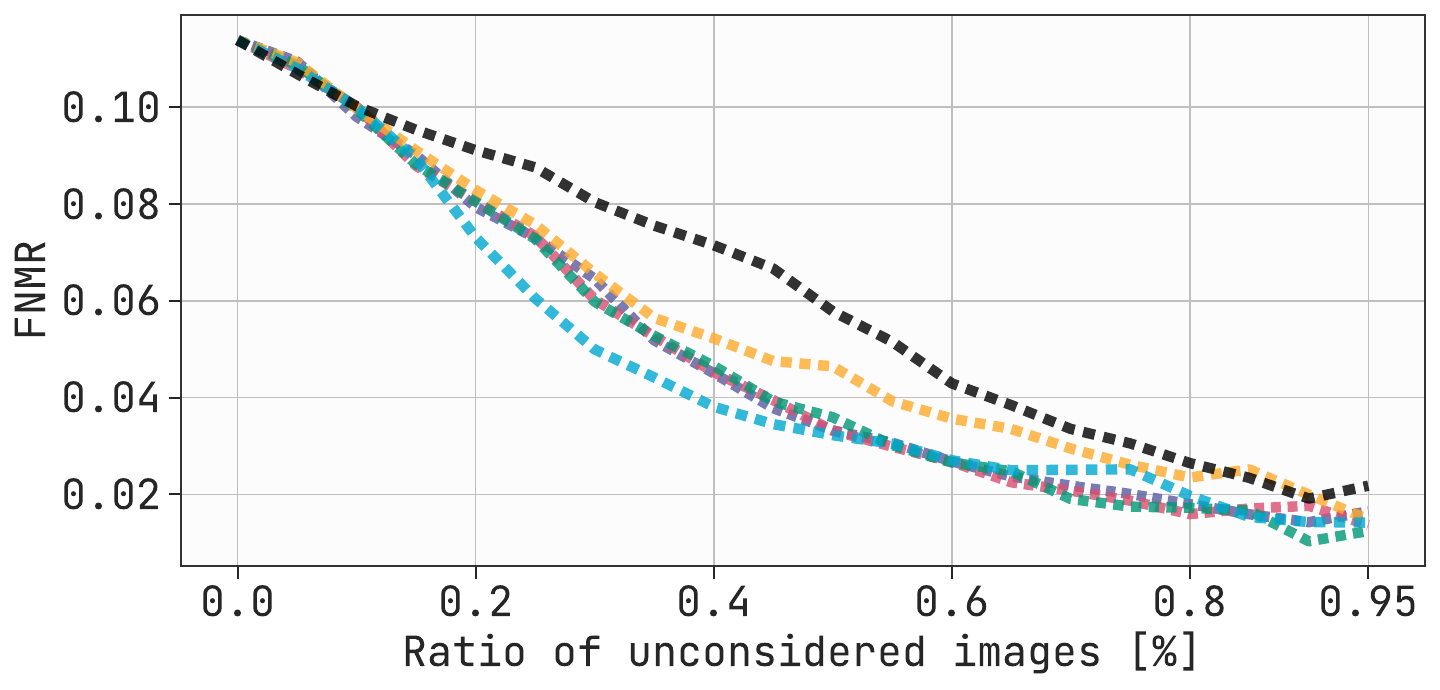}
		 \caption{CurricularFace \cite{curricularFace} Model, Adience \cite{Adience} Dataset \\ \grafiqs with ResNet50, FMR$=1e-4$}
	\end{subfigure}
\\
\caption{Error-versus-Discard Characteristic (EDC) curves for FNMR@FMR=$1e-3$ and FNMR@FMR=$1e-4$ of our proposed method using $\mathcal{L}_{\text{BNS}}$ as backpropagation loss and absolute sum as FIQ. The gradients at image level ($\phi=\mathcal{I}$), and block levels ($\phi=\text{B}1$ $-$ $\phi=\text{B}4$) are used to calculate FIQ. $\text{MSE}_{\text{BNS}}$ as FIQ is shown in black. Results shown on benchmark Adience \cite{Adience} using ArcFace, ElasticFace, MagFace, and, CurricularFace FR models. It is evident that the proposed \grafiqs method leads to lower verification error when images with lowest utility score estimated from gradient magnitudes are rejected. Furthermore, estimating FIQ by backpropagating $\mathcal{L}_{\text{BNS}}$ yields significantly better results than using $\text{MSE}_{\text{BNS}}$ directly.}
\vspace{-4mm}
\label{fig:iresnet50_supplementary_adience}
\end{figure*}

%% file: figures/fig_iresnet50_supplementary_agedb_30.tex
\begin{figure*}[h!]
\centering
	\begin{subfigure}[b]{0.9\textwidth}
		\centering
		\includegraphics[width=\textwidth]{figures/iresnet50_bn_overview/legend.pdf}
	\end{subfigure}
\\
	\begin{subfigure}[b]{0.48\textwidth}
		 \centering
		 \includegraphics[width=0.95\textwidth]{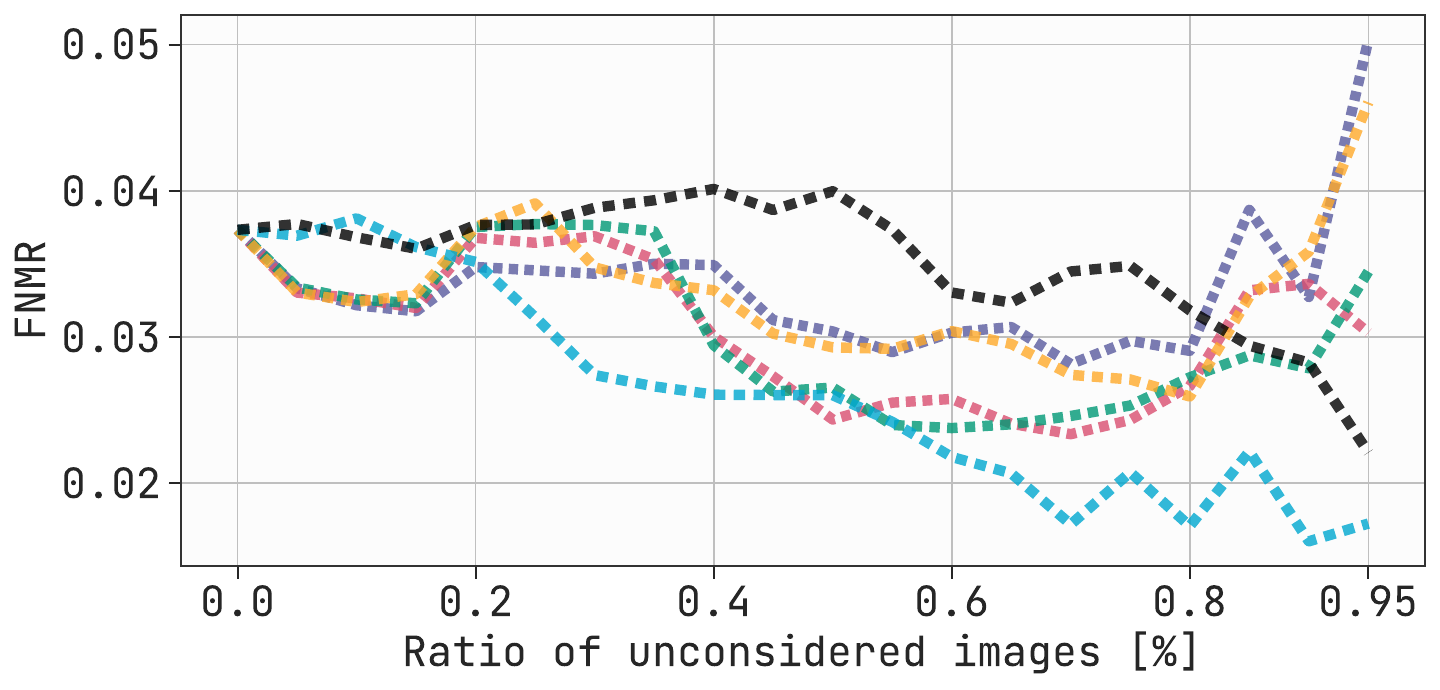}
		 \caption{ArcFace \cite{deng2019arcface} Model, AgeDB30 \cite{agedb} Dataset \\ \grafiqs with ResNet50, FMR$=1e-3$}
	\end{subfigure}
\hfill
	\begin{subfigure}[b]{0.48\textwidth}
		 \centering
		 \includegraphics[width=0.95\textwidth]{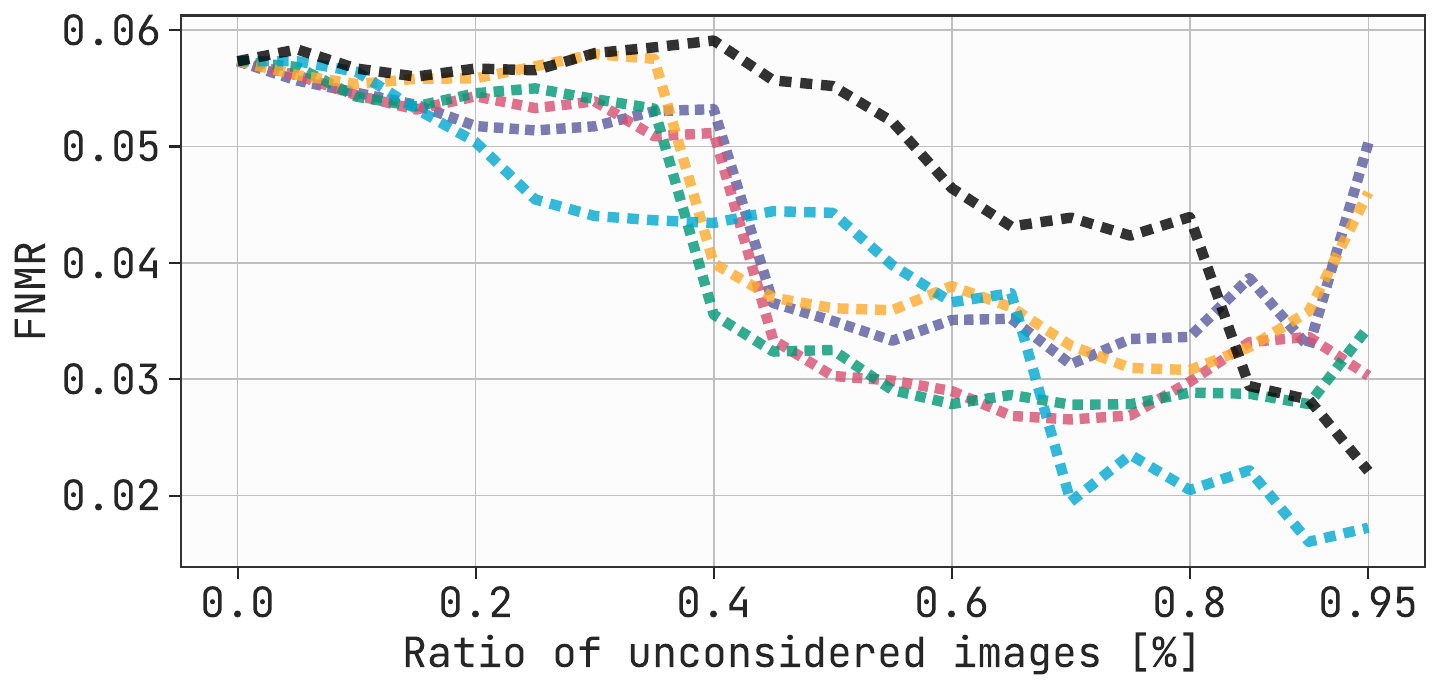}
		 \caption{ArcFace \cite{deng2019arcface} Model, AgeDB30 \cite{agedb} Dataset \\ \grafiqs with ResNet50, FMR$=1e-4$}
	\end{subfigure}
\\
	\begin{subfigure}[b]{0.48\textwidth}
		 \centering
		 \includegraphics[width=0.95\textwidth]{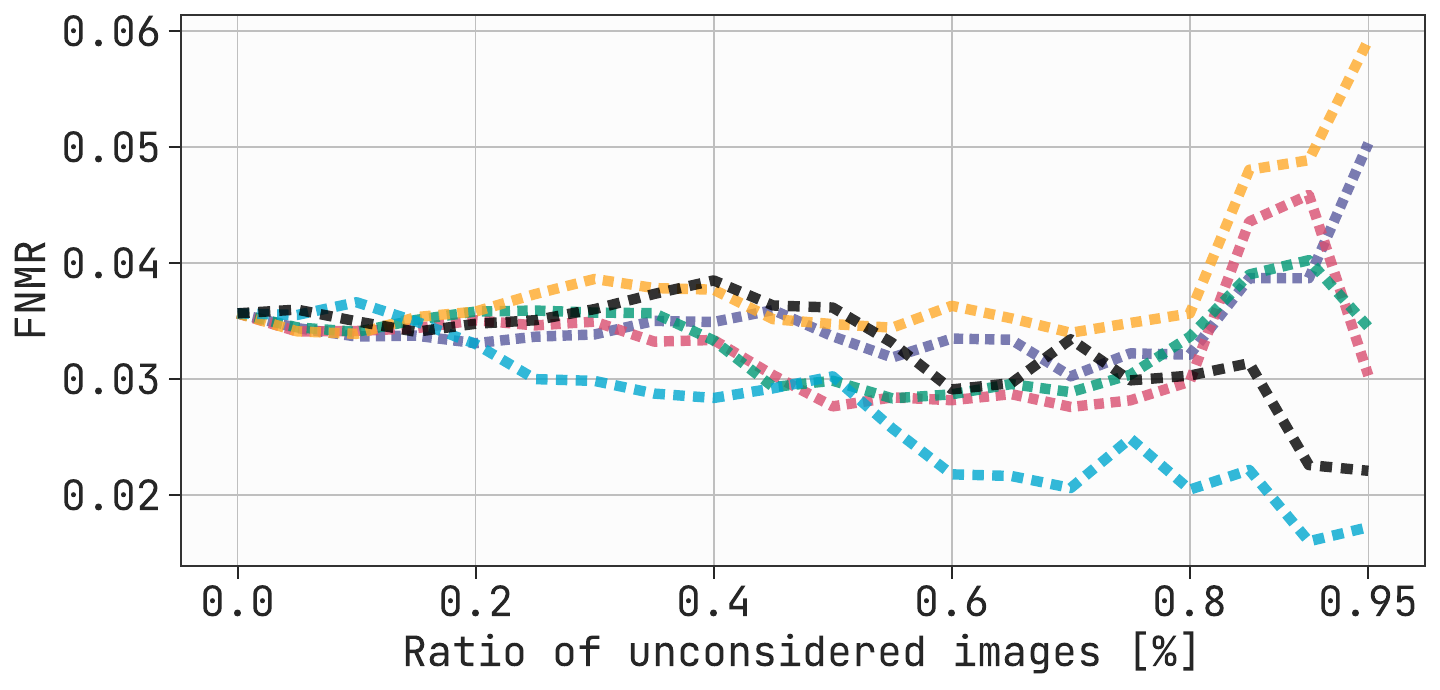}
		 \caption{ElasticFace \cite{elasticface} Model, AgeDB30 \cite{agedb} Dataset \\ \grafiqs with ResNet50, FMR$=1e-3$}
	\end{subfigure}
\hfill
	\begin{subfigure}[b]{0.48\textwidth}
		 \centering
		 \includegraphics[width=0.95\textwidth]{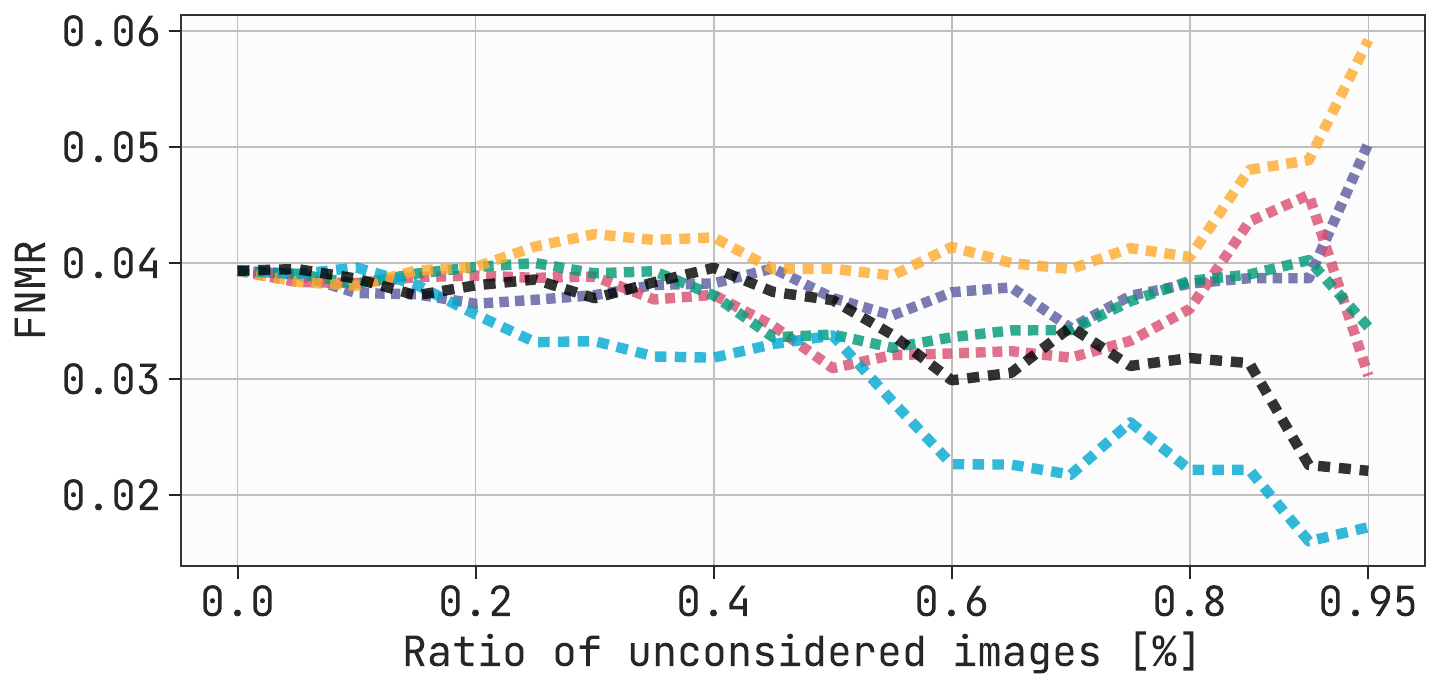}
		 \caption{ElasticFace \cite{elasticface} Model, AgeDB30 \cite{agedb} Dataset \\ \grafiqs with ResNet50, FMR$=1e-4$}
	\end{subfigure}
\\
	\begin{subfigure}[b]{0.48\textwidth}
		 \centering
		 \includegraphics[width=0.95\textwidth]{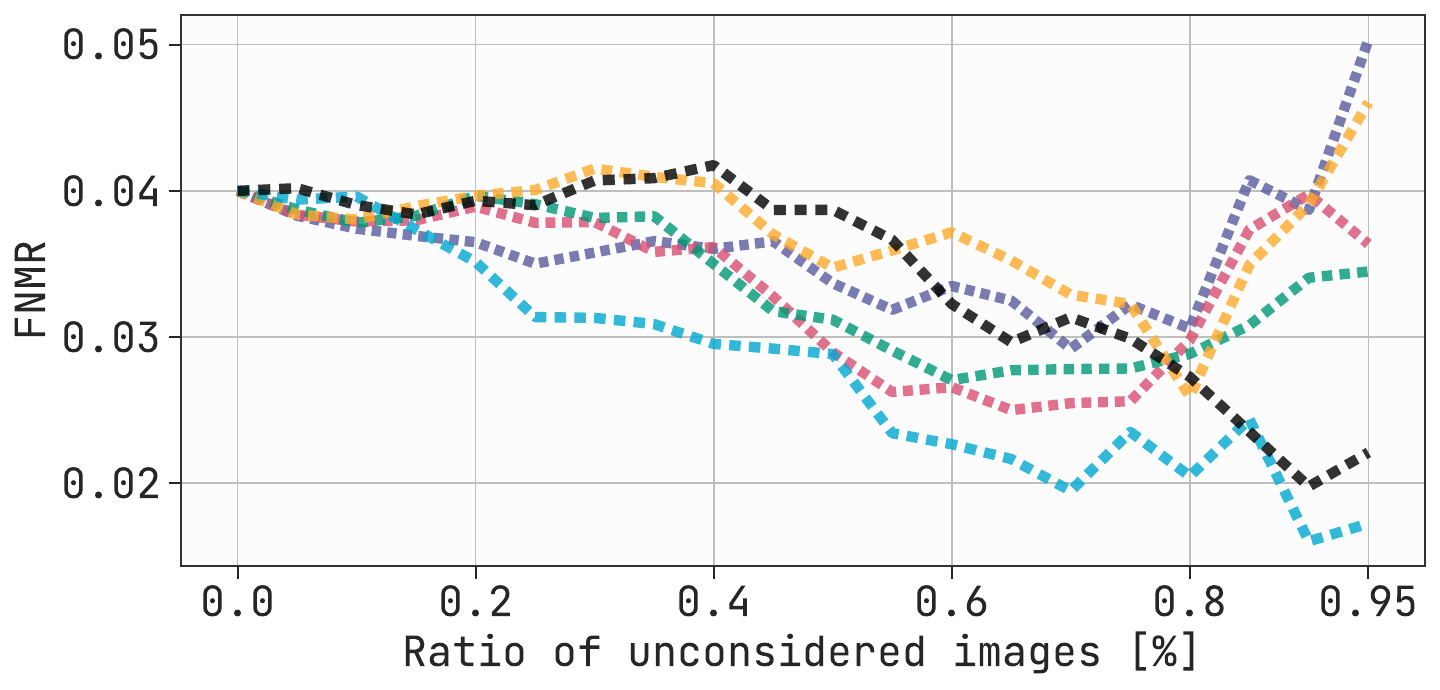}
		 \caption{MagFace \cite{meng_2021_magface} Model, AgeDB30 \cite{agedb} Dataset \\ \grafiqs with ResNet50, FMR$=1e-3$}
	\end{subfigure}
\hfill
	\begin{subfigure}[b]{0.48\textwidth}
		 \centering
		 \includegraphics[width=0.95\textwidth]{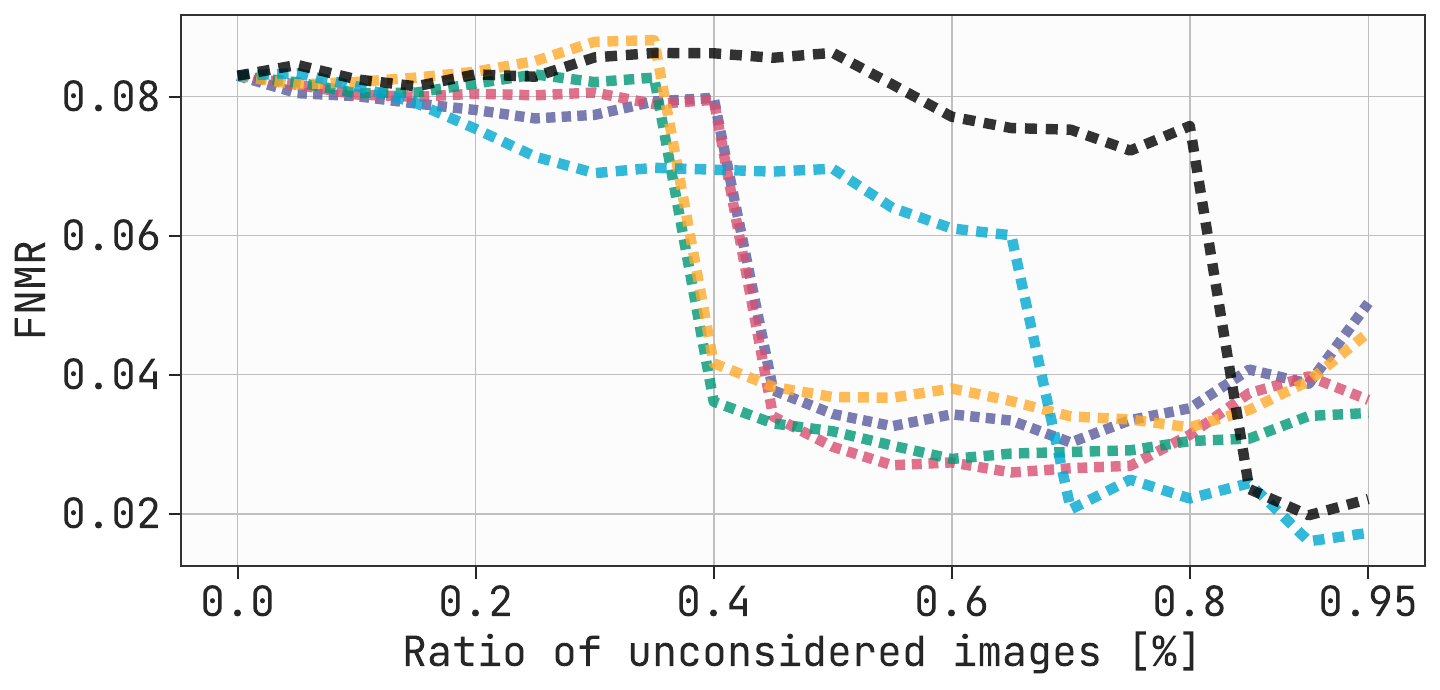}
		 \caption{MagFace \cite{meng_2021_magface} Model, AgeDB30 \cite{agedb} Dataset \\ \grafiqs with ResNet50, FMR$=1e-4$}
	\end{subfigure}
\\
	\begin{subfigure}[b]{0.48\textwidth}
		 \centering
		 \includegraphics[width=0.95\textwidth]{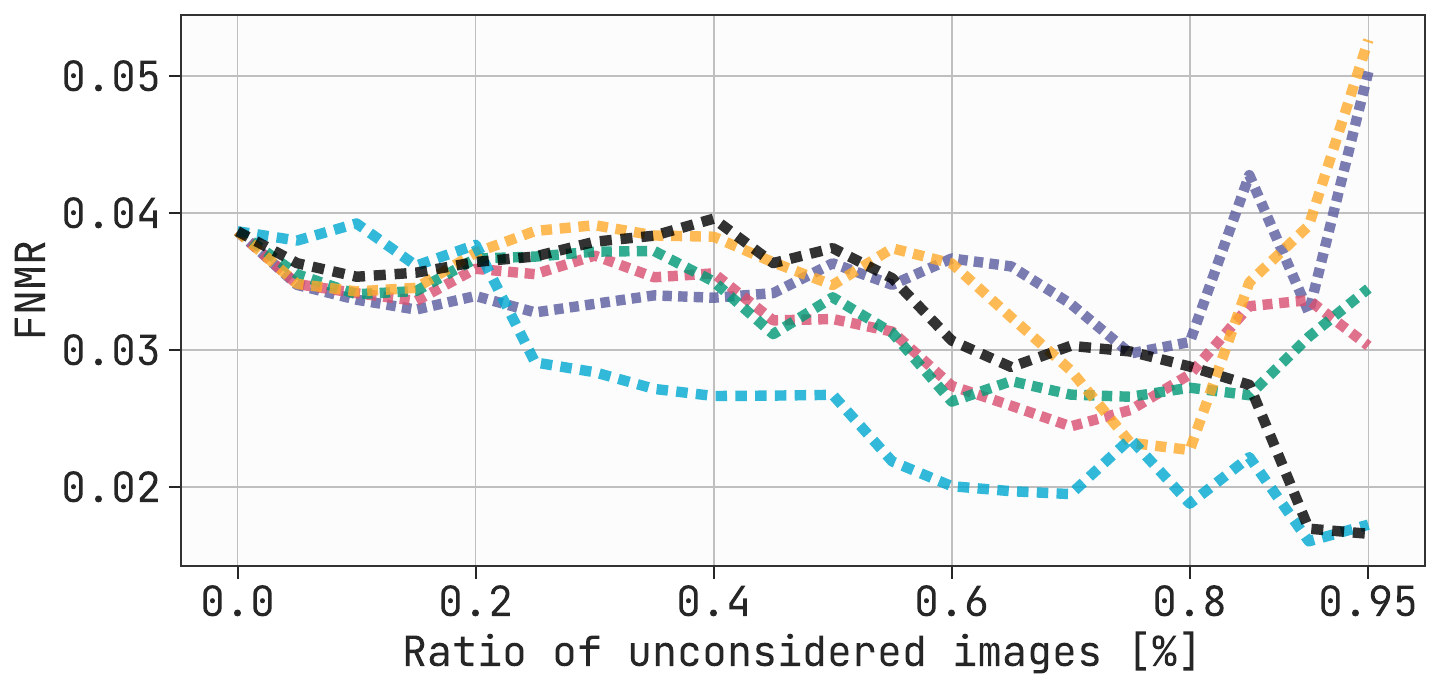}
		 \caption{CurricularFace \cite{curricularFace} Model, AgeDB30 \cite{agedb} Dataset \\ \grafiqs with ResNet50, FMR$=1e-3$}
	\end{subfigure}
\hfill
	\begin{subfigure}[b]{0.48\textwidth}
		 \centering
		 \includegraphics[width=0.95\textwidth]{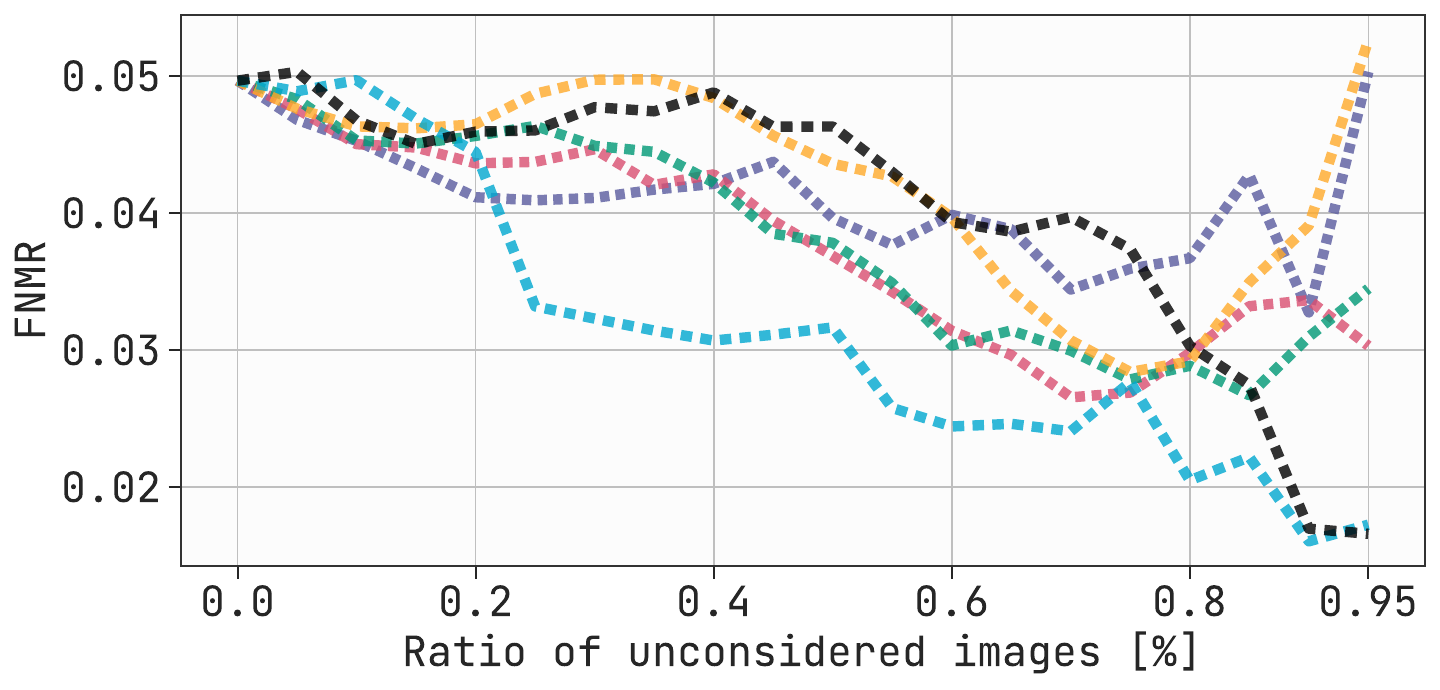}
		 \caption{CurricularFace \cite{curricularFace} Model, AgeDB30 \cite{agedb} Dataset \\ \grafiqs with ResNet50, FMR$=1e-4$}
	\end{subfigure}
\\
\caption{Error-versus-Discard Characteristic (EDC) curves for FNMR@FMR=$1e-3$ and FNMR@FMR=$1e-4$ of our proposed method using $\mathcal{L}_{\text{BNS}}$ as backpropagation loss and absolute sum as FIQ. The gradients at image level ($\phi=\mathcal{I}$), and block levels ($\phi=\text{B}1$ $-$ $\phi=\text{B}4$) are used to calculate FIQ. $\text{MSE}_{\text{BNS}}$ as FIQ is shown in black. Results shown on benchmark AgeDB30 \cite{agedb} using ArcFace, ElasticFace, MagFace, and, CurricularFace FR models. It is evident that the proposed \grafiqs method leads to lower verification error when images with lowest utility score estimated from gradient magnitudes are rejected. Furthermore, estimating FIQ by backpropagating $\mathcal{L}_{\text{BNS}}$ yields significantly better results than using $\text{MSE}_{\text{BNS}}$ directly.}
\vspace{-4mm}
\label{fig:iresnet50_supplementary_agedb_30}
\end{figure*}

%% file: figures/fig_iresnet50_supplementary_cfp_fp.tex
\begin{figure*}[h!]
\centering
	\begin{subfigure}[b]{0.9\textwidth}
		\centering
		\includegraphics[width=\textwidth]{figures/iresnet50_bn_overview/legend.pdf}
	\end{subfigure}
\\
	\begin{subfigure}[b]{0.48\textwidth}
		 \centering
		 \includegraphics[width=0.95\textwidth]{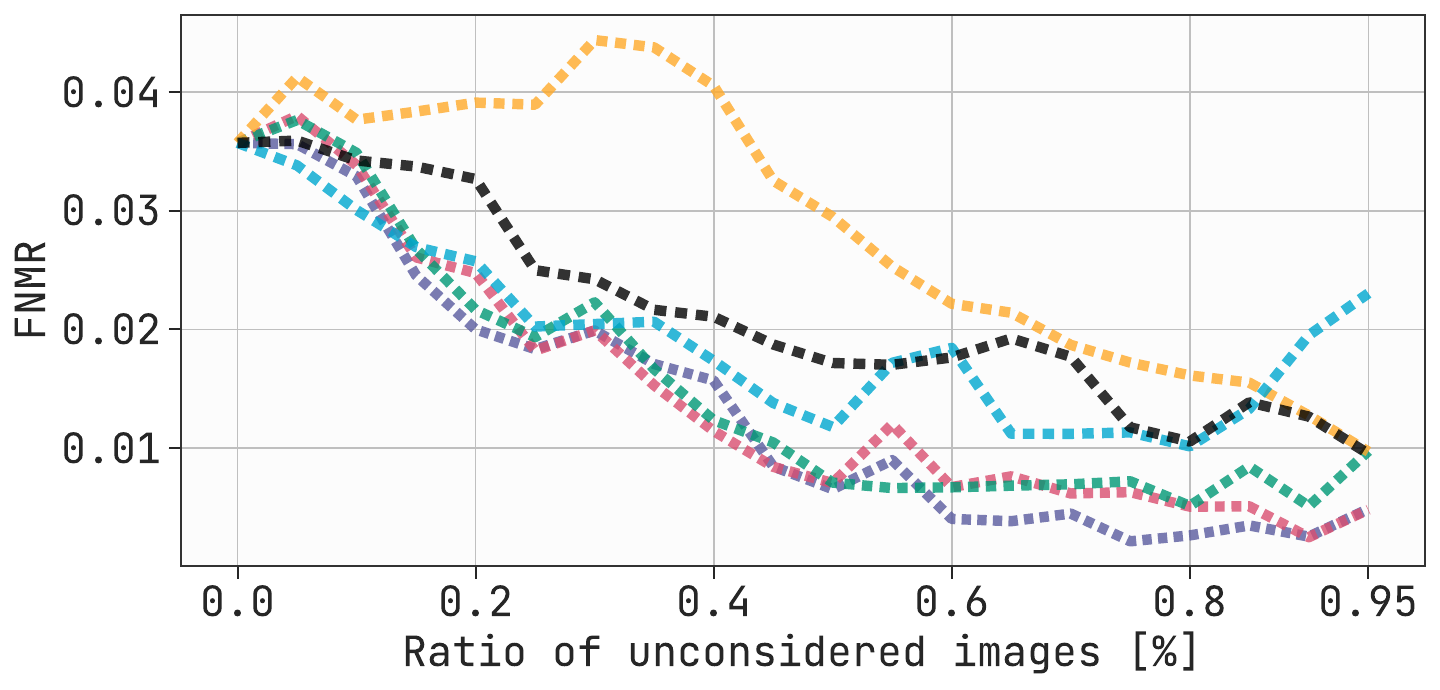}
		 \caption{ArcFace \cite{deng2019arcface} Model, CFP-FP \cite{cfp-fp} Dataset \\ \grafiqs with ResNet50, FMR$=1e-3$}
	\end{subfigure}
\hfill
	\begin{subfigure}[b]{0.48\textwidth}
		 \centering
		 \includegraphics[width=0.95\textwidth]{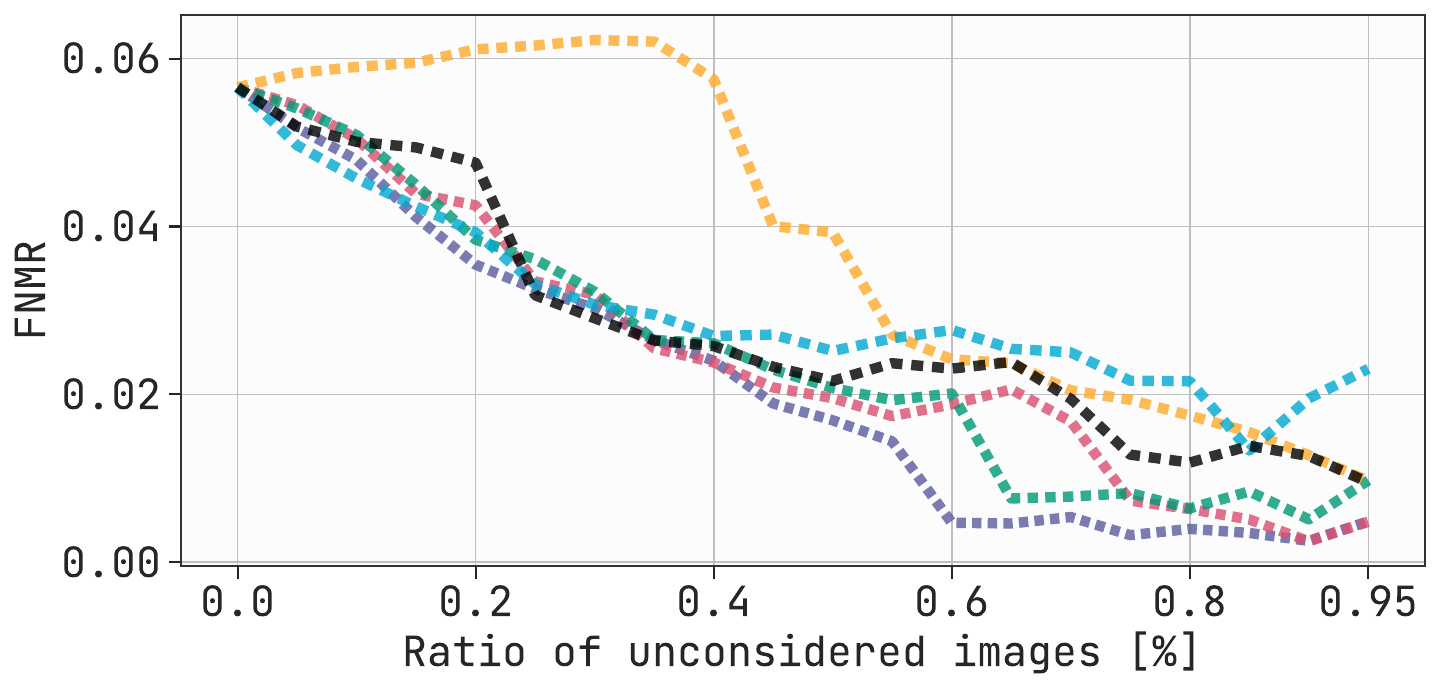}
		 \caption{ArcFace \cite{deng2019arcface} Model, CFP-FP \cite{cfp-fp} Dataset \\ \grafiqs with ResNet50, FMR$=1e-4$}
	\end{subfigure}
\\
	\begin{subfigure}[b]{0.48\textwidth}
		 \centering
		 \includegraphics[width=0.95\textwidth]{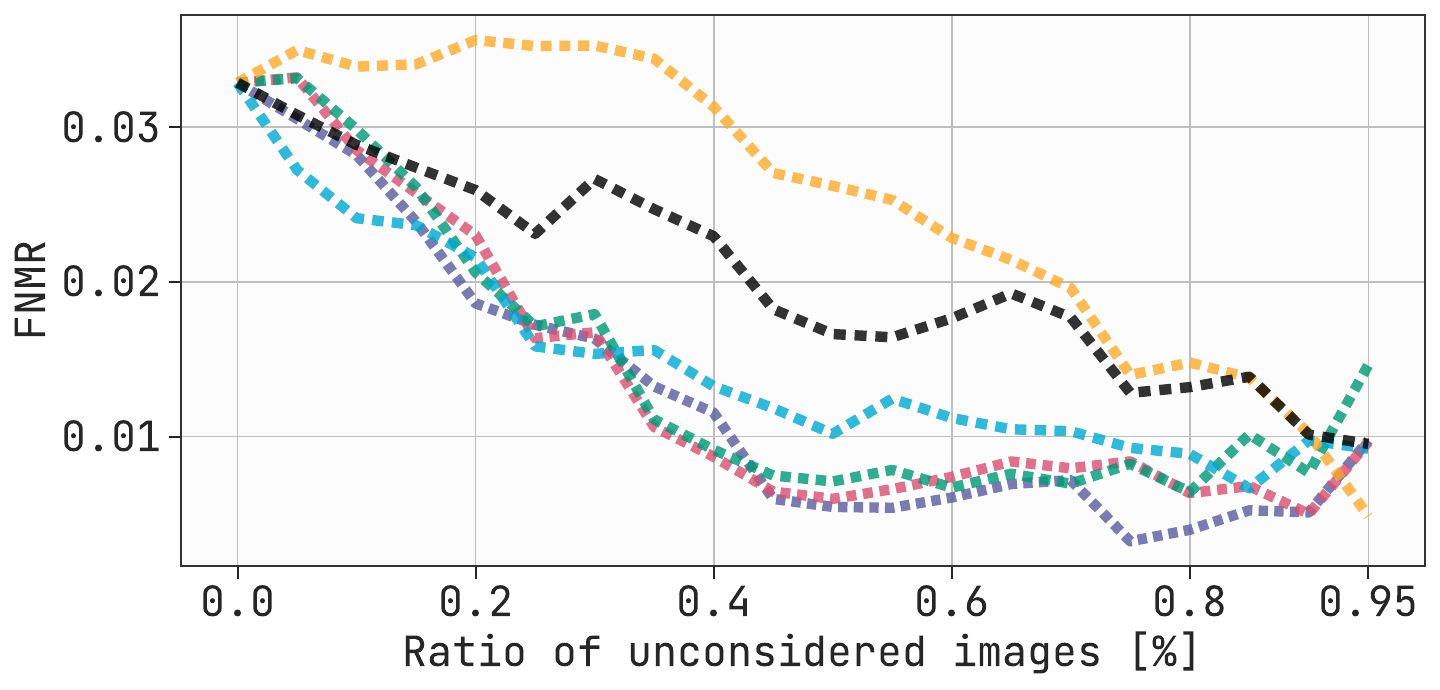}
		 \caption{ElasticFace \cite{elasticface} Model, CFP-FP \cite{cfp-fp} Dataset \\ \grafiqs with ResNet50, FMR$=1e-3$}
	\end{subfigure}
\hfill
	\begin{subfigure}[b]{0.48\textwidth}
		 \centering
		 \includegraphics[width=0.95\textwidth]{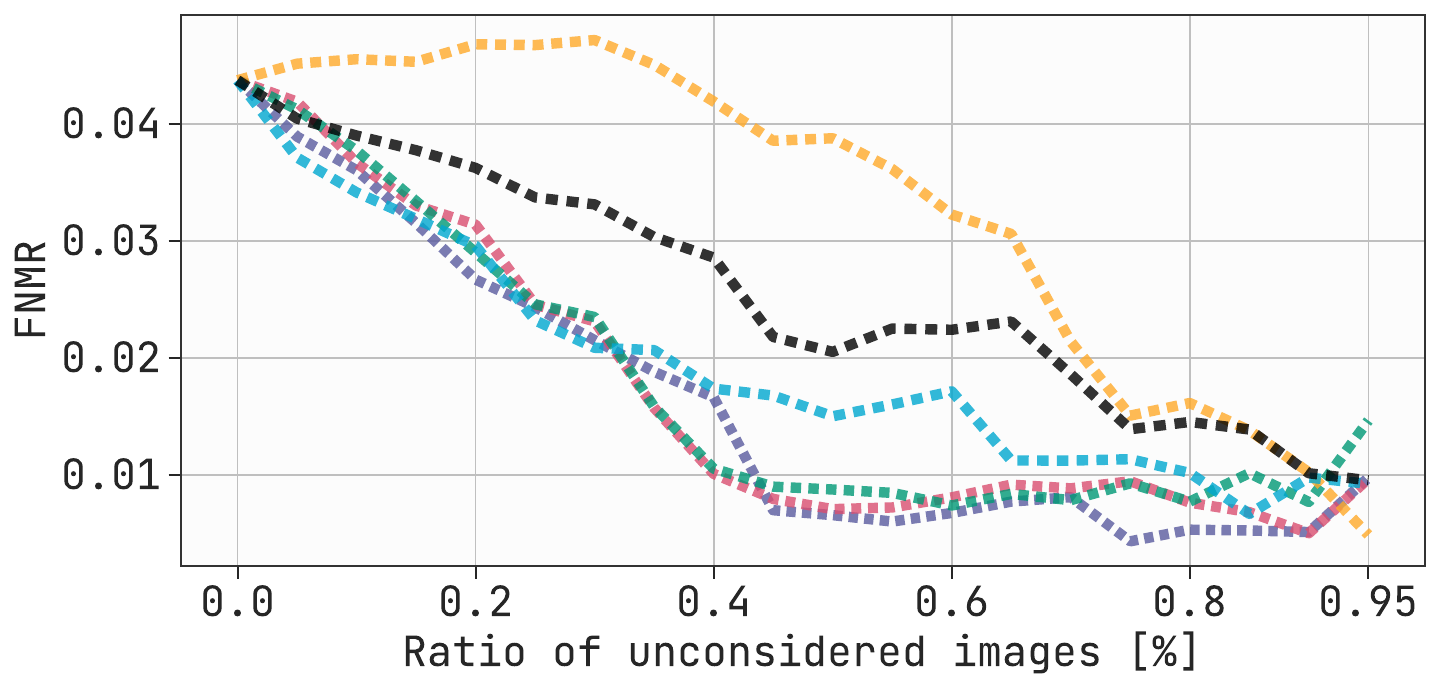}
		 \caption{ElasticFace \cite{elasticface} Model, CFP-FP \cite{cfp-fp} Dataset \\ \grafiqs with ResNet50, FMR$=1e-4$}
	\end{subfigure}
\\
	\begin{subfigure}[b]{0.48\textwidth}
		 \centering
		 \includegraphics[width=0.95\textwidth]{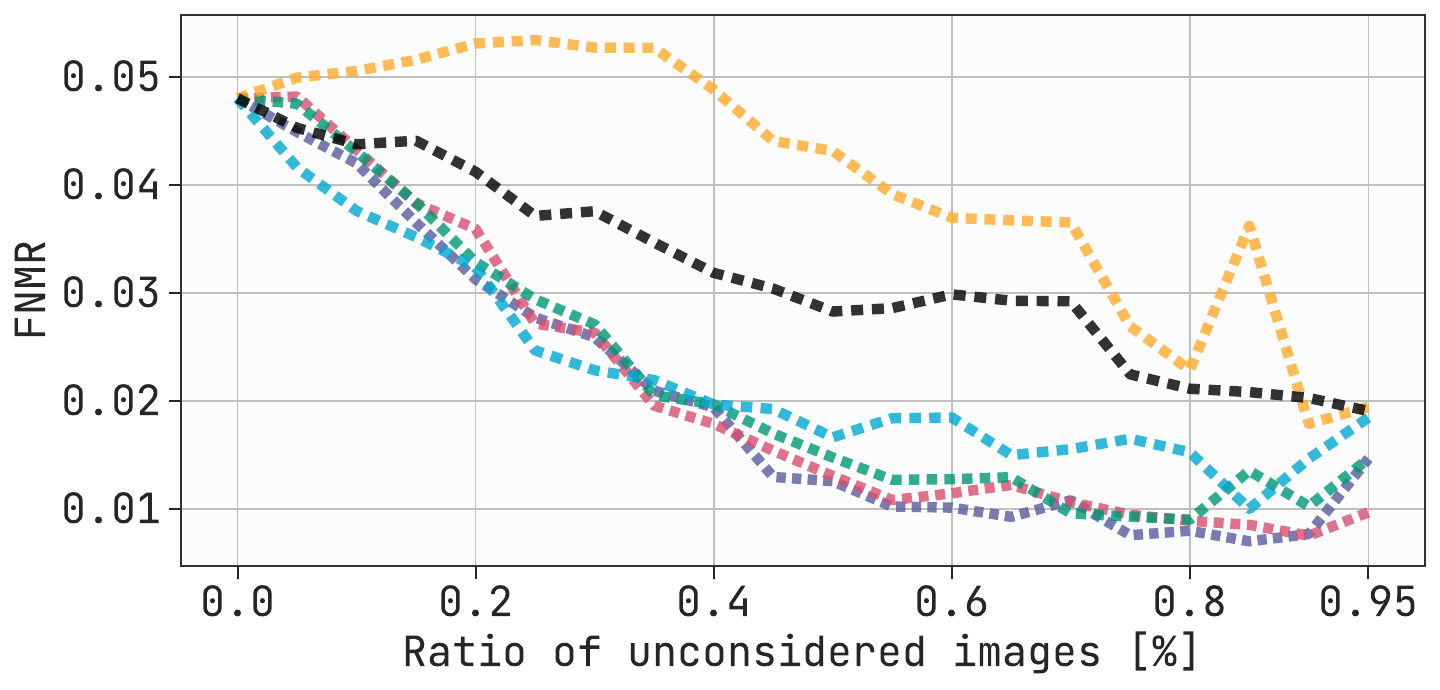}
		 \caption{MagFace \cite{meng_2021_magface} Model, CFP-FP \cite{cfp-fp} Dataset \\ \grafiqs with ResNet50, FMR$=1e-3$}
	\end{subfigure}
\hfill
	\begin{subfigure}[b]{0.48\textwidth}
		 \centering
		 \includegraphics[width=0.95\textwidth]{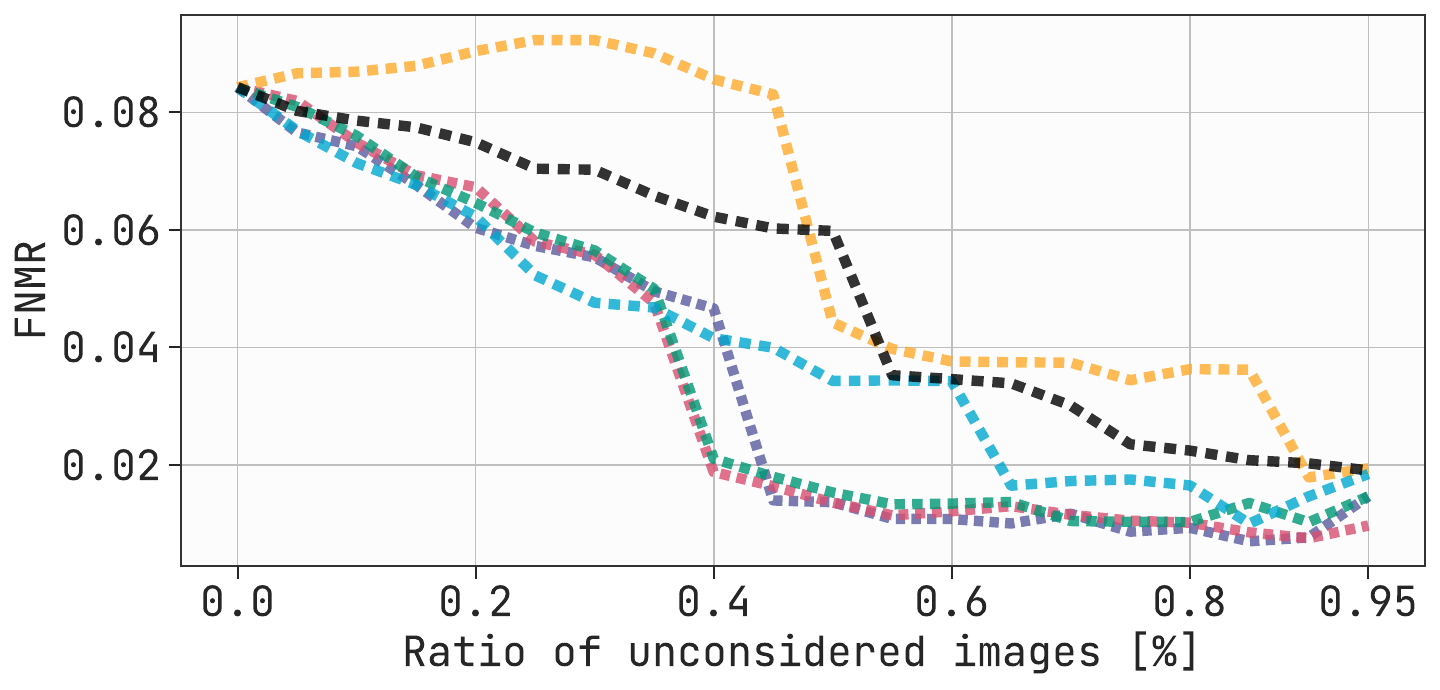}
		 \caption{MagFace \cite{meng_2021_magface} Model, CFP-FP \cite{cfp-fp} Dataset \\ \grafiqs with ResNet50, FMR$=1e-4$}
	\end{subfigure}
\\
	\begin{subfigure}[b]{0.48\textwidth}
		 \centering
		 \includegraphics[width=0.95\textwidth]{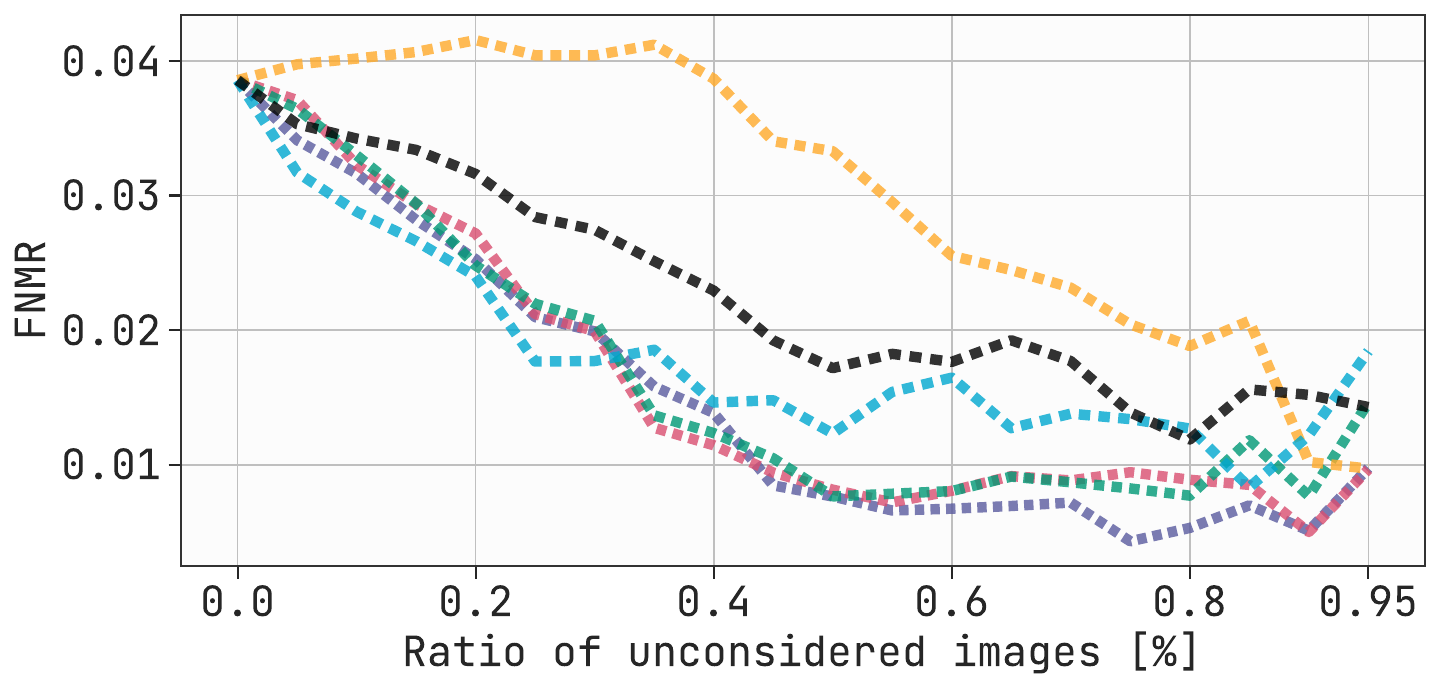}
		 \caption{CurricularFace \cite{curricularFace} Model, CFP-FP \cite{cfp-fp} Dataset \\ \grafiqs with ResNet50, FMR$=1e-3$}
	\end{subfigure}
\hfill
	\begin{subfigure}[b]{0.48\textwidth}
		 \centering
		 \includegraphics[width=0.95\textwidth]{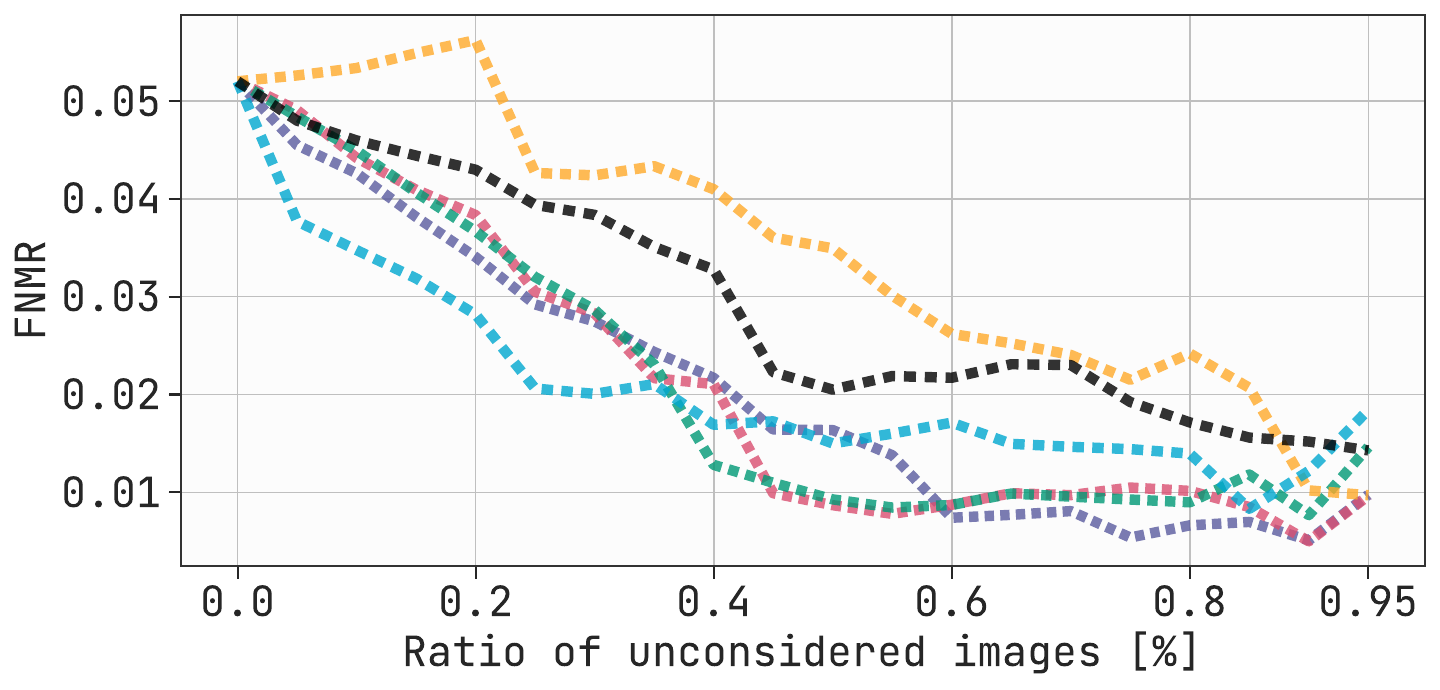}
		 \caption{CurricularFace \cite{curricularFace} Model, CFP-FP \cite{cfp-fp} Dataset \\ \grafiqs with ResNet50, FMR$=1e-4$}
	\end{subfigure}
\\
\caption{Error-versus-Discard Characteristic (EDC) curves for FNMR@FMR=$1e-3$ and FNMR@FMR=$1e-4$ of our proposed method using $\mathcal{L}_{\text{BNS}}$ as backpropagation loss and absolute sum as FIQ. The gradients at image level ($\phi=\mathcal{I}$), and block levels ($\phi=\text{B}1$ $-$ $\phi=\text{B}4$) are used to calculate FIQ. $\text{MSE}_{\text{BNS}}$ as FIQ is shown in black. Results shown on benchmark CFP-FP \cite{cfp-fp} using ArcFace, ElasticFace, MagFace, and, CurricularFace FR models. It is evident that the proposed \grafiqs method leads to lower verification error when images with lowest utility score estimated from gradient magnitudes are rejected. Furthermore, estimating FIQ by backpropagating $\mathcal{L}_{\text{BNS}}$ yields significantly better results than using $\text{MSE}_{\text{BNS}}$ directly.}
\vspace{-4mm}
\label{fig:iresnet50_supplementary_cfp_fp}
\end{figure*}

%% file: figures/fig_iresnet50_supplementary_lfw.tex
\begin{figure*}[h!]
\centering
	\begin{subfigure}[b]{0.9\textwidth}
		\centering
		\includegraphics[width=\textwidth]{figures/iresnet50_bn_overview/legend.pdf}
	\end{subfigure}
\\
	\begin{subfigure}[b]{0.48\textwidth}
		 \centering
		 \includegraphics[width=0.95\textwidth]{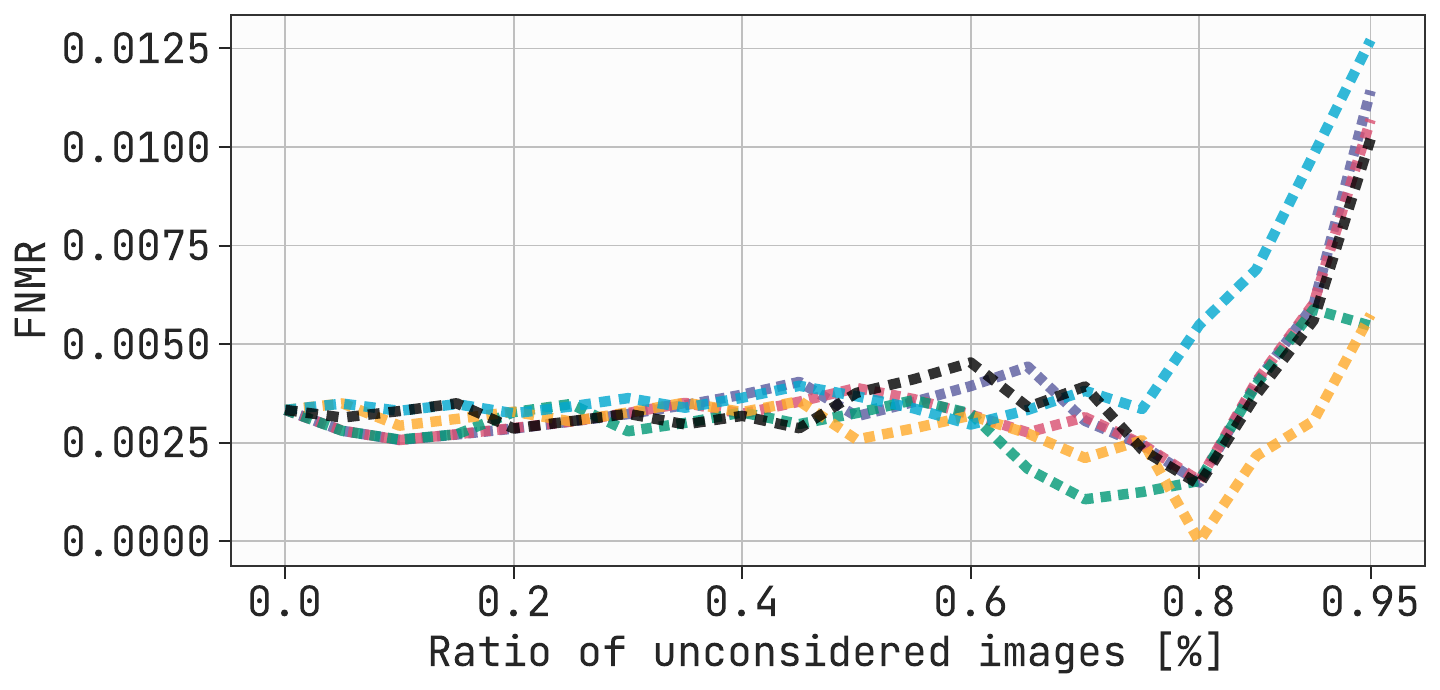}
		 \caption{ArcFace \cite{deng2019arcface} Model, LFW \cite{LFWTech} Dataset \\ \grafiqs with ResNet50, FMR$=1e-3$}
	\end{subfigure}
\hfill
	\begin{subfigure}[b]{0.48\textwidth}
		 \centering
		 \includegraphics[width=0.95\textwidth]{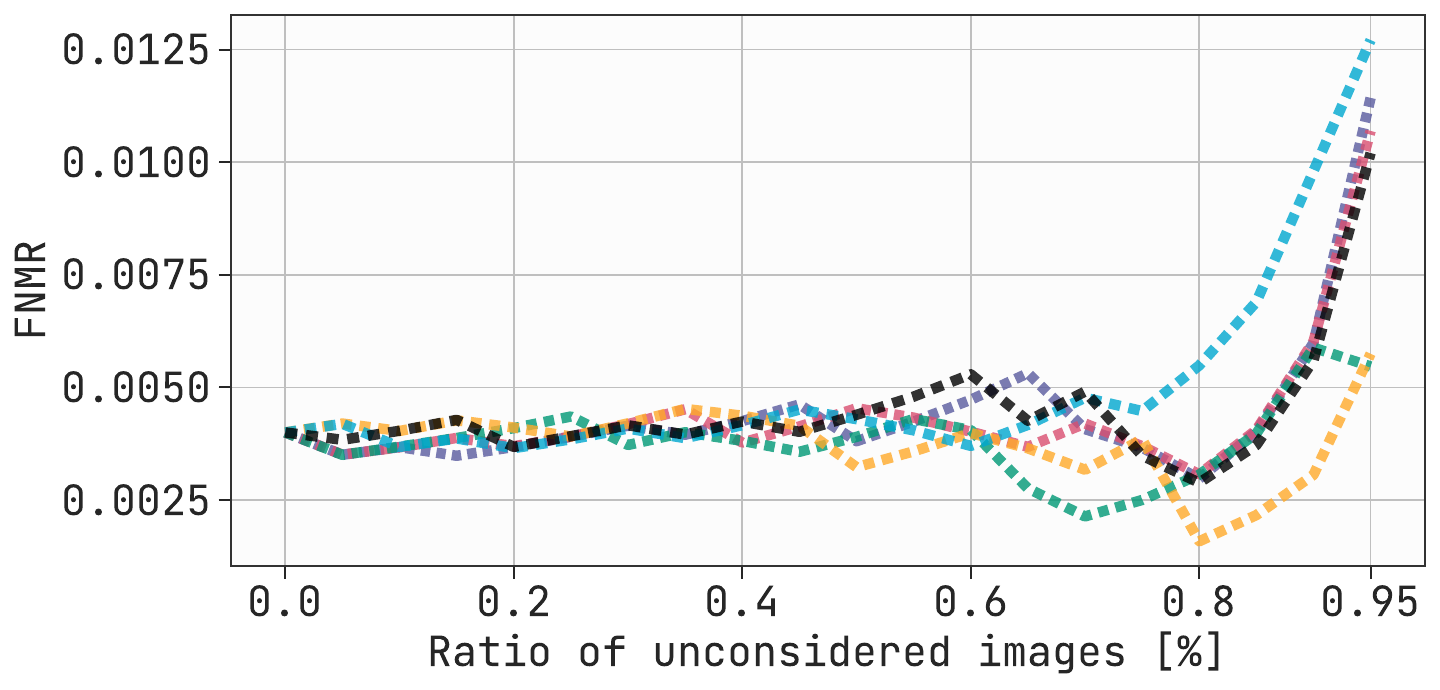}
		 \caption{ArcFace \cite{deng2019arcface} Model, LFW \cite{LFWTech} Dataset \\ \grafiqs with ResNet50, FMR$=1e-4$}
	\end{subfigure}
\\
	\begin{subfigure}[b]{0.48\textwidth}
		 \centering
		 \includegraphics[width=0.95\textwidth]{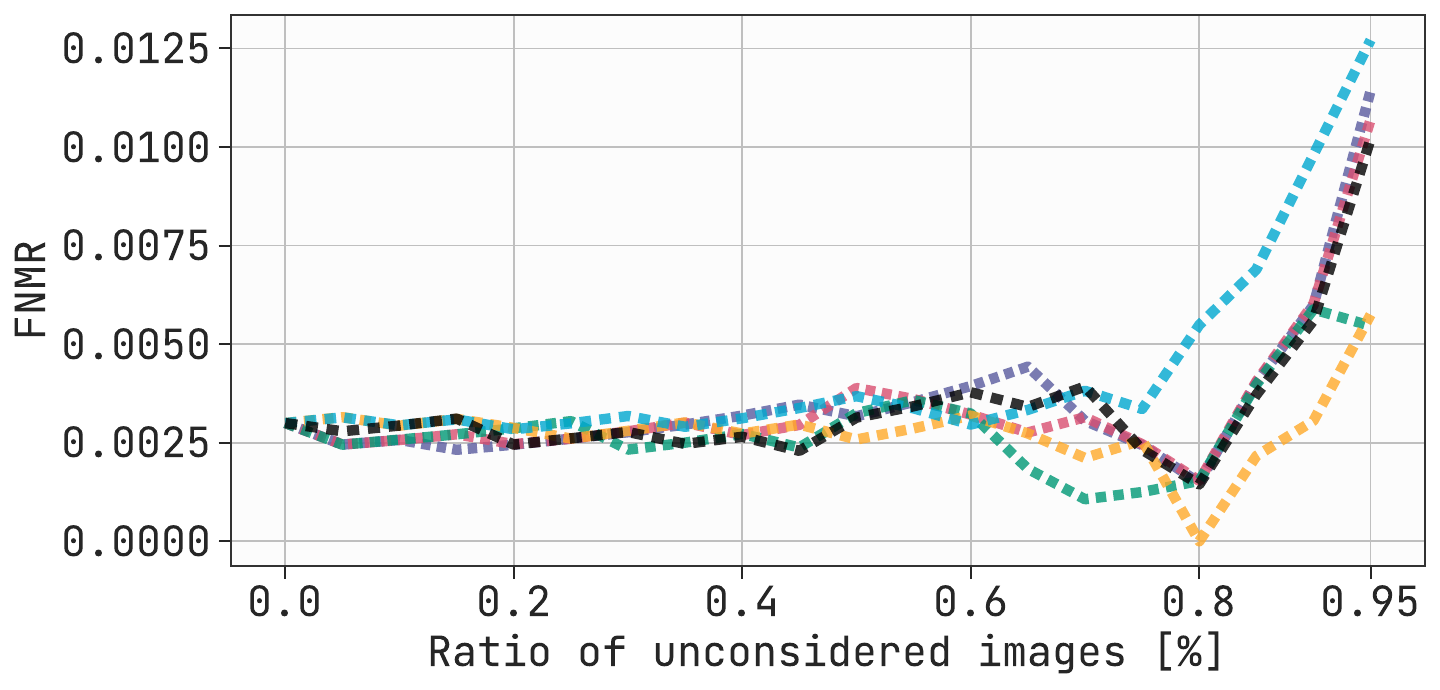}
		 \caption{ElasticFace \cite{elasticface} Model, LFW \cite{LFWTech} Dataset \\ \grafiqs with ResNet50, FMR$=1e-3$}
	\end{subfigure}
\hfill
	\begin{subfigure}[b]{0.48\textwidth}
		 \centering
		 \includegraphics[width=0.95\textwidth]{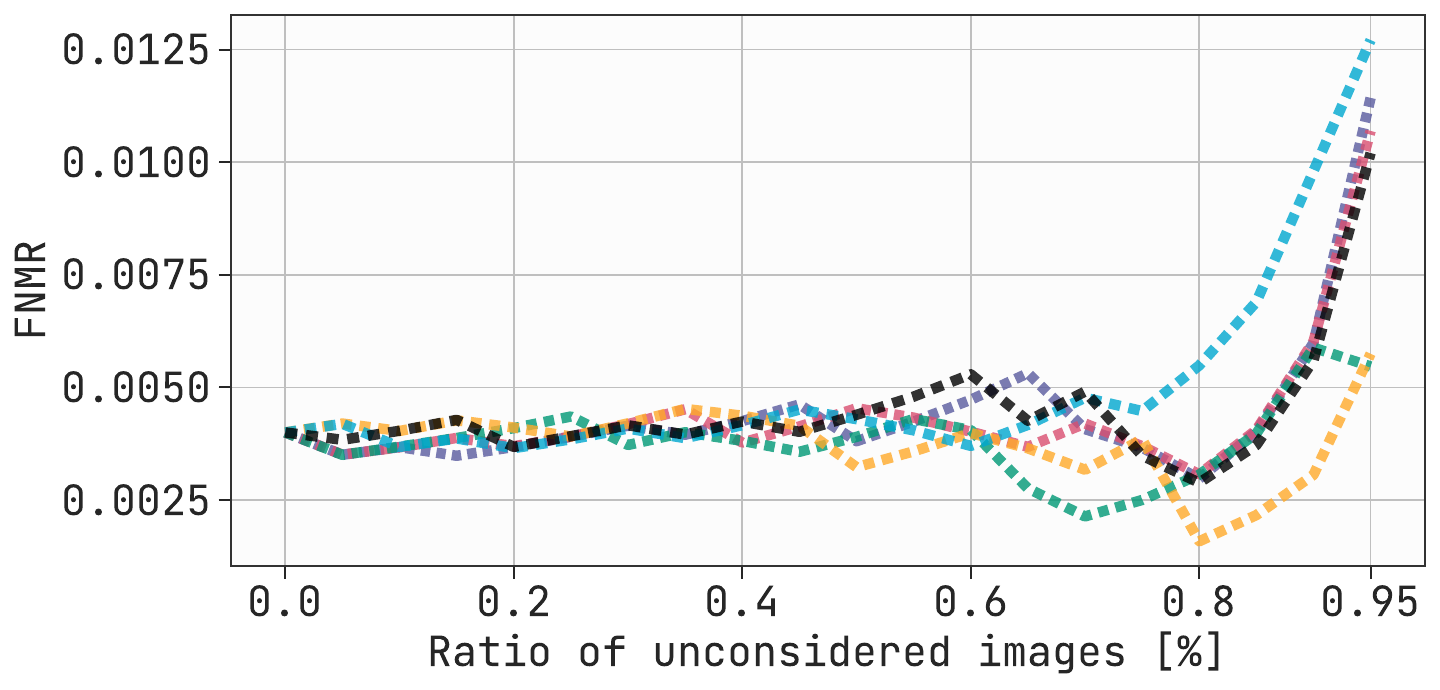}
		 \caption{ElasticFace \cite{elasticface} Model, LFW \cite{LFWTech} Dataset \\ \grafiqs with ResNet50, FMR$=1e-4$}
	\end{subfigure}
\\
	\begin{subfigure}[b]{0.48\textwidth}
		 \centering
		 \includegraphics[width=0.95\textwidth]{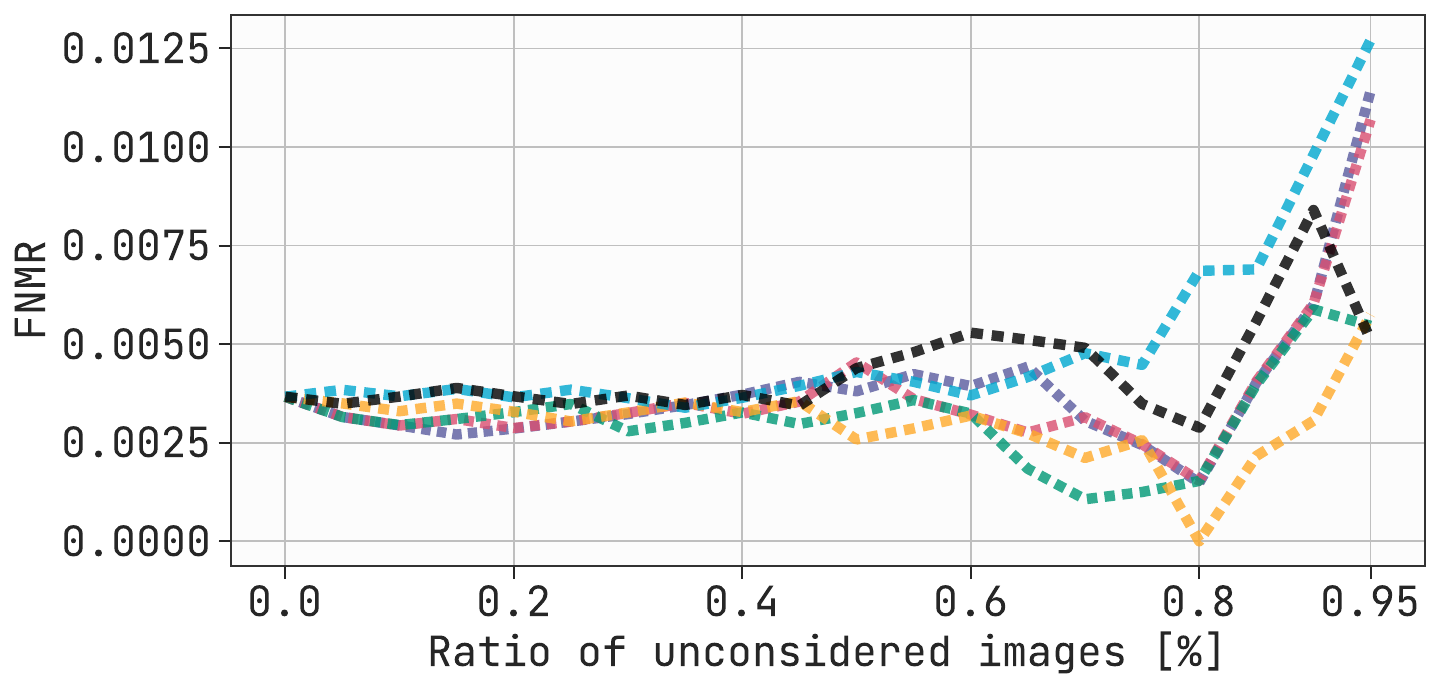}
		 \caption{MagFace \cite{meng_2021_magface} Model, LFW \cite{LFWTech} Dataset \\ \grafiqs with ResNet50, FMR$=1e-3$}
	\end{subfigure}
\hfill
	\begin{subfigure}[b]{0.48\textwidth}
		 \centering
		 \includegraphics[width=0.95\textwidth]{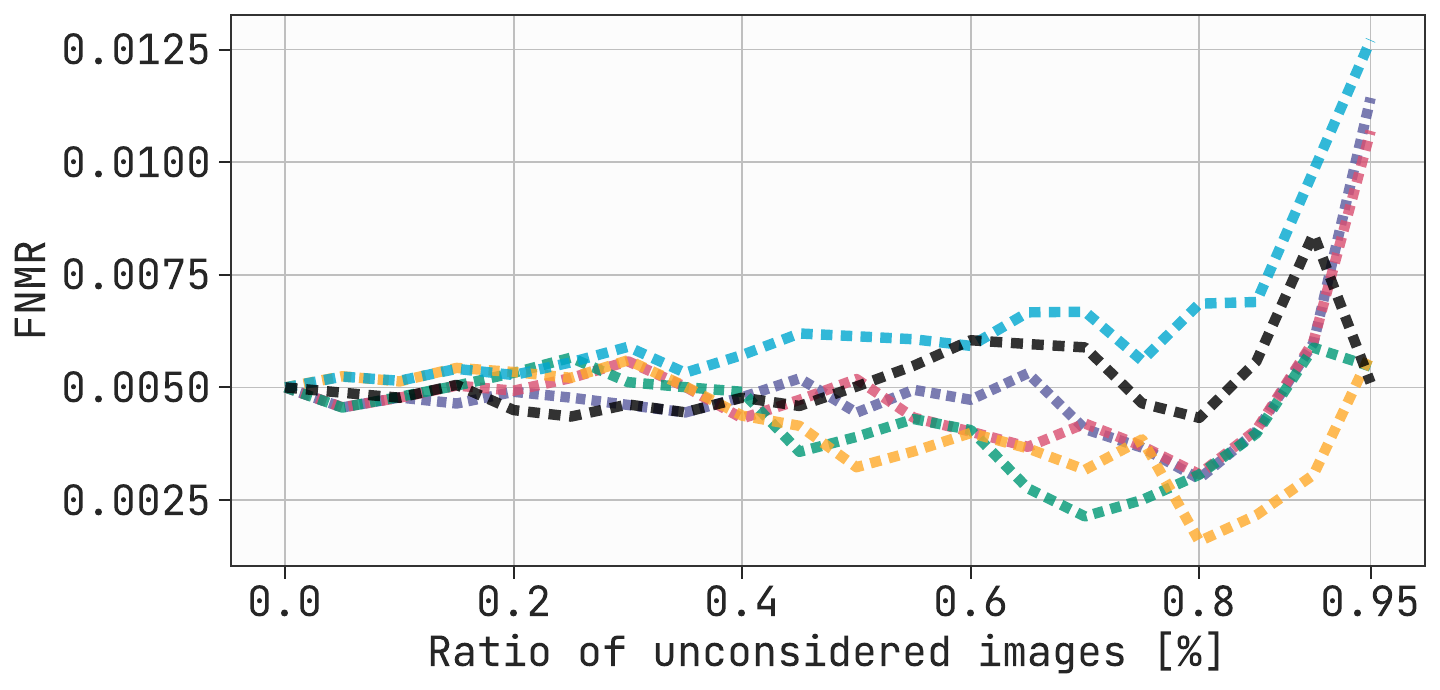}
		 \caption{MagFace \cite{meng_2021_magface} Model, LFW \cite{LFWTech} Dataset \\ \grafiqs with ResNet50, FMR$=1e-4$}
	\end{subfigure}
\\
	\begin{subfigure}[b]{0.48\textwidth}
		 \centering
		 \includegraphics[width=0.95\textwidth]{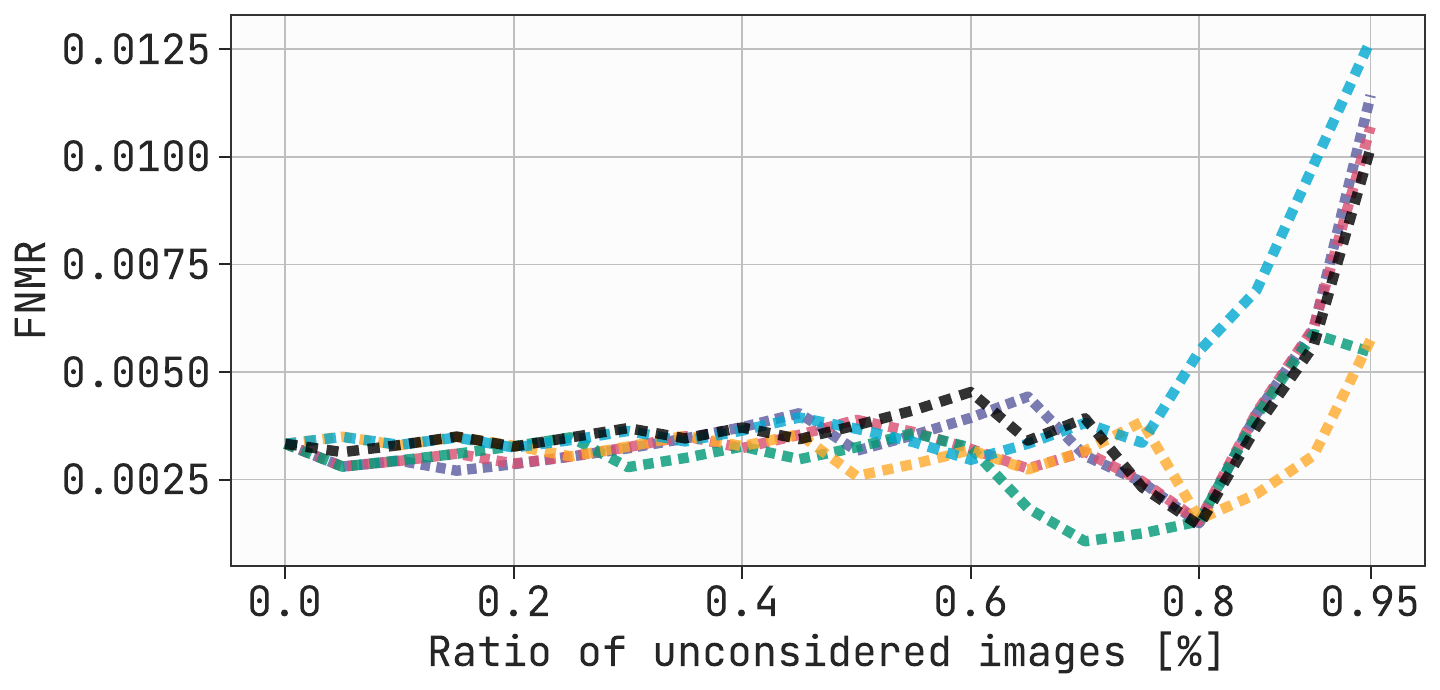}
		 \caption{CurricularFace \cite{curricularFace} Model, LFW \cite{LFWTech} Dataset \\ \grafiqs with ResNet50, FMR$=1e-3$}
	\end{subfigure}
\hfill
	\begin{subfigure}[b]{0.48\textwidth}
		 \centering
		 \includegraphics[width=0.95\textwidth]{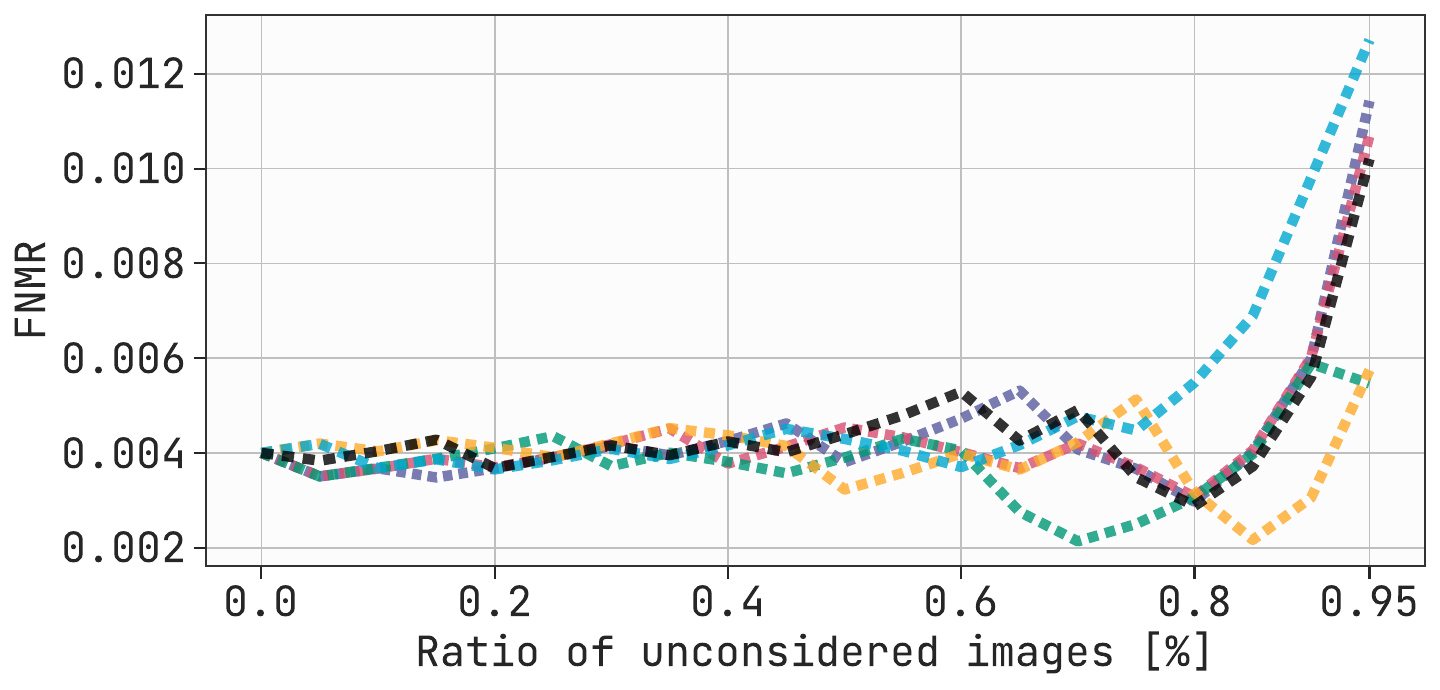}
		 \caption{CurricularFace \cite{curricularFace} Model, LFW \cite{LFWTech} Dataset \\ \grafiqs with ResNet50, FMR$=1e-4$}
	\end{subfigure}
\\
\caption{Error-versus-Discard Characteristic (EDC) curves for FNMR@FMR=$1e-3$ and FNMR@FMR=$1e-4$ of our proposed method using $\mathcal{L}_{\text{BNS}}$ as backpropagation loss and absolute sum as FIQ. The gradients at image level ($\phi=\mathcal{I}$), and block levels ($\phi=\text{B}1$ $-$ $\phi=\text{B}4$) are used to calculate FIQ. $\text{MSE}_{\text{BNS}}$ as FIQ is shown in black. Results shown on benchmark LFW \cite{LFWTech} using ArcFace, ElasticFace, MagFace, and, CurricularFace FR models. It is evident that the proposed \grafiqs method leads to lower verification error when images with lowest utility score estimated from gradient magnitudes are rejected. Furthermore, estimating FIQ by backpropagating $\mathcal{L}_{\text{BNS}}$ yields significantly better results than using $\text{MSE}_{\text{BNS}}$ directly.}
\vspace{-4mm}
\label{fig:iresnet50_supplementary_lfw}
\end{figure*}

%% file: figures/fig_iresnet50_supplementary_calfw.tex
\begin{figure*}[h!]
\centering
	\begin{subfigure}[b]{0.9\textwidth}
		\centering
		\includegraphics[width=\textwidth]{figures/iresnet50_bn_overview/legend.pdf}
	\end{subfigure}
\\
	\begin{subfigure}[b]{0.48\textwidth}
		 \centering
		 \includegraphics[width=0.95\textwidth]{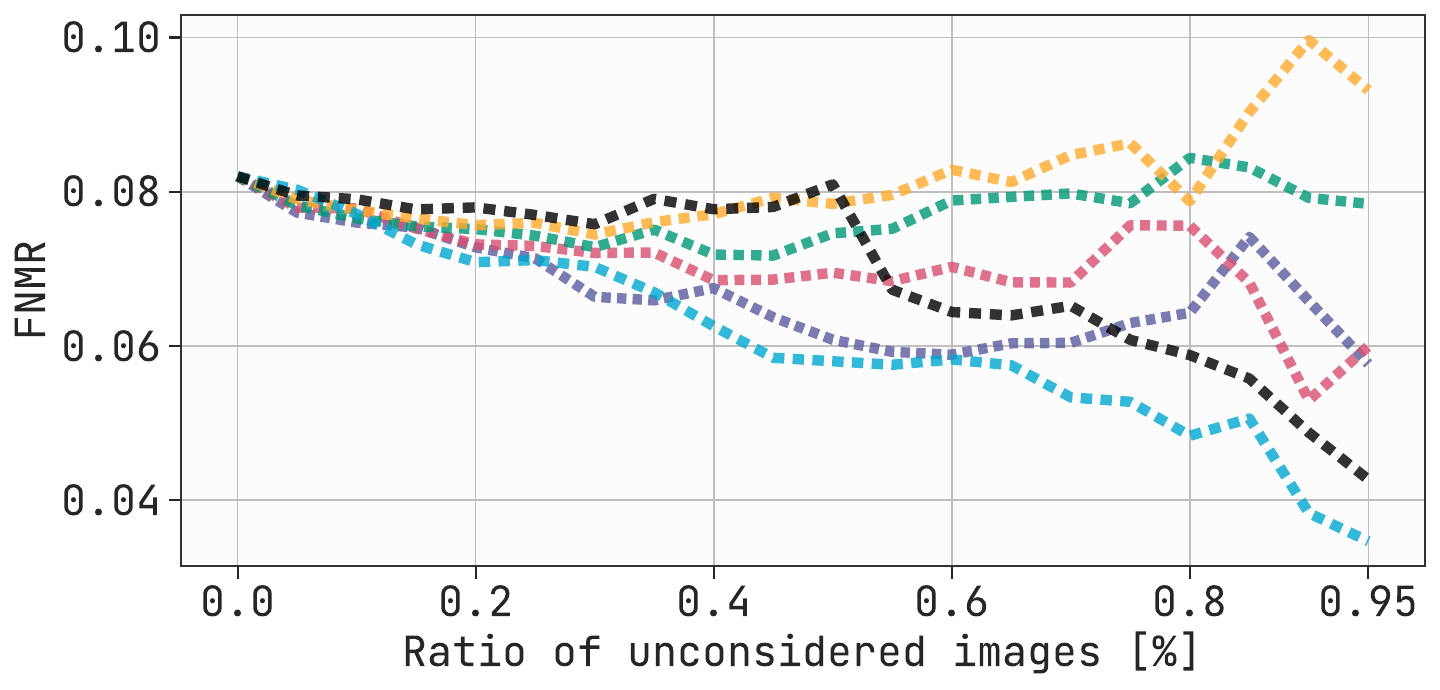}
		 \caption{ArcFace \cite{deng2019arcface} Model, CALFW \cite{CALFW} Dataset \\ \grafiqs with ResNet50, FMR$=1e-3$}
	\end{subfigure}
\hfill
	\begin{subfigure}[b]{0.48\textwidth}
		 \centering
		 \includegraphics[width=0.95\textwidth]{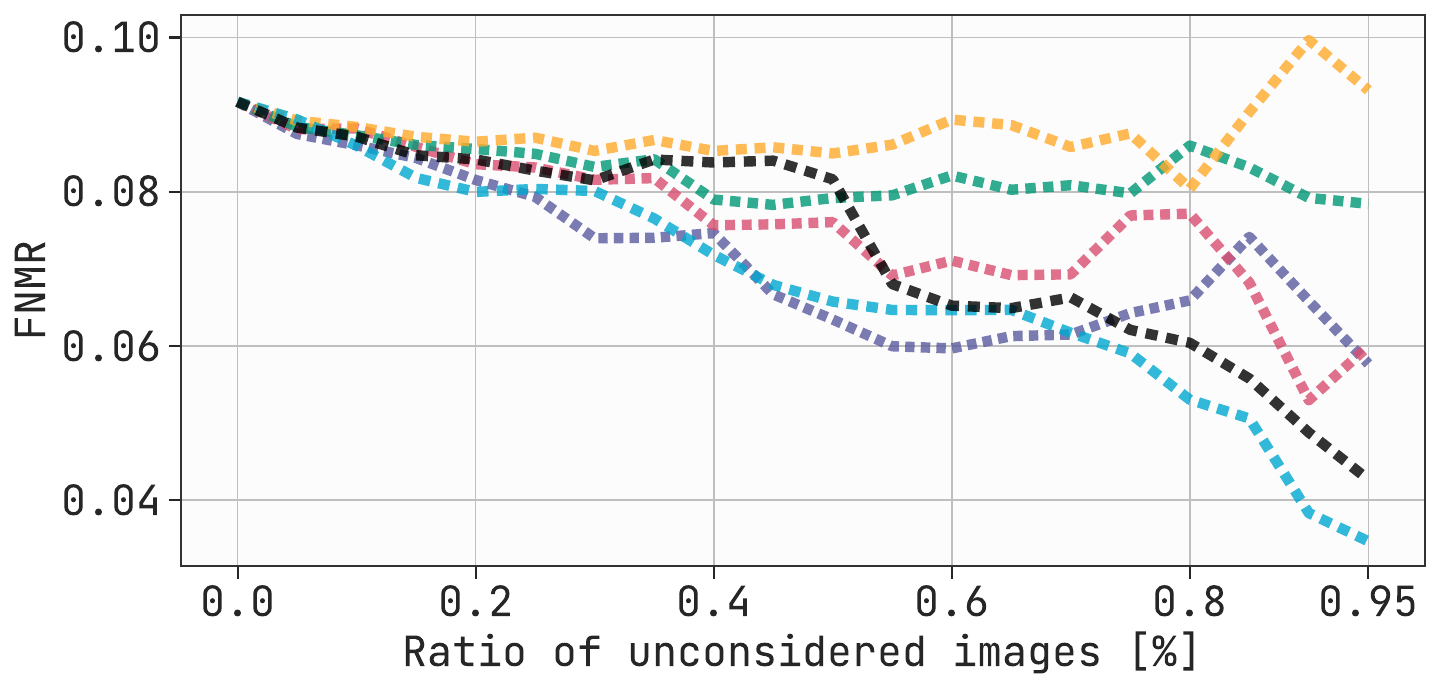}
		 \caption{ArcFace \cite{deng2019arcface} Model, CALFW \cite{CALFW} Dataset \\ \grafiqs with ResNet50, FMR$=1e-4$}
	\end{subfigure}
\\
	\begin{subfigure}[b]{0.48\textwidth}
		 \centering
		 \includegraphics[width=0.95\textwidth]{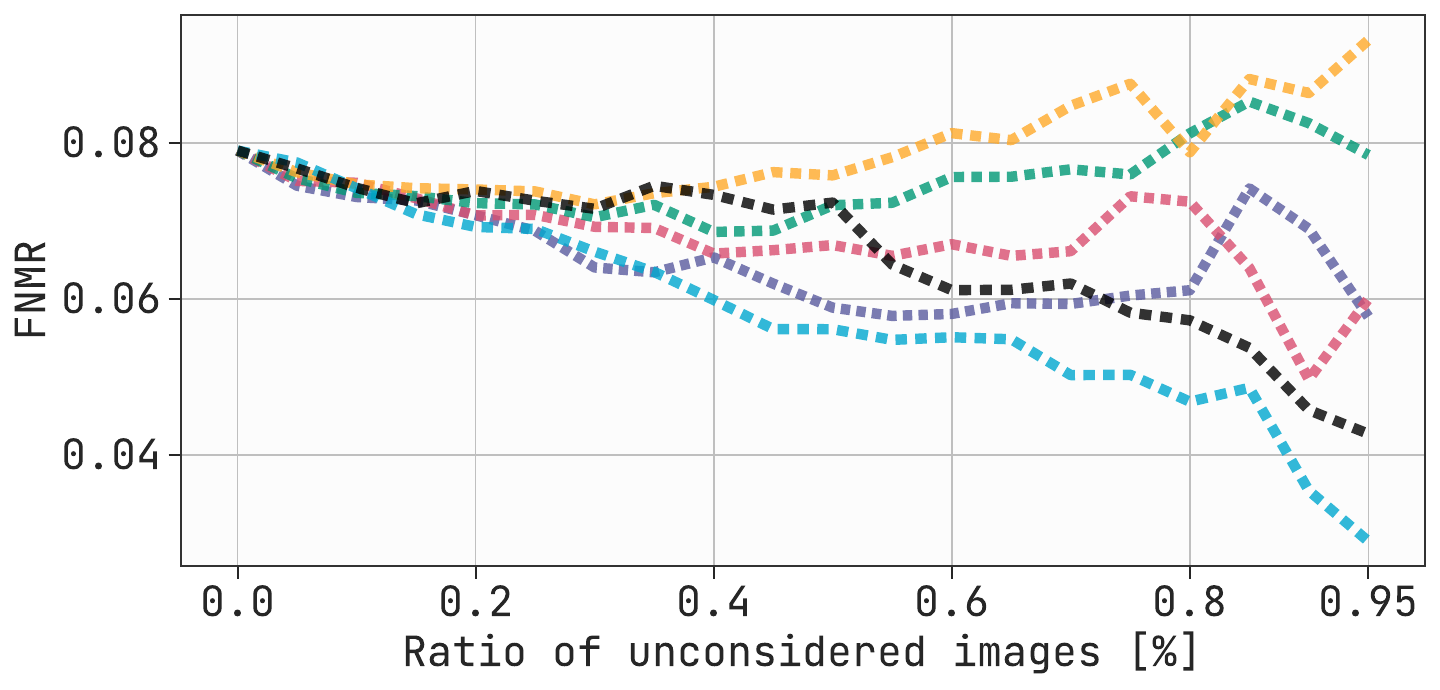}
		 \caption{ElasticFace \cite{elasticface} Model, CALFW \cite{CALFW} Dataset \\ \grafiqs with ResNet50, FMR$=1e-3$}
	\end{subfigure}
\hfill
	\begin{subfigure}[b]{0.48\textwidth}
		 \centering
		 \includegraphics[width=0.95\textwidth]{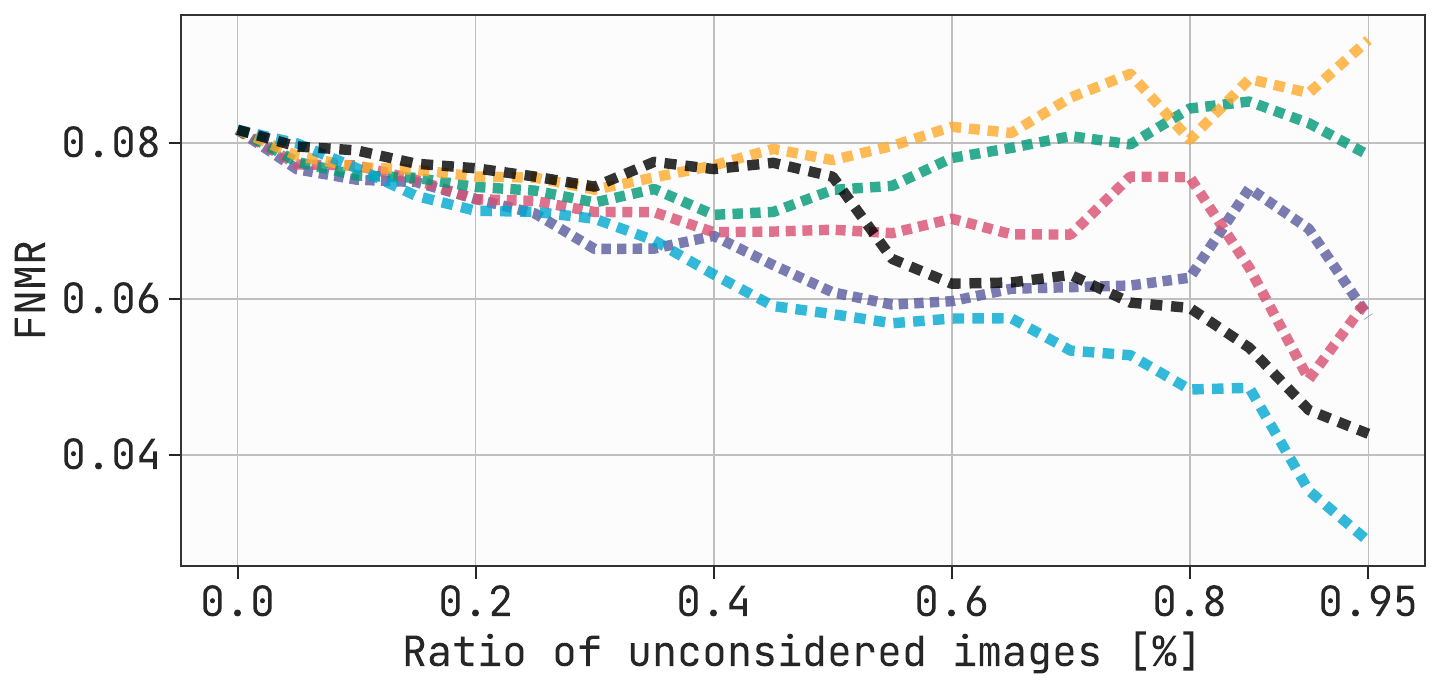}
		 \caption{ElasticFace \cite{elasticface} Model, CALFW \cite{CALFW} Dataset \\ \grafiqs with ResNet50, FMR$=1e-4$}
	\end{subfigure}
\\
	\begin{subfigure}[b]{0.48\textwidth}
		 \centering
		 \includegraphics[width=0.95\textwidth]{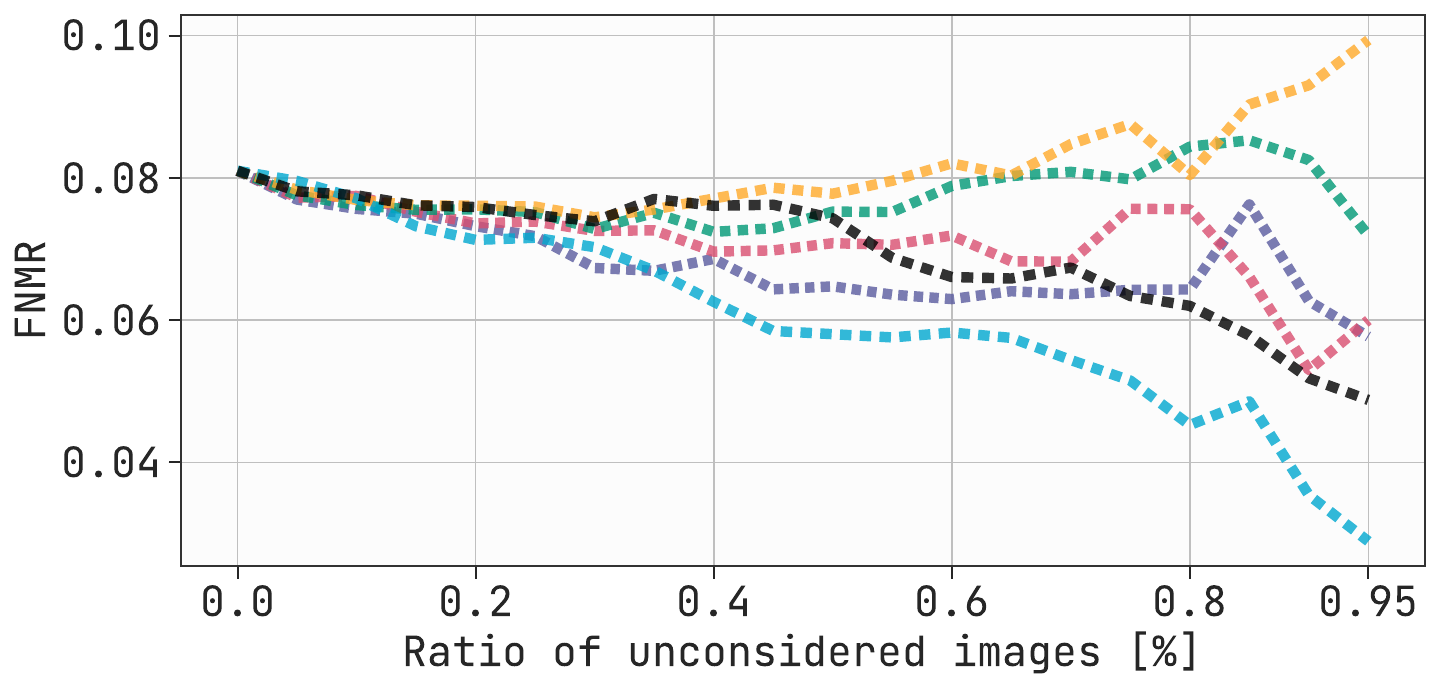}
		 \caption{MagFace \cite{meng_2021_magface} Model, CALFW \cite{CALFW} Dataset \\ \grafiqs with ResNet50, FMR$=1e-3$}
	\end{subfigure}
\hfill
	\begin{subfigure}[b]{0.48\textwidth}
		 \centering
		 \includegraphics[width=0.95\textwidth]{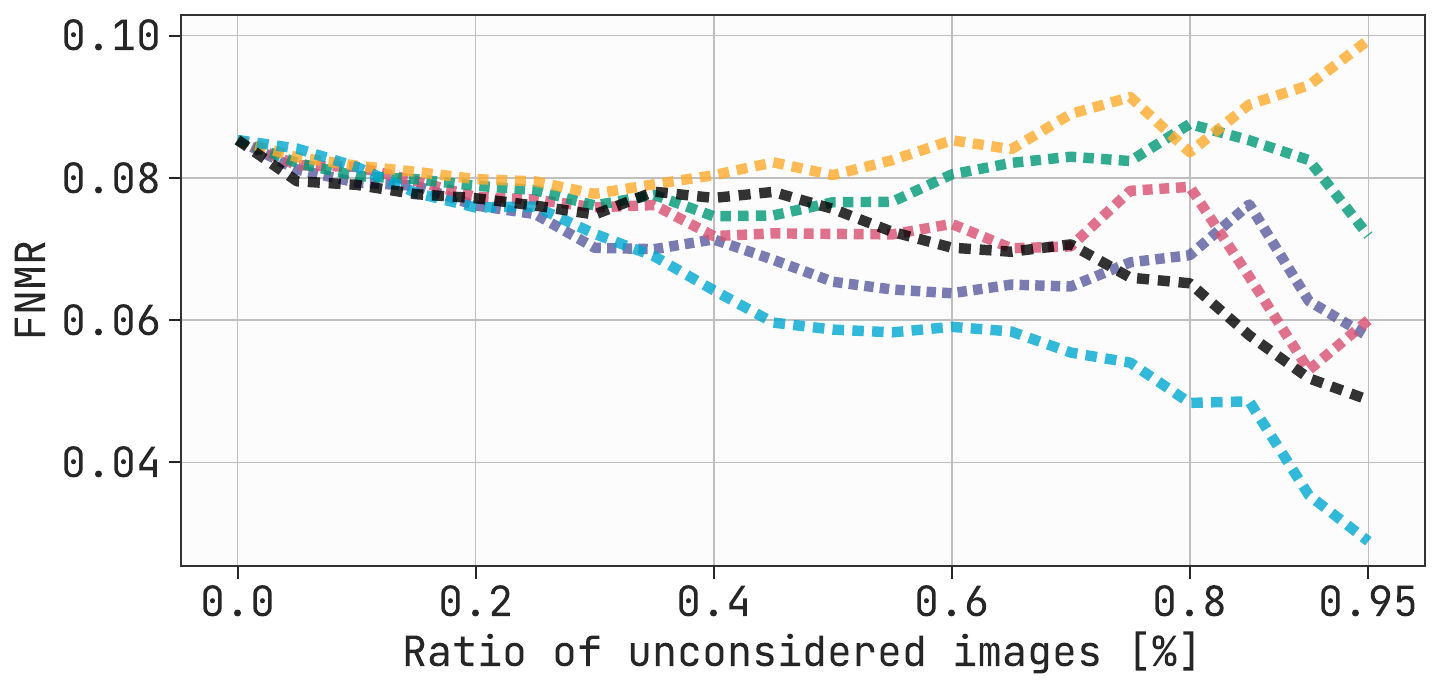}
		 \caption{MagFace \cite{meng_2021_magface} Model, CALFW \cite{CALFW} Dataset \\ \grafiqs with ResNet50, FMR$=1e-4$}
	\end{subfigure}
\\
	\begin{subfigure}[b]{0.48\textwidth}
		 \centering
		 \includegraphics[width=0.95\textwidth]{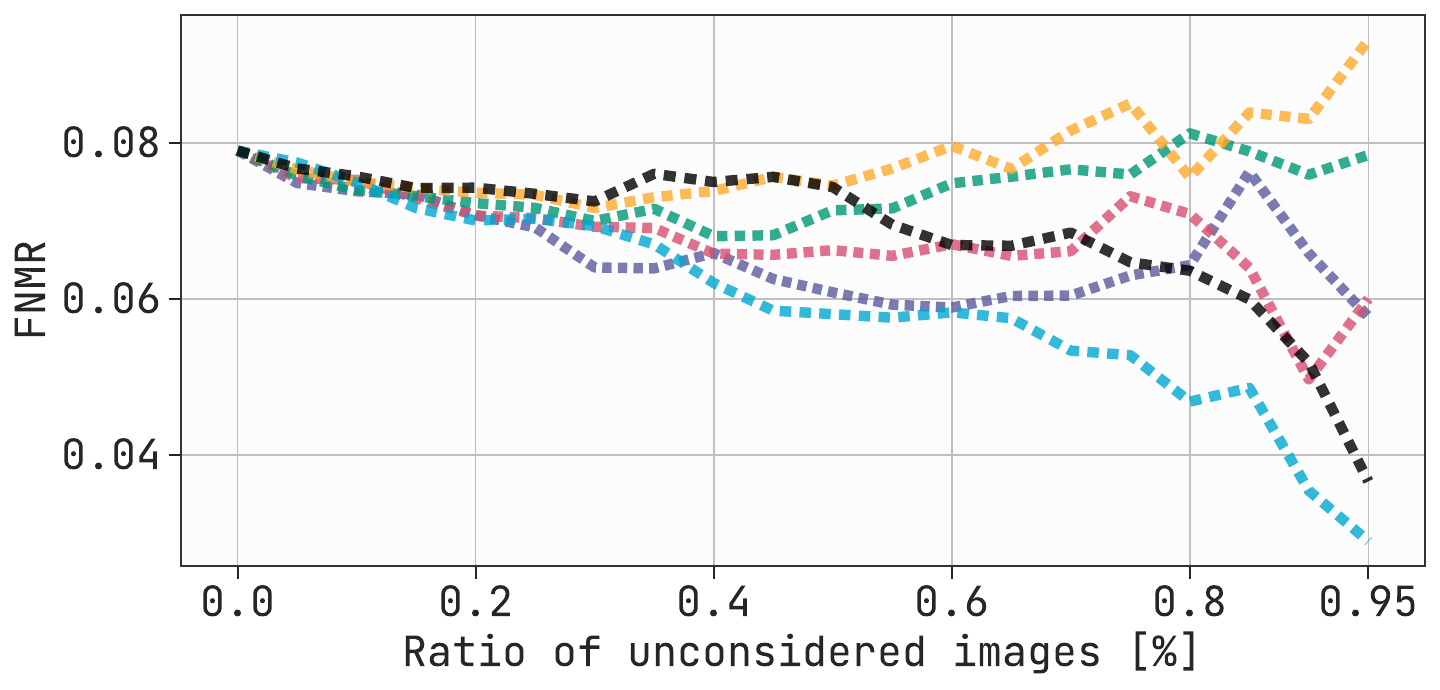}
		 \caption{CurricularFace \cite{curricularFace} Model, CALFW \cite{CALFW} Dataset \\ \grafiqs with ResNet50, FMR$=1e-3$}
	\end{subfigure}
\hfill
	\begin{subfigure}[b]{0.48\textwidth}
		 \centering
		 \includegraphics[width=0.95\textwidth]{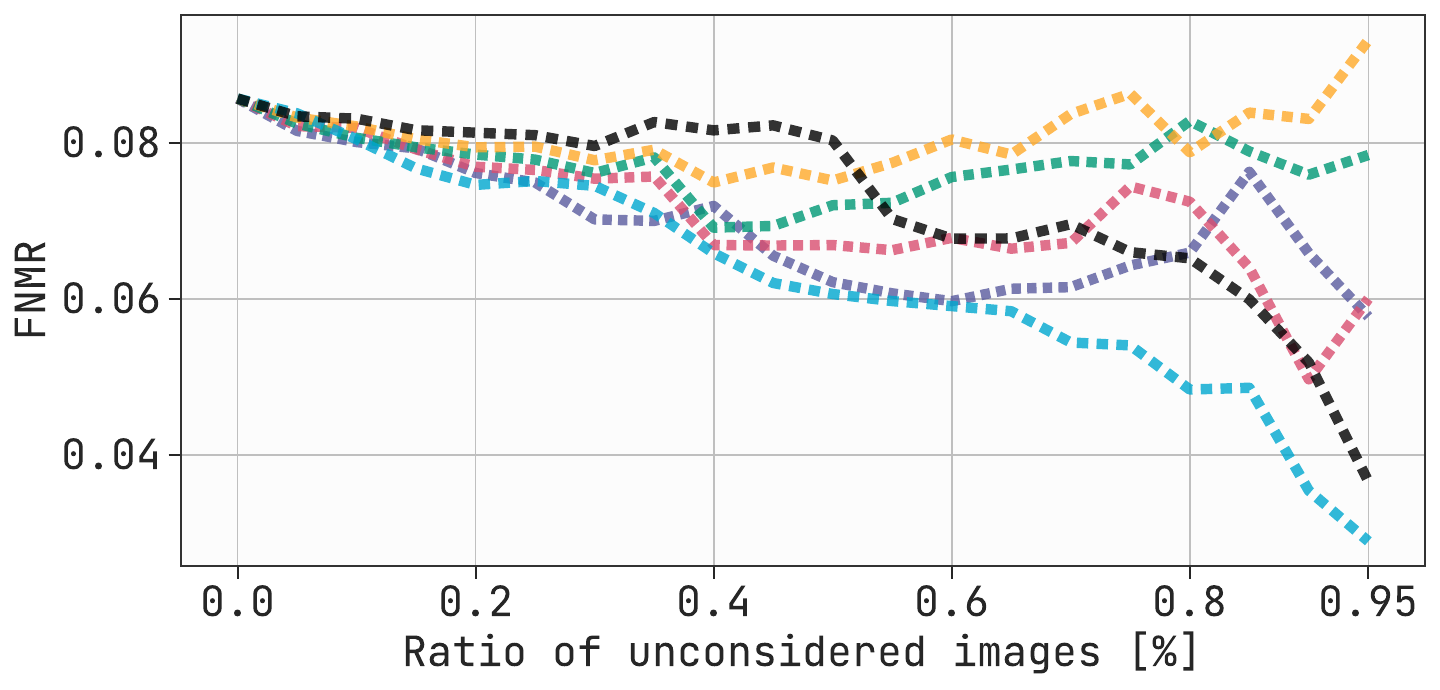}
		 \caption{CurricularFace \cite{curricularFace} Model, CALFW \cite{CALFW} Dataset \\ \grafiqs with ResNet50, FMR$=1e-4$}
	\end{subfigure}
\\
\caption{Error-versus-Discard Characteristic (EDC) curves for FNMR@FMR=$1e-3$ and FNMR@FMR=$1e-4$ of our proposed method using $\mathcal{L}_{\text{BNS}}$ as backpropagation loss and absolute sum as FIQ. The gradients at image level ($\phi=\mathcal{I}$), and block levels ($\phi=\text{B}1$ $-$ $\phi=\text{B}4$) are used to calculate FIQ. $\text{MSE}_{\text{BNS}}$ as FIQ is shown in black. Results shown on benchmark CALFW \cite{CALFW} using ArcFace, ElasticFace, MagFace, and, CurricularFace FR models. It is evident that the proposed \grafiqs method leads to lower verification error when images with lowest utility score estimated from gradient magnitudes are rejected. Furthermore, estimating FIQ by backpropagating $\mathcal{L}_{\text{BNS}}$ yields significantly better results than using $\text{MSE}_{\text{BNS}}$ directly.}
\vspace{-4mm}
\label{fig:iresnet50_supplementary_calfw}
\end{figure*}

%% file: figures/fig_iresnet50_supplementary_cplfw.tex
\begin{figure*}[h!]
\centering
	\begin{subfigure}[b]{0.9\textwidth}
		\centering
		\includegraphics[width=\textwidth]{figures/iresnet50_bn_overview/legend.pdf}
	\end{subfigure}
\\
	\begin{subfigure}[b]{0.48\textwidth}
		 \centering
		 \includegraphics[width=0.95\textwidth]{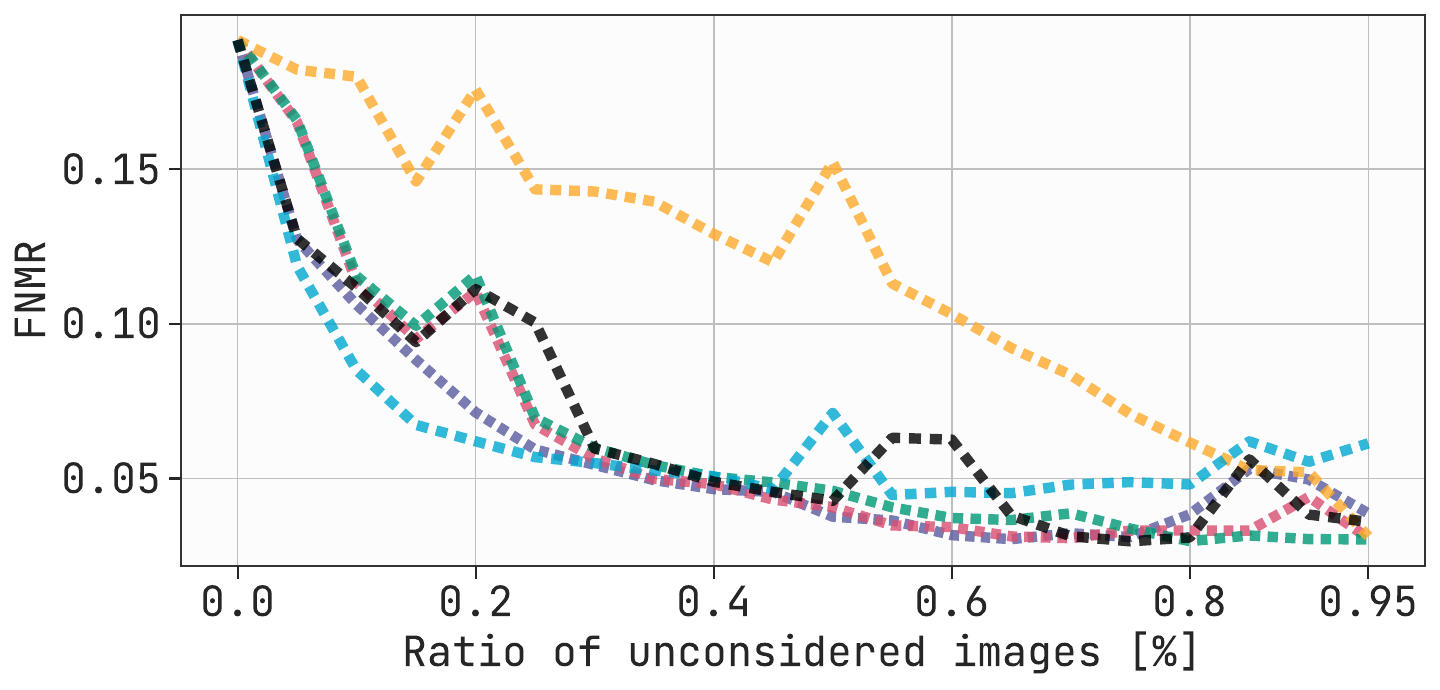}
		 \caption{ArcFace \cite{deng2019arcface} Model, CPLFW \cite{CPLFWTech} Dataset \\ \grafiqs with ResNet50, FMR$=1e-3$}
	\end{subfigure}
\hfill
	\begin{subfigure}[b]{0.48\textwidth}
		 \centering
		 \includegraphics[width=0.95\textwidth]{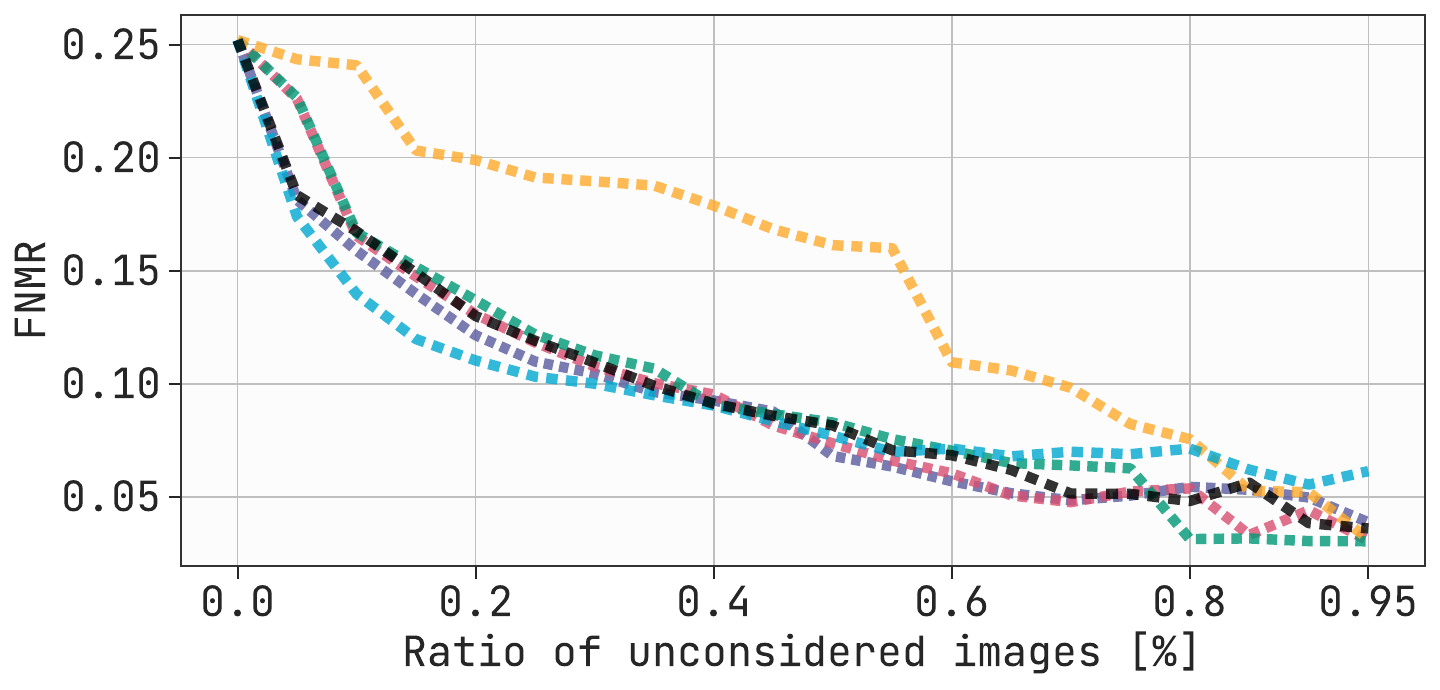}
		 \caption{ArcFace \cite{deng2019arcface} Model, CPLFW \cite{CPLFWTech} Dataset \\ \grafiqs with ResNet50, FMR$=1e-4$}
	\end{subfigure}
\\
	\begin{subfigure}[b]{0.48\textwidth}
		 \centering
		 \includegraphics[width=0.95\textwidth]{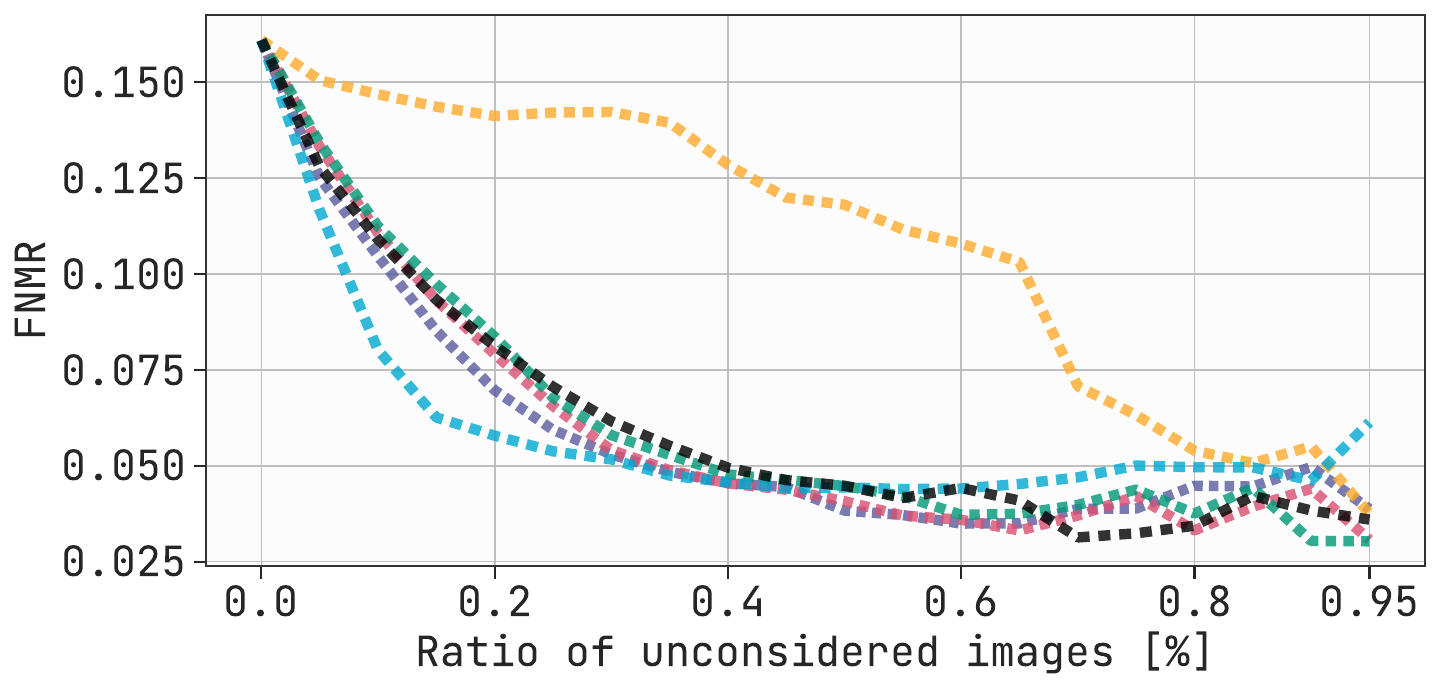}
		 \caption{ElasticFace \cite{elasticface} Model, CPLFW \cite{CPLFWTech} Dataset \\ \grafiqs with ResNet50, FMR$=1e-3$}
	\end{subfigure}
\hfill
	\begin{subfigure}[b]{0.48\textwidth}
		 \centering
		 \includegraphics[width=0.95\textwidth]{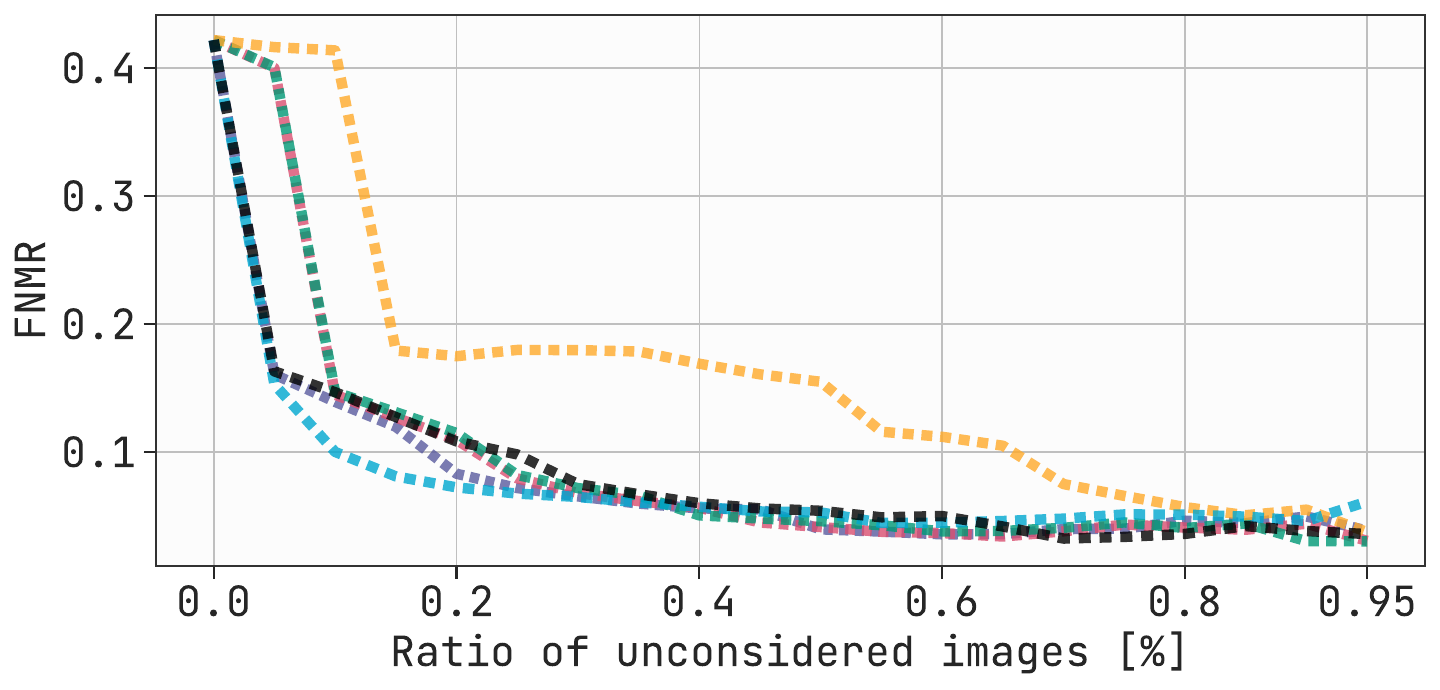}
		 \caption{ElasticFace \cite{elasticface} Model, CPLFW \cite{CPLFWTech} Dataset \\ \grafiqs with ResNet50, FMR$=1e-4$}
	\end{subfigure}
\\
	\begin{subfigure}[b]{0.48\textwidth}
		 \centering
		 \includegraphics[width=0.95\textwidth]{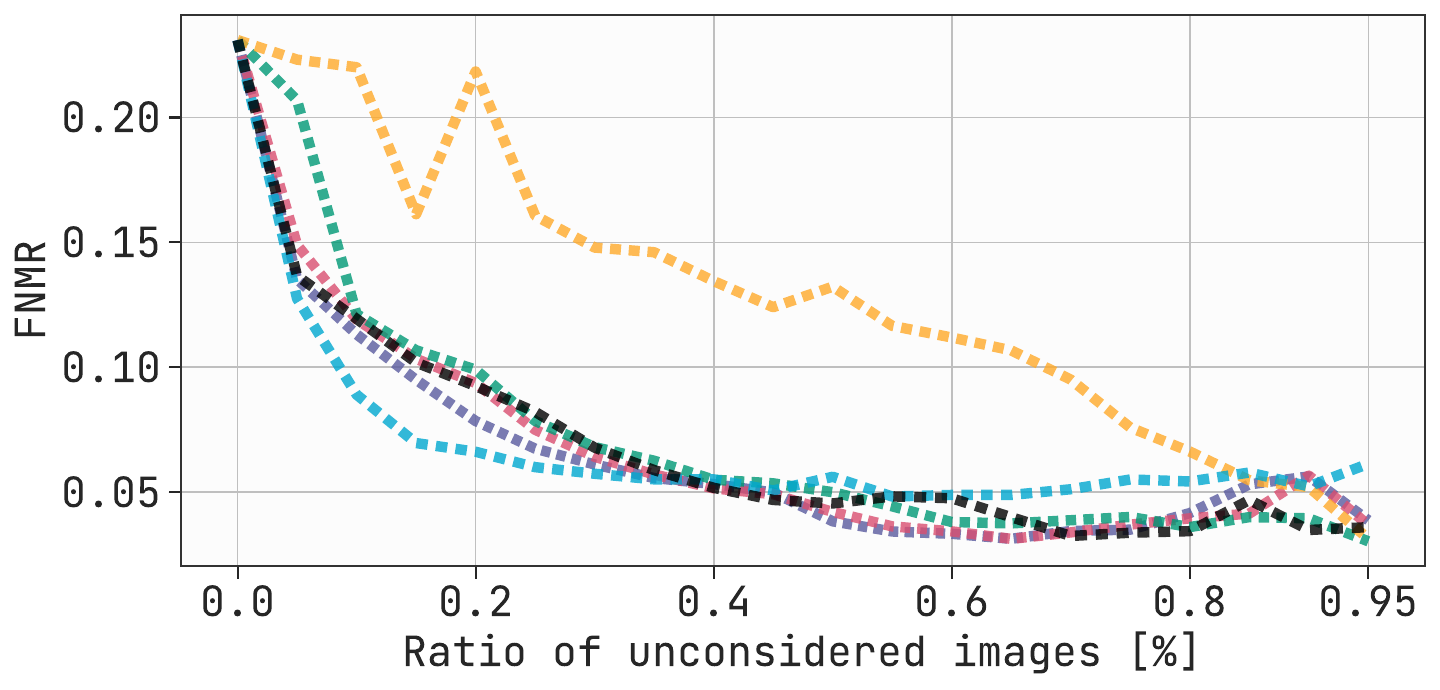}
		 \caption{MagFace \cite{meng_2021_magface} Model, CPLFW \cite{CPLFWTech} Dataset \\ \grafiqs with ResNet50, FMR$=1e-3$}
	\end{subfigure}
\hfill
	\begin{subfigure}[b]{0.48\textwidth}
		 \centering
		 \includegraphics[width=0.95\textwidth]{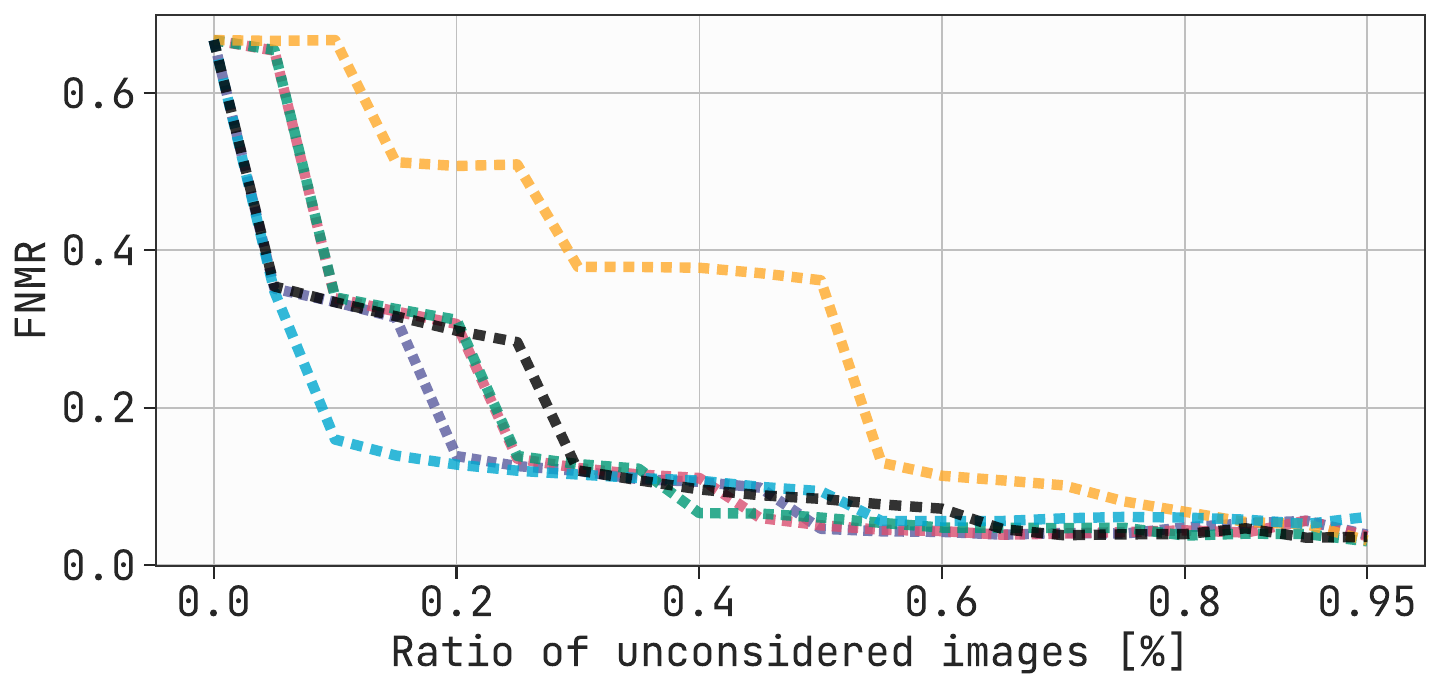}
		 \caption{MagFace \cite{meng_2021_magface} Model, CPLFW \cite{CPLFWTech} Dataset \\ \grafiqs with ResNet50, FMR$=1e-4$}
	\end{subfigure}
\\
	\begin{subfigure}[b]{0.48\textwidth}
		 \centering
		 \includegraphics[width=0.95\textwidth]{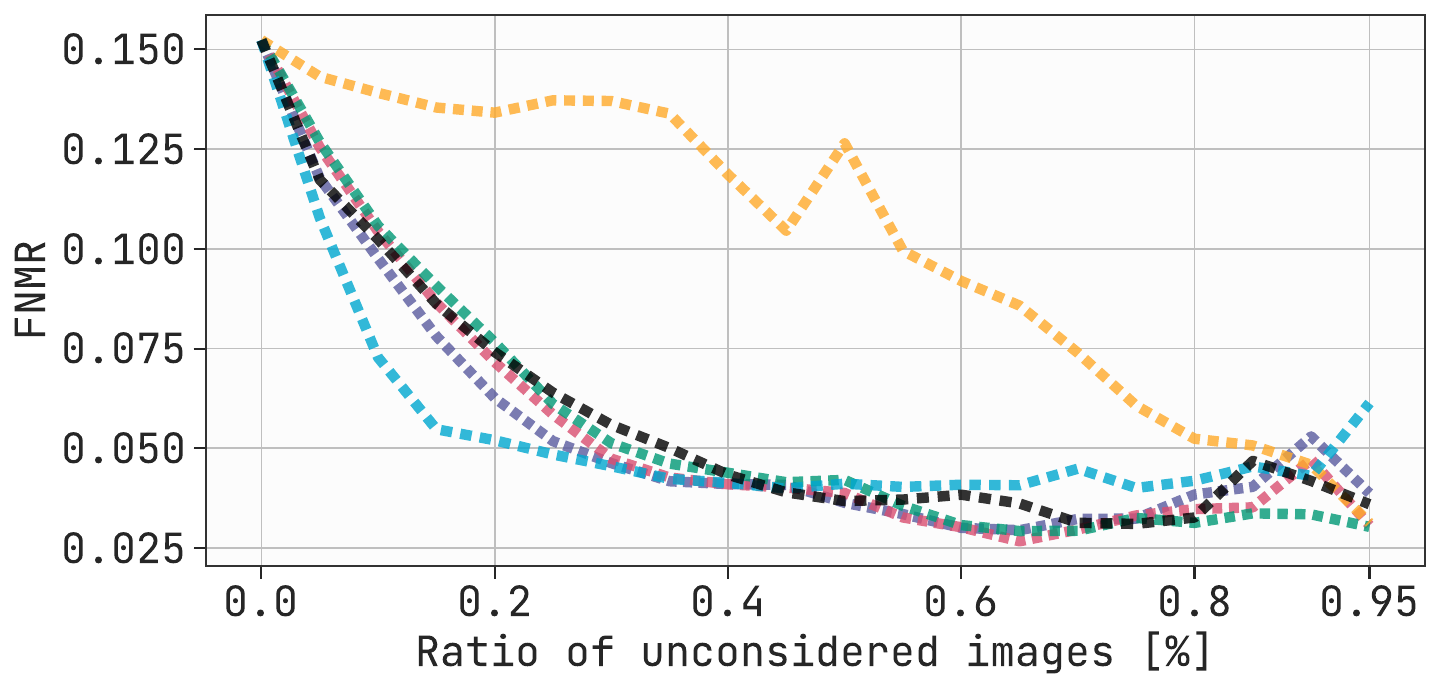}
		 \caption{CurricularFace \cite{curricularFace} Model, CPLFW \cite{CPLFWTech} Dataset \\ \grafiqs with ResNet50, FMR$=1e-3$}
	\end{subfigure}
\hfill
	\begin{subfigure}[b]{0.48\textwidth}
		 \centering
		 \includegraphics[width=0.95\textwidth]{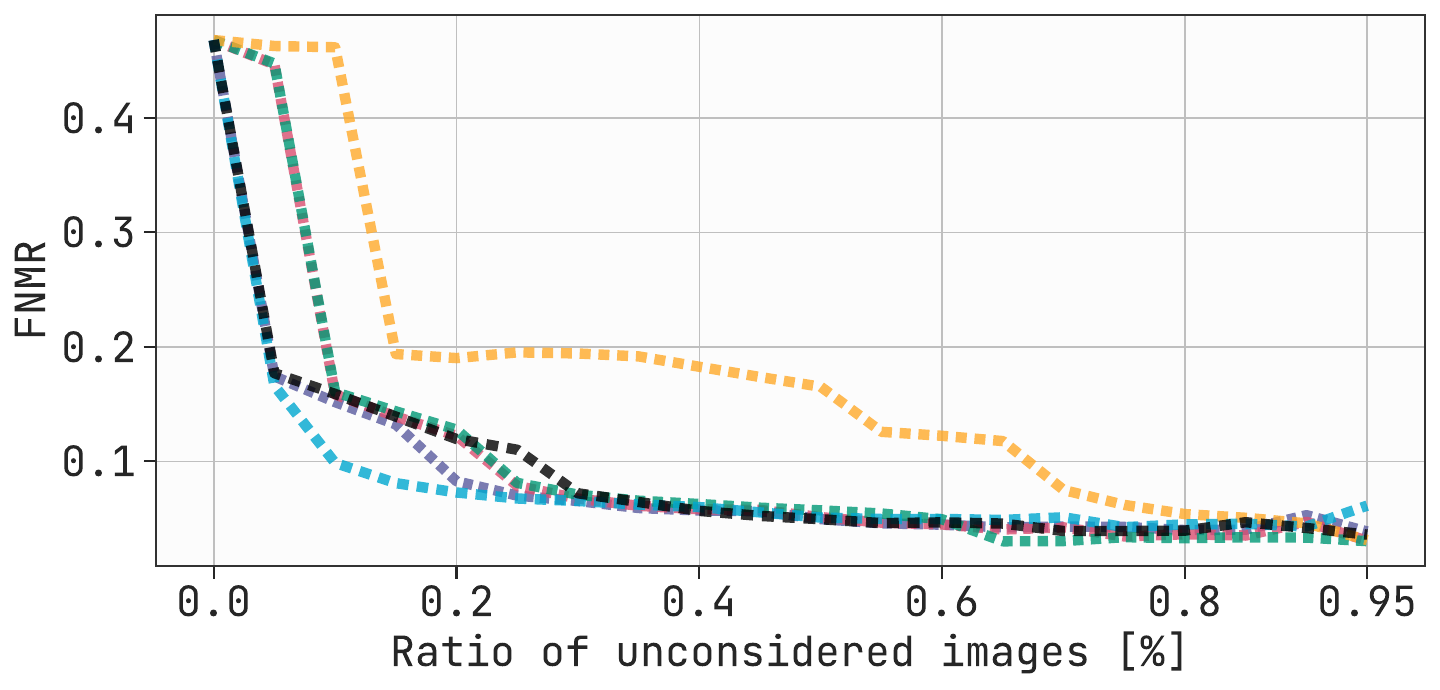}
		 \caption{CurricularFace \cite{curricularFace} Model, CPLFW \cite{CPLFWTech} Dataset \\ \grafiqs with ResNet50, FMR$=1e-4$}
	\end{subfigure}
\\
\caption{Error-versus-Discard Characteristic (EDC) curves for FNMR@FMR=$1e-3$ and FNMR@FMR=$1e-4$ of our proposed method using $\mathcal{L}_{\text{BNS}}$ as backpropagation loss and absolute sum as FIQ. The gradients at image level ($\phi=\mathcal{I}$), and block levels ($\phi=\text{B}1$ $-$ $\phi=\text{B}4$) are used to calculate FIQ. $\text{MSE}_{\text{BNS}}$ as FIQ is shown in black. Results shown on benchmark CPLFW \cite{CPLFWTech} using ArcFace, ElasticFace, MagFace, and, CurricularFace FR models. It is evident that the proposed \grafiqs method leads to lower verification error when images with lowest utility score estimated from gradient magnitudes are rejected. Furthermore, estimating FIQ by backpropagating $\mathcal{L}_{\text{BNS}}$ yields significantly better results than using $\text{MSE}_{\text{BNS}}$ directly.}
\vspace{-4mm}
\label{fig:iresnet50_supplementary_cplfw}
\end{figure*}

%% file: figures/fig_iresnet50_supplementary_XQLFW.tex
\begin{figure*}[h!]
\centering
	\begin{subfigure}[b]{0.9\textwidth}
		\centering
		\includegraphics[width=\textwidth]{figures/iresnet50_bn_overview/legend.pdf}
	\end{subfigure}
\\
	\begin{subfigure}[b]{0.48\textwidth}
		 \centering
		 \includegraphics[width=0.95\textwidth]{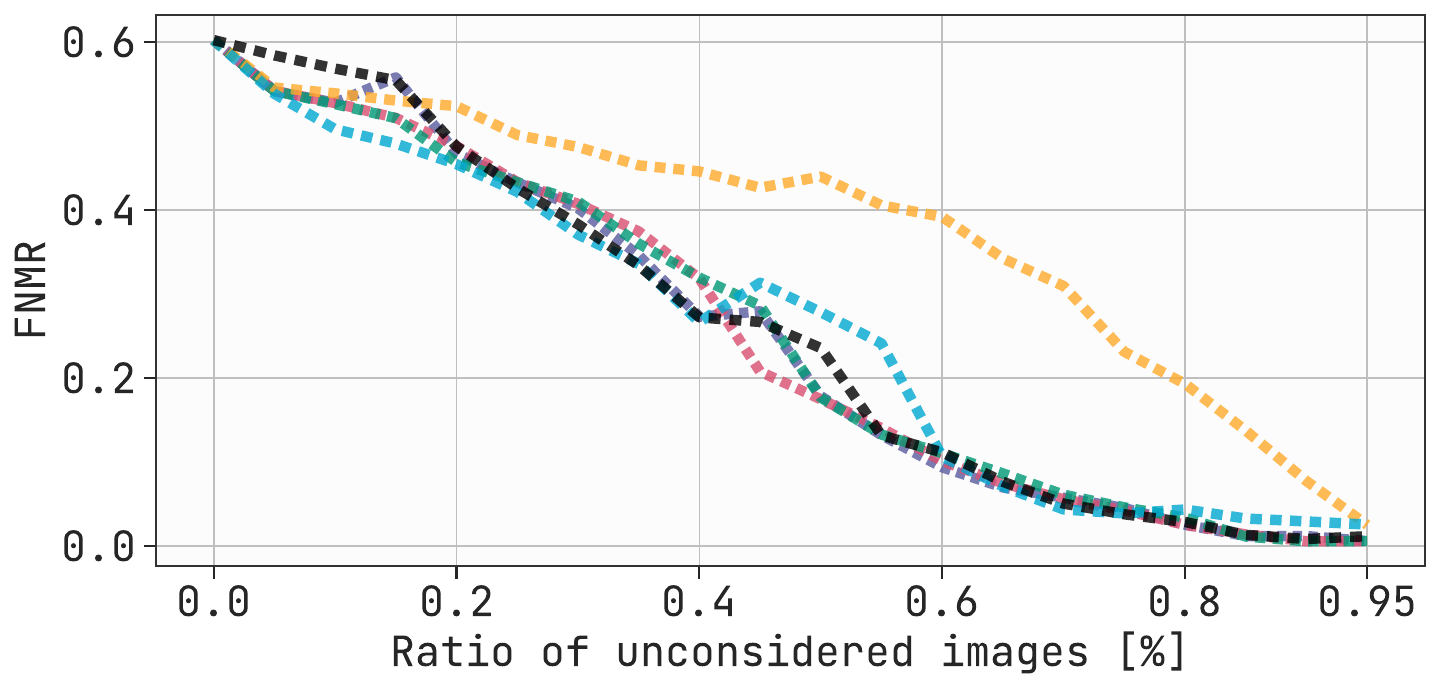}
		 \caption{ArcFace \cite{deng2019arcface} Model, XQLFW \cite{XQLFW} Dataset \\ \grafiqs with ResNet50, FMR$=1e-3$}
	\end{subfigure}
\hfill
	\begin{subfigure}[b]{0.48\textwidth}
		 \centering
		 \includegraphics[width=0.95\textwidth]{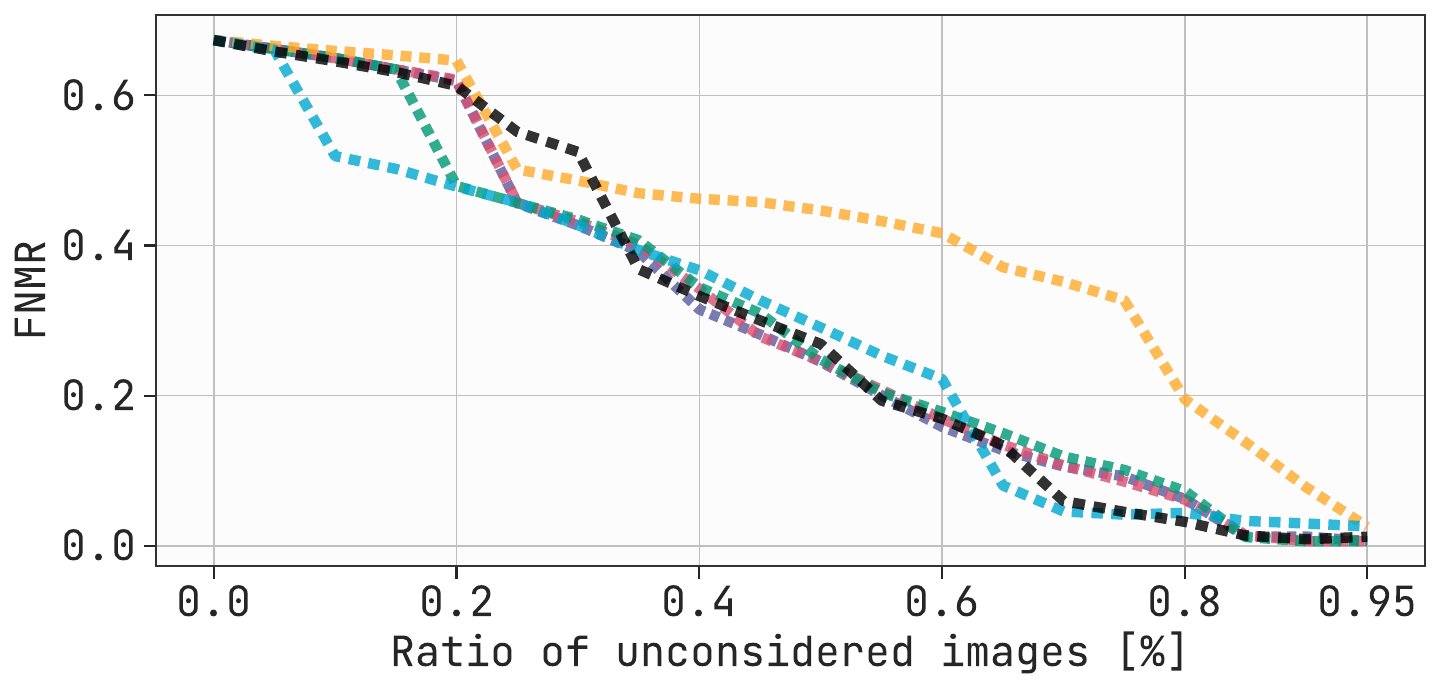}
		 \caption{ArcFace \cite{deng2019arcface} Model, XQLFW \cite{XQLFW} Dataset \\ \grafiqs with ResNet50, FMR$=1e-4$}
	\end{subfigure}
\\
	\begin{subfigure}[b]{0.48\textwidth}
		 \centering
		 \includegraphics[width=0.95\textwidth]{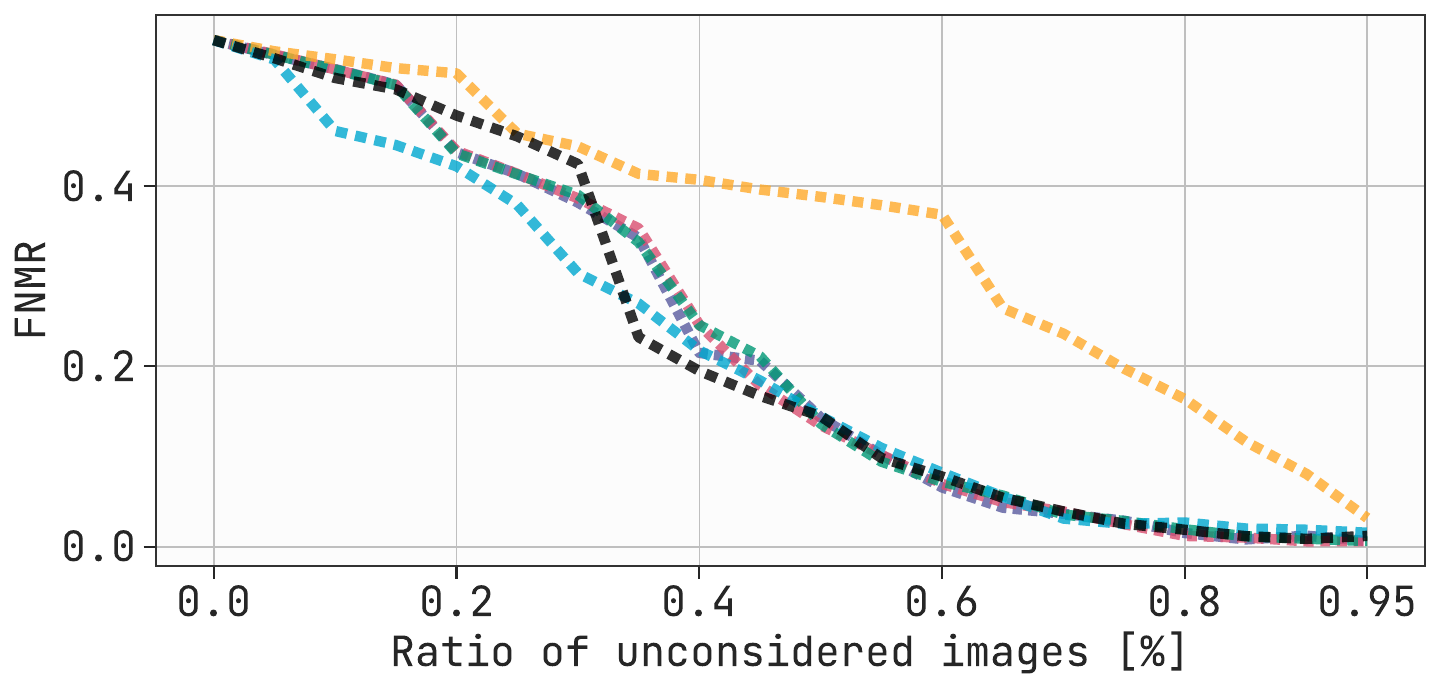}
		 \caption{ElasticFace \cite{elasticface} Model, XQLFW \cite{XQLFW} Dataset \\ \grafiqs with ResNet50, FMR$=1e-3$}
	\end{subfigure}
\hfill
	\begin{subfigure}[b]{0.48\textwidth}
		 \centering
		 \includegraphics[width=0.95\textwidth]{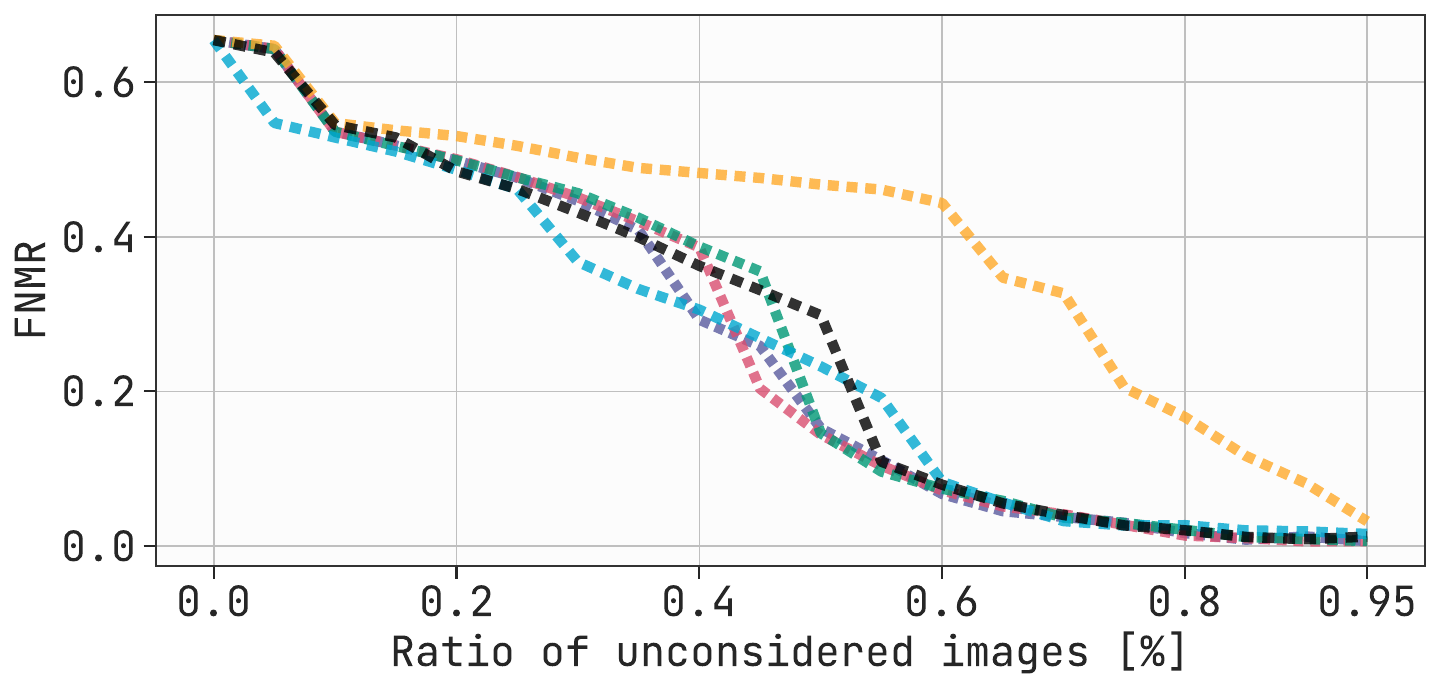}
		 \caption{ElasticFace \cite{elasticface} Model, XQLFW \cite{XQLFW} Dataset \\ \grafiqs with ResNet50, FMR$=1e-4$}
	\end{subfigure}
\\
	\begin{subfigure}[b]{0.48\textwidth}
		 \centering
		 \includegraphics[width=0.95\textwidth]{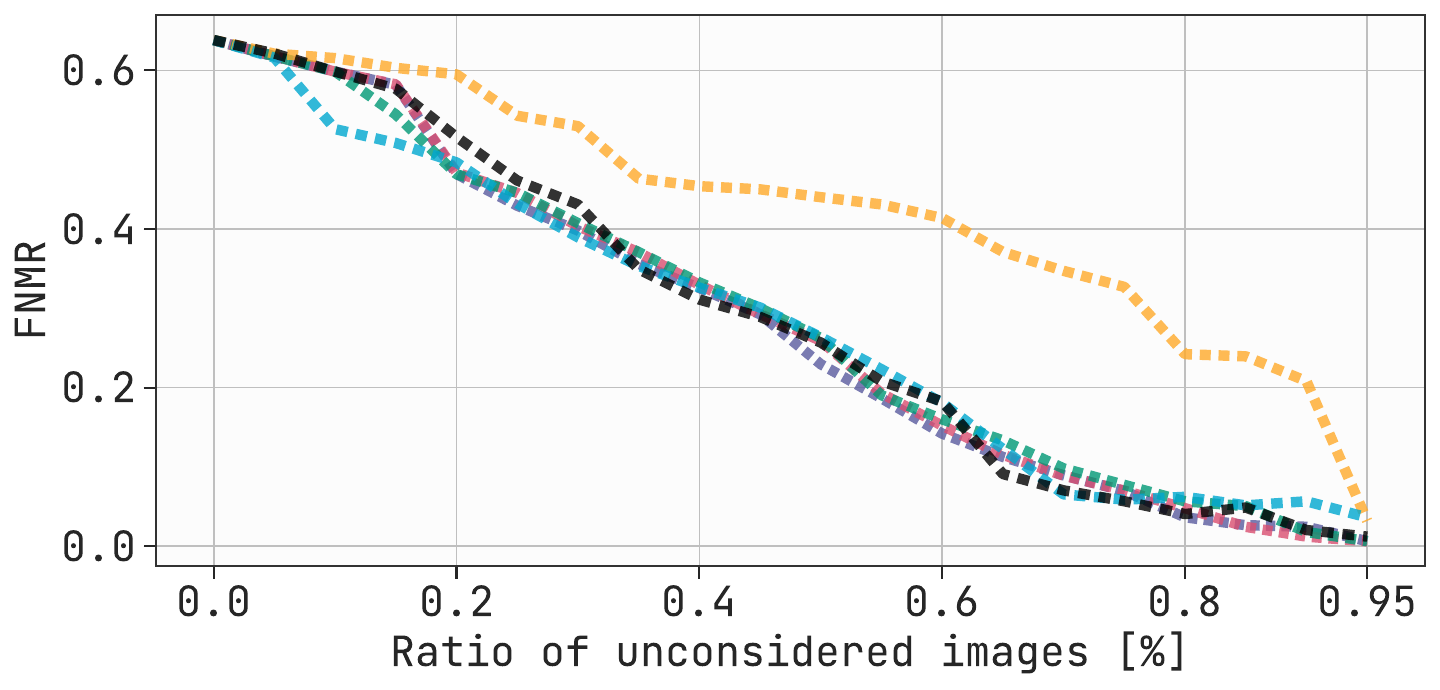}
		 \caption{MagFace \cite{meng_2021_magface} Model, XQLFW \cite{XQLFW} Dataset \\ \grafiqs with ResNet50, FMR$=1e-3$}
	\end{subfigure}
\hfill
	\begin{subfigure}[b]{0.48\textwidth}
		 \centering
		 \includegraphics[width=0.95\textwidth]{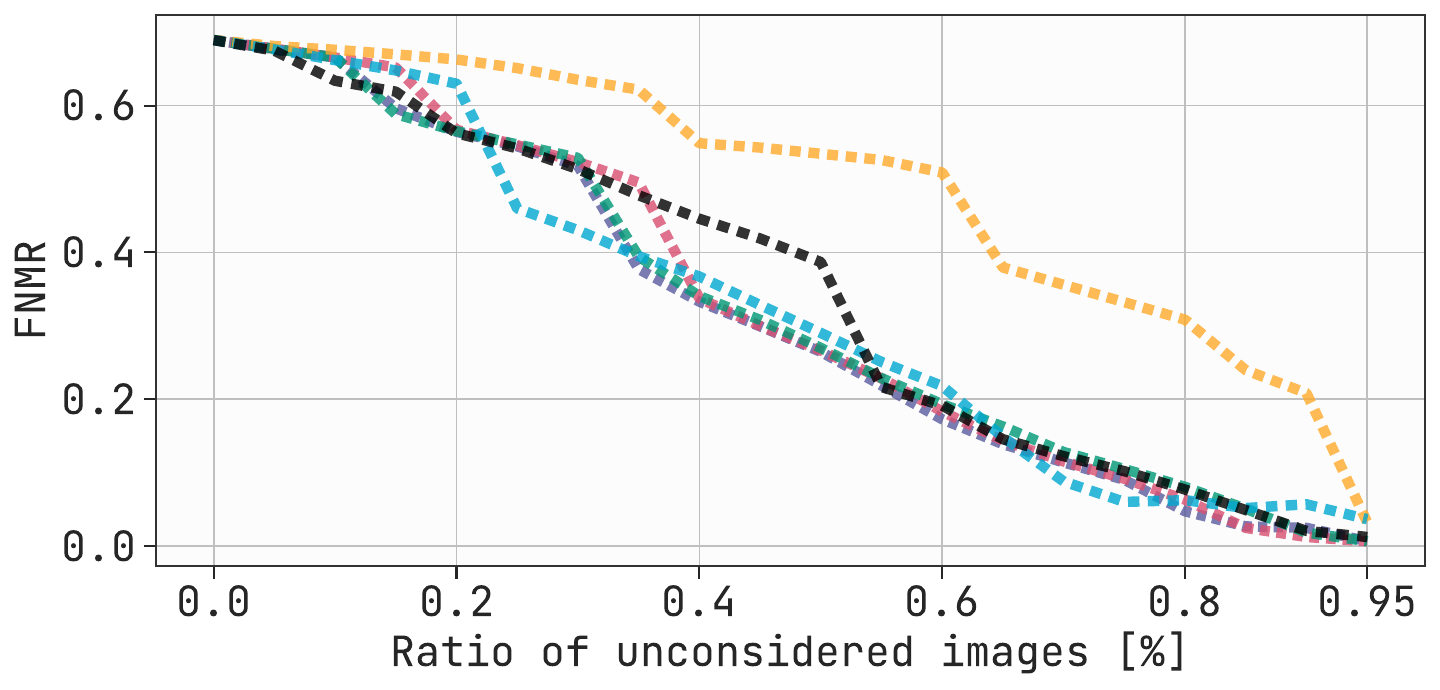}
		 \caption{MagFace \cite{meng_2021_magface} Model, XQLFW \cite{XQLFW} Dataset \\ \grafiqs with ResNet50, FMR$=1e-4$}
	\end{subfigure}
\\
	\begin{subfigure}[b]{0.48\textwidth}
		 \centering
		 \includegraphics[width=0.95\textwidth]{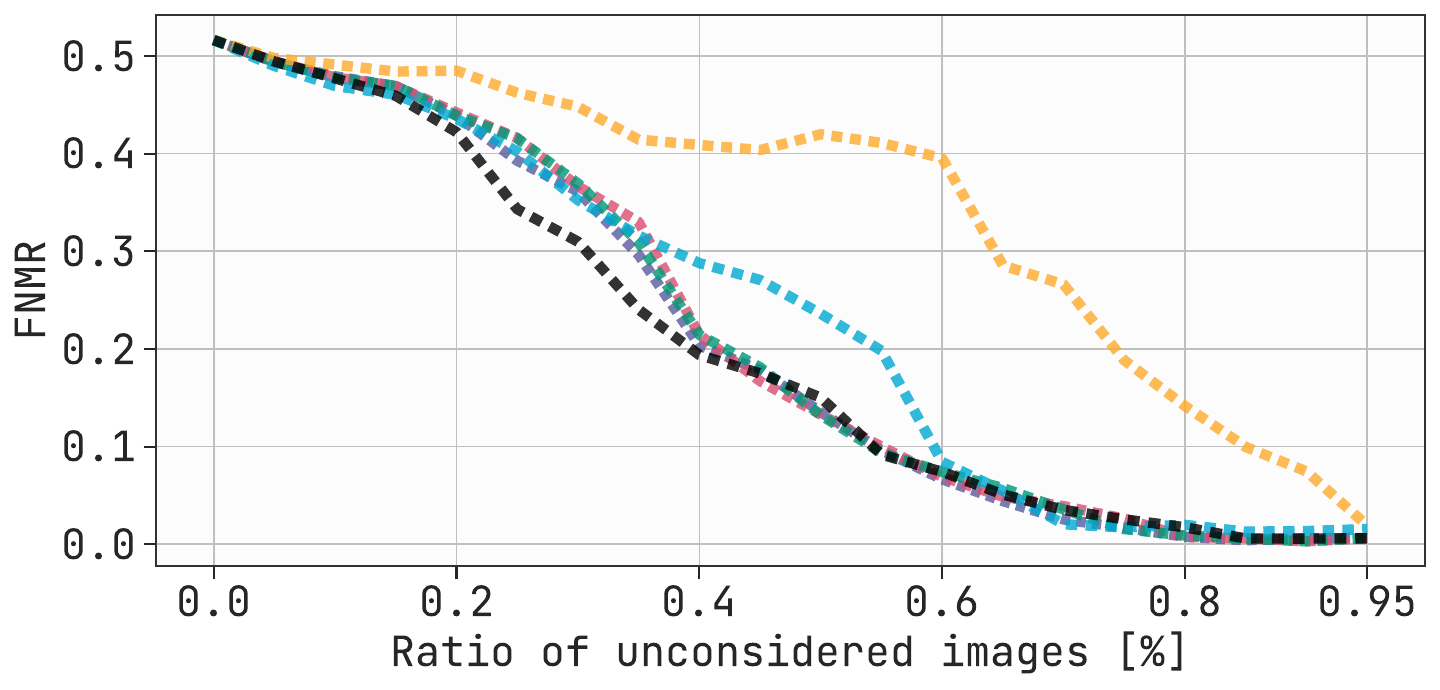}
		 \caption{CurricularFace \cite{curricularFace} Model, XQLFW \cite{XQLFW} Dataset \\ \grafiqs with ResNet50, FMR$=1e-3$}
	\end{subfigure}
\hfill
	\begin{subfigure}[b]{0.48\textwidth}
		 \centering
		 \includegraphics[width=0.95\textwidth]{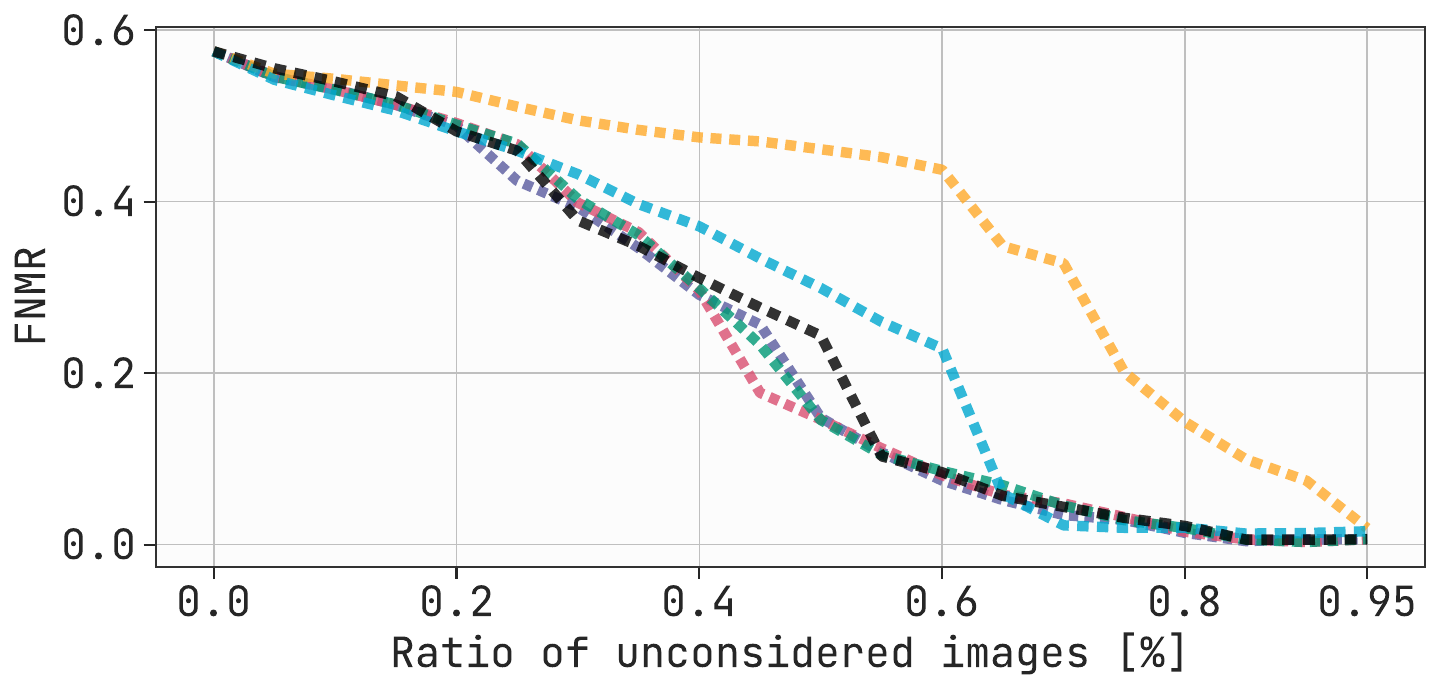}
		 \caption{CurricularFace \cite{curricularFace} Model, XQLFW \cite{XQLFW} Dataset \\ \grafiqs with ResNet50, FMR$=1e-4$}
	\end{subfigure}
\\
\caption{Error-versus-Discard Characteristic (EDC) curves for FNMR@FMR=$1e-3$ and FNMR@FMR=$1e-4$ of our proposed method using $\mathcal{L}_{\text{BNS}}$ as backpropagation loss and absolute sum as FIQ. The gradients at image level ($\phi=\mathcal{I}$), and block levels ($\phi=\text{B}1$ $-$ $\phi=\text{B}4$) are used to calculate FIQ. $\text{MSE}_{\text{BNS}}$ as FIQ is shown in black. Results shown on benchmark XQLFW \cite{XQLFW} using ArcFace, ElasticFace, MagFace, and, CurricularFace FR models. It is evident that the proposed \grafiqs method leads to lower verification error in most cases when images with lowest utility score estimated from gradient magnitudes are rejected. Furthermore, estimating FIQ by backpropagating $\mathcal{L}_{\text{BNS}}$ yields on average better results than using $\text{MSE}_{\text{BNS}}$ directly.}
\vspace{-4mm}
\label{fig:iresnet50_supplementary_XQLFW}
\end{figure*}

%% file: figures/fig_iresnet100_supplementary_adience.tex
\begin{figure*}[h!]
\centering
	\begin{subfigure}[b]{0.9\textwidth}
		\centering
		\includegraphics[width=\textwidth]{figures/iresnet100_bn_overview/legend.pdf}
	\end{subfigure}
\\
	\begin{subfigure}[b]{0.48\textwidth}
		 \centering
		 \includegraphics[width=0.95\textwidth]{figures/iresnet100_bn_overview/adience/ArcFaceModel/iresnet100_bn_combined_ArcFaceModel_adience_fnmr3.pdf}
		 \caption{ArcFace \cite{deng2019arcface} Model, Adience \cite{Adience} Dataset \\ \grafiqs with ResNet100, FMR$=1e-3$}
	\end{subfigure}
\hfill
	\begin{subfigure}[b]{0.48\textwidth}
		 \centering
		 \includegraphics[width=0.95\textwidth]{figures/iresnet100_bn_overview/adience/ArcFaceModel/iresnet100_bn_combined_ArcFaceModel_adience_fnmr4.pdf}
		 \caption{ArcFace \cite{deng2019arcface} Model, Adience \cite{Adience} Dataset \\ \grafiqs with ResNet100, FMR$=1e-4$}
	\end{subfigure}
\\
	\begin{subfigure}[b]{0.48\textwidth}
		 \centering
		 \includegraphics[width=0.95\textwidth]{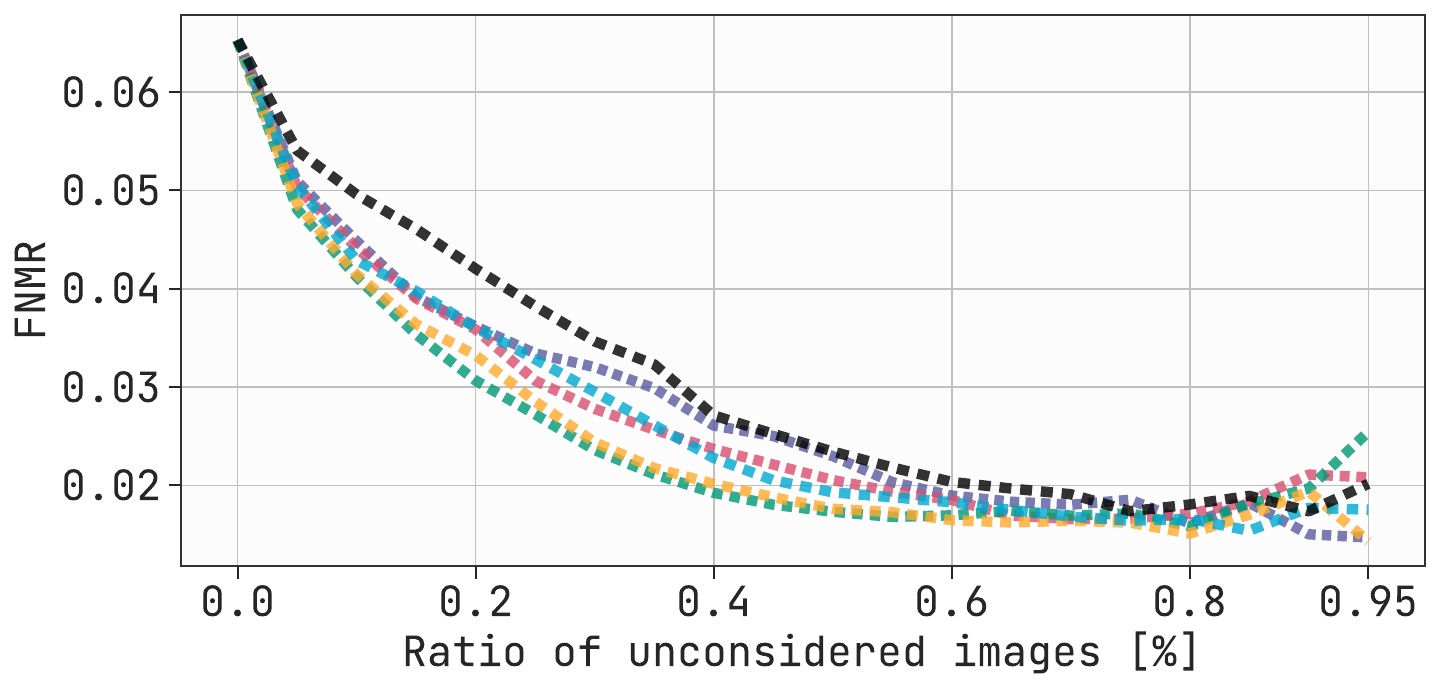}
		 \caption{ElasticFace \cite{elasticface} Model, Adience \cite{Adience} Dataset \\ \grafiqs with ResNet100, FMR$=1e-3$}
	\end{subfigure}
\hfill
	\begin{subfigure}[b]{0.48\textwidth}
		 \centering
		 \includegraphics[width=0.95\textwidth]{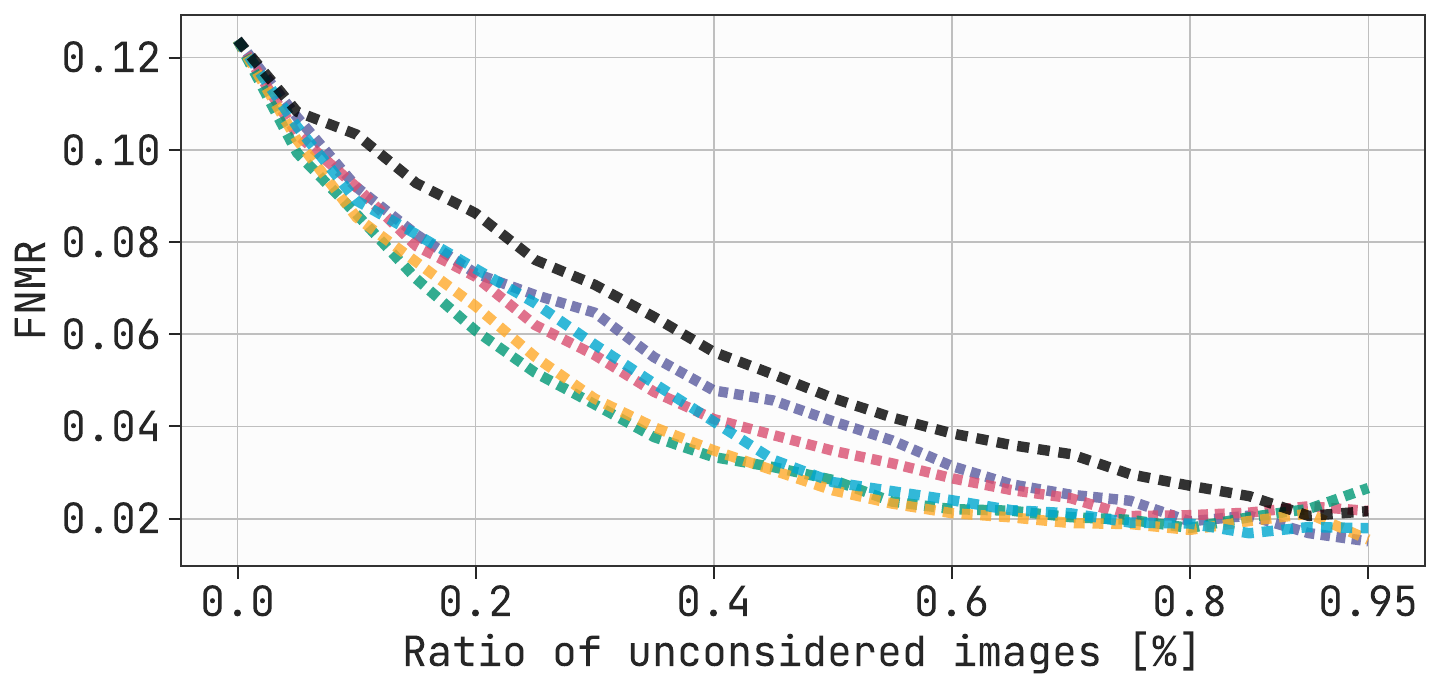}
		 \caption{ElasticFace \cite{elasticface} Model, Adience \cite{Adience} Dataset \\ \grafiqs with ResNet100, FMR$=1e-4$}
	\end{subfigure}
\\
	\begin{subfigure}[b]{0.48\textwidth}
		 \centering
		 \includegraphics[width=0.95\textwidth]{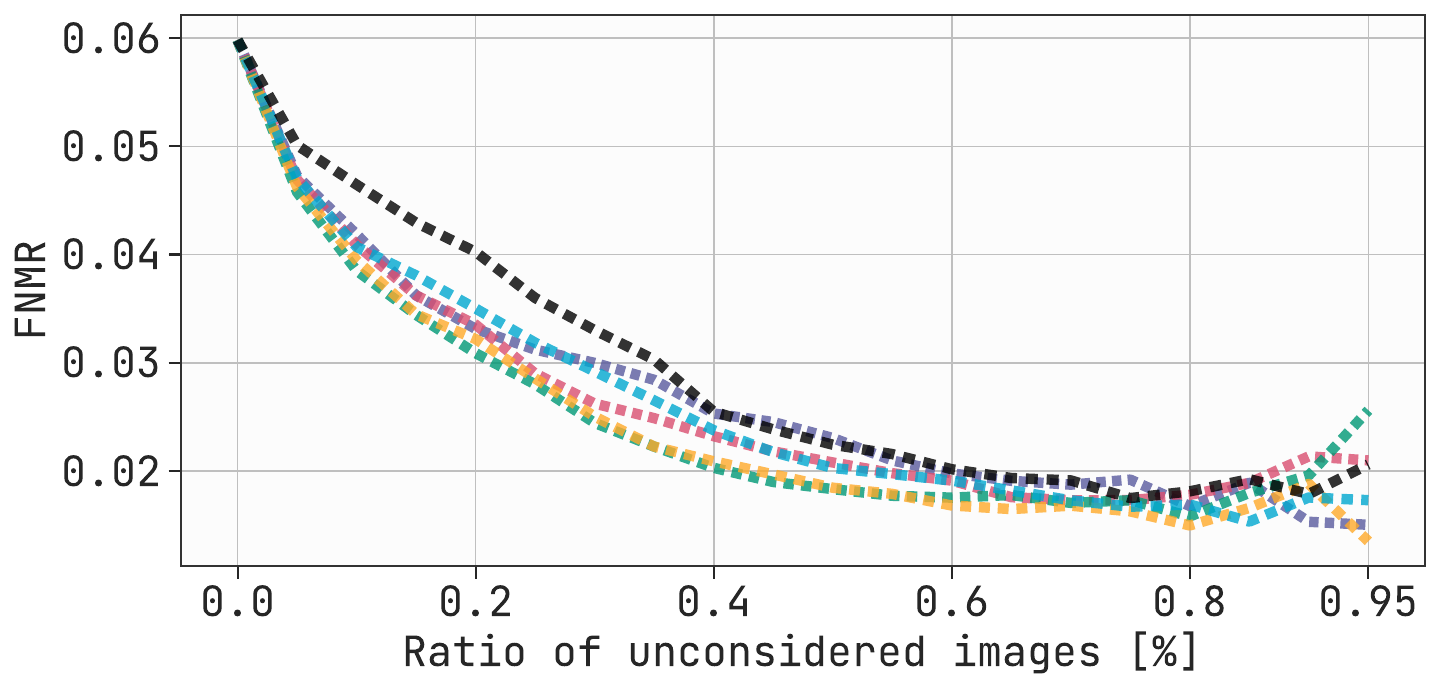}
		 \caption{MagFace \cite{meng_2021_magface} Model, Adience \cite{Adience} Dataset \\ \grafiqs with ResNet100, FMR$=1e-3$}
	\end{subfigure}
\hfill
	\begin{subfigure}[b]{0.48\textwidth}
		 \centering
		 \includegraphics[width=0.95\textwidth]{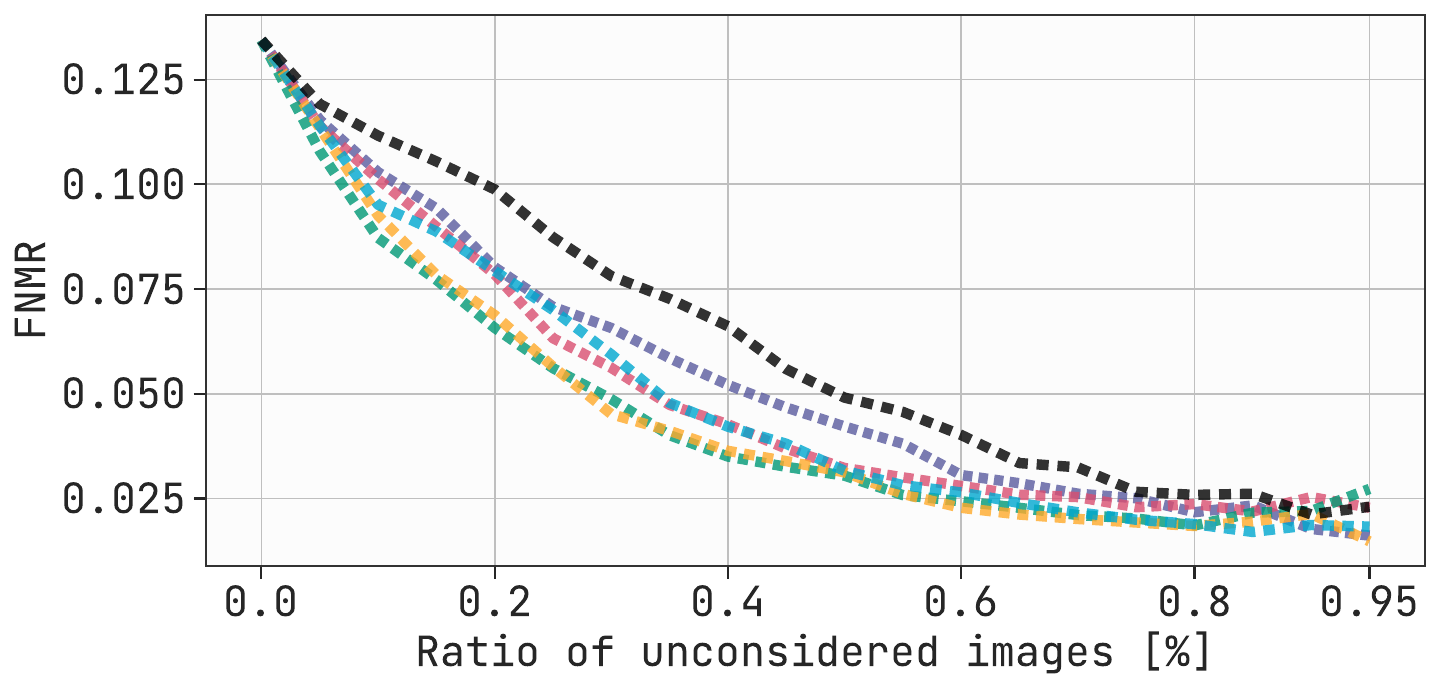}
		 \caption{MagFace \cite{meng_2021_magface} Model, Adience \cite{Adience} Dataset \\ \grafiqs with ResNet100, FMR$=1e-4$}
	\end{subfigure}
\\
	\begin{subfigure}[b]{0.48\textwidth}
		 \centering
		 \includegraphics[width=0.95\textwidth]{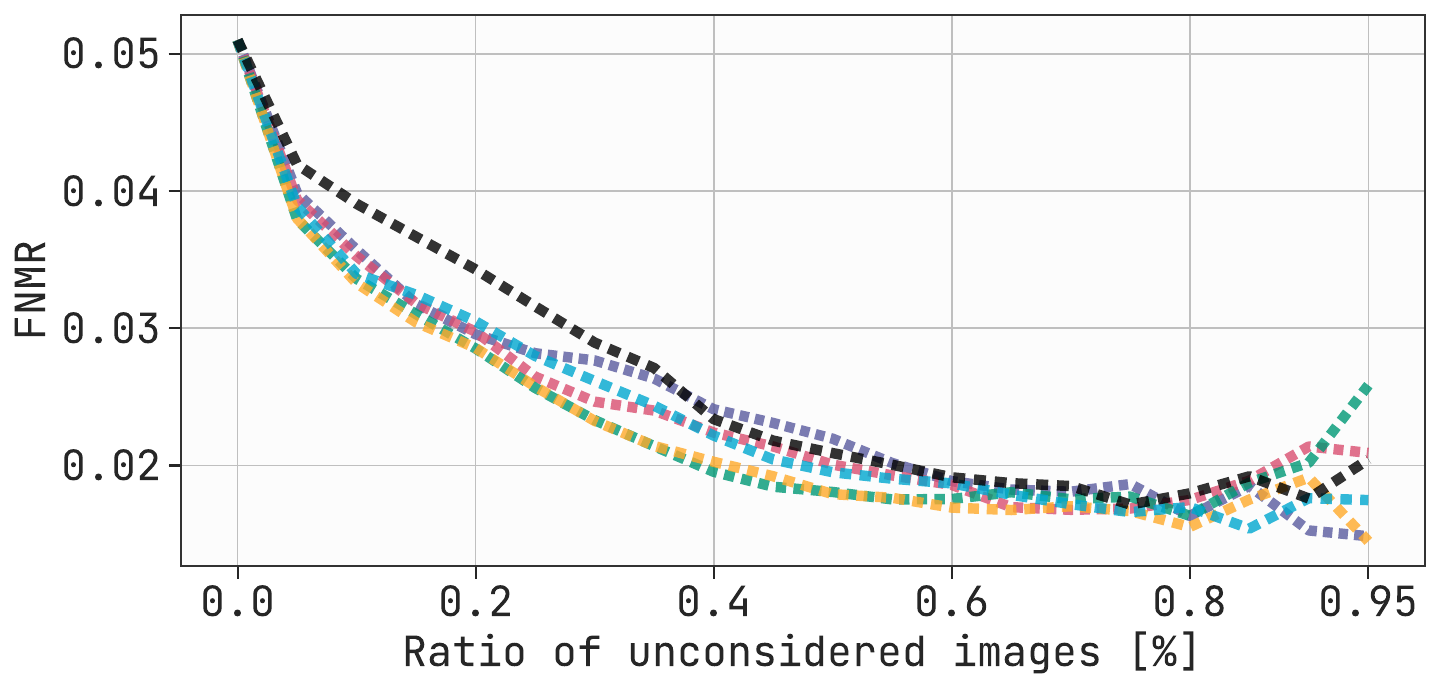}
		 \caption{CurricularFace \cite{curricularFace} Model, Adience \cite{Adience} Dataset \\ \grafiqs with ResNet100, FMR$=1e-3$}
	\end{subfigure}
\hfill
	\begin{subfigure}[b]{0.48\textwidth}
		 \centering
		 \includegraphics[width=0.95\textwidth]{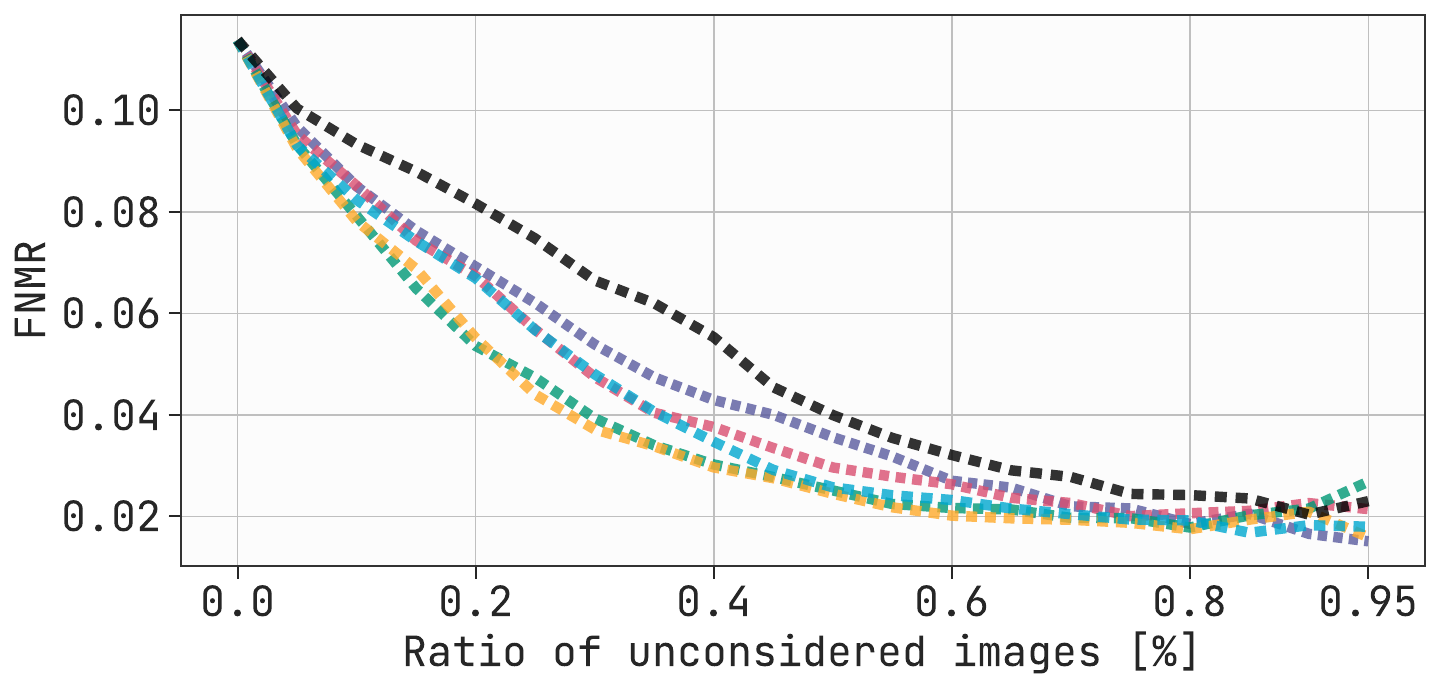}
		 \caption{CurricularFace \cite{curricularFace} Model, Adience \cite{Adience} Dataset \\ \grafiqs with ResNet100, FMR$=1e-4$}
	\end{subfigure}
\\
\caption{Error-versus-Discard Characteristic (EDC) curves for FNMR@FMR=$1e-3$ and FNMR@FMR=$1e-4$ of our proposed method using $\mathcal{L}_{\text{BNS}}$ as backpropagation loss and absolute sum as FIQ. The gradients at image level ($\phi=\mathcal{I}$), and block levels ($\phi=\text{B}1$ $-$ $\phi=\text{B}4$) are used to calculate FIQ. $\text{MSE}_{\text{BNS}}$ as FIQ is shown in black. Results shown on benchmark Adience \cite{Adience} using ArcFace, ElasticFace, MagFace, and, CurricularFace FR models. It is evident that the proposed \grafiqs method leads to lower verification error when images with lowest utility score estimated from gradient magnitudes are rejected. Furthermore, estimating FIQ by backpropagating $\mathcal{L}_{\text{BNS}}$ yields significantly better results than using $\text{MSE}_{\text{BNS}}$ directly.}
\vspace{-4mm}
\label{fig:iresnet100_supplementary_adience}
\end{figure*}

%% file: figures/fig_iresnet100_supplementary_agedb_30.tex
\begin{figure*}[h!]
\centering
	\begin{subfigure}[b]{0.9\textwidth}
		\centering
		\includegraphics[width=\textwidth]{figures/iresnet100_bn_overview/legend.pdf}
	\end{subfigure}
\\
	\begin{subfigure}[b]{0.48\textwidth}
		 \centering
		 \includegraphics[width=0.95\textwidth]{figures/iresnet100_bn_overview/agedb_30/ArcFaceModel/iresnet100_bn_combined_ArcFaceModel_agedb_30_fnmr3.pdf}
		 \caption{ArcFace \cite{deng2019arcface} Model, AgeDB30 \cite{agedb} Dataset \\ \grafiqs with ResNet100, FMR$=1e-3$}
	\end{subfigure}
\hfill
	\begin{subfigure}[b]{0.48\textwidth}
		 \centering
		 \includegraphics[width=0.95\textwidth]{figures/iresnet100_bn_overview/agedb_30/ArcFaceModel/iresnet100_bn_combined_ArcFaceModel_agedb_30_fnmr4.pdf}
		 \caption{ArcFace \cite{deng2019arcface} Model, AgeDB30 \cite{agedb} Dataset \\ \grafiqs with ResNet100, FMR$=1e-4$}
	\end{subfigure}
\\
	\begin{subfigure}[b]{0.48\textwidth}
		 \centering
		 \includegraphics[width=0.95\textwidth]{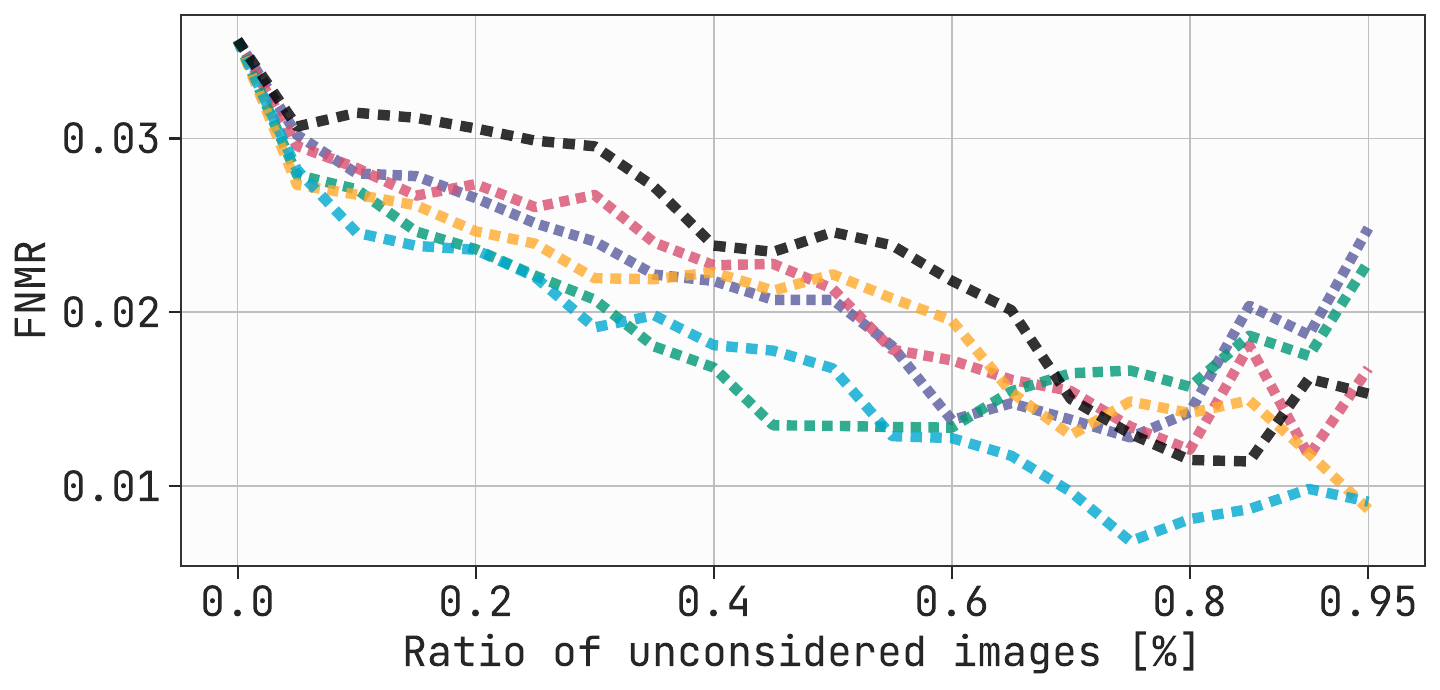}
		 \caption{ElasticFace \cite{elasticface} Model, AgeDB30 \cite{agedb} Dataset \\ \grafiqs with ResNet100, FMR$=1e-3$}
	\end{subfigure}
\hfill
	\begin{subfigure}[b]{0.48\textwidth}
		 \centering
		 \includegraphics[width=0.95\textwidth]{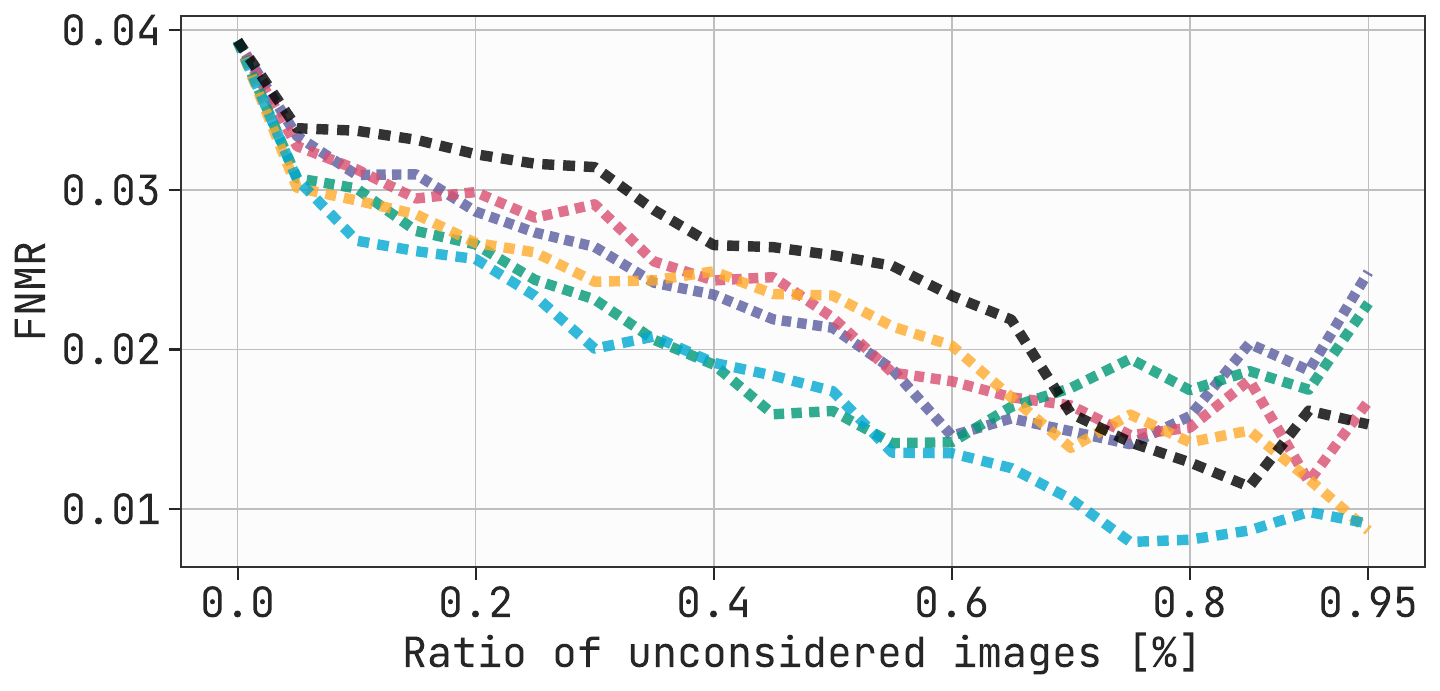}
		 \caption{ElasticFace \cite{elasticface} Model, AgeDB30 \cite{agedb} Dataset \\ \grafiqs with ResNet100, FMR$=1e-4$}
	\end{subfigure}
\\
	\begin{subfigure}[b]{0.48\textwidth}
		 \centering
		 \includegraphics[width=0.95\textwidth]{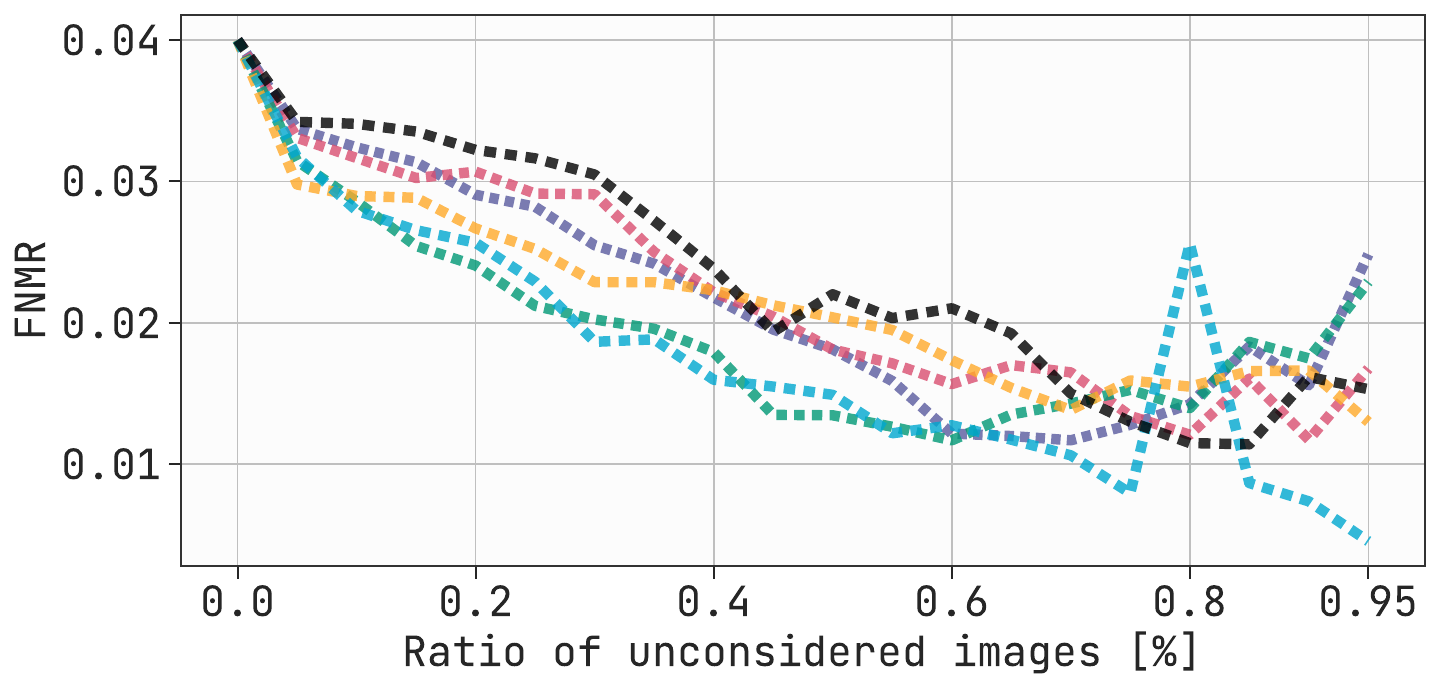}
		 \caption{MagFace \cite{meng_2021_magface} Model, AgeDB30 \cite{agedb} Dataset \\ \grafiqs with ResNet100, FMR$=1e-3$}
	\end{subfigure}
\hfill
	\begin{subfigure}[b]{0.48\textwidth}
		 \centering
		 \includegraphics[width=0.95\textwidth]{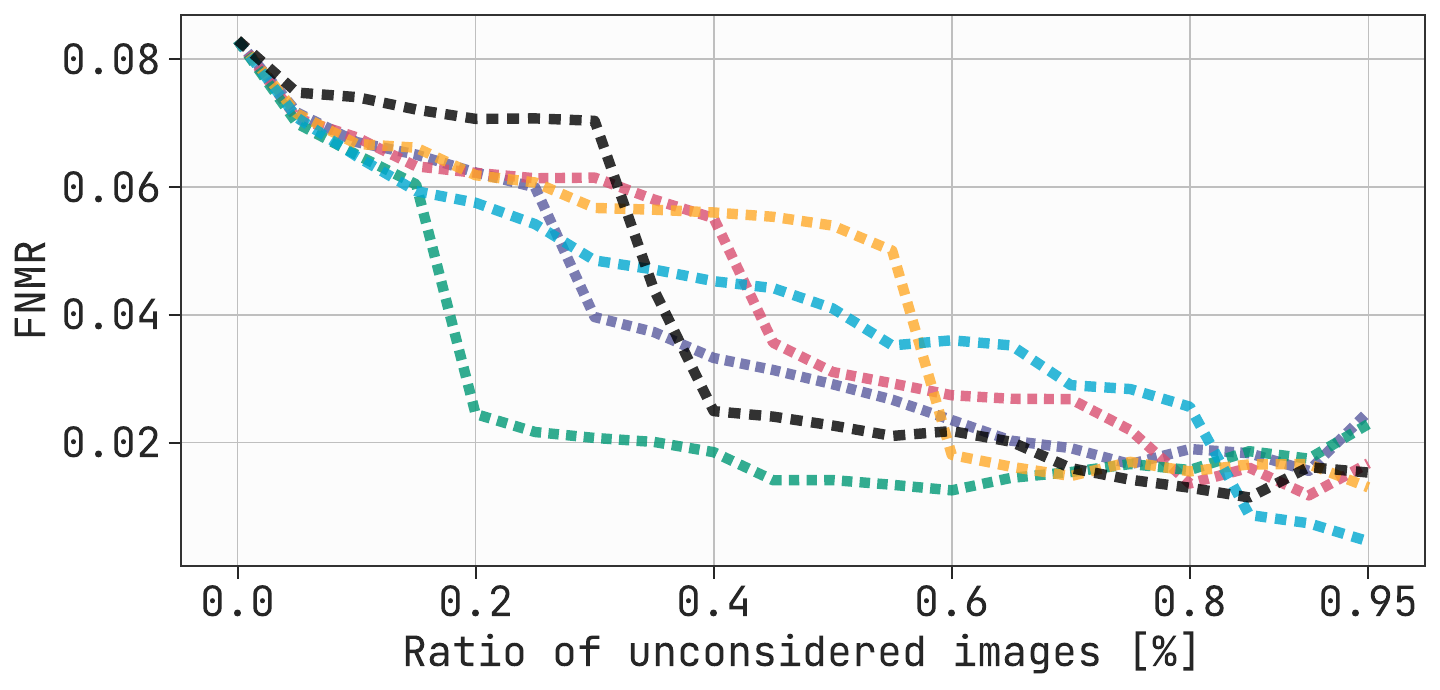}
		 \caption{MagFace \cite{meng_2021_magface} Model, AgeDB30 \cite{agedb} Dataset \\ \grafiqs with ResNet100, FMR$=1e-4$}
	\end{subfigure}
\\
	\begin{subfigure}[b]{0.48\textwidth}
		 \centering
		 \includegraphics[width=0.95\textwidth]{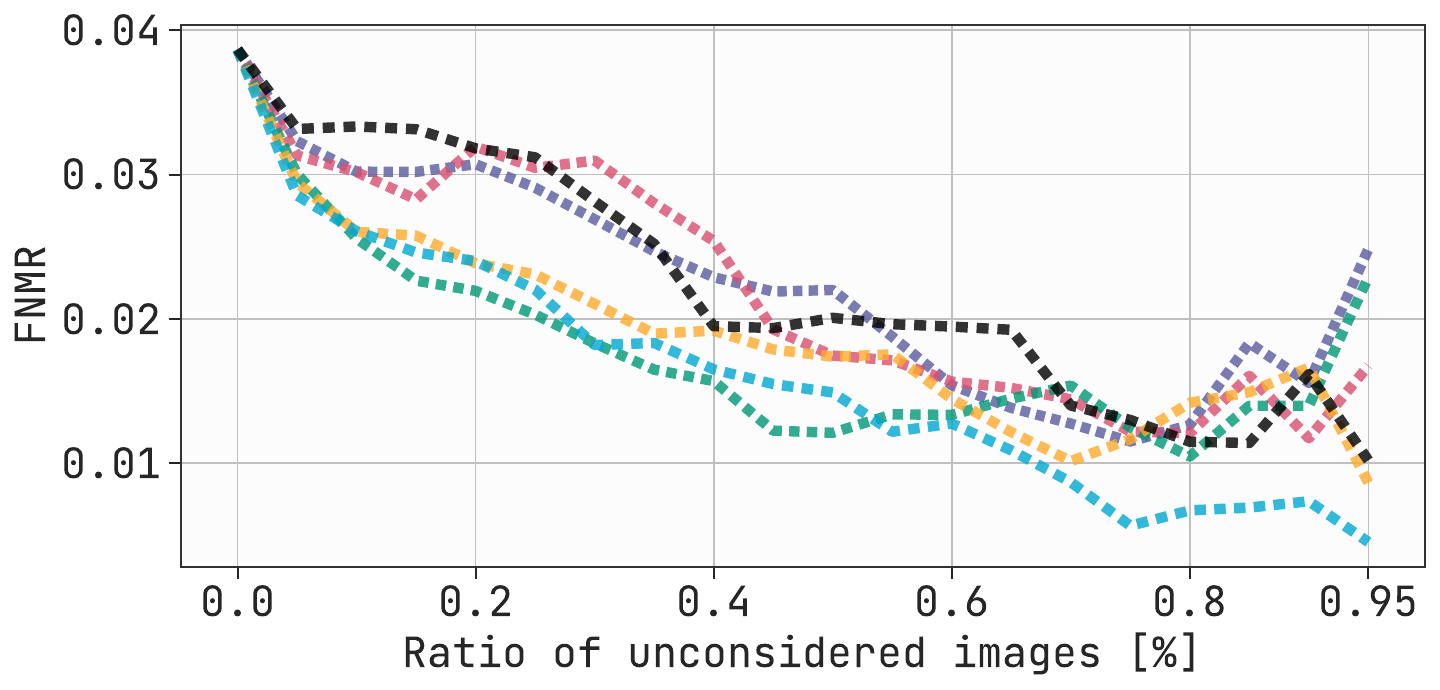}
		 \caption{CurricularFace \cite{curricularFace} Model, AgeDB30 \cite{agedb} Dataset \\ \grafiqs with ResNet100, FMR$=1e-3$}
	\end{subfigure}
\hfill
	\begin{subfigure}[b]{0.48\textwidth}
		 \centering
		 \includegraphics[width=0.95\textwidth]{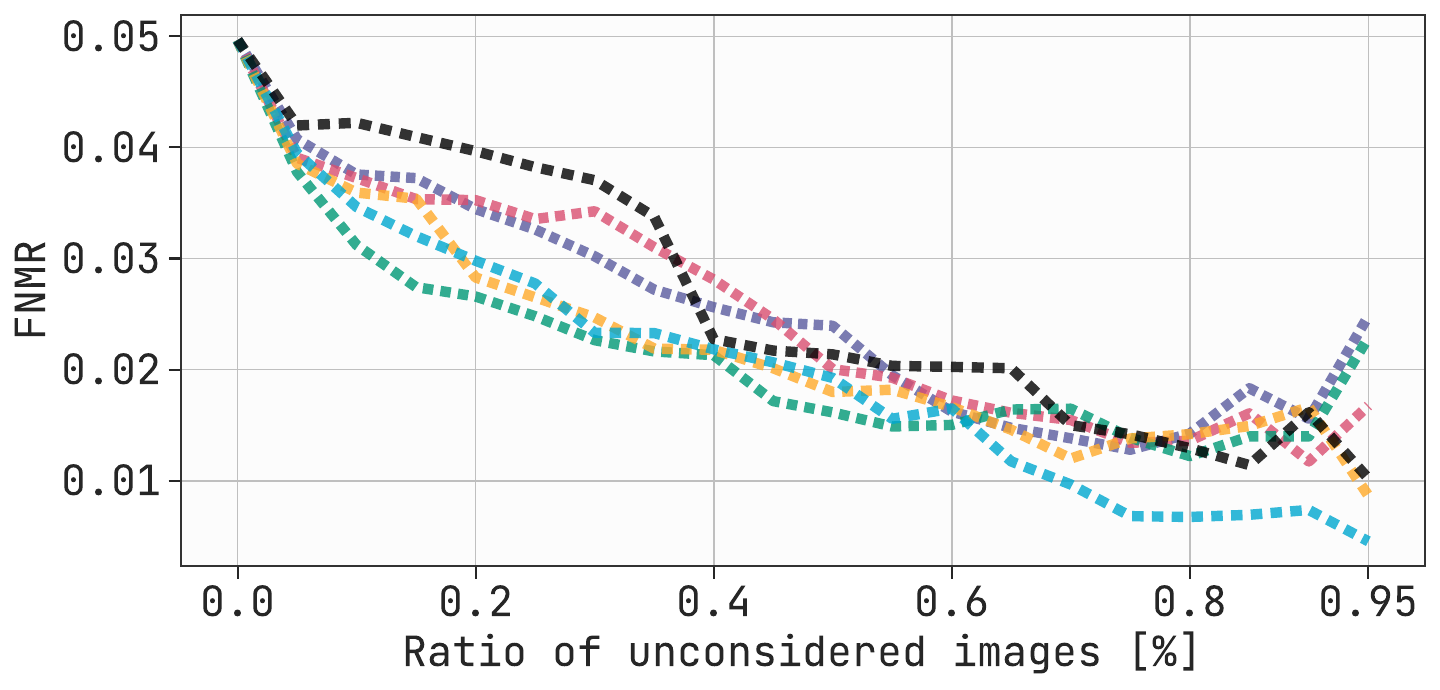}
		 \caption{CurricularFace \cite{curricularFace} Model, AgeDB30 \cite{agedb} Dataset \\ \grafiqs with ResNet100, FMR$=1e-4$}
	\end{subfigure}
\\
\caption{Error-versus-Discard Characteristic (EDC) curves for FNMR@FMR=$1e-3$ and FNMR@FMR=$1e-4$ of our proposed method using $\mathcal{L}_{\text{BNS}}$ as backpropagation loss and absolute sum as FIQ. The gradients at image level ($\phi=\mathcal{I}$), and block levels ($\phi=\text{B}1$ $-$ $\phi=\text{B}4$) are used to calculate FIQ. $\text{MSE}_{\text{BNS}}$ as FIQ is shown in black. Results shown on benchmark AgeDB30 \cite{agedb} using ArcFace, ElasticFace, MagFace, and, CurricularFace FR models. It is evident that the proposed \grafiqs method leads to lower verification error when images with lowest utility score estimated from gradient magnitudes are rejected. Furthermore, estimating FIQ by backpropagating $\mathcal{L}_{\text{BNS}}$ yields significantly better results than using $\text{MSE}_{\text{BNS}}$ directly.}
\vspace{-4mm}
\label{fig:iresnet100_supplementary_agedb_30}
\end{figure*}

%% file: figures/fig_iresnet100_supplementary_cfp_fp.tex
\begin{figure*}[h!]
\centering
	\begin{subfigure}[b]{0.9\textwidth}
		\centering
		\includegraphics[width=\textwidth]{figures/iresnet100_bn_overview/legend.pdf}
	\end{subfigure}
\\
	\begin{subfigure}[b]{0.48\textwidth}
		 \centering
		 \includegraphics[width=0.95\textwidth]{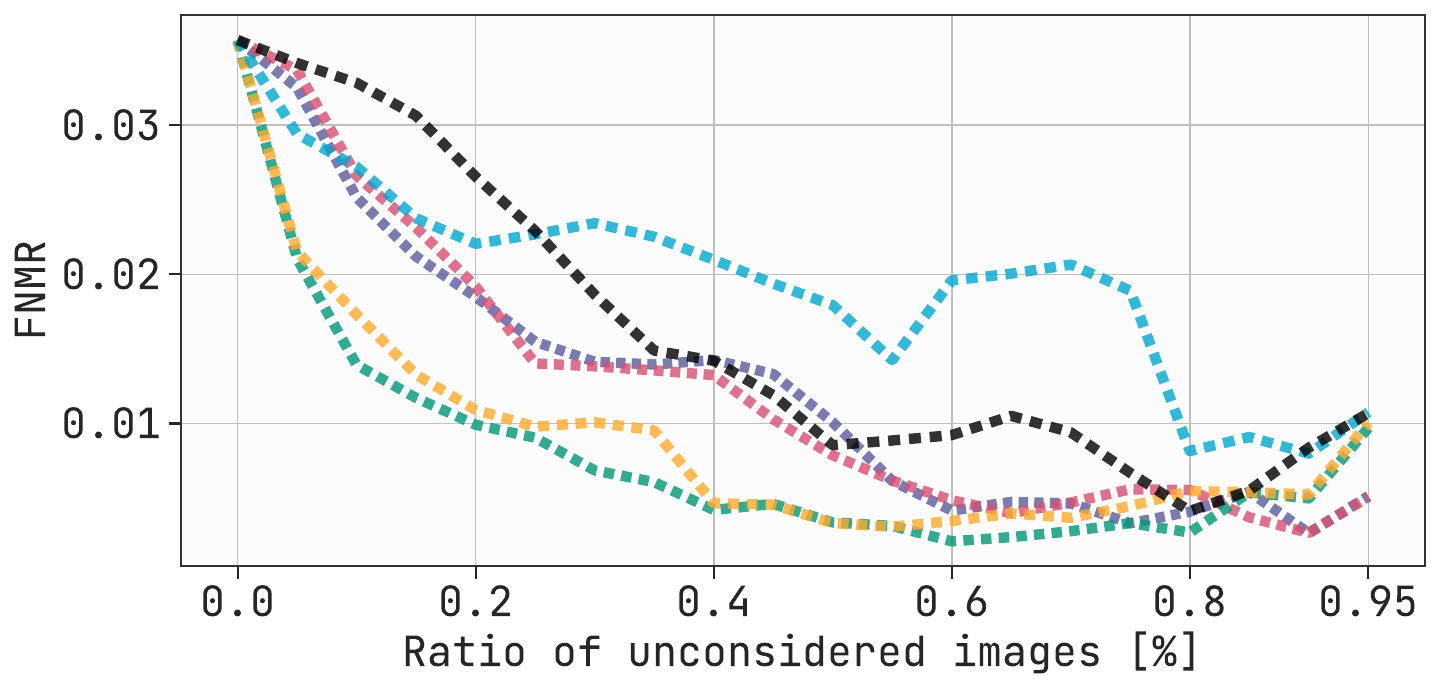}
		 \caption{ArcFace \cite{deng2019arcface} Model, CFP-FP \cite{cfp-fp} Dataset \\ \grafiqs with ResNet100, FMR$=1e-3$}
	\end{subfigure}
\hfill
	\begin{subfigure}[b]{0.48\textwidth}
		 \centering
		 \includegraphics[width=0.95\textwidth]{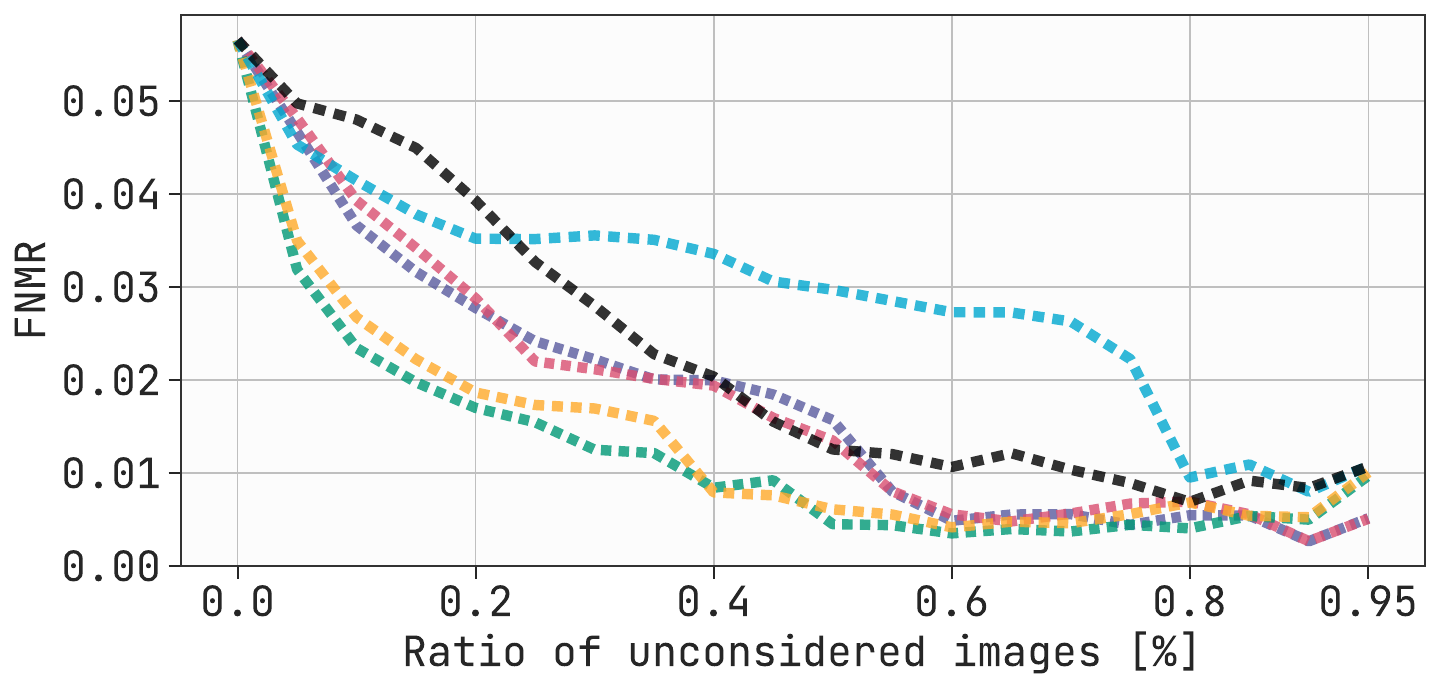}
		 \caption{ArcFace \cite{deng2019arcface} Model, CFP-FP \cite{cfp-fp} Dataset \\ \grafiqs with ResNet100, FMR$=1e-4$}
	\end{subfigure}
\\
	\begin{subfigure}[b]{0.48\textwidth}
		 \centering
		 \includegraphics[width=0.95\textwidth]{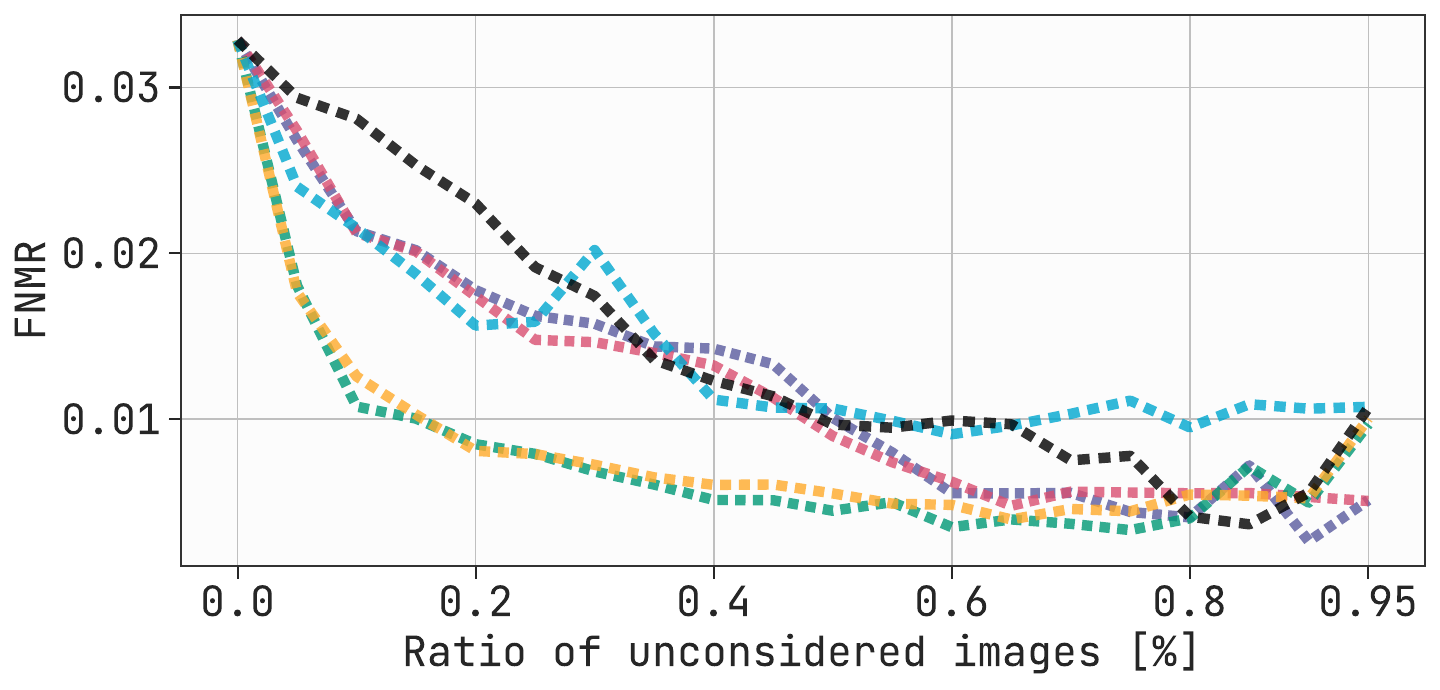}
		 \caption{ElasticFace \cite{elasticface} Model, CFP-FP \cite{cfp-fp} Dataset \\ \grafiqs with ResNet100, FMR$=1e-3$}
	\end{subfigure}
\hfill
	\begin{subfigure}[b]{0.48\textwidth}
		 \centering
		 \includegraphics[width=0.95\textwidth]{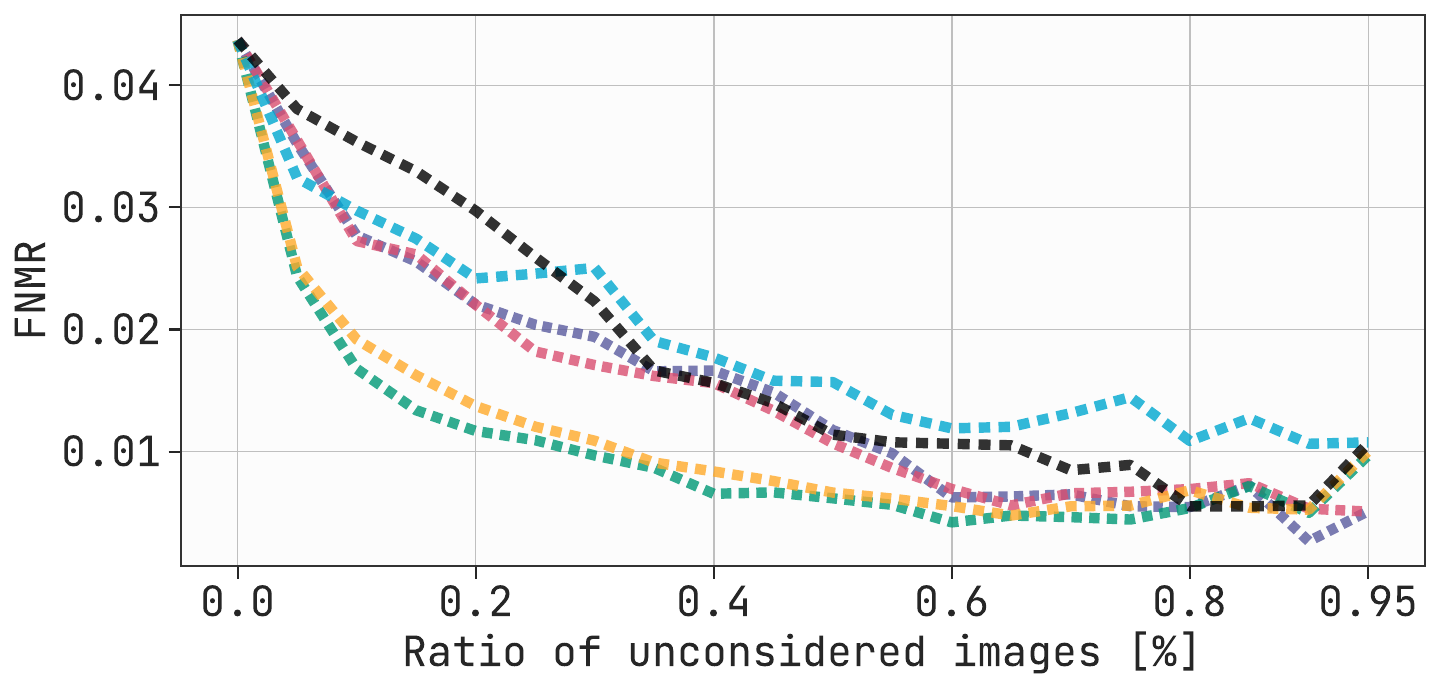}
		 \caption{ElasticFace \cite{elasticface} Model, CFP-FP \cite{cfp-fp} Dataset \\ \grafiqs with ResNet100, FMR$=1e-4$}
	\end{subfigure}
\\
	\begin{subfigure}[b]{0.48\textwidth}
		 \centering
		 \includegraphics[width=0.95\textwidth]{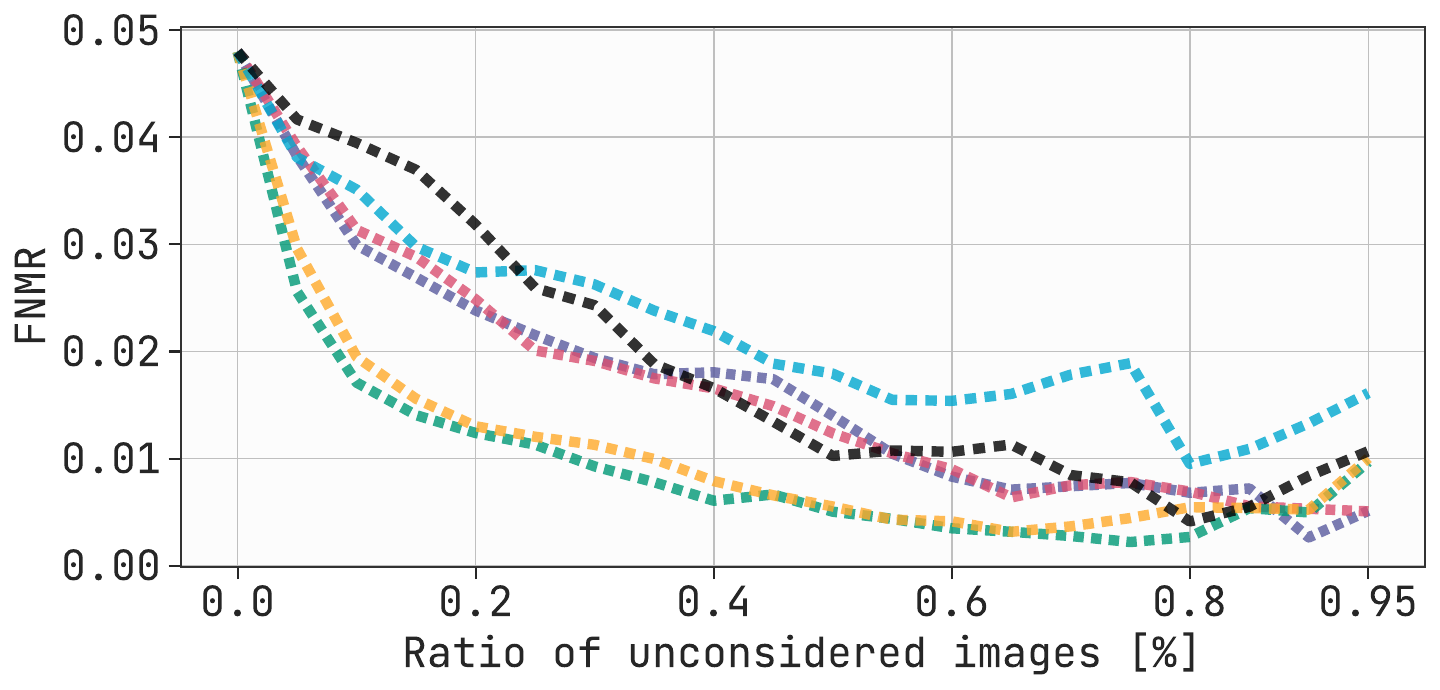}
		 \caption{MagFace \cite{meng_2021_magface} Model, CFP-FP \cite{cfp-fp} Dataset \\ \grafiqs with ResNet100, FMR$=1e-3$}
	\end{subfigure}
\hfill
	\begin{subfigure}[b]{0.48\textwidth}
		 \centering
		 \includegraphics[width=0.95\textwidth]{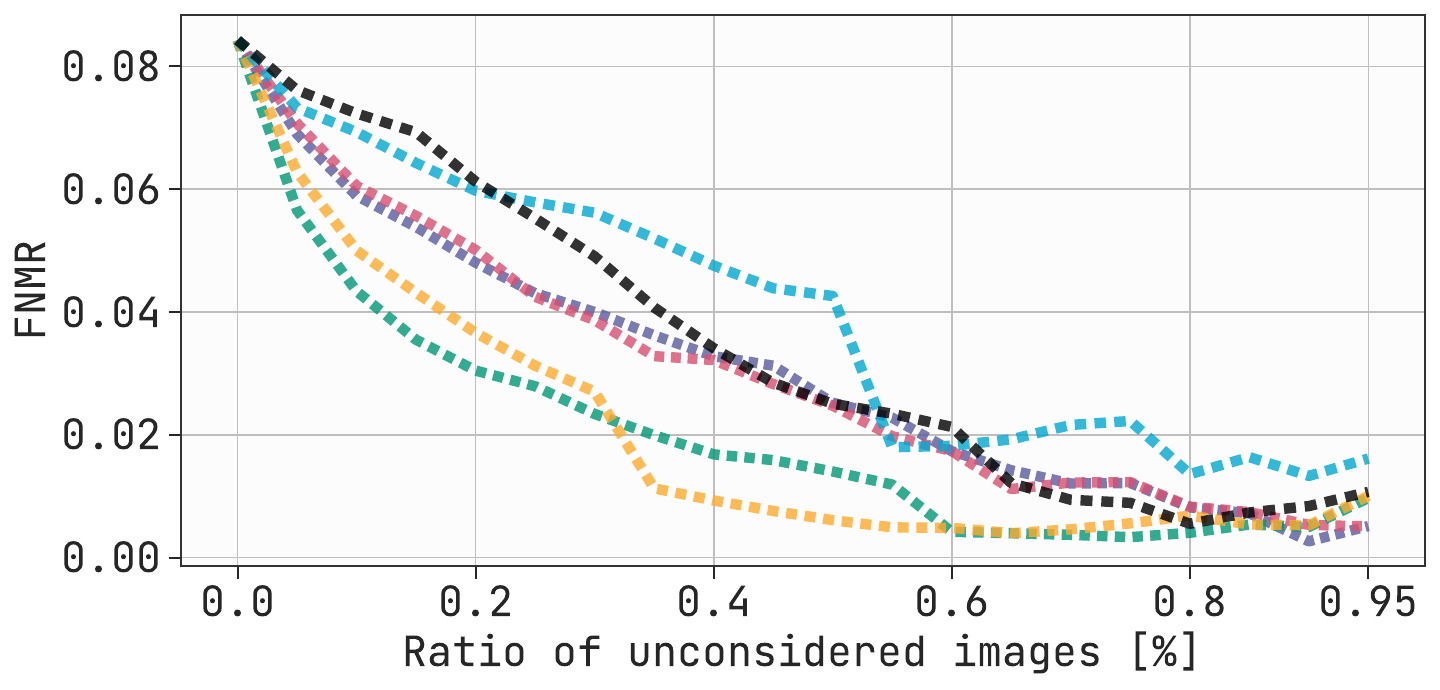}
		 \caption{MagFace \cite{meng_2021_magface} Model, CFP-FP \cite{cfp-fp} Dataset \\ \grafiqs with ResNet100, FMR$=1e-4$}
	\end{subfigure}
\\
	\begin{subfigure}[b]{0.48\textwidth}
		 \centering
		 \includegraphics[width=0.95\textwidth]{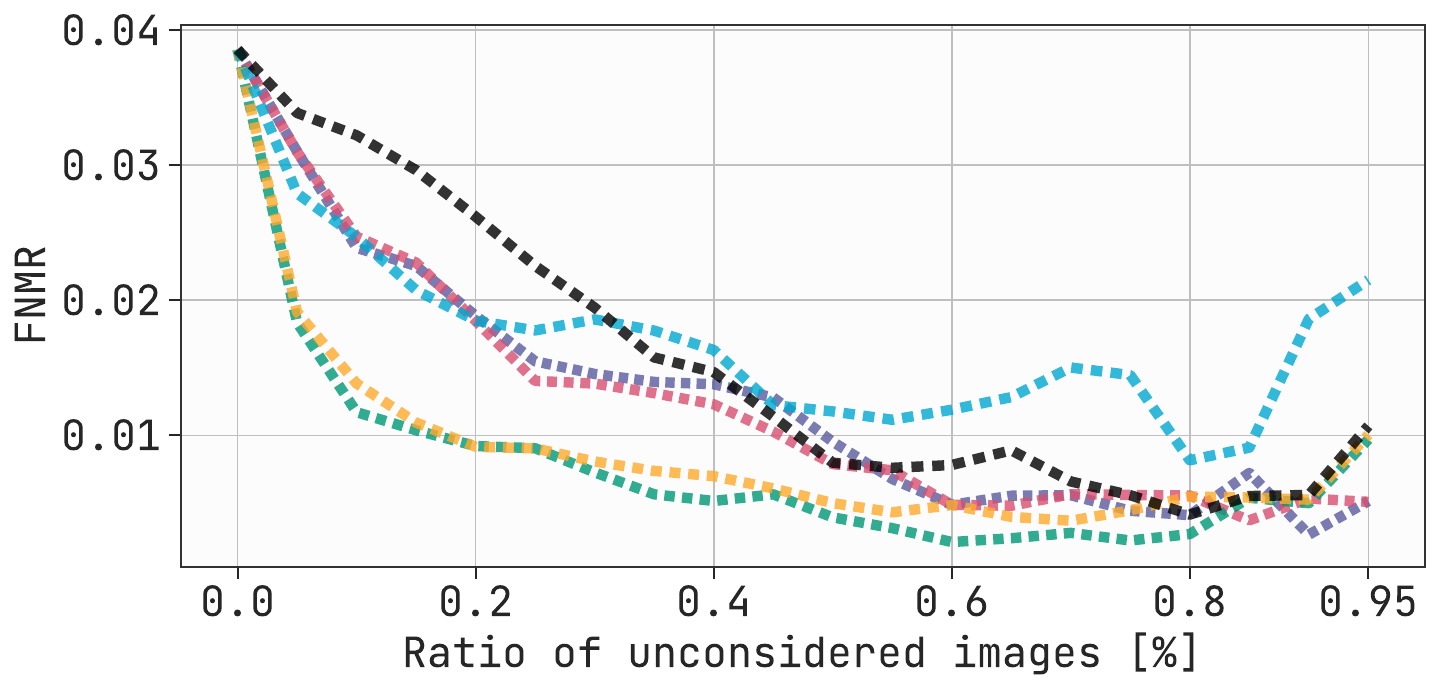}
		 \caption{CurricularFace \cite{curricularFace} Model, CFP-FP \cite{cfp-fp} Dataset \\ \grafiqs with ResNet100, FMR$=1e-3$}
	\end{subfigure}
\hfill
	\begin{subfigure}[b]{0.48\textwidth}
		 \centering
		 \includegraphics[width=0.95\textwidth]{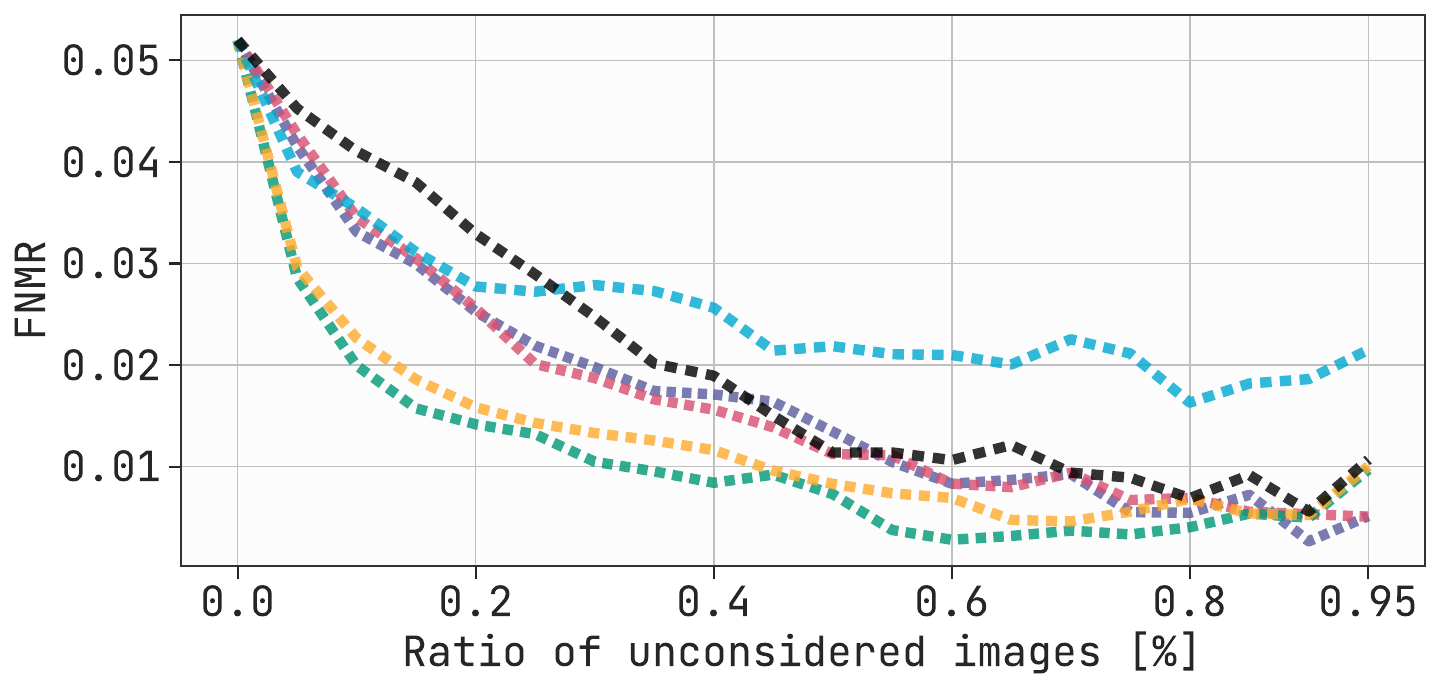}
		 \caption{CurricularFace \cite{curricularFace} Model, CFP-FP \cite{cfp-fp} Dataset \\ \grafiqs with ResNet100, FMR$=1e-4$}
	\end{subfigure}
\\
\caption{Error-versus-Discard Characteristic (EDC) curves for FNMR@FMR=$1e-3$ and FNMR@FMR=$1e-4$ of our proposed method using $\mathcal{L}_{\text{BNS}}$ as backpropagation loss and absolute sum as FIQ. The gradients at image level ($\phi=\mathcal{I}$), and block levels ($\phi=\text{B}1$ $-$ $\phi=\text{B}4$) are used to calculate FIQ. $\text{MSE}_{\text{BNS}}$ as FIQ is shown in black. Results shown on benchmark CFP-FP \cite{cfp-fp} using ArcFace, ElasticFace, MagFace, and, CurricularFace FR models. It is evident that the proposed \grafiqs method leads to lower verification error when images with lowest utility score estimated from gradient magnitudes are rejected. Furthermore, estimating FIQ by backpropagating $\mathcal{L}_{\text{BNS}}$ yields significantly better results than using $\text{MSE}_{\text{BNS}}$ directly.}
\vspace{-4mm}
\label{fig:iresnet100_supplementary_cfp_fp}
\end{figure*}

%% file: figures/fig_iresnet100_supplementary_lfw.tex
\begin{figure*}[h!]
\centering
	\begin{subfigure}[b]{0.9\textwidth}
		\centering
		\includegraphics[width=\textwidth]{figures/iresnet100_bn_overview/legend.pdf}
	\end{subfigure}
\\
	\begin{subfigure}[b]{0.48\textwidth}
		 \centering
		 \includegraphics[width=0.95\textwidth]{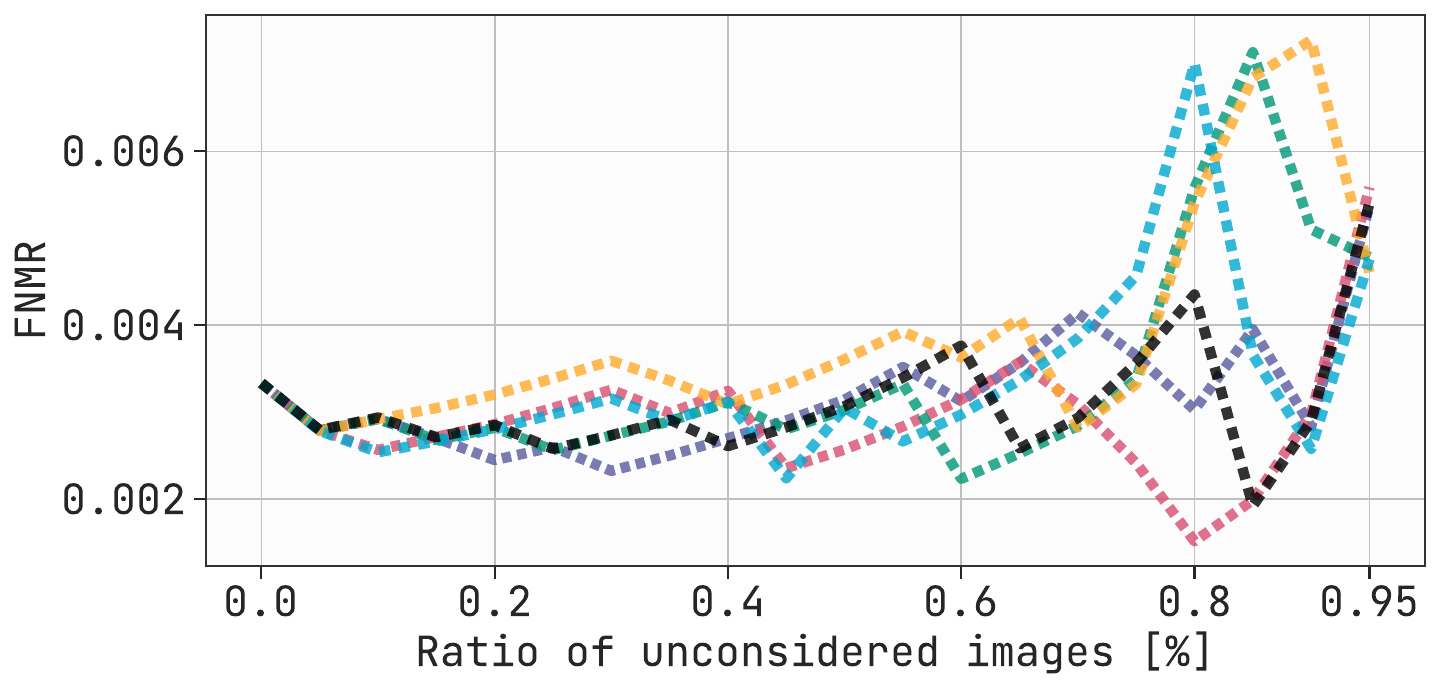}
		 \caption{ArcFace \cite{deng2019arcface} Model, LFW \cite{LFWTech} Dataset \\ \grafiqs with ResNet100, FMR$=1e-3$}
	\end{subfigure}
\hfill
	\begin{subfigure}[b]{0.48\textwidth}
		 \centering
		 \includegraphics[width=0.95\textwidth]{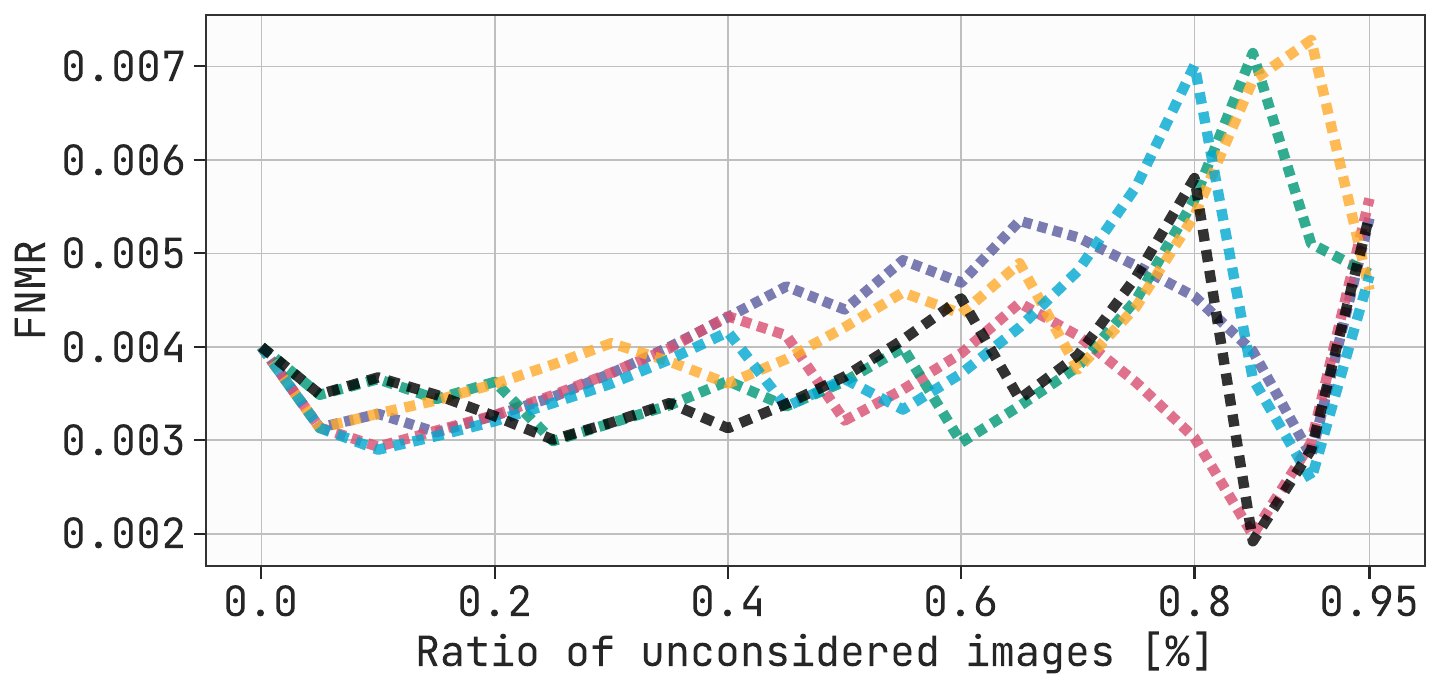}
		 \caption{ArcFace \cite{deng2019arcface} Model, LFW \cite{LFWTech} Dataset \\ \grafiqs with ResNet100, FMR$=1e-4$}
	\end{subfigure}
\\
	\begin{subfigure}[b]{0.48\textwidth}
		 \centering
		 \includegraphics[width=0.95\textwidth]{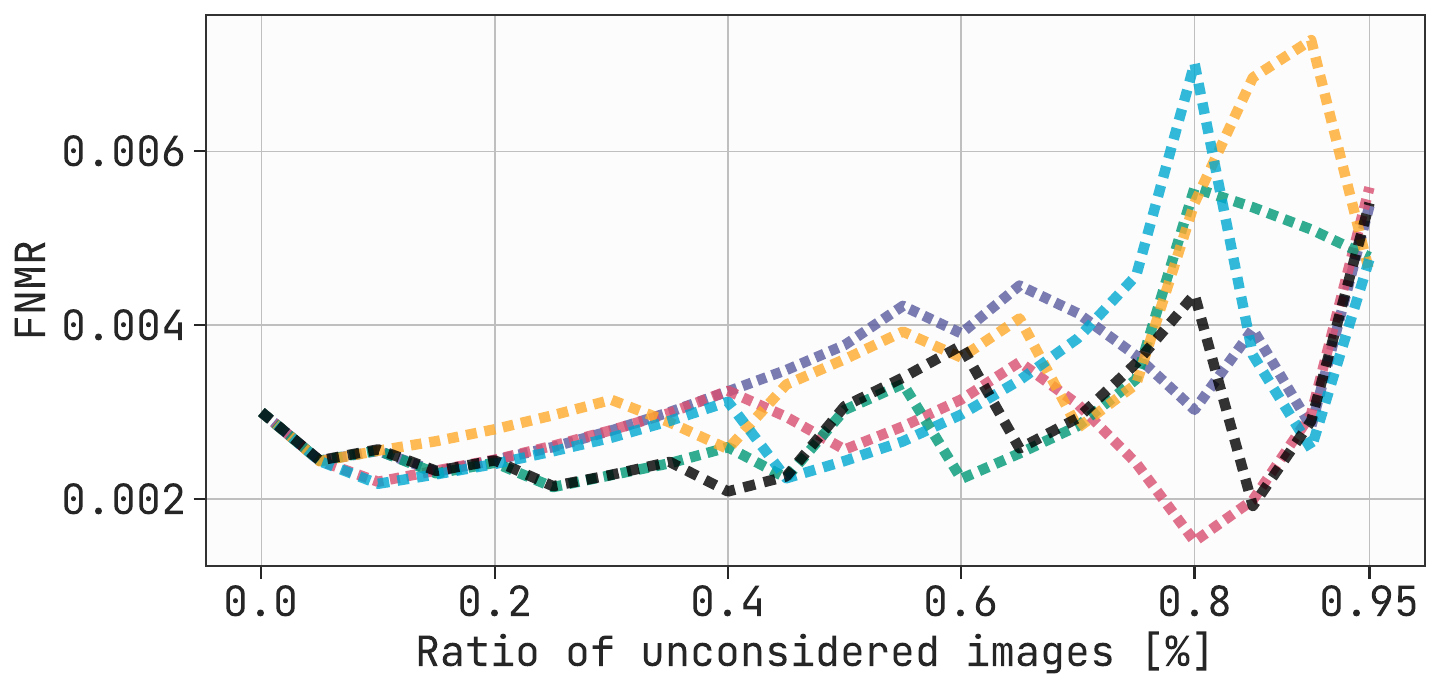}
		 \caption{ElasticFace \cite{elasticface} Model, LFW \cite{LFWTech} Dataset \\ \grafiqs with ResNet100, FMR$=1e-3$}
	\end{subfigure}
\hfill
	\begin{subfigure}[b]{0.48\textwidth}
		 \centering
		 \includegraphics[width=0.95\textwidth]{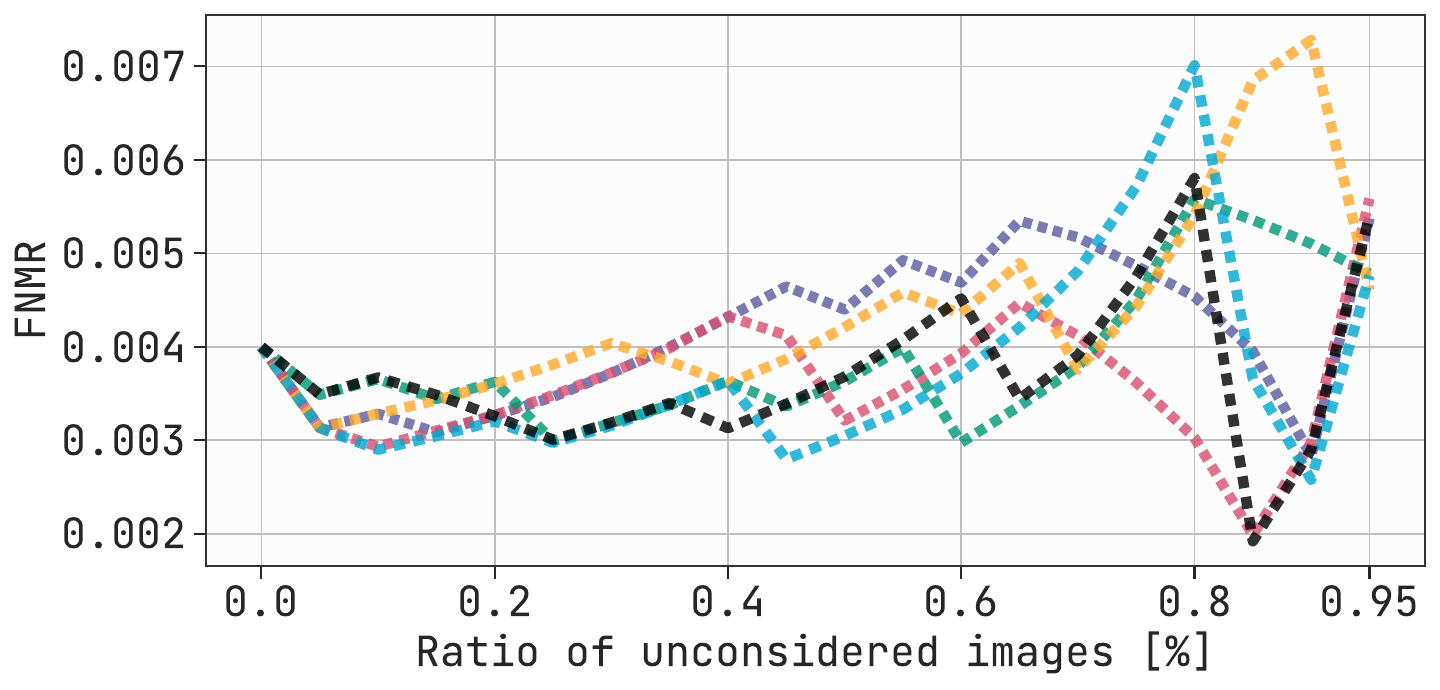}
		 \caption{ElasticFace \cite{elasticface} Model, LFW \cite{LFWTech} Dataset \\ \grafiqs with ResNet100, FMR$=1e-4$}
	\end{subfigure}
\\
	\begin{subfigure}[b]{0.48\textwidth}
		 \centering
		 \includegraphics[width=0.95\textwidth]{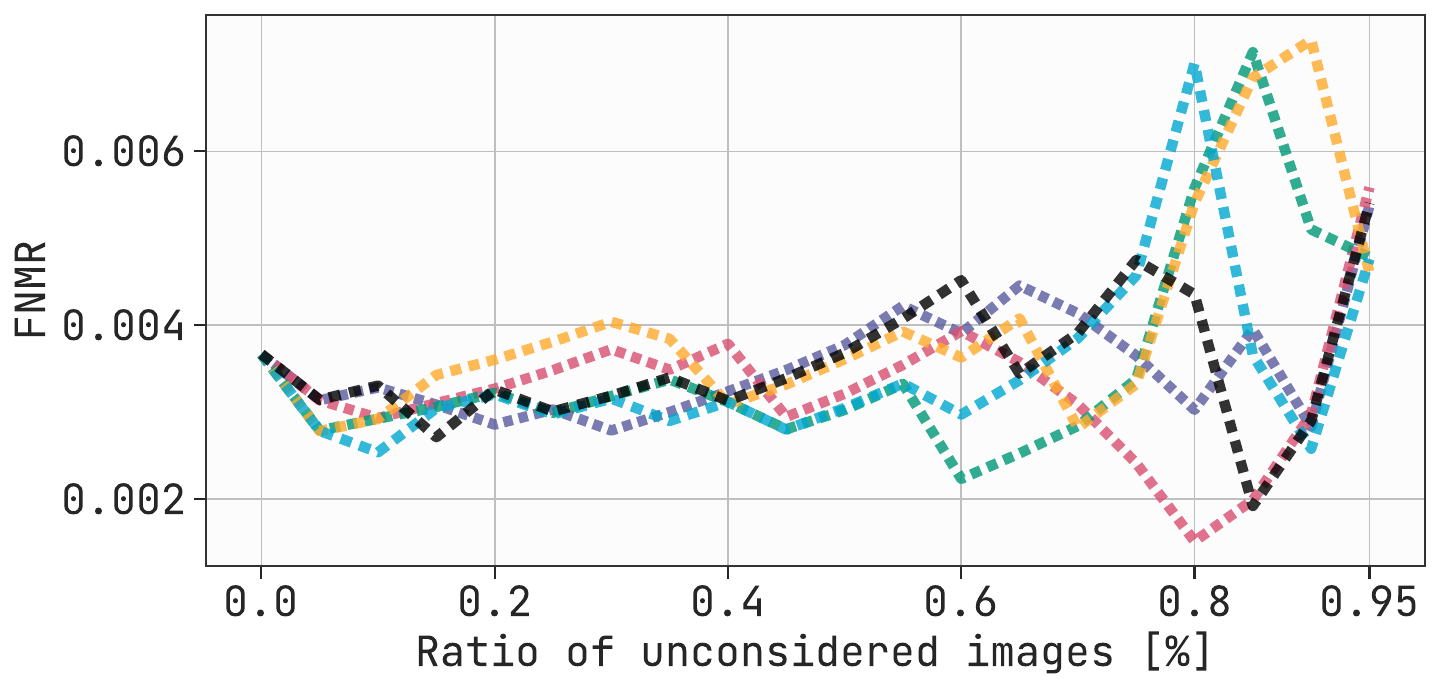}
		 \caption{MagFace \cite{meng_2021_magface} Model, LFW \cite{LFWTech} Dataset \\ \grafiqs with ResNet100, FMR$=1e-3$}
	\end{subfigure}
\hfill
	\begin{subfigure}[b]{0.48\textwidth}
		 \centering
		 \includegraphics[width=0.95\textwidth]{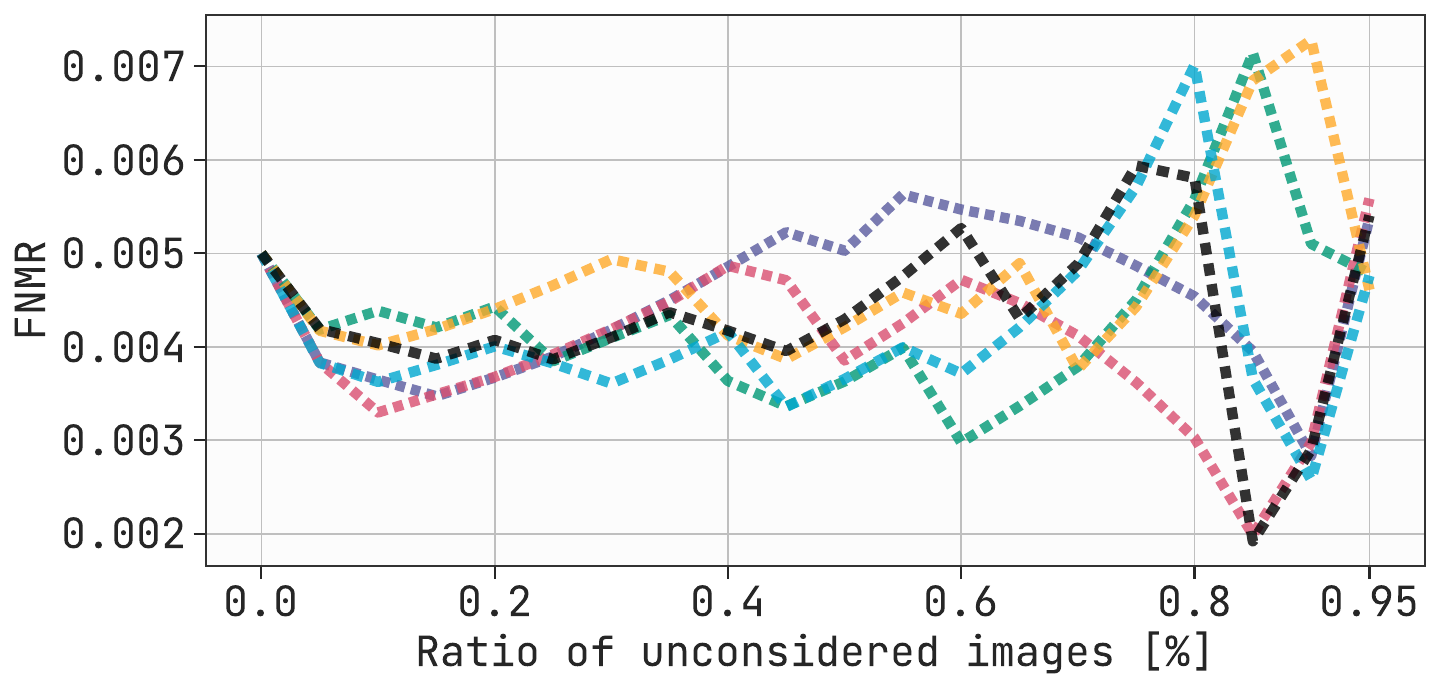}
		 \caption{MagFace \cite{meng_2021_magface} Model, LFW \cite{LFWTech} Dataset \\ \grafiqs with ResNet100, FMR$=1e-4$}
	\end{subfigure}
\\
	\begin{subfigure}[b]{0.48\textwidth}
		 \centering
		 \includegraphics[width=0.95\textwidth]{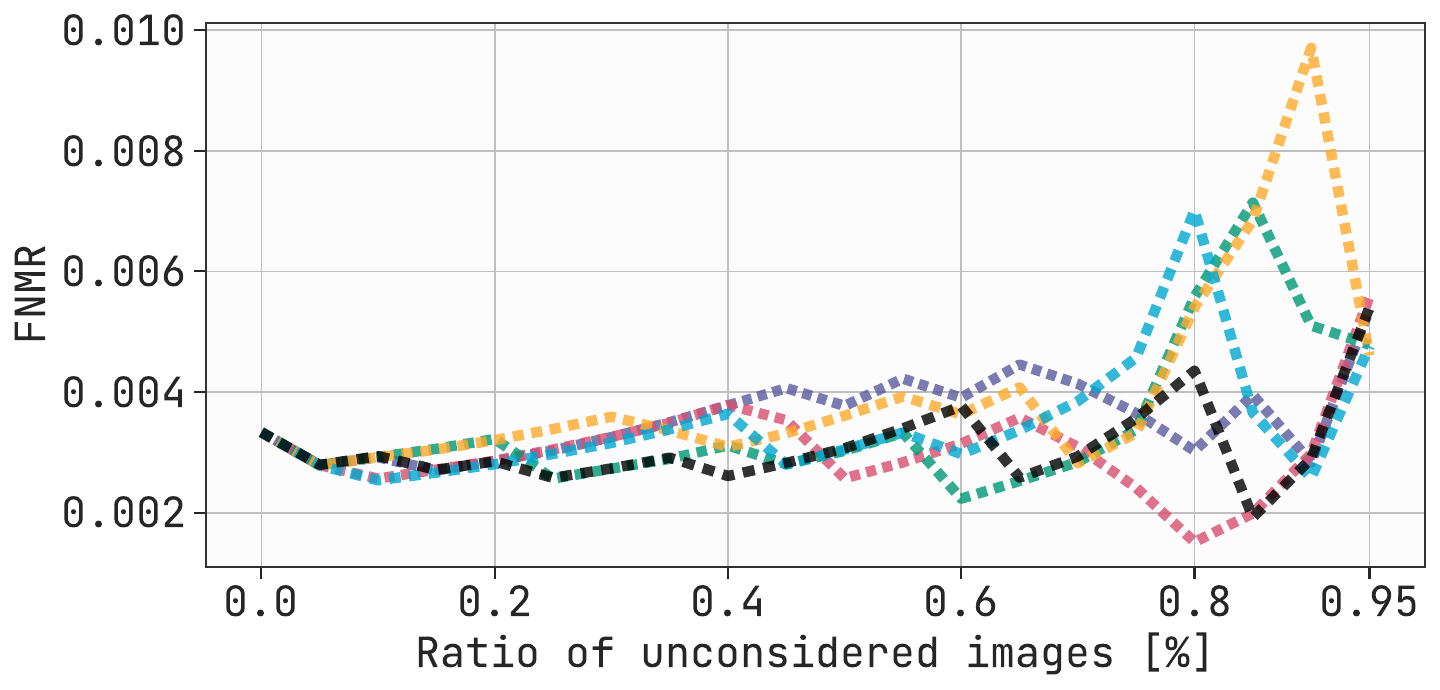}
		 \caption{CurricularFace \cite{curricularFace} Model, LFW \cite{LFWTech} Dataset \\ \grafiqs with ResNet100, FMR$=1e-3$}
	\end{subfigure}
\hfill
	\begin{subfigure}[b]{0.48\textwidth}
		 \centering
		 \includegraphics[width=0.95\textwidth]{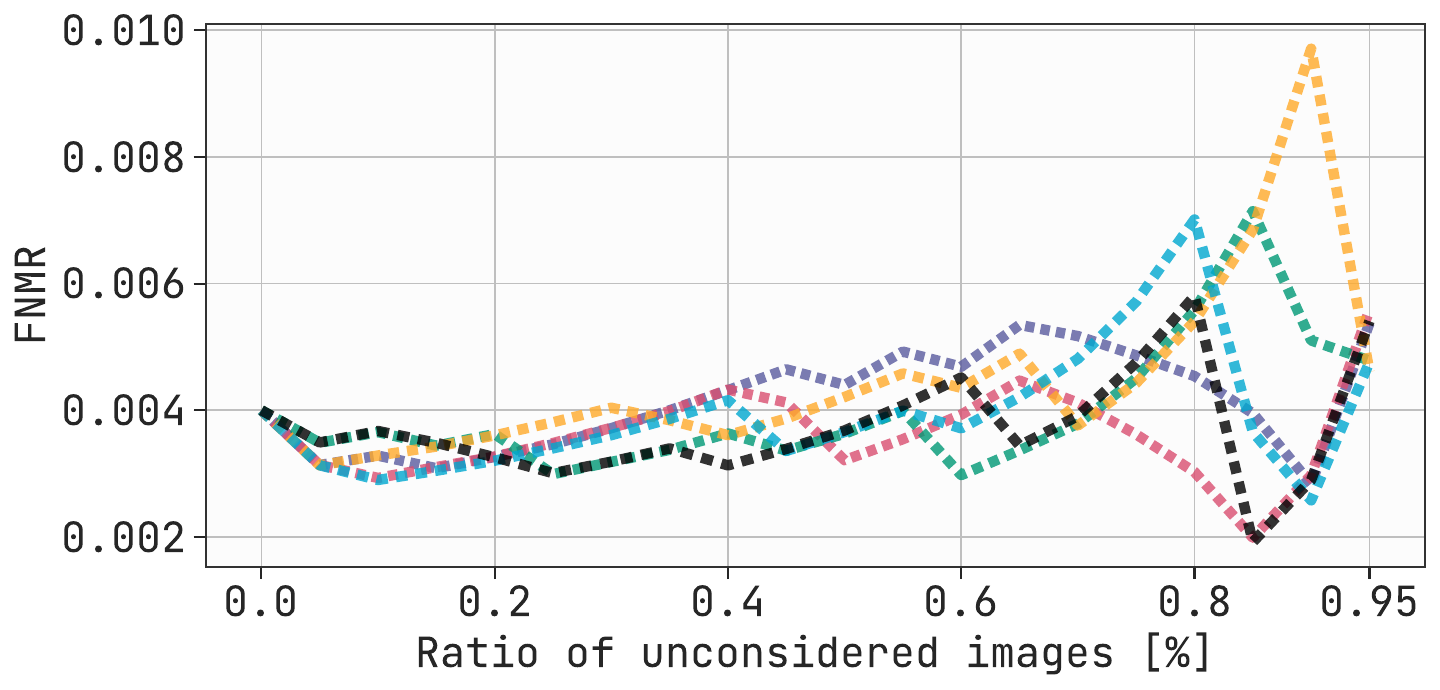}
		 \caption{CurricularFace \cite{curricularFace} Model, LFW \cite{LFWTech} Dataset \\ \grafiqs with ResNet100, FMR$=1e-4$}
	\end{subfigure}
\\
\caption{Error-versus-Discard Characteristic (EDC) curves for FNMR@FMR=$1e-3$ and FNMR@FMR=$1e-4$ of our proposed method using $\mathcal{L}_{\text{BNS}}$ as backpropagation loss and absolute sum as FIQ. The gradients at image level ($\phi=\mathcal{I}$), and block levels ($\phi=\text{B}1$ $-$ $\phi=\text{B}4$) are used to calculate FIQ. $\text{MSE}_{\text{BNS}}$ as FIQ is shown in black. Results shown on benchmark LFW \cite{LFWTech} using ArcFace, ElasticFace, MagFace, and, CurricularFace FR models. It is evident that the proposed \grafiqs method leads to lower verification error when images with lowest utility score estimated from gradient magnitudes are rejected. Furthermore, estimating FIQ by backpropagating $\mathcal{L}_{\text{BNS}}$ yields significantly better results than using $\text{MSE}_{\text{BNS}}$ directly.}
\vspace{-4mm}
\label{fig:iresnet100_supplementary_lfw}
\end{figure*}

%% file: figures/fig_iresnet100_supplementary_calfw.tex
\begin{figure*}[h!]
\centering
	\begin{subfigure}[b]{0.9\textwidth}
		\centering
		\includegraphics[width=\textwidth]{figures/iresnet100_bn_overview/legend.pdf}
	\end{subfigure}
\\
	\begin{subfigure}[b]{0.48\textwidth}
		 \centering
		 \includegraphics[width=0.95\textwidth]{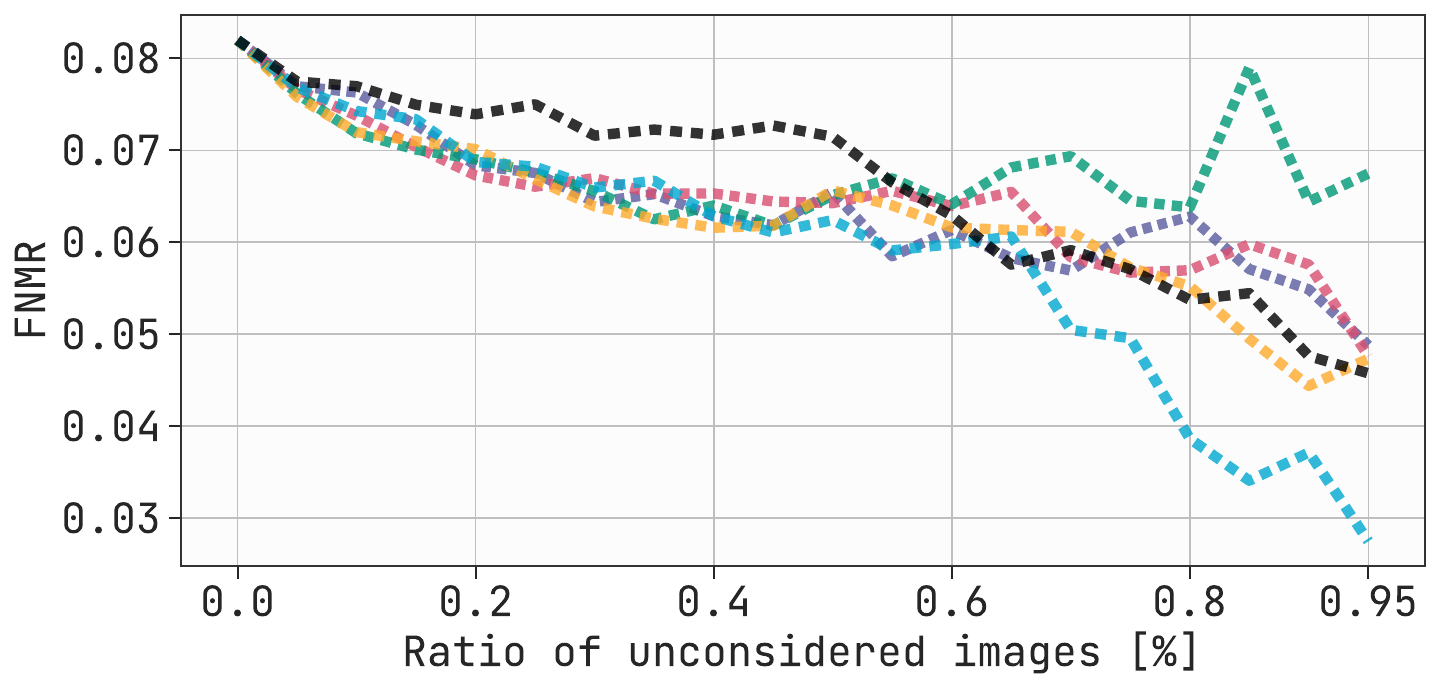}
		 \caption{ArcFace \cite{deng2019arcface} Model, CALFW \cite{CALFW} Dataset \\ \grafiqs with ResNet100, FMR$=1e-3$}
	\end{subfigure}
\hfill
	\begin{subfigure}[b]{0.48\textwidth}
		 \centering
		 \includegraphics[width=0.95\textwidth]{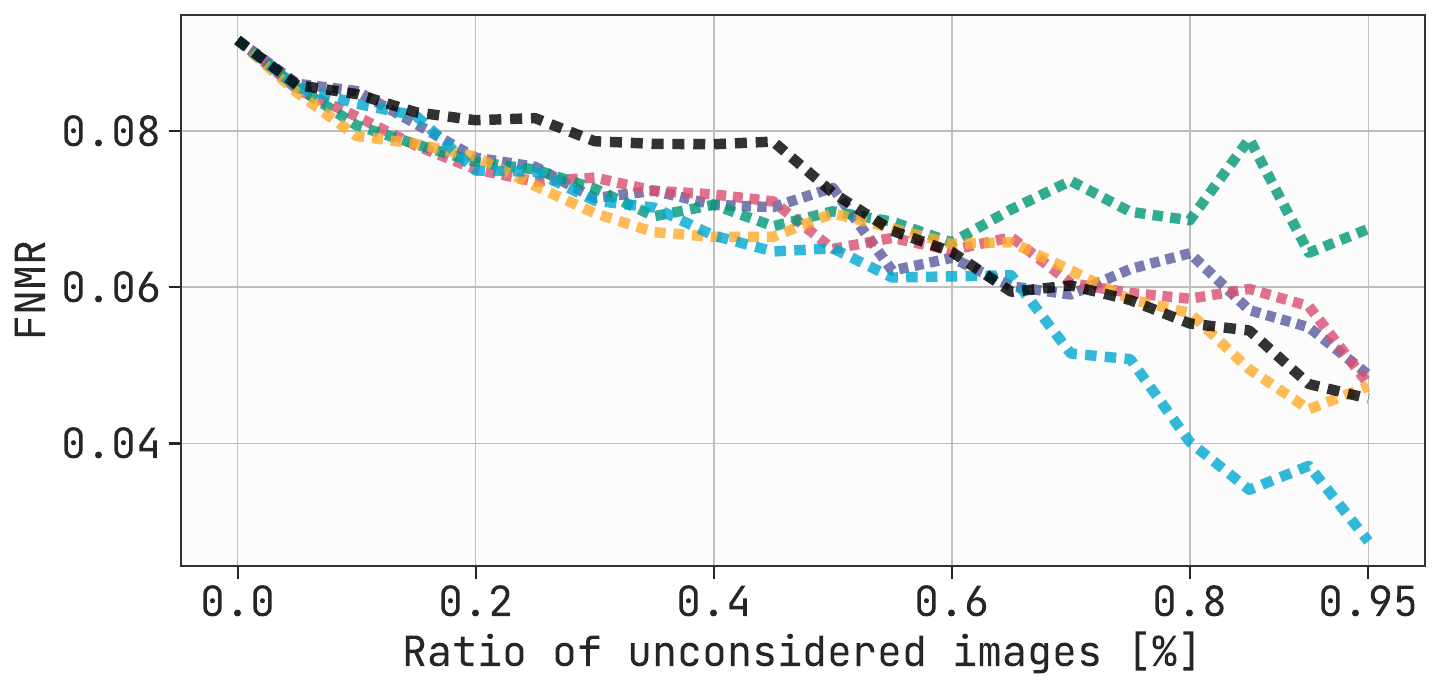}
		 \caption{ArcFace \cite{deng2019arcface} Model, CALFW \cite{CALFW} Dataset \\ \grafiqs with ResNet100, FMR$=1e-4$}
	\end{subfigure}
\\
	\begin{subfigure}[b]{0.48\textwidth}
		 \centering
		 \includegraphics[width=0.95\textwidth]{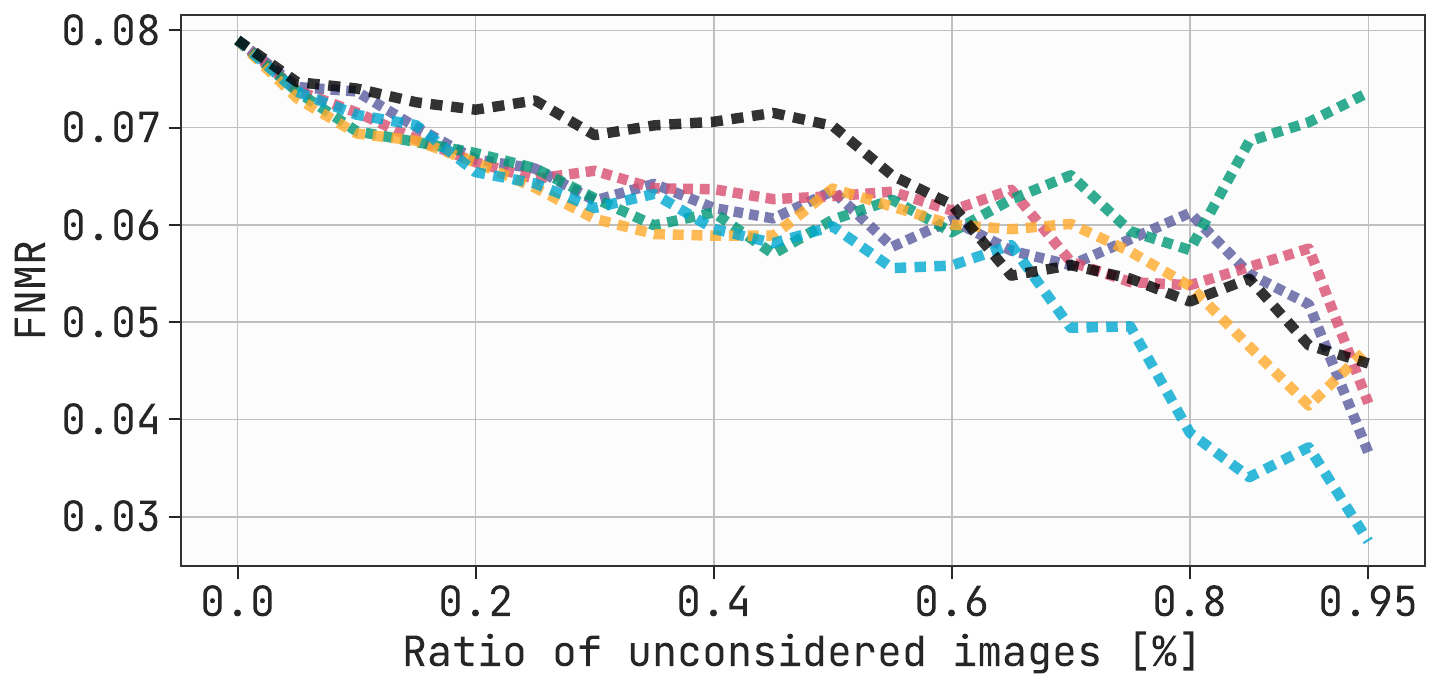}
		 \caption{ElasticFace \cite{elasticface} Model, CALFW \cite{CALFW} Dataset \\ \grafiqs with ResNet100, FMR$=1e-3$}
	\end{subfigure}
\hfill
	\begin{subfigure}[b]{0.48\textwidth}
		 \centering
		 \includegraphics[width=0.95\textwidth]{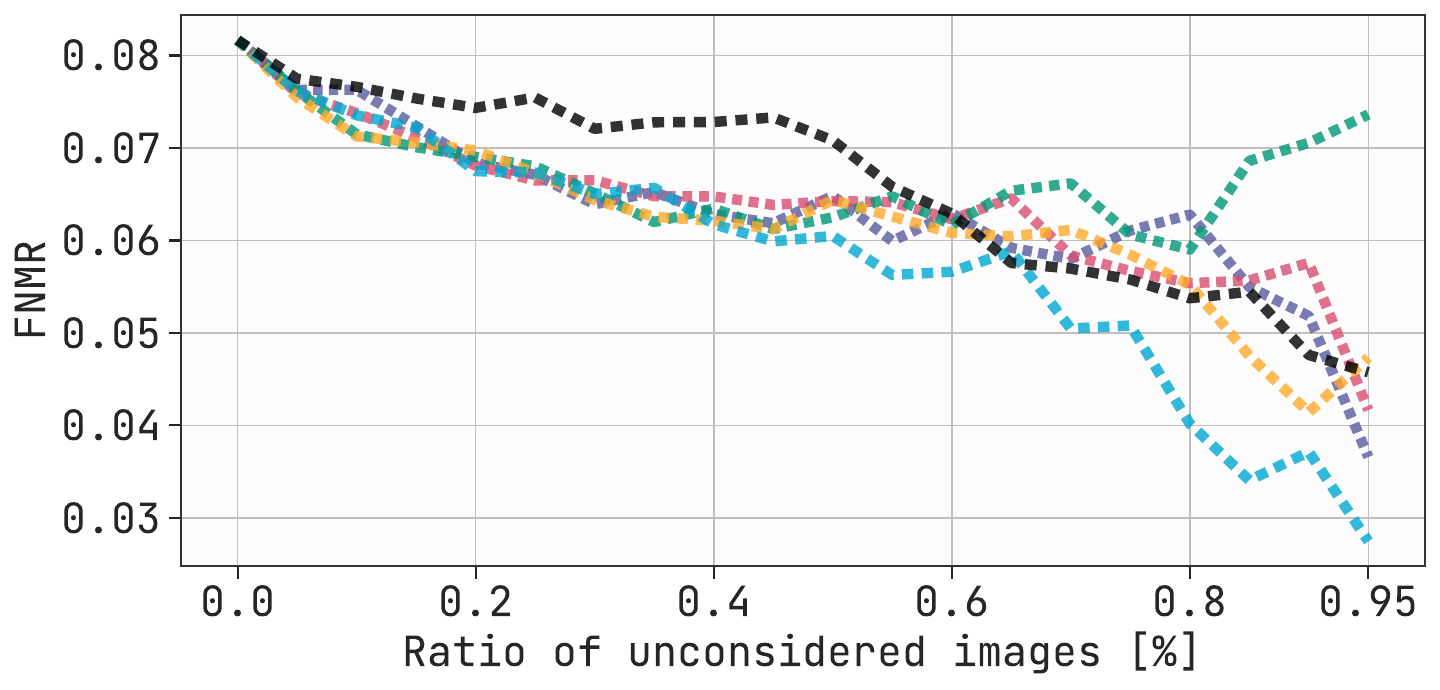}
		 \caption{ElasticFace \cite{elasticface} Model, CALFW \cite{CALFW} Dataset \\ \grafiqs with ResNet100, FMR$=1e-4$}
	\end{subfigure}
\\
	\begin{subfigure}[b]{0.48\textwidth}
		 \centering
		 \includegraphics[width=0.95\textwidth]{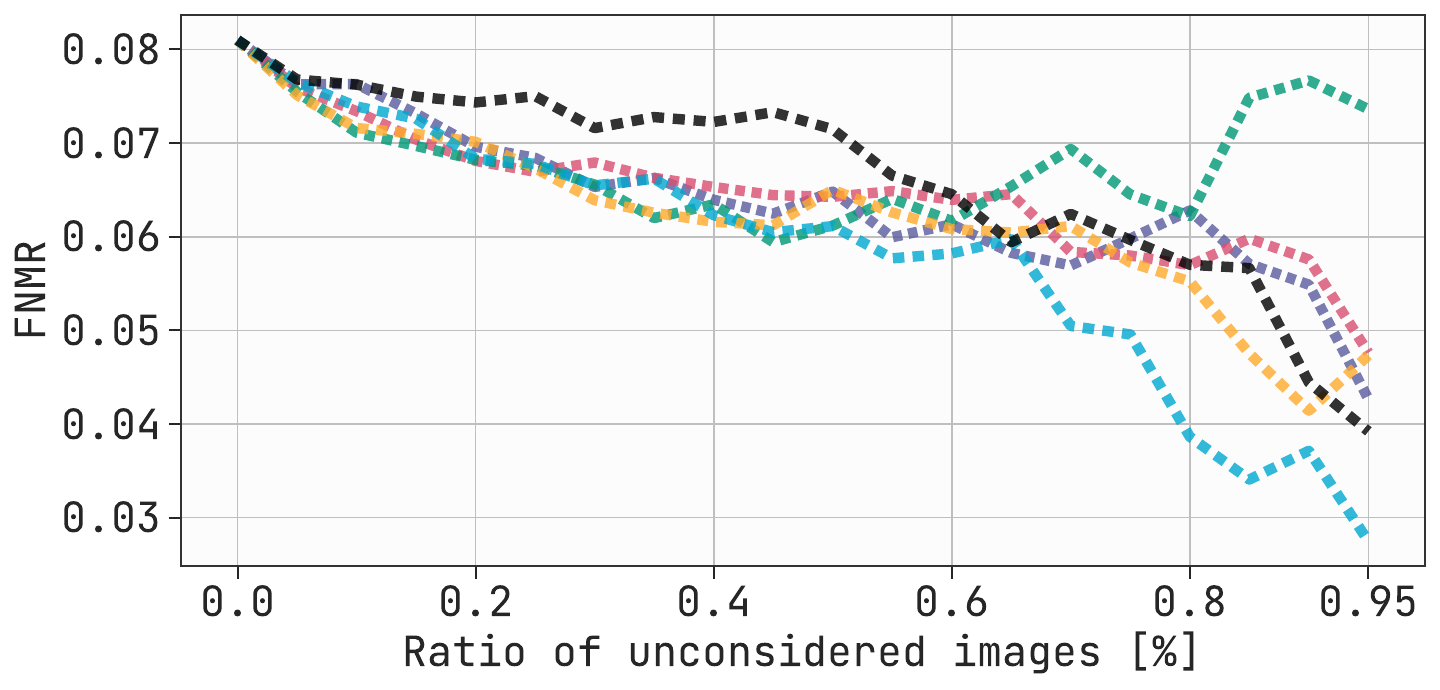}
		 \caption{MagFace \cite{meng_2021_magface} Model, CALFW \cite{CALFW} Dataset \\ \grafiqs with ResNet100, FMR$=1e-3$}
	\end{subfigure}
\hfill
	\begin{subfigure}[b]{0.48\textwidth}
		 \centering
		 \includegraphics[width=0.95\textwidth]{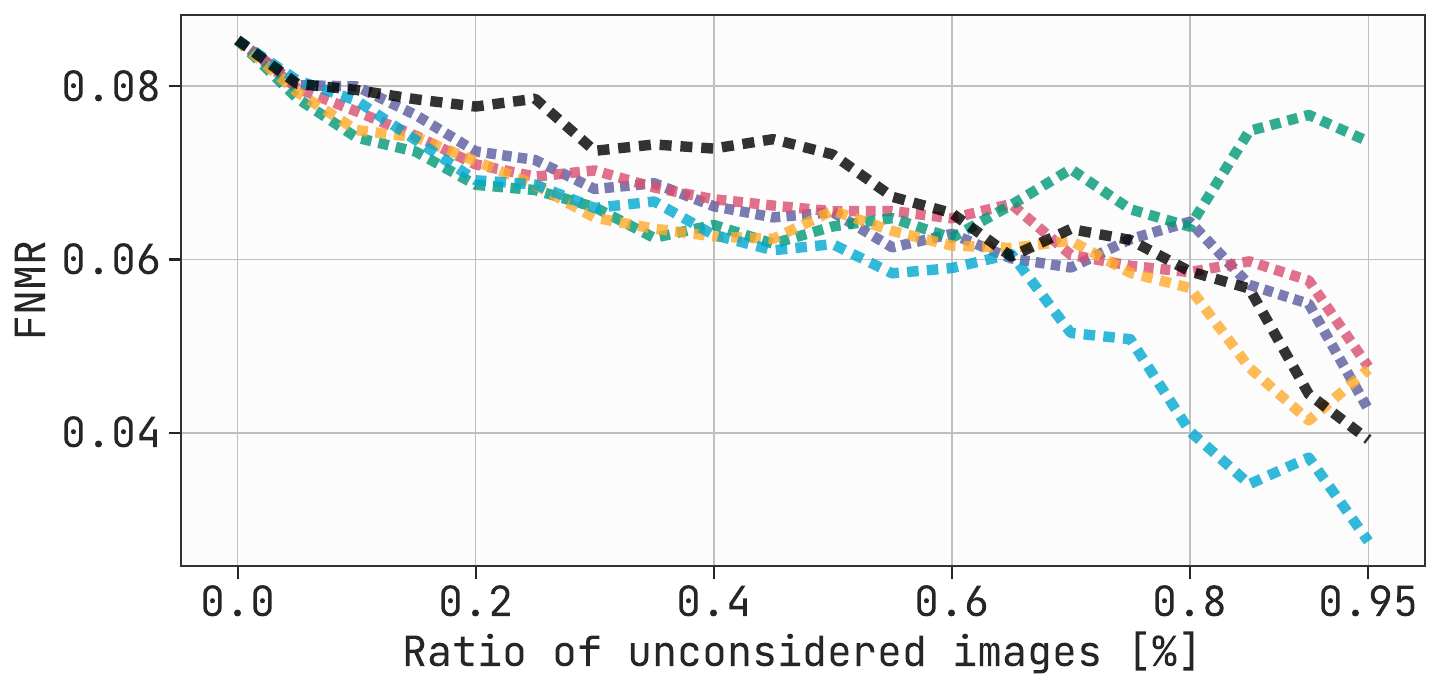}
		 \caption{MagFace \cite{meng_2021_magface} Model, CALFW \cite{CALFW} Dataset \\ \grafiqs with ResNet100, FMR$=1e-4$}
	\end{subfigure}
\\
	\begin{subfigure}[b]{0.48\textwidth}
		 \centering
		 \includegraphics[width=0.95\textwidth]{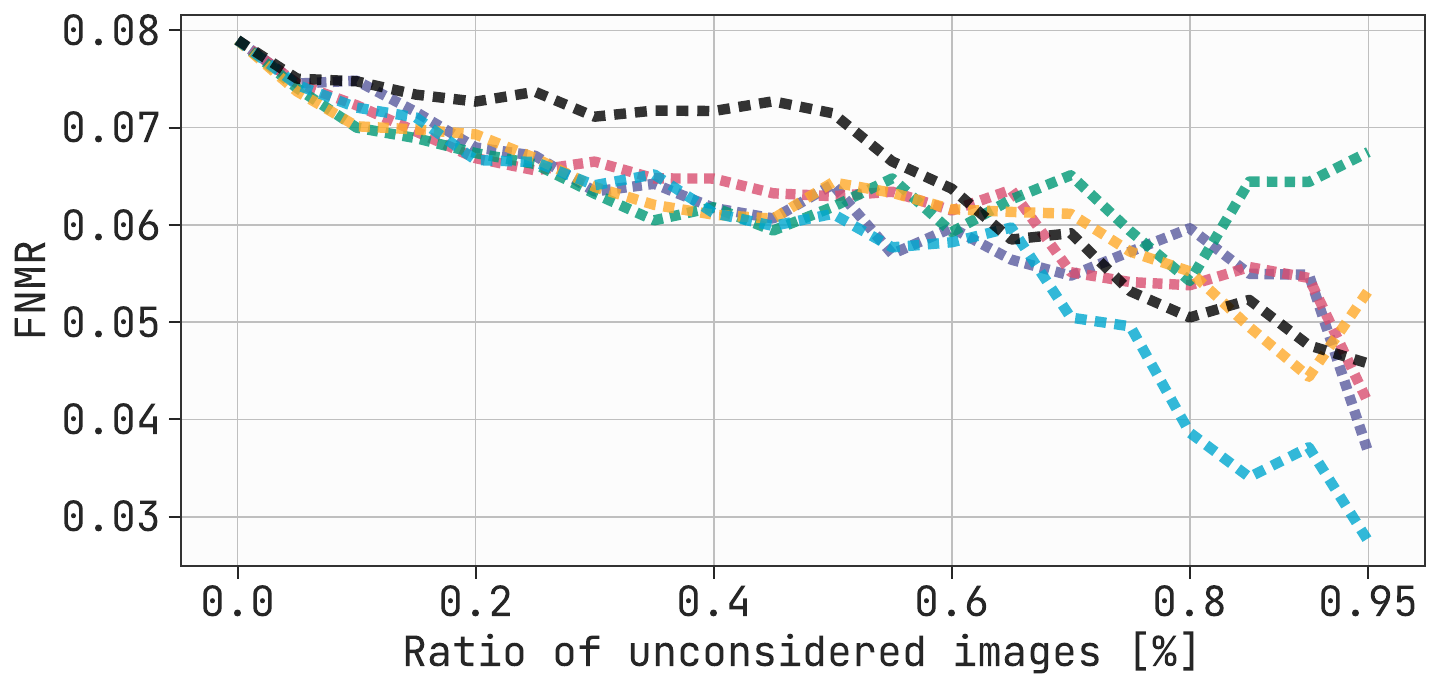}
		 \caption{CurricularFace \cite{curricularFace} Model, CALFW \cite{CALFW} Dataset \\ \grafiqs with ResNet100, FMR$=1e-3$}
	\end{subfigure}
\hfill
	\begin{subfigure}[b]{0.48\textwidth}
		 \centering
		 \includegraphics[width=0.95\textwidth]{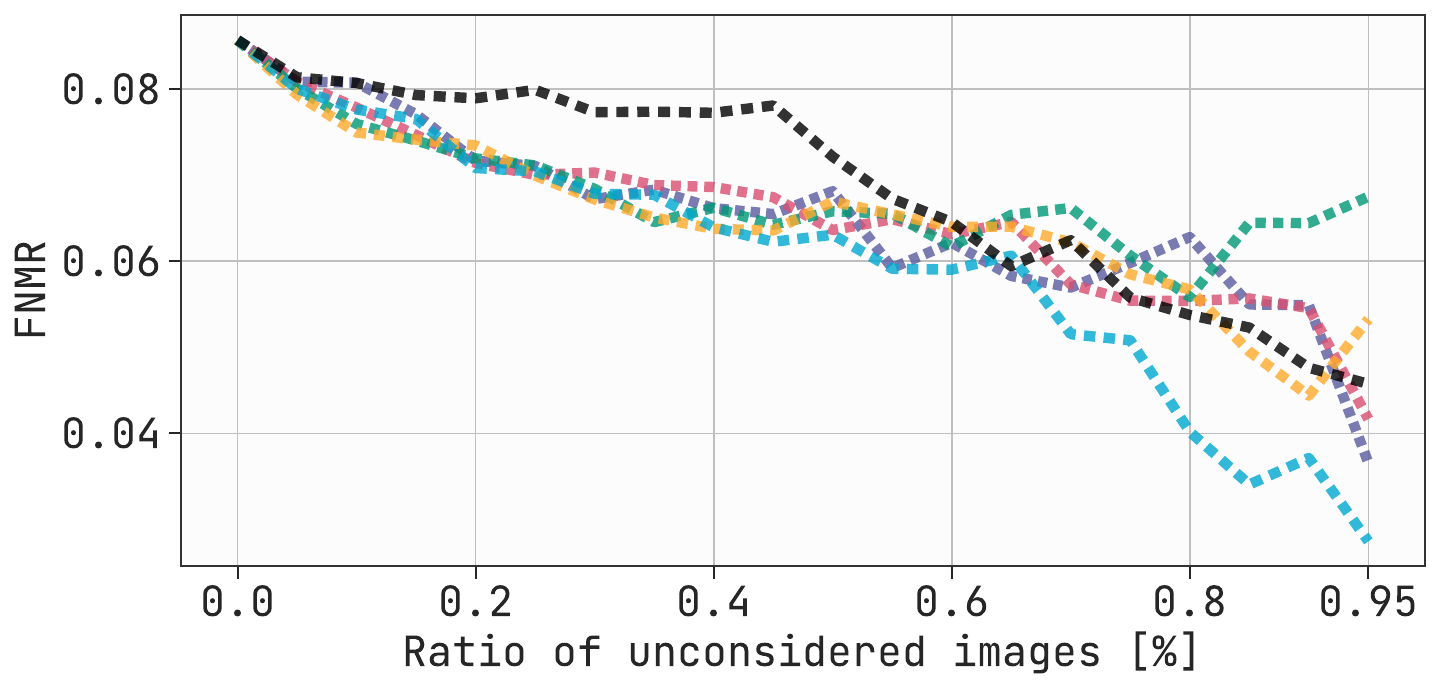}
		 \caption{CurricularFace \cite{curricularFace} Model, CALFW \cite{CALFW} Dataset \\ \grafiqs with ResNet100, FMR$=1e-4$}
	\end{subfigure}
\\
\caption{Error-versus-Discard Characteristic (EDC) curves for FNMR@FMR=$1e-3$ and FNMR@FMR=$1e-4$ of our proposed method using $\mathcal{L}_{\text{BNS}}$ as backpropagation loss and absolute sum as FIQ. The gradients at image level ($\phi=\mathcal{I}$), and block levels ($\phi=\text{B}1$ $-$ $\phi=\text{B}4$) are used to calculate FIQ. $\text{MSE}_{\text{BNS}}$ as FIQ is shown in black. Results shown on benchmark CALFW \cite{CALFW} using ArcFace, ElasticFace, MagFace, and, CurricularFace FR models. It is evident that the proposed \grafiqs method leads to lower verification error when images with lowest utility score estimated from gradient magnitudes are rejected. Furthermore, estimating FIQ by backpropagating $\mathcal{L}_{\text{BNS}}$ yields significantly better results than using $\text{MSE}_{\text{BNS}}$ directly.}
\vspace{-4mm}
\label{fig:iresnet100_supplementary_calfw}
\end{figure*}

%% file: figures/fig_iresnet100_supplementary_cplfw.tex
\begin{figure*}[h!]
\centering
	\begin{subfigure}[b]{0.9\textwidth}
		\centering
		\includegraphics[width=\textwidth]{figures/iresnet100_bn_overview/legend.pdf}
	\end{subfigure}
\\
	\begin{subfigure}[b]{0.48\textwidth}
		 \centering
		 \includegraphics[width=0.95\textwidth]{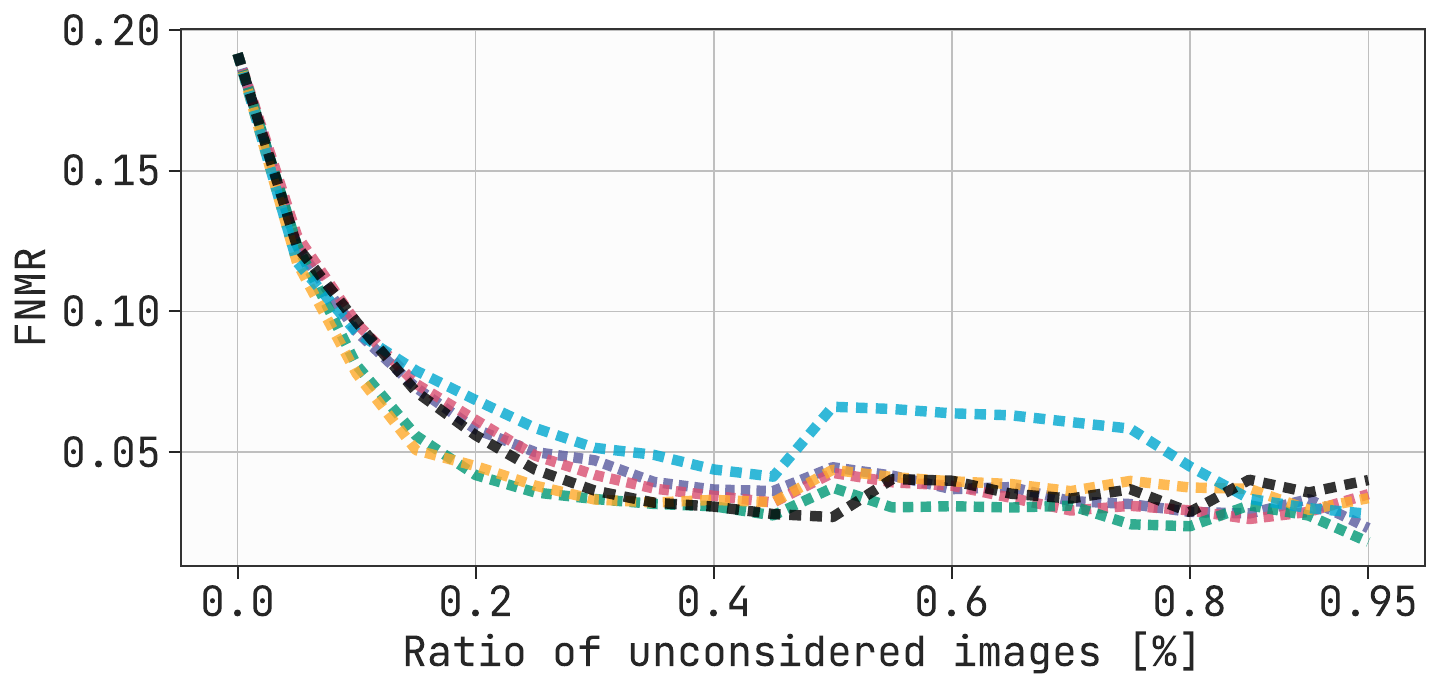}
		 \caption{ArcFace \cite{deng2019arcface} Model, CPLFW \cite{CPLFWTech} Dataset \\ \grafiqs with ResNet100, FMR$=1e-3$}
	\end{subfigure}
\hfill
	\begin{subfigure}[b]{0.48\textwidth}
		 \centering
		 \includegraphics[width=0.95\textwidth]{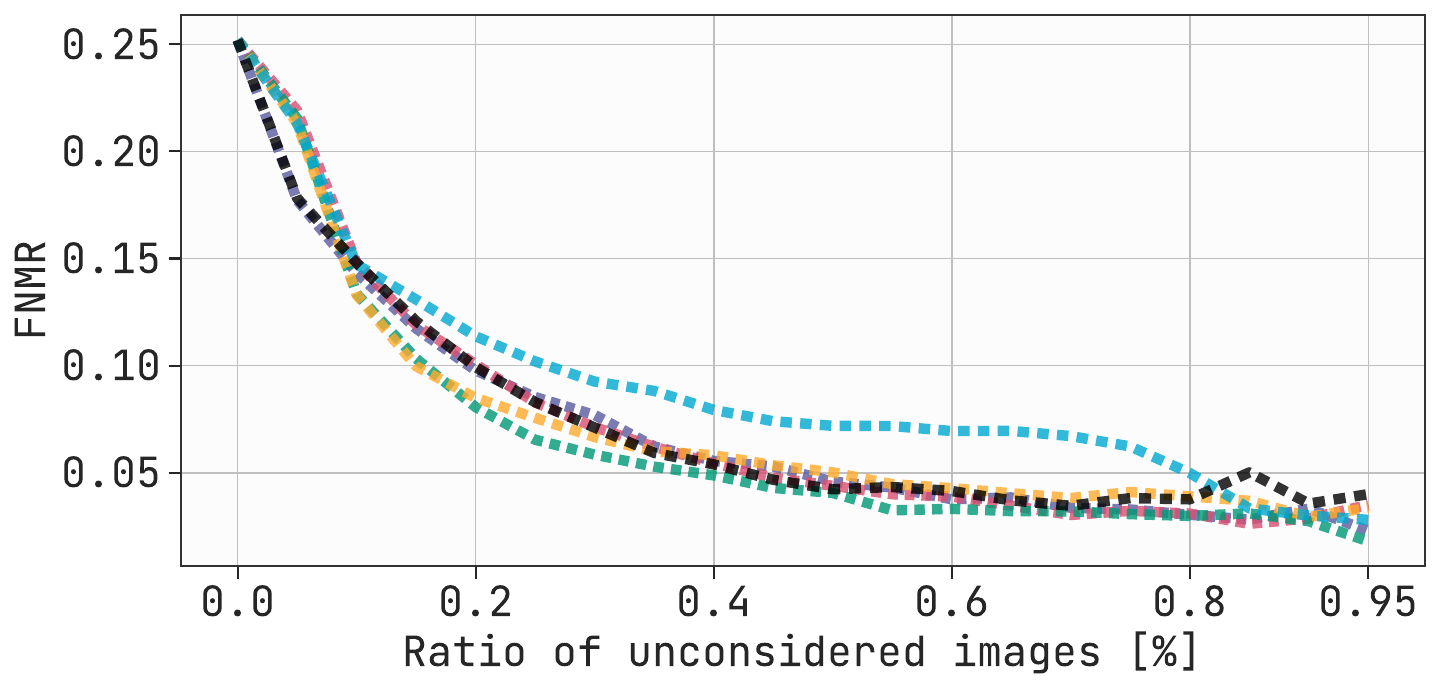}
		 \caption{ArcFace \cite{deng2019arcface} Model, CPLFW \cite{CPLFWTech} Dataset \\ \grafiqs with ResNet100, FMR$=1e-4$}
	\end{subfigure}
\\
	\begin{subfigure}[b]{0.48\textwidth}
		 \centering
		 \includegraphics[width=0.95\textwidth]{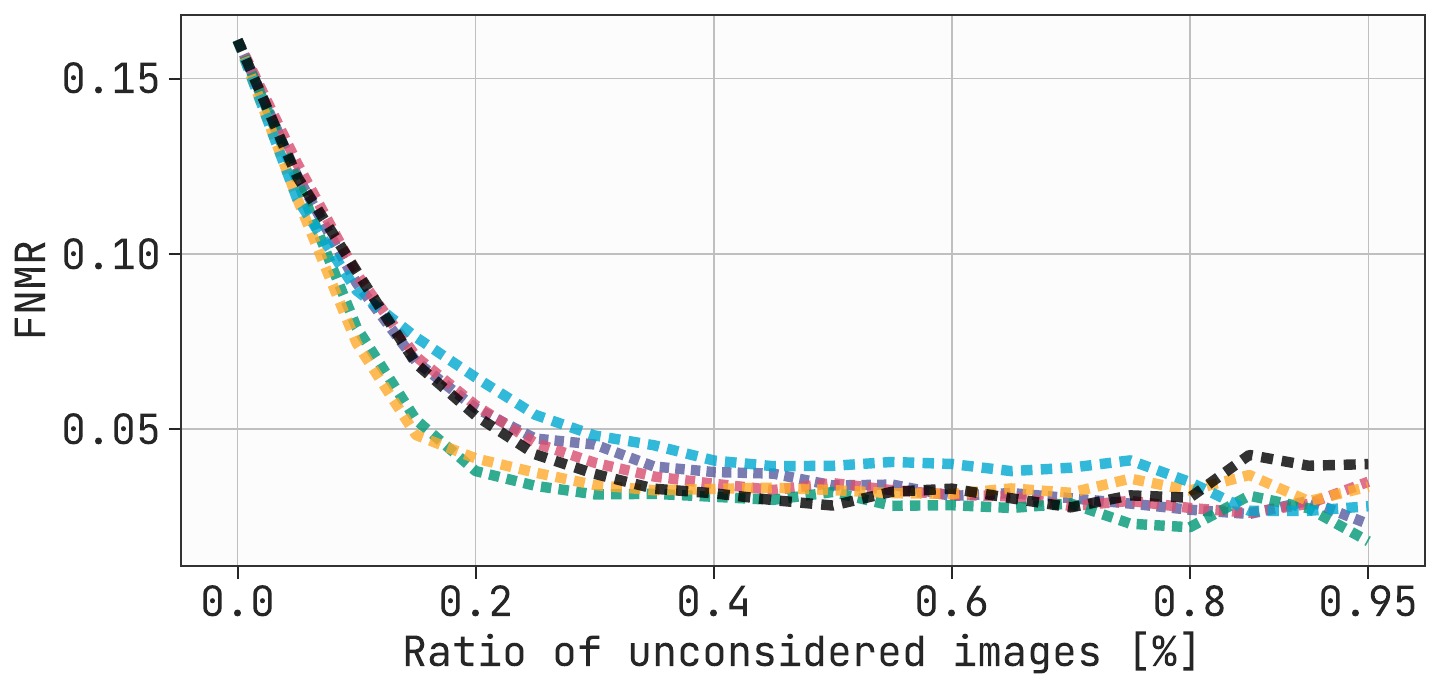}
		 \caption{ElasticFace \cite{elasticface} Model, CPLFW \cite{CPLFWTech} Dataset \\ \grafiqs with ResNet100, FMR$=1e-3$}
	\end{subfigure}
\hfill
	\begin{subfigure}[b]{0.48\textwidth}
		 \centering
		 \includegraphics[width=0.95\textwidth]{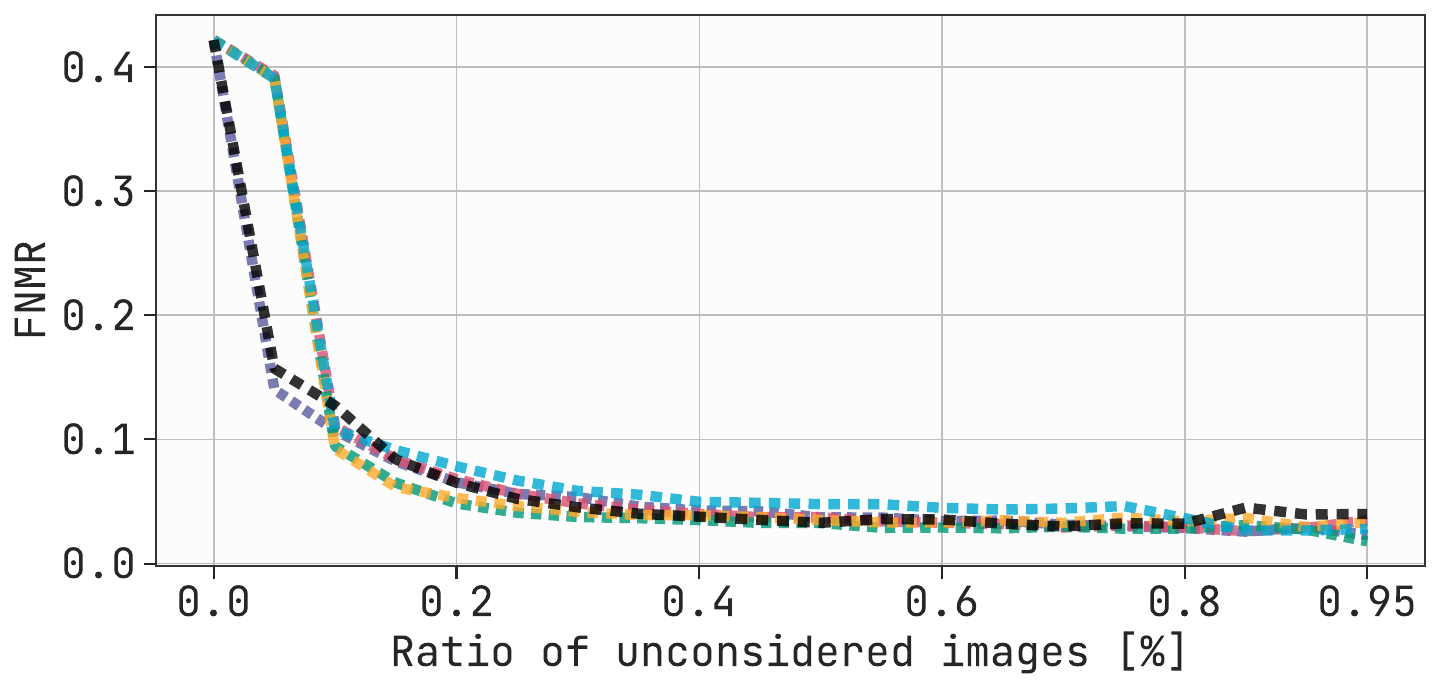}
		 \caption{ElasticFace \cite{elasticface} Model, CPLFW \cite{CPLFWTech} Dataset \\ \grafiqs with ResNet100, FMR$=1e-4$}
	\end{subfigure}
\\
	\begin{subfigure}[b]{0.48\textwidth}
		 \centering
		 \includegraphics[width=0.95\textwidth]{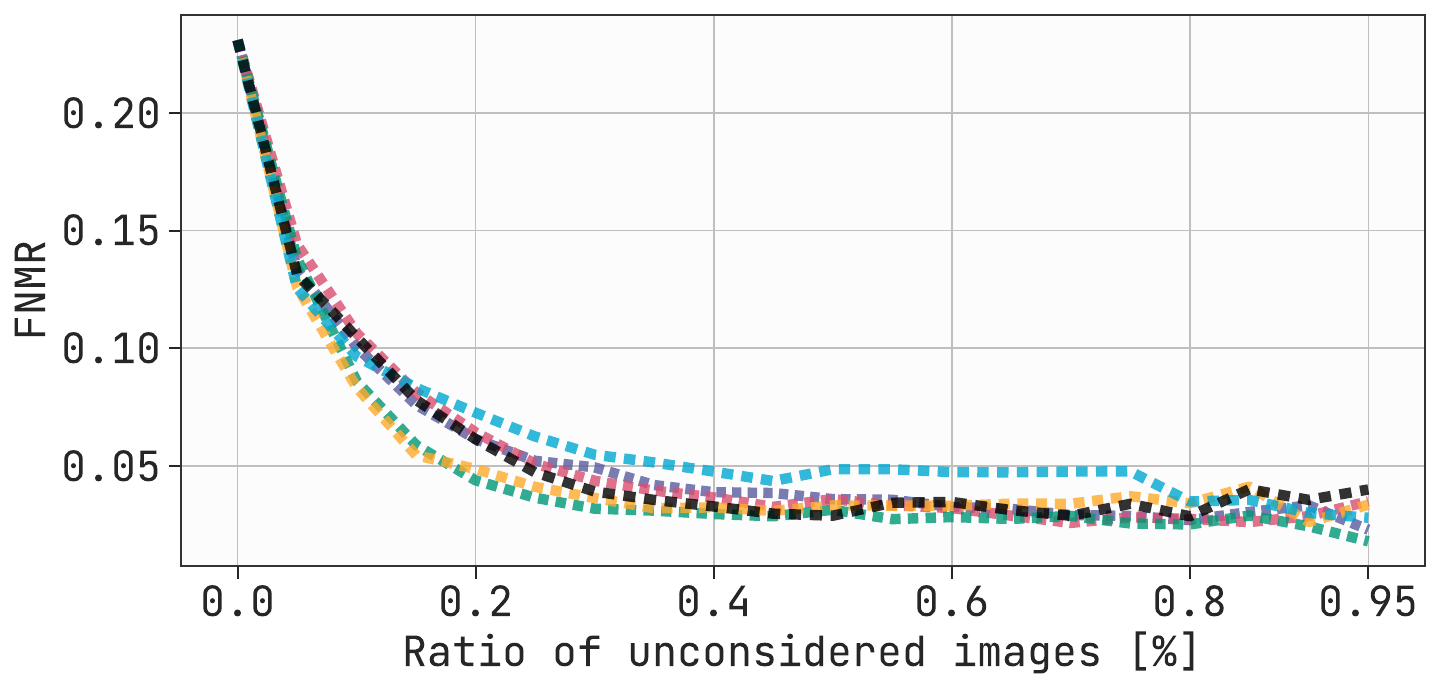}
		 \caption{MagFace \cite{meng_2021_magface} Model, CPLFW \cite{CPLFWTech} Dataset \\ \grafiqs with ResNet100, FMR$=1e-3$}
	\end{subfigure}
\hfill
	\begin{subfigure}[b]{0.48\textwidth}
		 \centering
		 \includegraphics[width=0.95\textwidth]{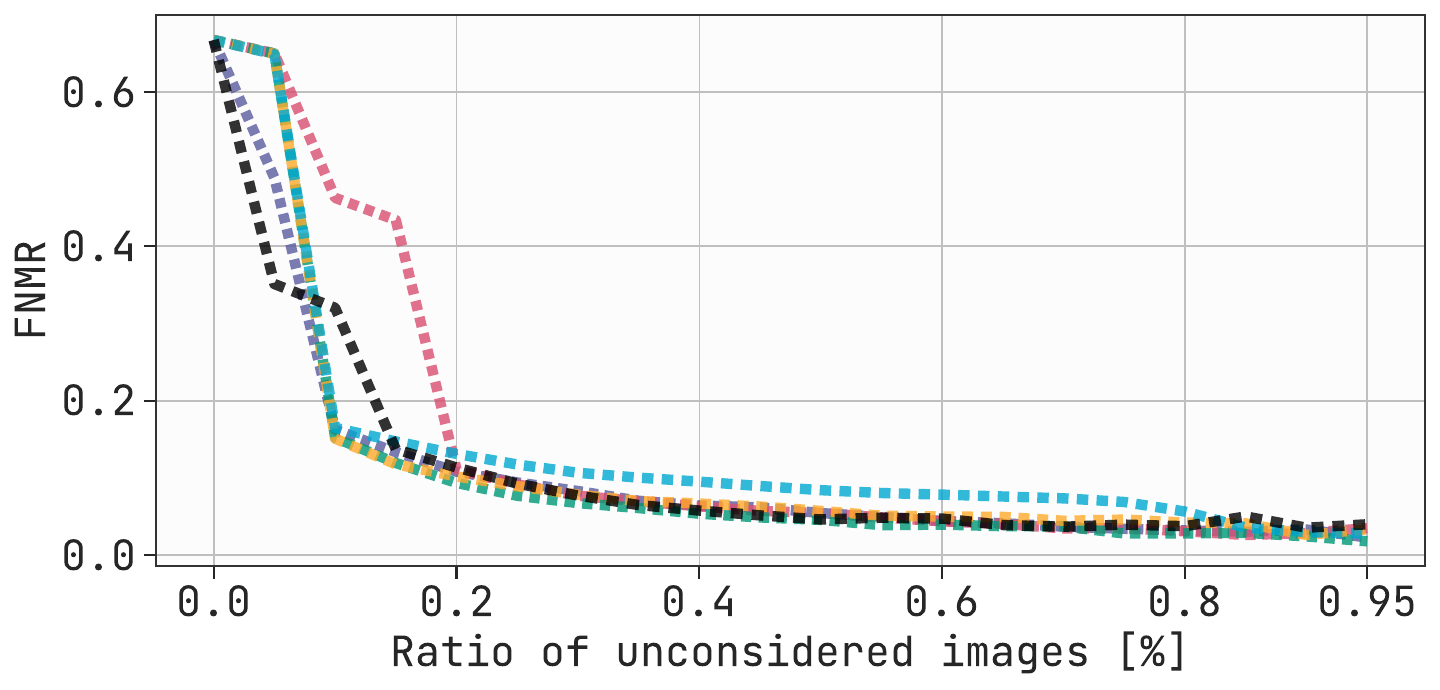}
		 \caption{MagFace \cite{meng_2021_magface} Model, CPLFW \cite{CPLFWTech} Dataset \\ \grafiqs with ResNet100, FMR$=1e-4$}
	\end{subfigure}
\\
	\begin{subfigure}[b]{0.48\textwidth}
		 \centering
		 \includegraphics[width=0.95\textwidth]{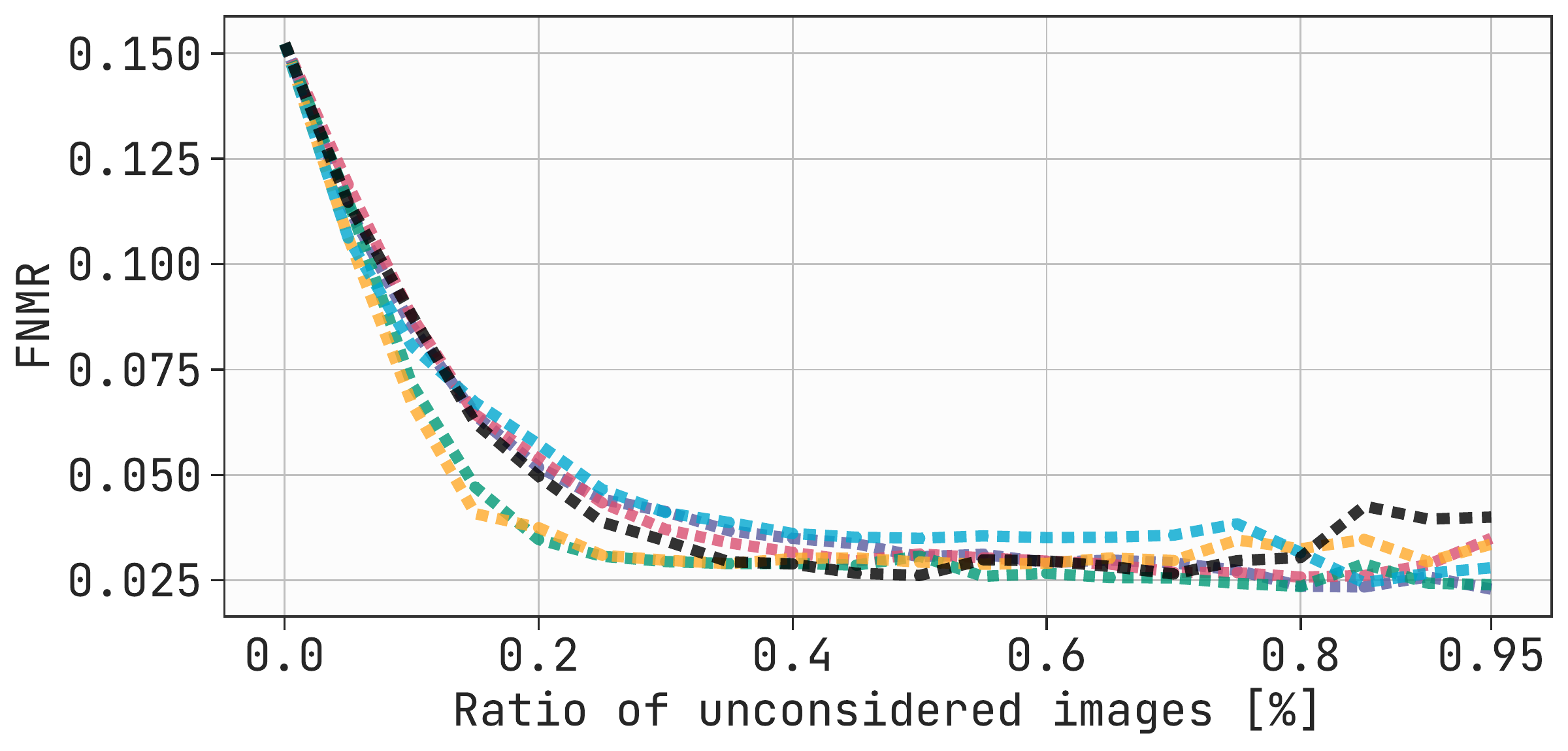}
		 \caption{CurricularFace \cite{curricularFace} Model, CPLFW \cite{CPLFWTech} Dataset \\ \grafiqs with ResNet100, FMR$=1e-3$}
	\end{subfigure}
\hfill
	\begin{subfigure}[b]{0.48\textwidth}
		 \centering
		 \includegraphics[width=0.95\textwidth]{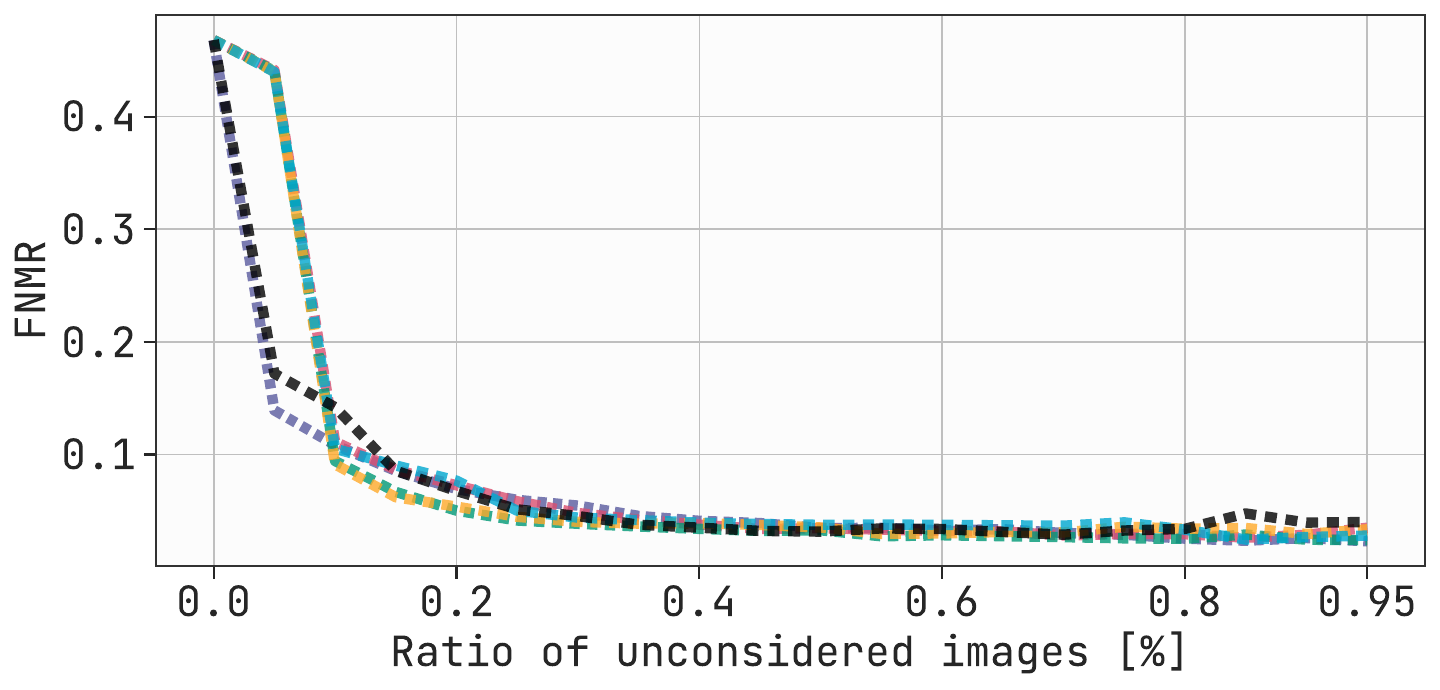}
		 \caption{CurricularFace \cite{curricularFace} Model, CPLFW \cite{CPLFWTech} Dataset \\ \grafiqs with ResNet100, FMR$=1e-4$}
	\end{subfigure}
\\
\caption{Error-versus-Discard Characteristic (EDC) curves for FNMR@FMR=$1e-3$ and FNMR@FMR=$1e-4$ of our proposed method using $\mathcal{L}_{\text{BNS}}$ as backpropagation loss and absolute sum as FIQ. The gradients at image level ($\phi=\mathcal{I}$), and block levels ($\phi=\text{B}1$ $-$ $\phi=\text{B}4$) are used to calculate FIQ. $\text{MSE}_{\text{BNS}}$ as FIQ is shown in black. Results shown on benchmark CPLFW \cite{CPLFWTech} using ArcFace, ElasticFace, MagFace, and, CurricularFace FR models. It is evident that the proposed \grafiqs method leads to lower verification error when images with lowest utility score estimated from gradient magnitudes are rejected. Furthermore, estimating FIQ by backpropagating $\mathcal{L}_{\text{BNS}}$ yields significantly better results than using $\text{MSE}_{\text{BNS}}$ directly.}
\vspace{-4mm}
\label{fig:iresnet100_supplementary_cplfw}
\end{figure*}

%% file: figures/fig_iresnet100_supplementary_XQLFW.tex
\begin{figure*}[h!]
\centering
	\begin{subfigure}[b]{0.9\textwidth}
		\centering
		\includegraphics[width=\textwidth]{figures/iresnet100_bn_overview/legend.pdf}
	\end{subfigure}
\\
	\begin{subfigure}[b]{0.48\textwidth}
		 \centering
		 \includegraphics[width=0.95\textwidth]{figures/iresnet100_bn_overview/XQLFW/ArcFaceModel/iresnet100_bn_combined_ArcFaceModel_XQLFW_fnmr3.pdf}
		 \caption{ArcFace \cite{deng2019arcface} Model, XQLFW \cite{XQLFW} Dataset \\ \grafiqs with ResNet100, FMR$=1e-3$}
	\end{subfigure}
\hfill
	\begin{subfigure}[b]{0.48\textwidth}
		 \centering
		 \includegraphics[width=0.95\textwidth]{figures/iresnet100_bn_overview/XQLFW/ArcFaceModel/iresnet100_bn_combined_ArcFaceModel_XQLFW_fnmr4.pdf}
		 \caption{ArcFace \cite{deng2019arcface} Model, XQLFW \cite{XQLFW} Dataset \\ \grafiqs with ResNet100, FMR$=1e-4$}
	\end{subfigure}
\\
	\begin{subfigure}[b]{0.48\textwidth}
		 \centering
		 \includegraphics[width=0.95\textwidth]{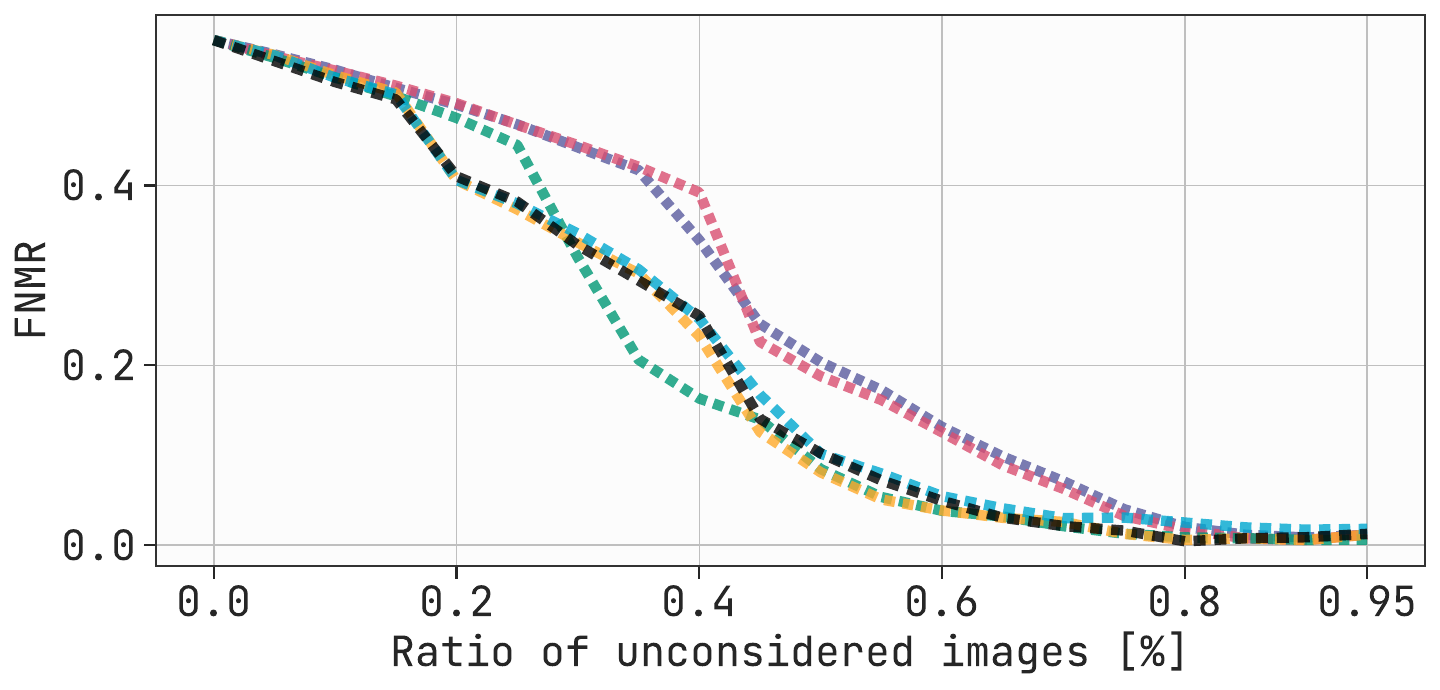}
		 \caption{ElasticFace \cite{elasticface} Model, XQLFW \cite{XQLFW} Dataset \\ \grafiqs with ResNet100, FMR$=1e-3$}
	\end{subfigure}
\hfill
	\begin{subfigure}[b]{0.48\textwidth}
		 \centering
		 \includegraphics[width=0.95\textwidth]{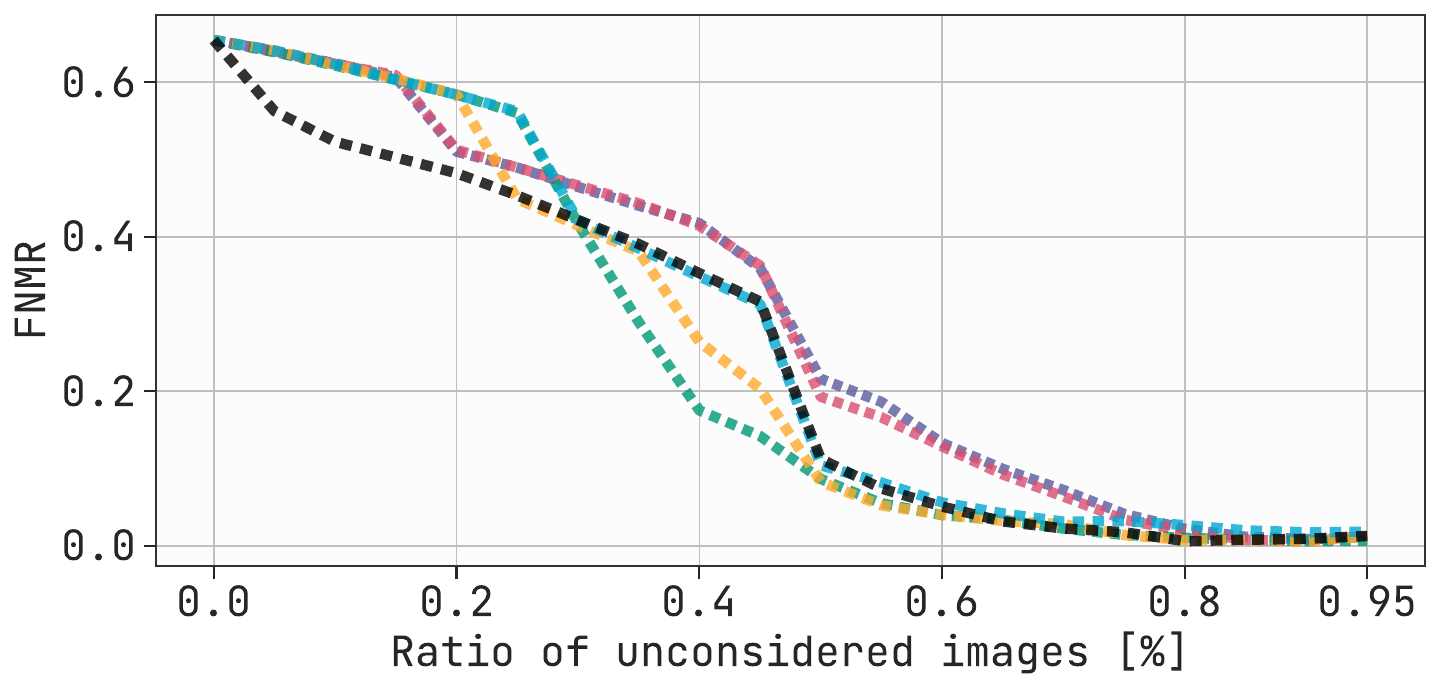}
		 \caption{ElasticFace \cite{elasticface} Model, XQLFW \cite{XQLFW} Dataset \\ \grafiqs with ResNet100, FMR$=1e-4$}
	\end{subfigure}
\\
	\begin{subfigure}[b]{0.48\textwidth}
		 \centering
		 \includegraphics[width=0.95\textwidth]{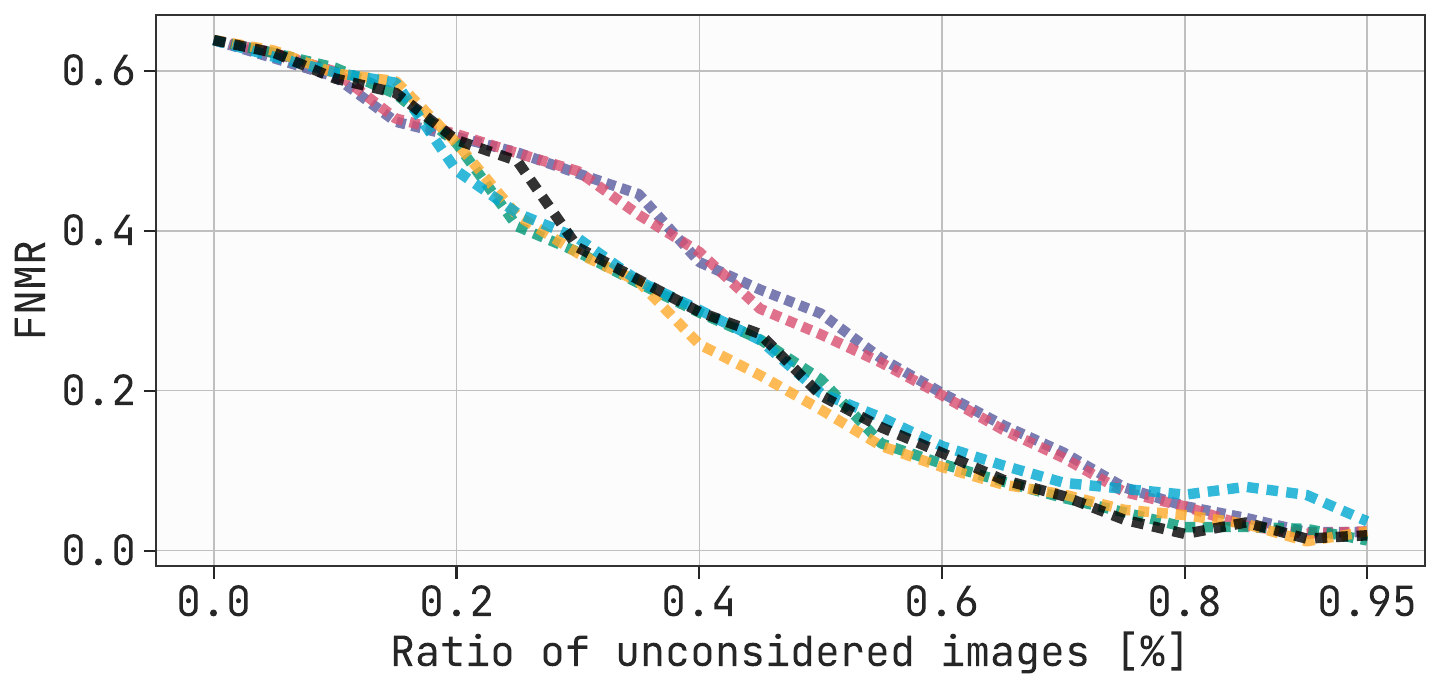}
		 \caption{MagFace \cite{meng_2021_magface} Model, XQLFW \cite{XQLFW} Dataset \\ \grafiqs with ResNet100, FMR$=1e-3$}
	\end{subfigure}
\hfill
	\begin{subfigure}[b]{0.48\textwidth}
		 \centering
		 \includegraphics[width=0.95\textwidth]{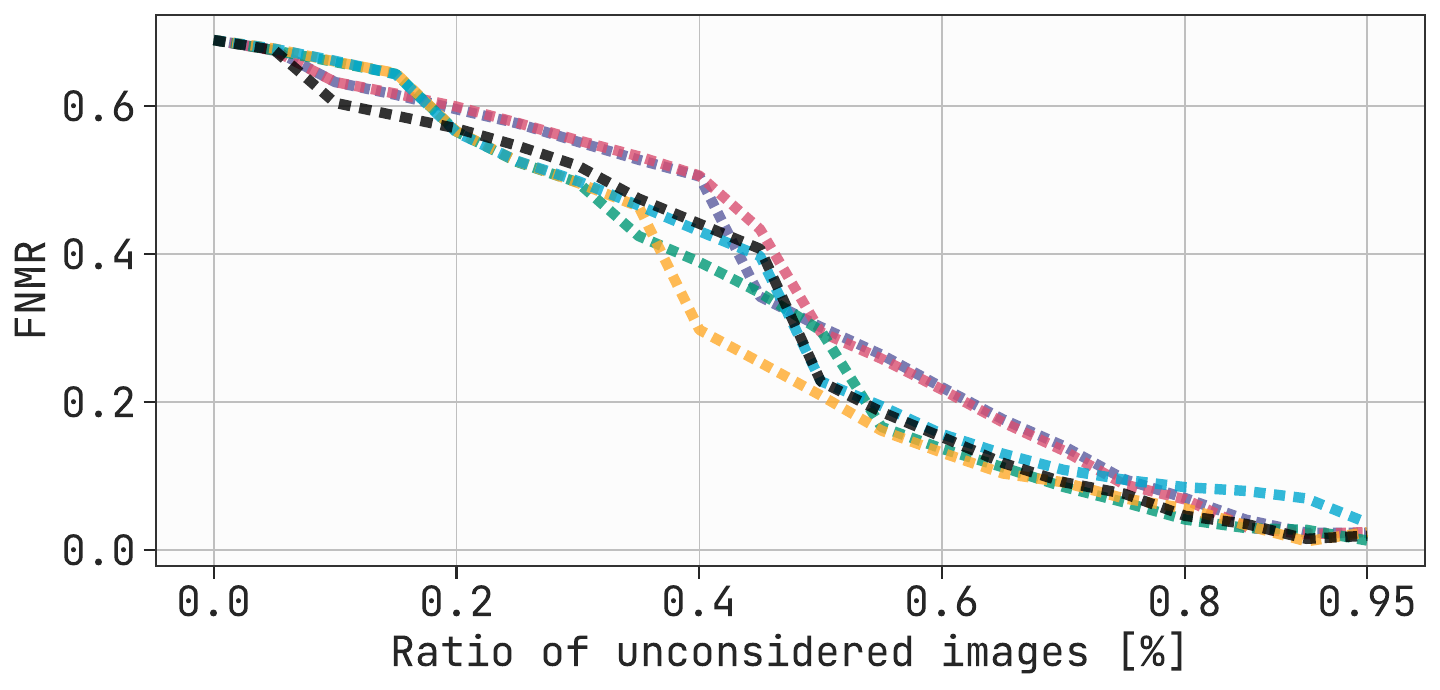}
		 \caption{MagFace \cite{meng_2021_magface} Model, XQLFW \cite{XQLFW} Dataset \\ \grafiqs with ResNet100, FMR$=1e-4$}
	\end{subfigure}
\\
	\begin{subfigure}[b]{0.48\textwidth}
		 \centering
		 \includegraphics[width=0.95\textwidth]{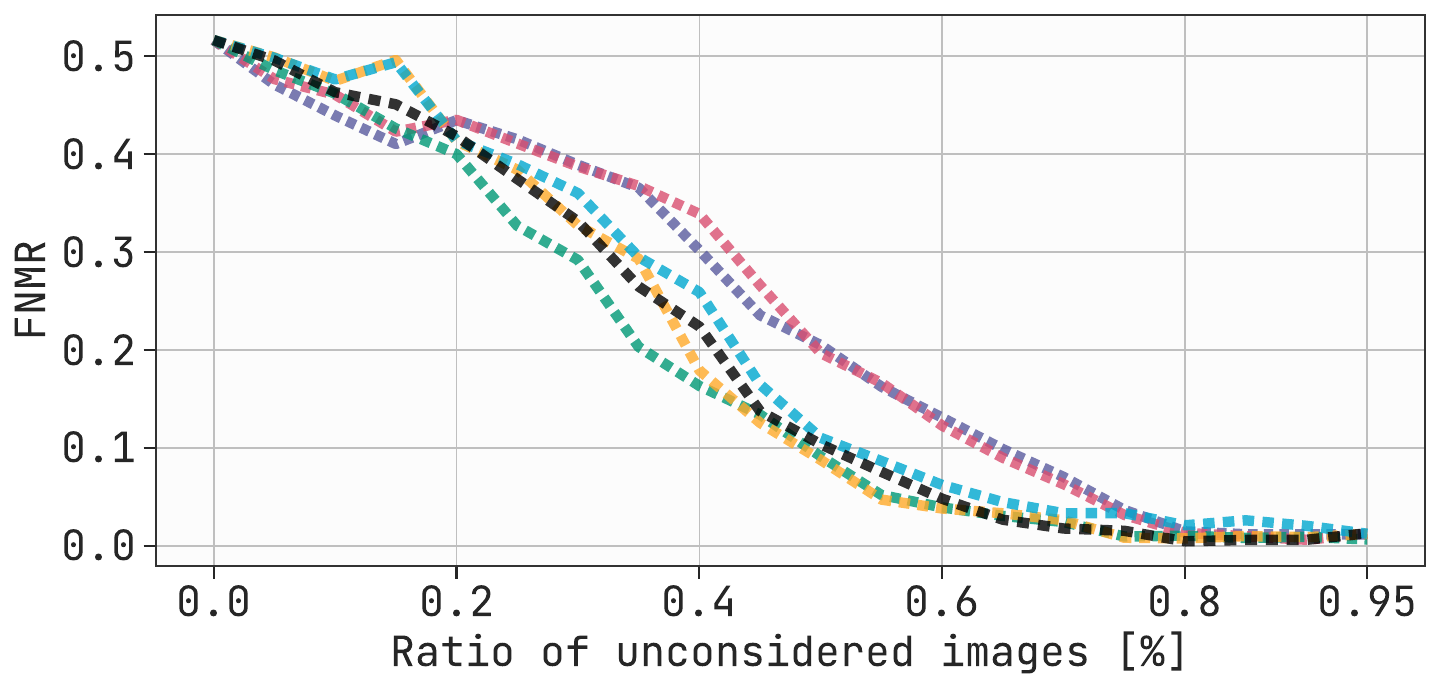}
		 \caption{CurricularFace \cite{curricularFace} Model, XQLFW \cite{XQLFW} Dataset \\ \grafiqs with ResNet100, FMR$=1e-3$}
	\end{subfigure}
\hfill
	\begin{subfigure}[b]{0.48\textwidth}
		 \centering
		 \includegraphics[width=0.95\textwidth]{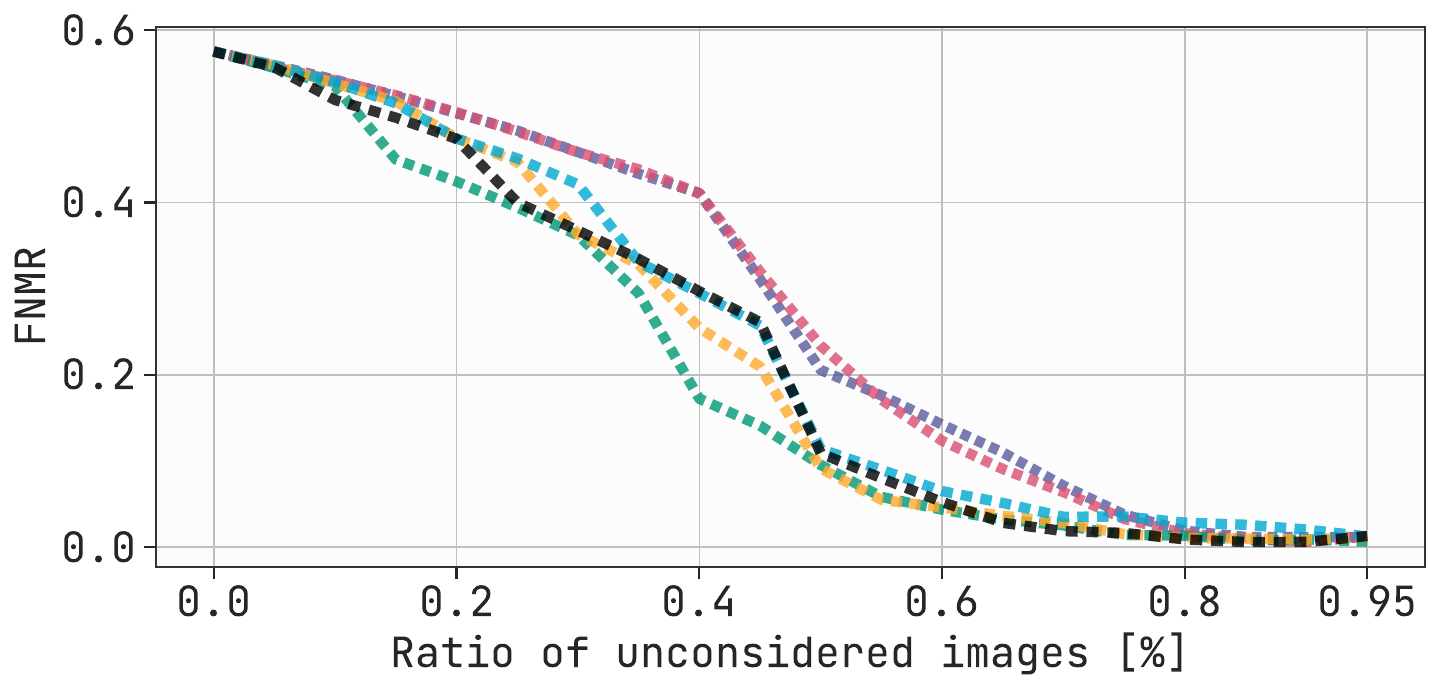}
		 \caption{CurricularFace \cite{curricularFace} Model, XQLFW \cite{XQLFW} Dataset \\ \grafiqs with ResNet100, FMR$=1e-4$}
	\end{subfigure}
\\
\caption{Error-versus-Discard Characteristic (EDC) curves for FNMR@FMR=$1e-3$ and FNMR@FMR=$1e-4$ of our proposed method using $\mathcal{L}_{\text{BNS}}$ as backpropagation loss and absolute sum as FIQ. The gradients at image level ($\phi=\mathcal{I}$), and block levels ($\phi=\text{B}1$ $-$ $\phi=\text{B}4$) are used to calculate FIQ. $\text{MSE}_{\text{BNS}}$ as FIQ is shown in black. Results shown on benchmark XQLFW \cite{XQLFW} using ArcFace, ElasticFace, MagFace, and, CurricularFace FR models. It is evident that the proposed \grafiqs method leads to lower verification error when images with lowest utility score estimated from gradient magnitudes are rejected. Furthermore, estimating FIQ by backpropagating $\mathcal{L}_{\text{BNS}}$ yields significantly better results than using $\text{MSE}_{\text{BNS}}$ directly.}
\vspace{-4mm}
\label{fig:iresnet100_supplementary_XQLFW}
\end{figure*}

%% file: figures/fig_iresnet100_supplementary_sota_adience.tex
\begin{figure*}[h!]
\centering
	\begin{subfigure}[b]{0.95\textwidth}
		\centering
		\includegraphics[width=\textwidth]{figures/iresnet100_bn_sota/legend.pdf}
	\end{subfigure}
\\
	\begin{subfigure}[b]{0.48\textwidth}
		 \centering
		 \includegraphics[width=0.95\textwidth]{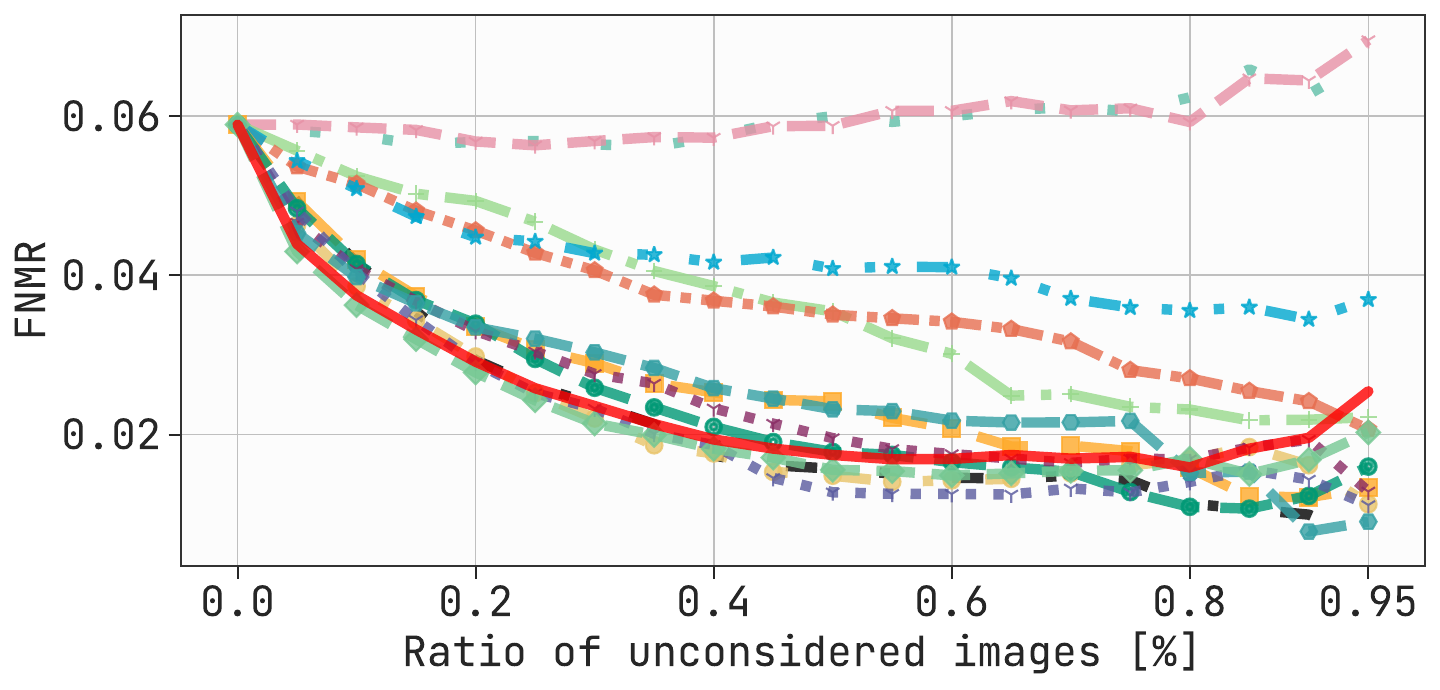}
		 \caption{ArcFace \cite{deng2019arcface} Model, Adience \cite{Adience} Dataset \\ \grafiqs with ResNet100, FMR$=1e-3$}
	\end{subfigure}
\hfill
	\begin{subfigure}[b]{0.48\textwidth}
		 \centering
		 \includegraphics[width=0.95\textwidth]{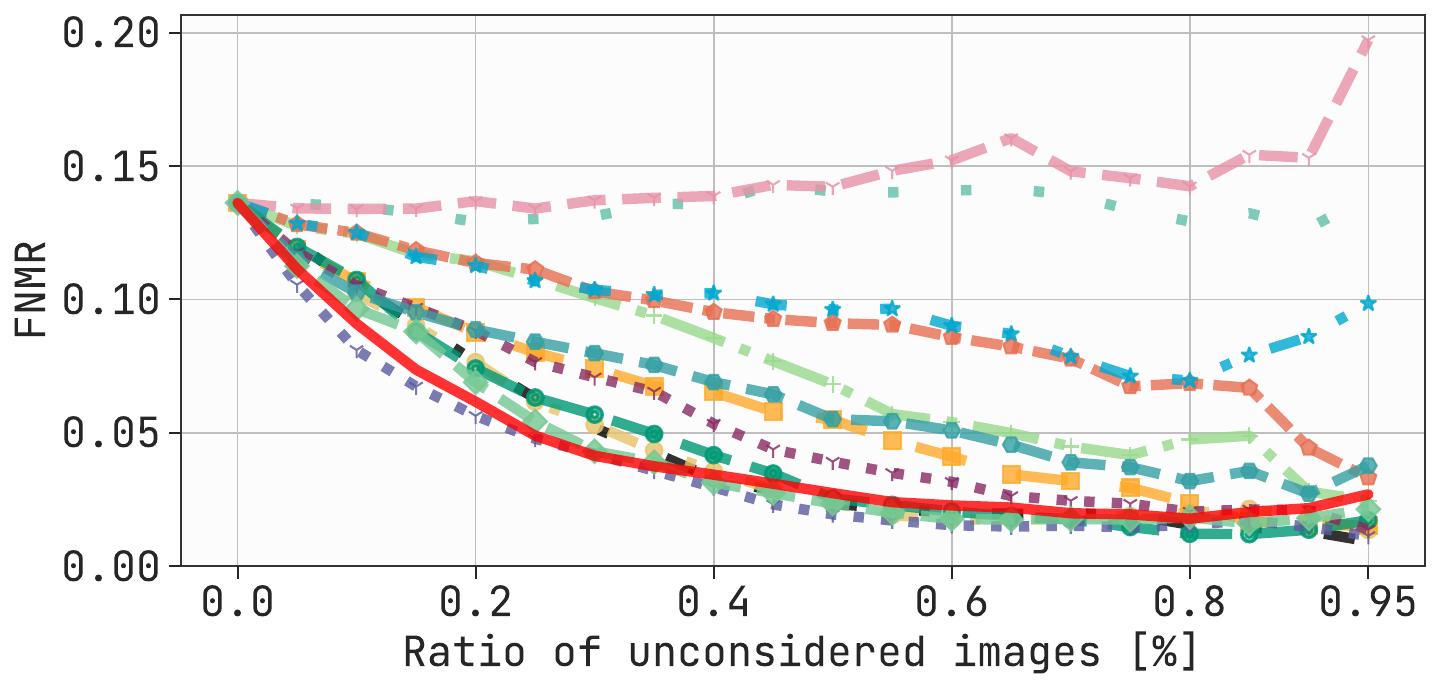}
		 \caption{ArcFace \cite{deng2019arcface} Model, Adience \cite{Adience} Dataset \\ \grafiqs with ResNet100, FMR$=1e-4$}
	\end{subfigure}
\\
	\begin{subfigure}[b]{0.48\textwidth}
		 \centering
		 \includegraphics[width=0.95\textwidth]{figures/iresnet100_bn_sota/adience/ElasticFaceModel/model_ElasticFaceModel_adience_fnmr3_sota.pdf}
		 \caption{ElasticFace \cite{elasticface} Model, Adience \cite{Adience} Dataset \\ \grafiqs with ResNet100, FMR$=1e-3$}
	\end{subfigure}
\hfill
	\begin{subfigure}[b]{0.48\textwidth}
		 \centering
		 \includegraphics[width=0.95\textwidth]{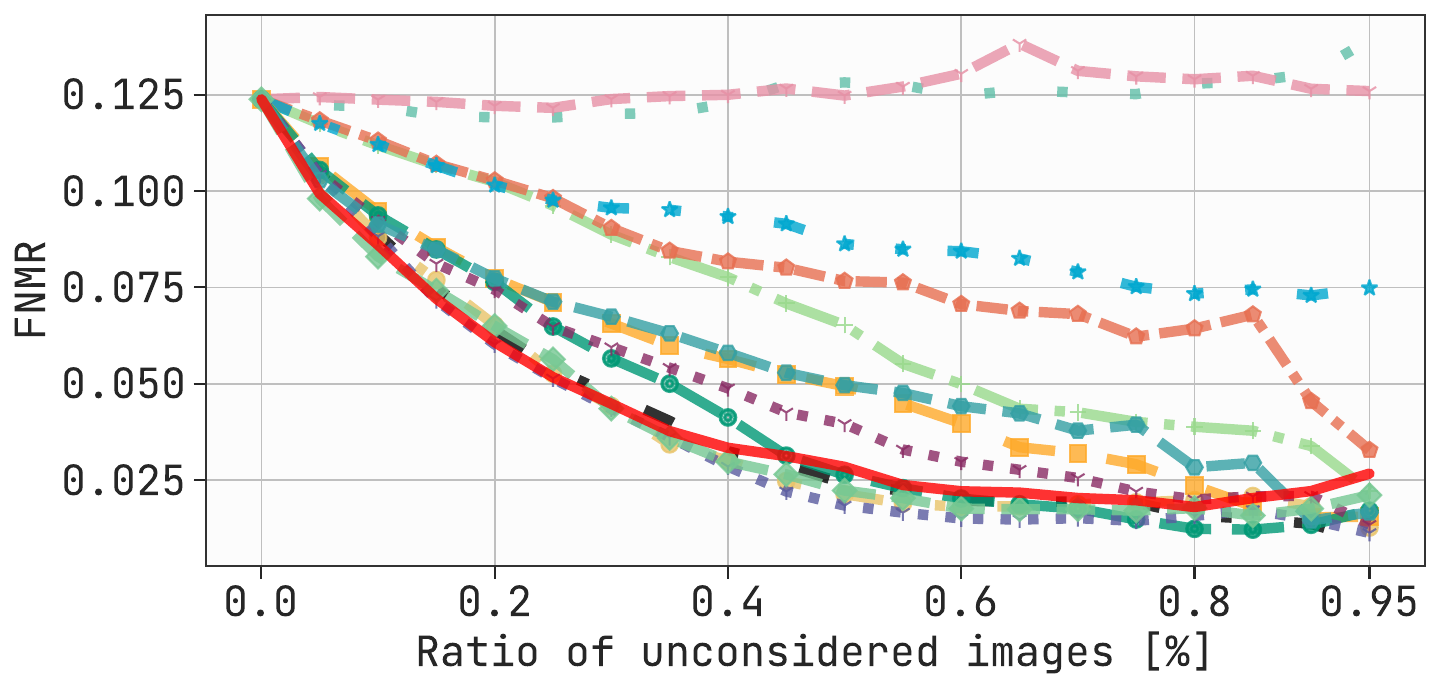}
		 \caption{ElasticFace \cite{elasticface} Model, Adience \cite{Adience} Dataset \\ \grafiqs with ResNet100, FMR$=1e-4$}
	\end{subfigure}
\\
	\begin{subfigure}[b]{0.48\textwidth}
		 \centering
		 \includegraphics[width=0.95\textwidth]{figures/iresnet100_bn_sota/adience/MagFaceModel/model_MagFaceModel_adience_fnmr3_sota.pdf}
		 \caption{MagFace \cite{meng_2021_magface} Model, Adience \cite{Adience} Dataset \\ \grafiqs with ResNet100, FMR$=1e-3$}
	\end{subfigure}
\hfill
	\begin{subfigure}[b]{0.48\textwidth}
		 \centering
		 \includegraphics[width=0.95\textwidth]{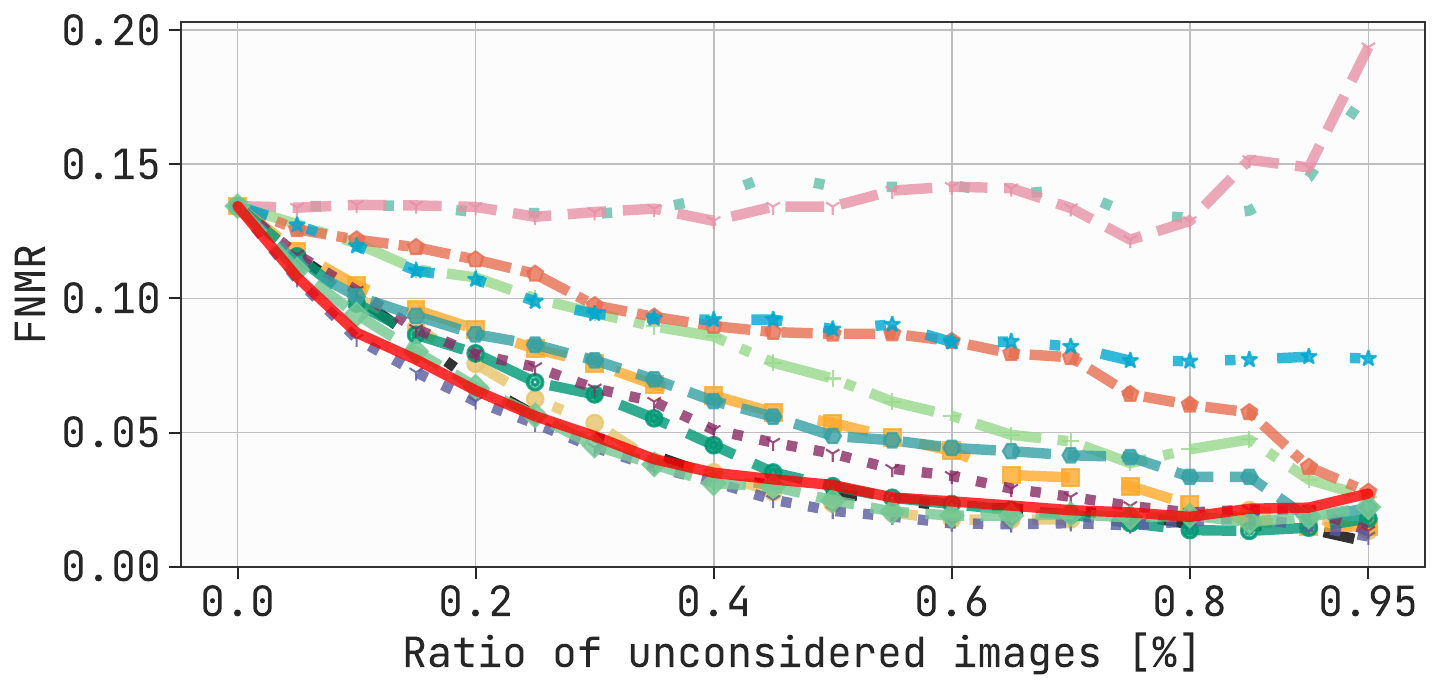}
		 \caption{MagFace \cite{meng_2021_magface} Model, Adience \cite{Adience} Dataset \\ \grafiqs with ResNet100, FMR$=1e-4$}
	\end{subfigure}
\\
	\begin{subfigure}[b]{0.48\textwidth}
		 \centering
		 \includegraphics[width=0.95\textwidth]{figures/iresnet100_bn_sota/adience/CurricularFaceModel/model_CurricularFaceModel_adience_fnmr3_sota.pdf}
		 \caption{CurricularFace \cite{curricularFace} Model, Adience \cite{Adience} Dataset \\ \grafiqs with ResNet100, FMR$=1e-3$}
	\end{subfigure}
\hfill
	\begin{subfigure}[b]{0.48\textwidth}
		 \centering
		 \includegraphics[width=0.95\textwidth]{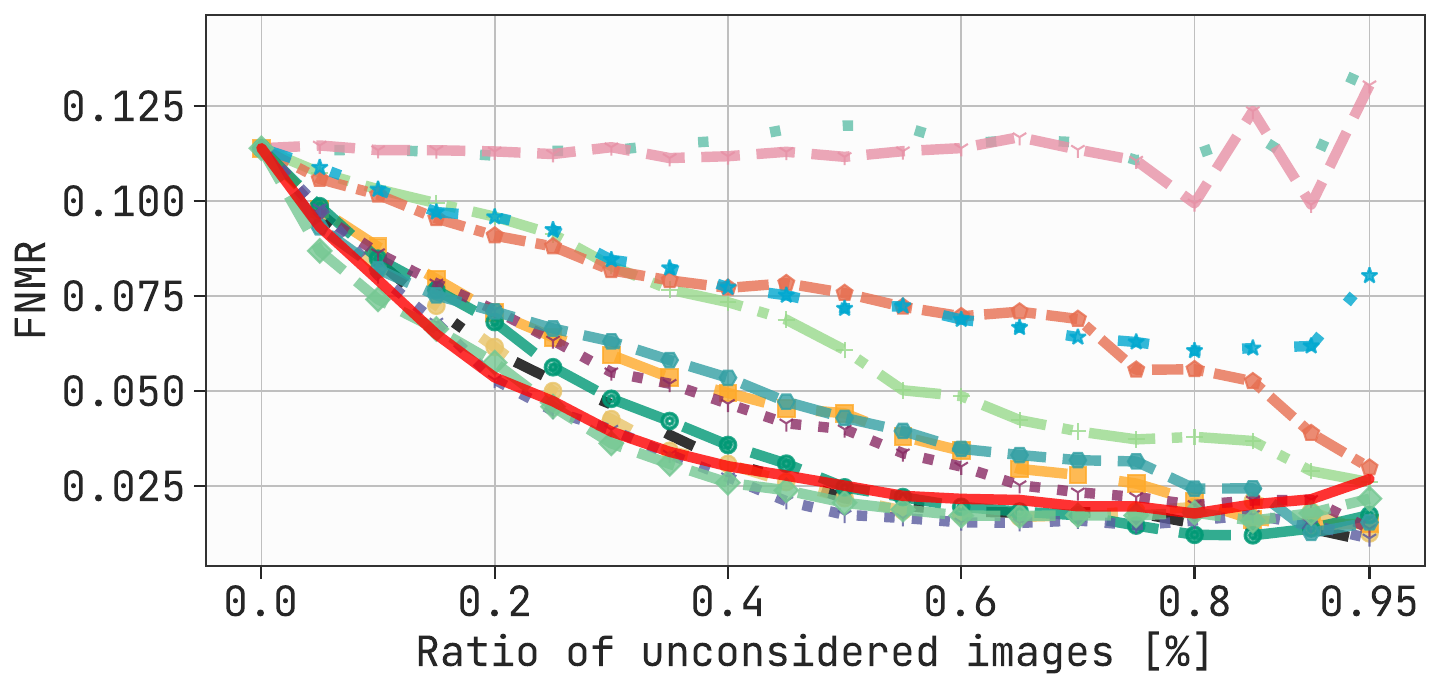}
		 \caption{CurricularFace \cite{curricularFace} Model, Adience \cite{Adience} Dataset \\ \grafiqs with ResNet100, FMR$=1e-4$}
	\end{subfigure}
\\
\caption{EDC curves for FNMR@FMR=$1e-3$ and FNMR@FMR=$1e-4$ on dataset Adience \cite{Adience} using ArcFace, ElasticFace, MagFace, and CurricularFace FR models. The proposed \grafiqs method, shown in \textcolor{red}{solid red}, utilizes gradient magnitudes and it is reported using the best setting from Table 2 in the main paper.}
\vspace{-4mm}
\label{fig:iresnet100_supplementary_sota_adience}
\end{figure*}

%% file: figures/fig_iresnet100_supplementary_sota_agedb_30.tex
\begin{figure*}[h!]
\centering
	\begin{subfigure}[b]{0.95\textwidth}
		\centering
		\includegraphics[width=\textwidth]{figures/iresnet100_bn_sota/legend.pdf}
	\end{subfigure}
\\
	\begin{subfigure}[b]{0.48\textwidth}
		 \centering
		 \includegraphics[width=0.95\textwidth]{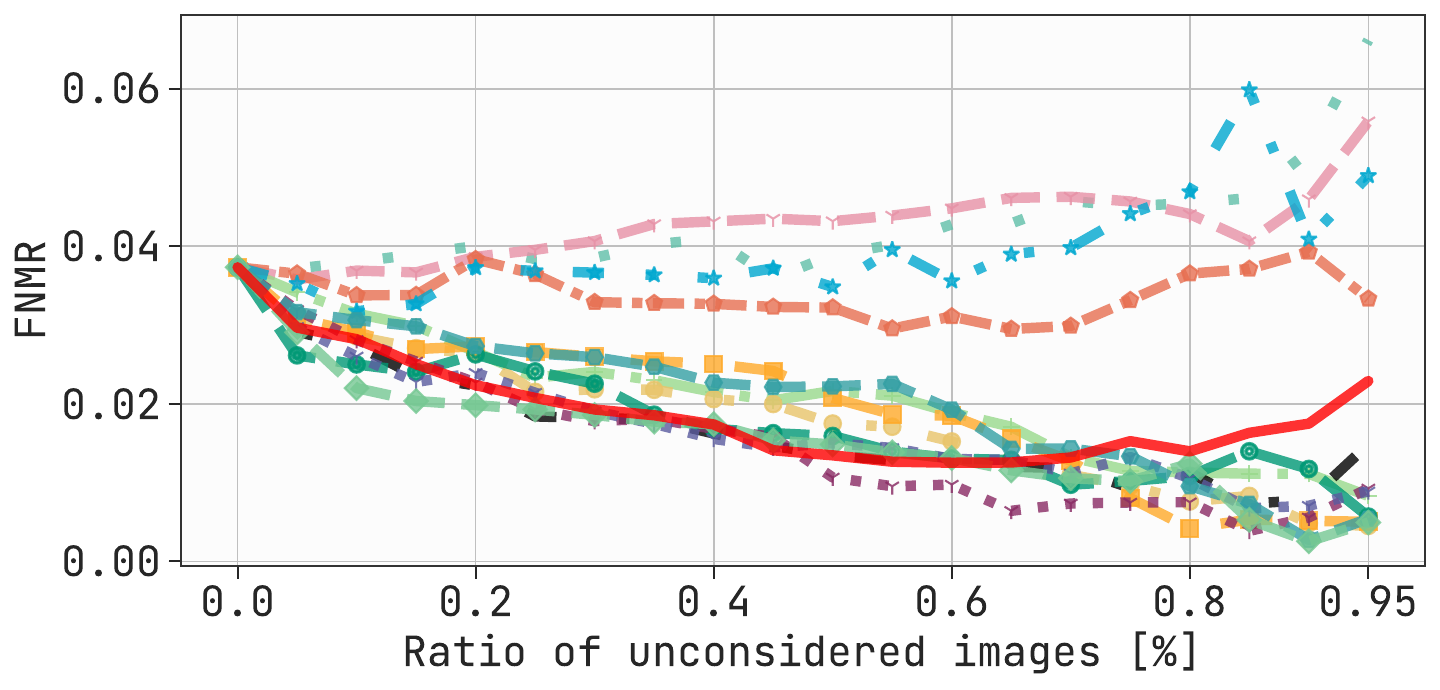}
		 \caption{ArcFace \cite{deng2019arcface} Model, AgeDB30 \cite{agedb} Dataset \\ \grafiqs with ResNet100, FMR$=1e-3$}
	\end{subfigure}
\hfill
	\begin{subfigure}[b]{0.48\textwidth}
		 \centering
		 \includegraphics[width=0.95\textwidth]{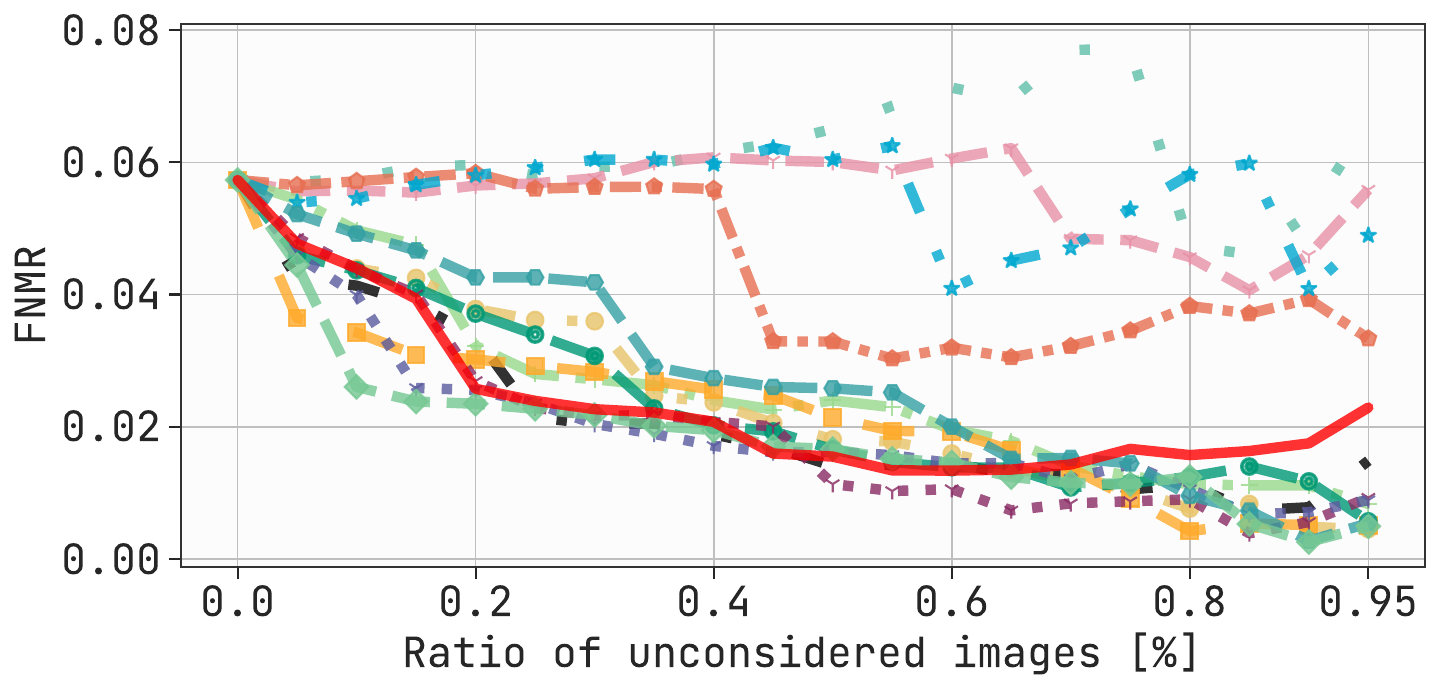}
		 \caption{ArcFace \cite{deng2019arcface} Model, AgeDB30 \cite{agedb} Dataset \\ \grafiqs with ResNet100, FMR$=1e-4$}
	\end{subfigure}
\\
	\begin{subfigure}[b]{0.48\textwidth}
		 \centering
		 \includegraphics[width=0.95\textwidth]{figures/iresnet100_bn_sota/agedb_30/ElasticFaceModel/model_ElasticFaceModel_agedb_30_fnmr3_sota.pdf}
		 \caption{ElasticFace \cite{elasticface} Model, AgeDB30 \cite{agedb} Dataset \\ \grafiqs with ResNet100, FMR$=1e-3$}
	\end{subfigure}
\hfill
	\begin{subfigure}[b]{0.48\textwidth}
		 \centering
		 \includegraphics[width=0.95\textwidth]{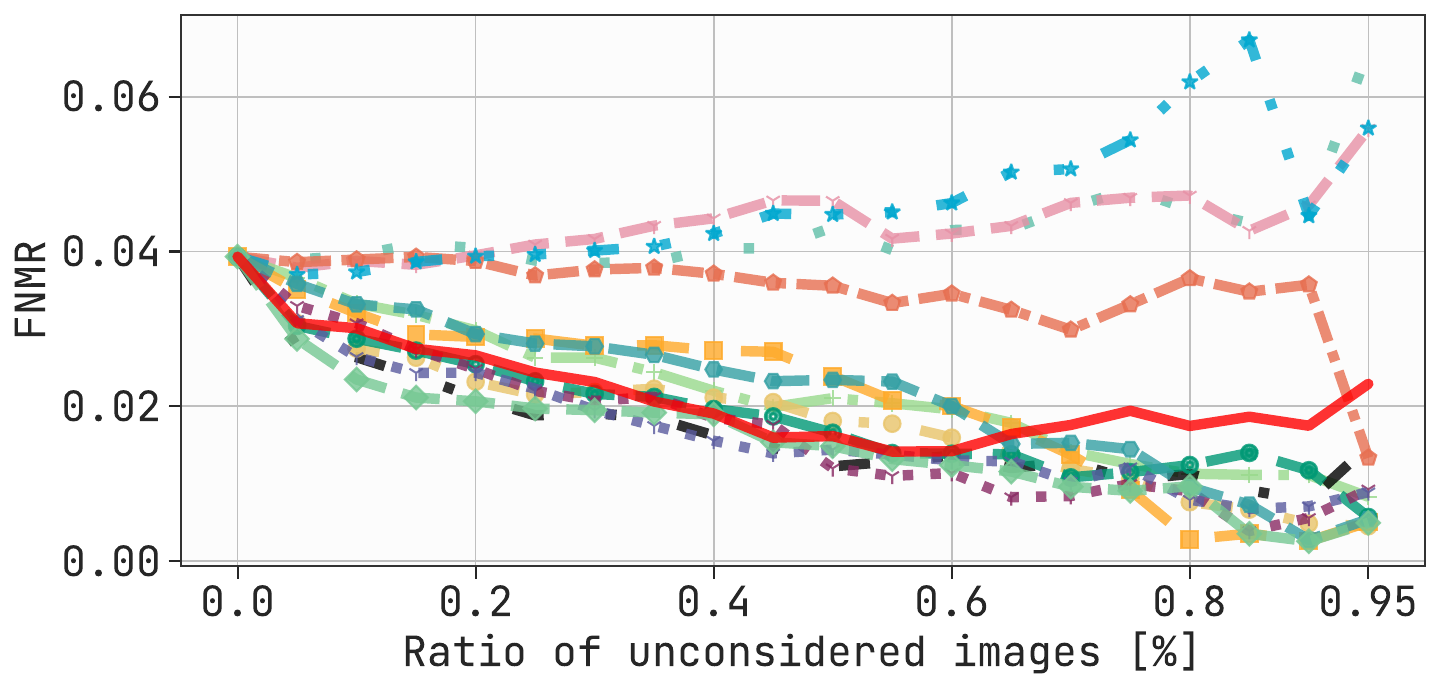}
		 \caption{ElasticFace \cite{elasticface} Model, AgeDB30 \cite{agedb} Dataset \\ \grafiqs with ResNet100, FMR$=1e-4$}
	\end{subfigure}
\\
	\begin{subfigure}[b]{0.48\textwidth}
		 \centering
		 \includegraphics[width=0.95\textwidth]{figures/iresnet100_bn_sota/agedb_30/MagFaceModel/model_MagFaceModel_agedb_30_fnmr3_sota.pdf}
		 \caption{MagFace \cite{meng_2021_magface} Model, AgeDB30 \cite{agedb} Dataset \\ \grafiqs with ResNet100, FMR$=1e-3$}
	\end{subfigure}
\hfill
	\begin{subfigure}[b]{0.48\textwidth}
		 \centering
		 \includegraphics[width=0.95\textwidth]{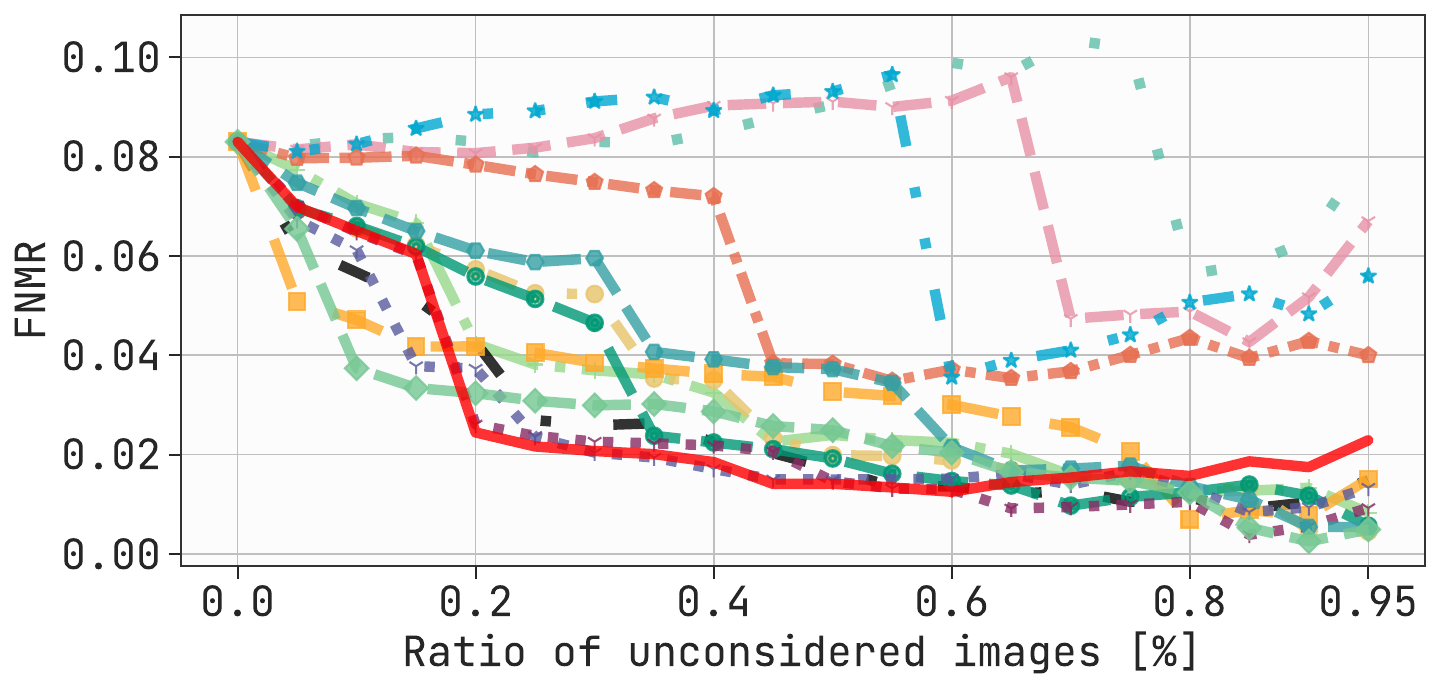}
		 \caption{MagFace \cite{meng_2021_magface} Model, AgeDB30 \cite{agedb} Dataset \\ \grafiqs with ResNet100, FMR$=1e-4$}
	\end{subfigure}
\\
	\begin{subfigure}[b]{0.48\textwidth}
		 \centering
		 \includegraphics[width=0.95\textwidth]{figures/iresnet100_bn_sota/agedb_30/CurricularFaceModel/model_CurricularFaceModel_agedb_30_fnmr3_sota.pdf}
		 \caption{CurricularFace \cite{curricularFace} Model, AgeDB30 \cite{agedb} Dataset \\ \grafiqs with ResNet100, FMR$=1e-3$}
	\end{subfigure}
\hfill
	\begin{subfigure}[b]{0.48\textwidth}
		 \centering
		 \includegraphics[width=0.95\textwidth]{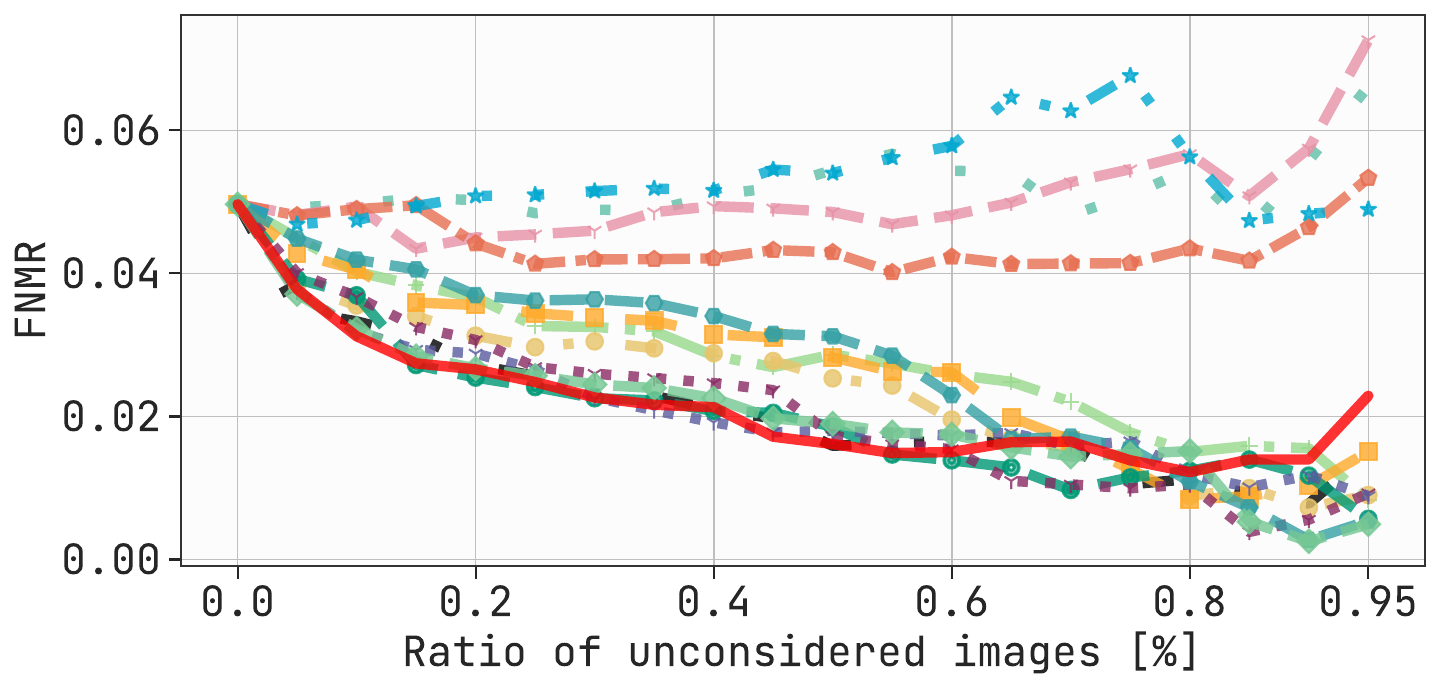}
		 \caption{CurricularFace \cite{curricularFace} Model, AgeDB30 \cite{agedb} Dataset \\ \grafiqs with ResNet100, FMR$=1e-4$}
	\end{subfigure}
\\
\caption{EDC curves for FNMR@FMR=$1e-3$ and FNMR@FMR=$1e-4$ on dataset AgeDB30 \cite{agedb} using ArcFace, ElasticFace, MagFace, and CurricularFace FR models. The proposed \grafiqs method, shown in \textcolor{red}{solid red}, utilizes gradient magnitudes and it is reported using the best setting from Table 2 in the main paper.}
\vspace{-4mm}
\label{fig:iresnet100_supplementary_sota_agedb_30}
\end{figure*}

%% file: figures/fig_iresnet100_supplementary_sota_cfp_fp.tex
\begin{figure*}[h!]
\centering
	\begin{subfigure}[b]{0.95\textwidth}
		\centering
		\includegraphics[width=\textwidth]{figures/iresnet100_bn_sota/legend.pdf}
	\end{subfigure}
\\
	\begin{subfigure}[b]{0.48\textwidth}
		 \centering
		 \includegraphics[width=0.95\textwidth]{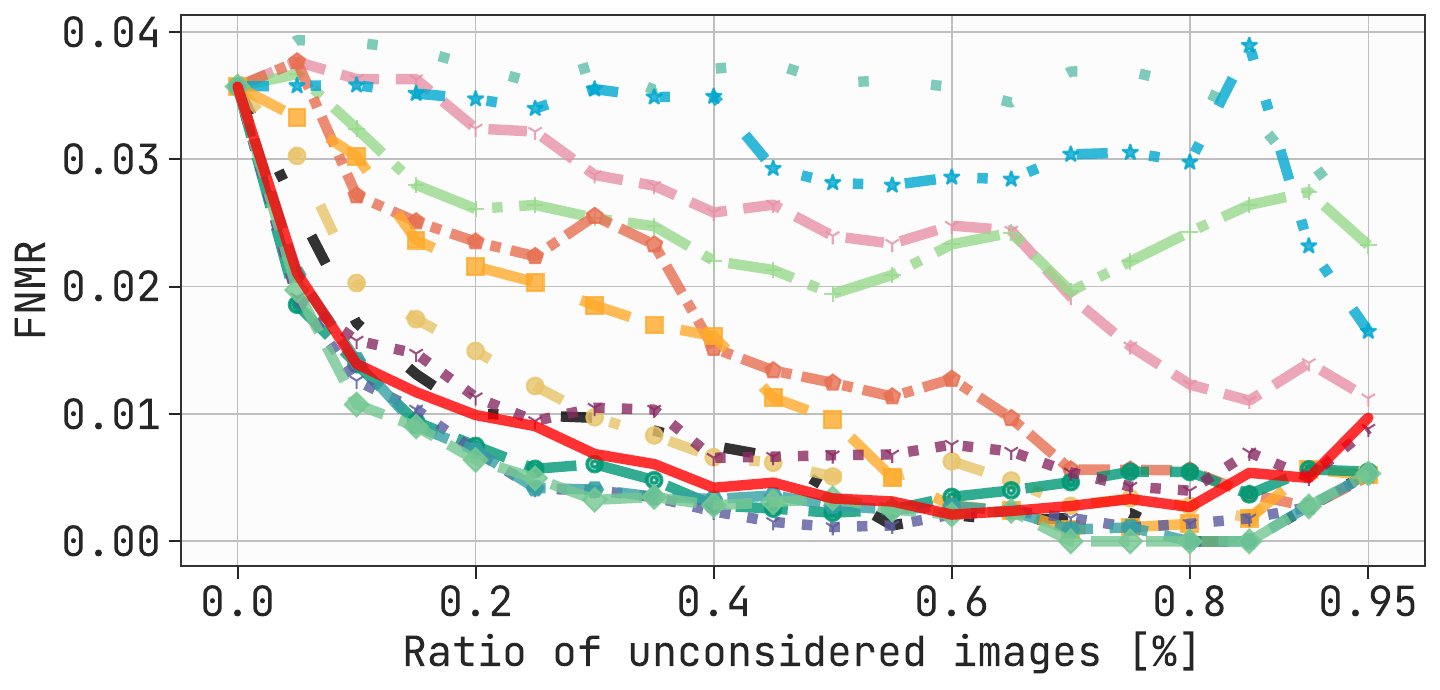}
		 \caption{ArcFace \cite{deng2019arcface} Model, CFP-FP \cite{cfp-fp} Dataset \\ \grafiqs with ResNet100, FMR$=1e-3$}
	\end{subfigure}
\hfill
	\begin{subfigure}[b]{0.48\textwidth}
		 \centering
		 \includegraphics[width=0.95\textwidth]{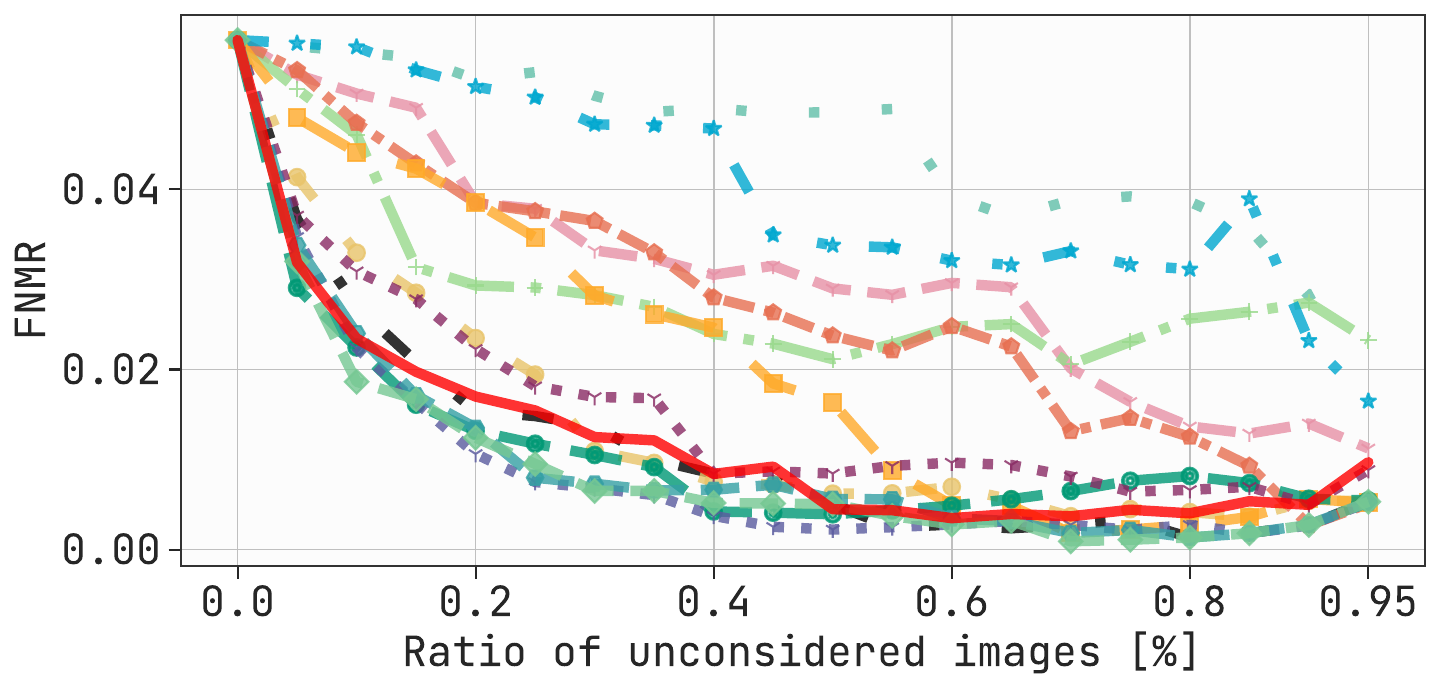}
		 \caption{ArcFace \cite{deng2019arcface} Model, CFP-FP \cite{cfp-fp} Dataset \\ \grafiqs with ResNet100, FMR$=1e-4$}
	\end{subfigure}
\\
	\begin{subfigure}[b]{0.48\textwidth}
		 \centering
		 \includegraphics[width=0.95\textwidth]{figures/iresnet100_bn_sota/cfp_fp/ElasticFaceModel/model_ElasticFaceModel_cfp_fp_fnmr3_sota.pdf}
		 \caption{ElasticFace \cite{elasticface} Model, CFP-FP \cite{cfp-fp} Dataset \\ \grafiqs with ResNet100, FMR$=1e-3$}
	\end{subfigure}
\hfill
	\begin{subfigure}[b]{0.48\textwidth}
		 \centering
		 \includegraphics[width=0.95\textwidth]{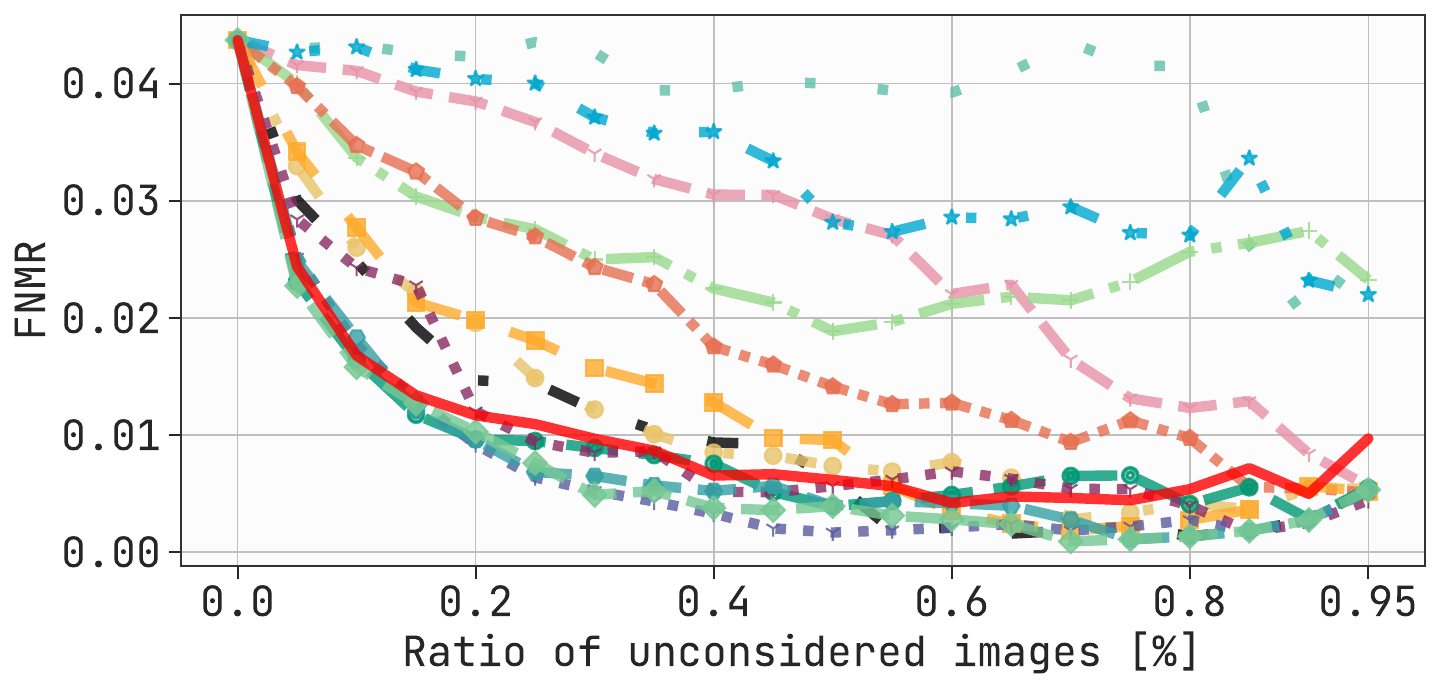}
		 \caption{ElasticFace \cite{elasticface} Model, CFP-FP \cite{cfp-fp} Dataset \\ \grafiqs with ResNet100, FMR$=1e-4$}
	\end{subfigure}
\\
	\begin{subfigure}[b]{0.48\textwidth}
		 \centering
		 \includegraphics[width=0.95\textwidth]{figures/iresnet100_bn_sota/cfp_fp/MagFaceModel/model_MagFaceModel_cfp_fp_fnmr3_sota.pdf}
		 \caption{MagFace \cite{meng_2021_magface} Model, CFP-FP \cite{cfp-fp} Dataset \\ \grafiqs with ResNet100, FMR$=1e-3$}
	\end{subfigure}
\hfill
	\begin{subfigure}[b]{0.48\textwidth}
		 \centering
		 \includegraphics[width=0.95\textwidth]{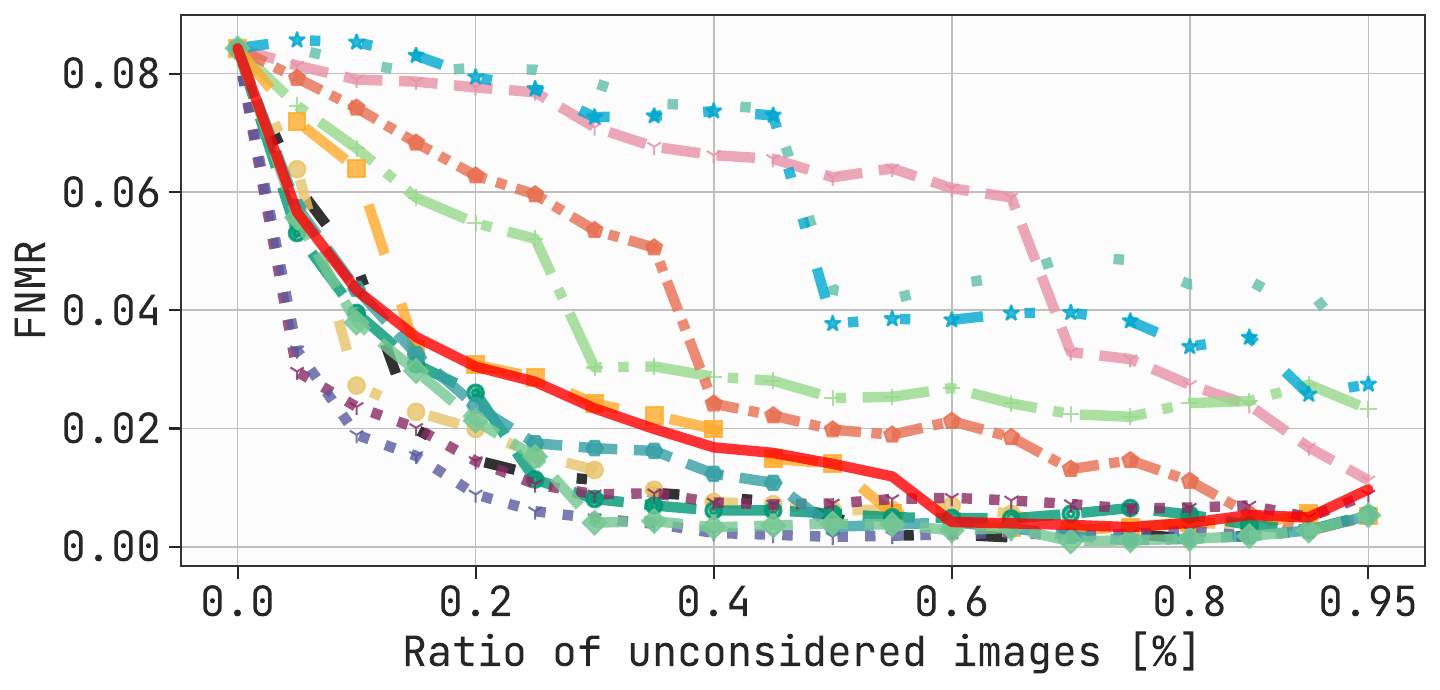}
		 \caption{MagFace \cite{meng_2021_magface} Model, CFP-FP \cite{cfp-fp} Dataset \\ \grafiqs with ResNet100, FMR$=1e-4$}
	\end{subfigure}
\\
	\begin{subfigure}[b]{0.48\textwidth}
		 \centering
		 \includegraphics[width=0.95\textwidth]{figures/iresnet100_bn_sota/cfp_fp/CurricularFaceModel/model_CurricularFaceModel_cfp_fp_fnmr3_sota.pdf}
		 \caption{CurricularFace \cite{curricularFace} Model, CFP-FP \cite{cfp-fp} Dataset \\ \grafiqs with ResNet100, FMR$=1e-3$}
	\end{subfigure}
\hfill
	\begin{subfigure}[b]{0.48\textwidth}
		 \centering
		 \includegraphics[width=0.95\textwidth]{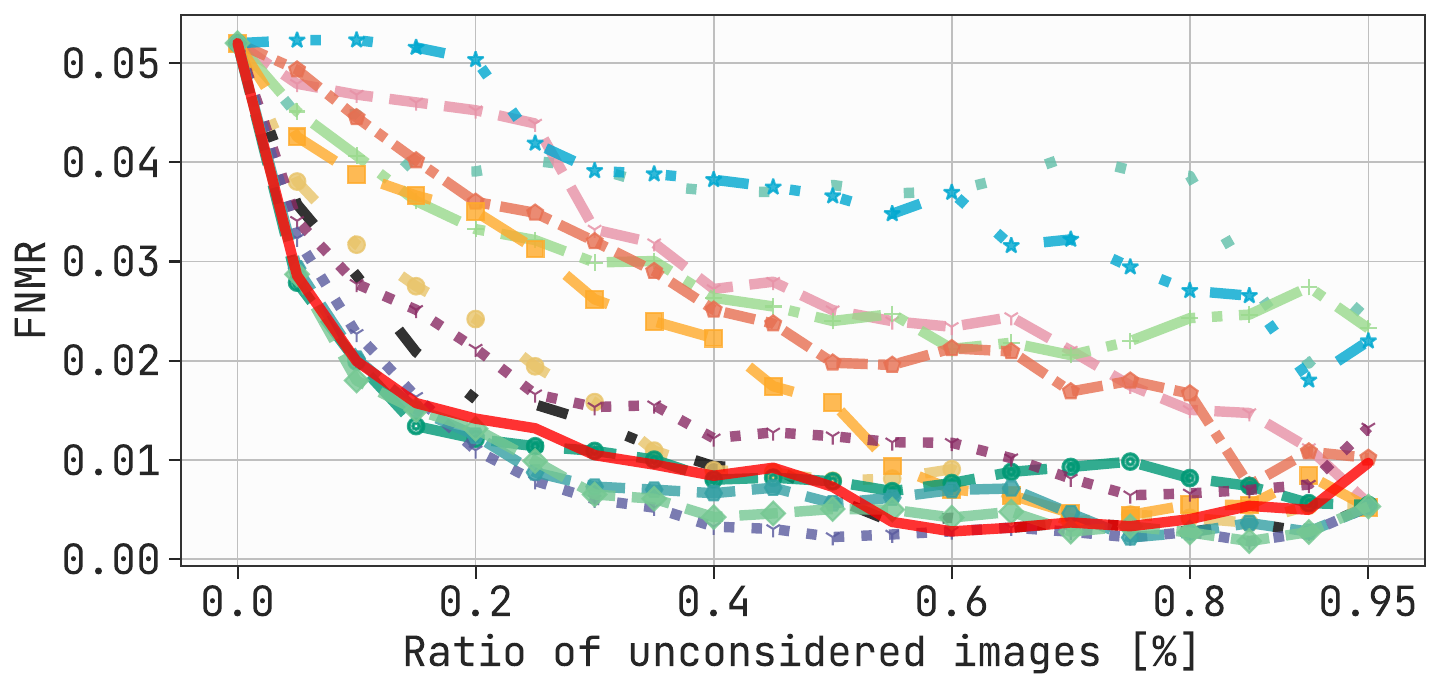}
		 \caption{CurricularFace \cite{curricularFace} Model, CFP-FP \cite{cfp-fp} Dataset \\ \grafiqs with ResNet100, FMR$=1e-4$}
	\end{subfigure}
\\
\caption{EDC curves for FNMR@FMR=$1e-3$ and FNMR@FMR=$1e-4$ on dataset CFP-FP \cite{cfp-fp} using ArcFace, ElasticFace, MagFace, and CurricularFace FR models. The proposed \grafiqs method, shown in \textcolor{red}{solid red}, utilizes gradient magnitudes and it is reported using the best setting from Table 2 in the main paper.}
\vspace{-4mm}
\label{fig:iresnet100_supplementary_sota_cfp_fp}
\end{figure*}

%% file: figures/fig_iresnet100_supplementary_sota_lfw.tex
\begin{figure*}[h!]
\centering
	\begin{subfigure}[b]{0.95\textwidth}
		\centering
		\includegraphics[width=\textwidth]{figures/iresnet100_bn_sota/legend.pdf}
	\end{subfigure}
\\
	\begin{subfigure}[b]{0.48\textwidth}
		 \centering
		 \includegraphics[width=0.95\textwidth]{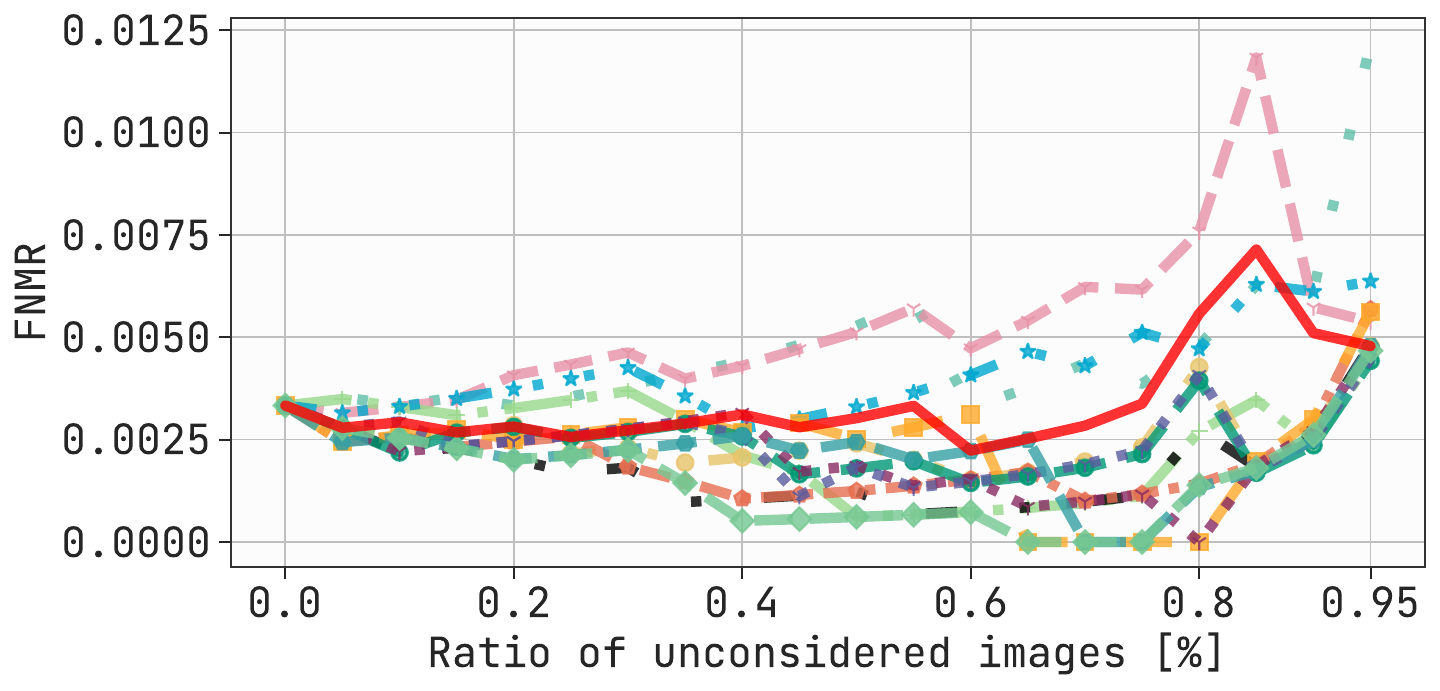}
		 \caption{ArcFace \cite{deng2019arcface} Model, LFW \cite{LFWTech} Dataset \\ \grafiqs with ResNet100, FMR$=1e-3$}
	\end{subfigure}
\hfill
	\begin{subfigure}[b]{0.48\textwidth}
		 \centering
		 \includegraphics[width=0.95\textwidth]{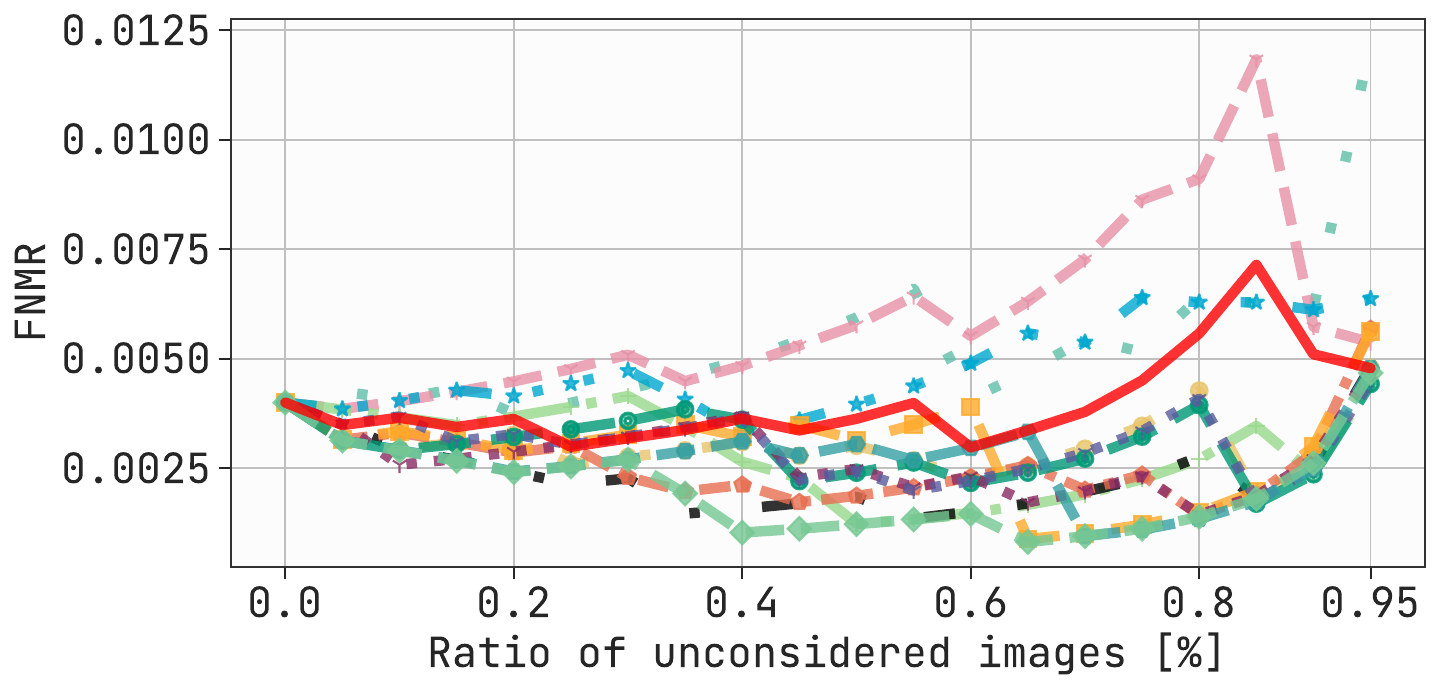}
		 \caption{ArcFace \cite{deng2019arcface} Model, LFW \cite{LFWTech} Dataset \\ \grafiqs with ResNet100, FMR$=1e-4$}
	\end{subfigure}
\\
	\begin{subfigure}[b]{0.48\textwidth}
		 \centering
		 \includegraphics[width=0.95\textwidth]{figures/iresnet100_bn_sota/lfw/ElasticFaceModel/model_ElasticFaceModel_lfw_fnmr3_sota.pdf}
		 \caption{ElasticFace \cite{elasticface} Model, LFW \cite{LFWTech} Dataset \\ \grafiqs with ResNet100, FMR$=1e-3$}
	\end{subfigure}
\hfill
	\begin{subfigure}[b]{0.48\textwidth}
		 \centering
		 \includegraphics[width=0.95\textwidth]{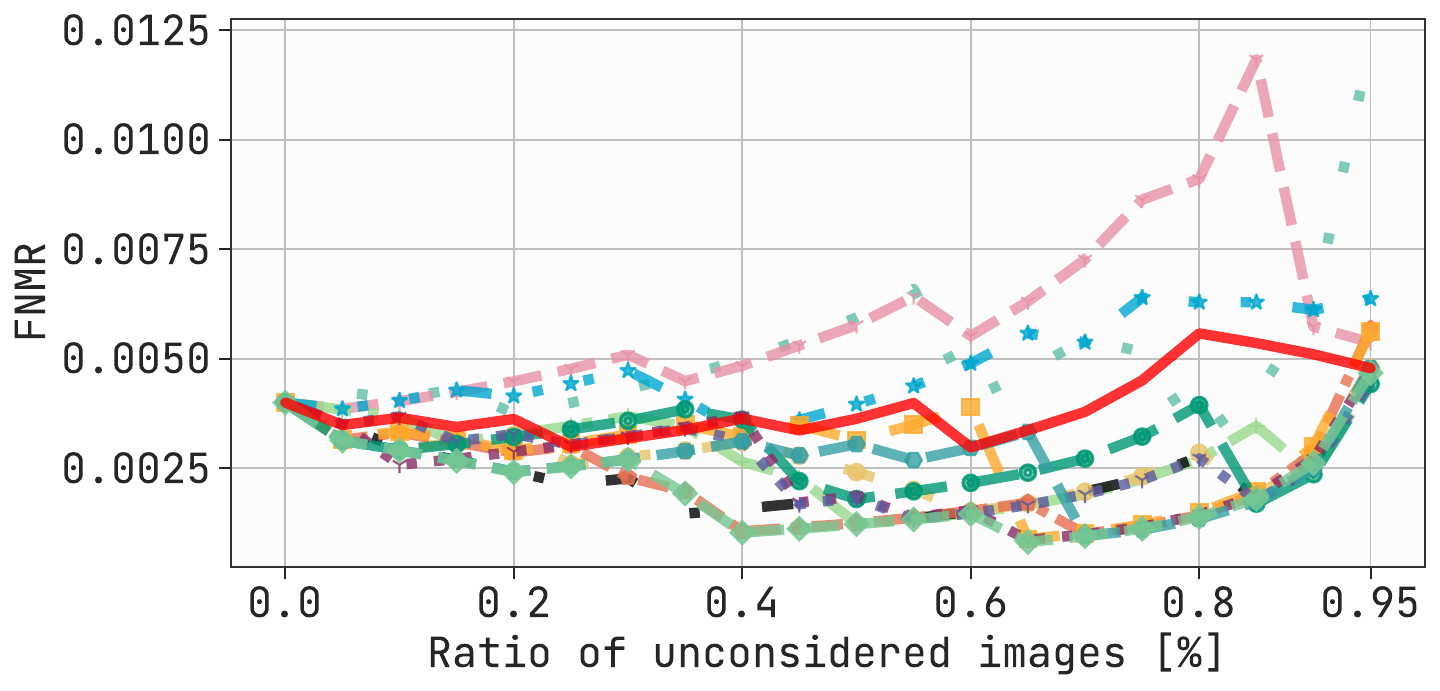}
		 \caption{ElasticFace \cite{elasticface} Model, LFW \cite{LFWTech} Dataset \\ \grafiqs with ResNet100, FMR$=1e-4$}
	\end{subfigure}
\\
	\begin{subfigure}[b]{0.48\textwidth}
		 \centering
		 \includegraphics[width=0.95\textwidth]{figures/iresnet100_bn_sota/lfw/MagFaceModel/model_MagFaceModel_lfw_fnmr3_sota.pdf}
		 \caption{MagFace \cite{meng_2021_magface} Model, LFW \cite{LFWTech} Dataset \\ \grafiqs with ResNet100, FMR$=1e-3$}
	\end{subfigure}
\hfill
	\begin{subfigure}[b]{0.48\textwidth}
		 \centering
		 \includegraphics[width=0.95\textwidth]{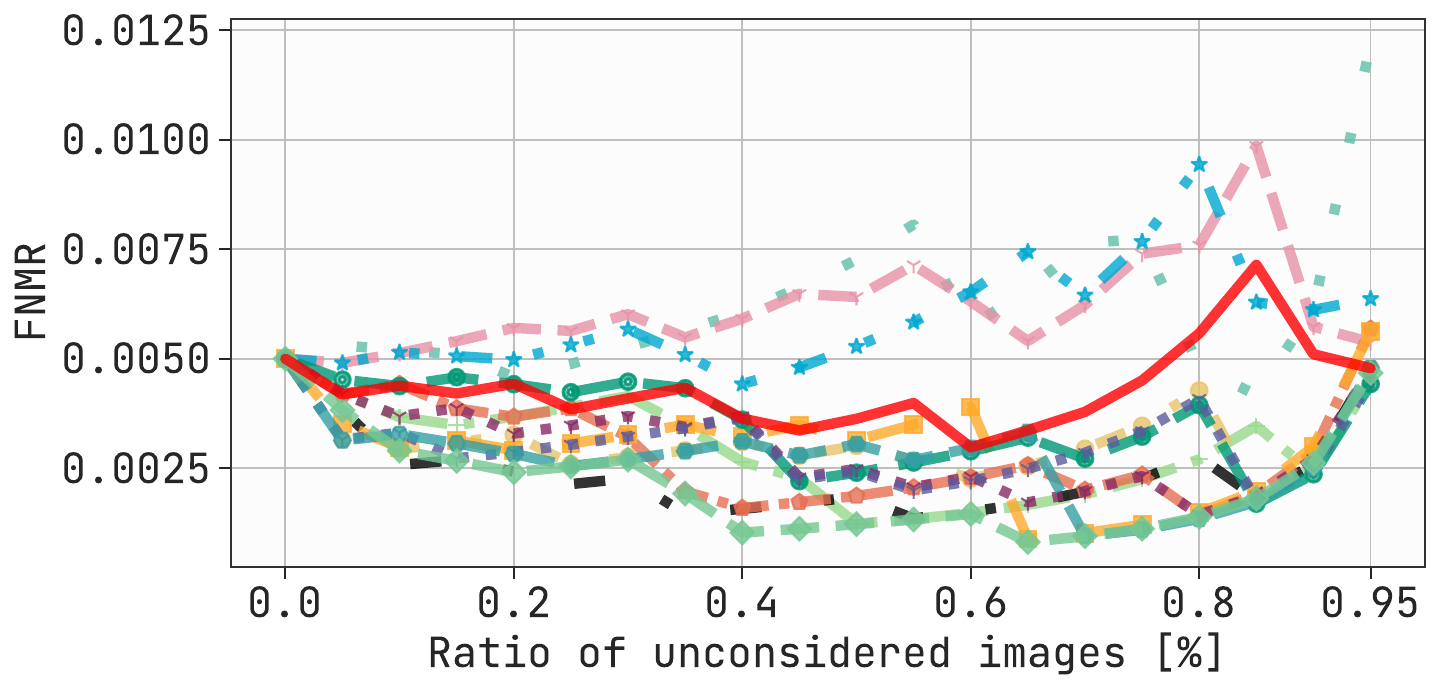}
		 \caption{MagFace \cite{meng_2021_magface} Model, LFW \cite{LFWTech} Dataset \\ \grafiqs with ResNet100, FMR$=1e-4$}
	\end{subfigure}
\\
	\begin{subfigure}[b]{0.48\textwidth}
		 \centering
		 \includegraphics[width=0.95\textwidth]{figures/iresnet100_bn_sota/lfw/CurricularFaceModel/model_CurricularFaceModel_lfw_fnmr3_sota.pdf}
		 \caption{CurricularFace \cite{curricularFace} Model, LFW \cite{LFWTech} Dataset \\ \grafiqs with ResNet100, FMR$=1e-3$}
	\end{subfigure}
\hfill
	\begin{subfigure}[b]{0.48\textwidth}
		 \centering
		 \includegraphics[width=0.95\textwidth]{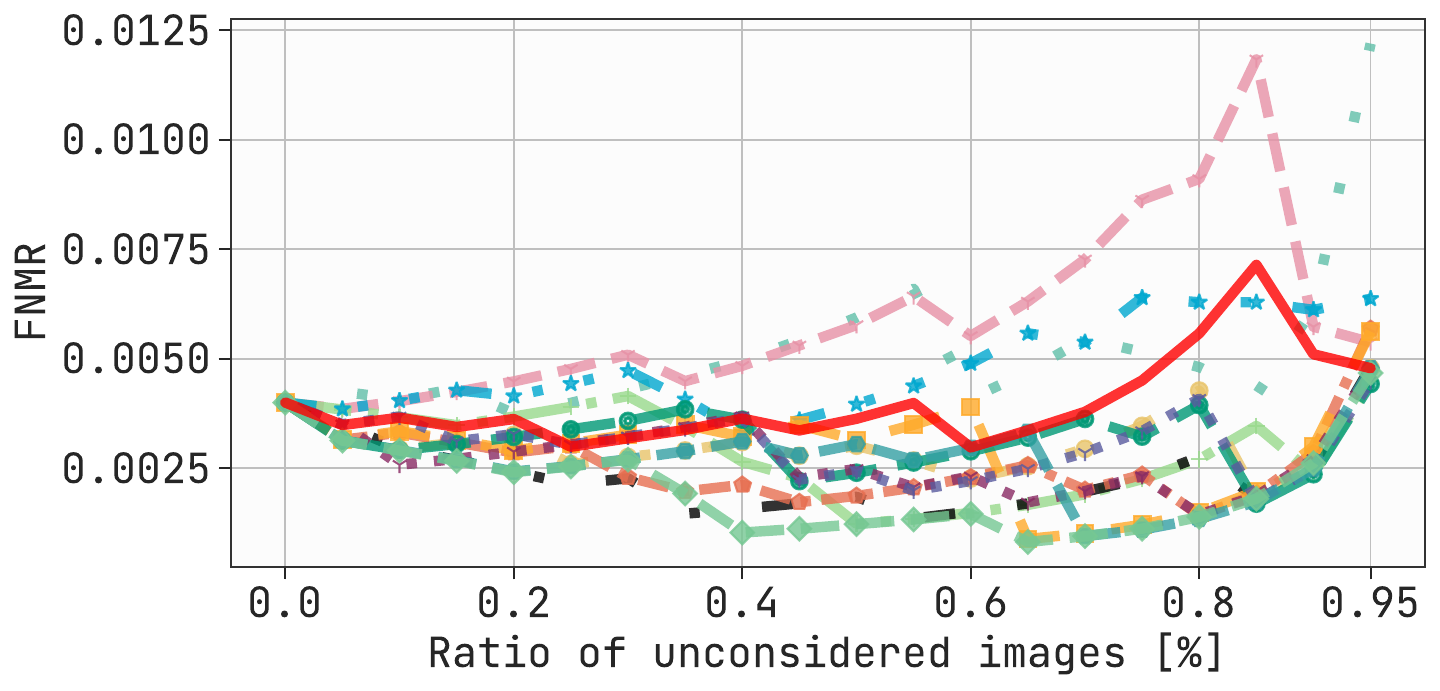}
		 \caption{CurricularFace \cite{curricularFace} Model, LFW \cite{LFWTech} Dataset \\ \grafiqs with ResNet100, FMR$=1e-4$}
	\end{subfigure}
\\
\caption{EDC curves for FNMR@FMR=$1e-3$ and FNMR@FMR=$1e-4$ on dataset LFW \cite{LFWTech} using ArcFace, ElasticFace, MagFace, and CurricularFace FR models. The proposed \grafiqs method, shown in \textcolor{red}{solid red}, utilizes gradient magnitudes and it is reported using the best setting from Table 2 in the main paper.}
\vspace{-4mm}
\label{fig:iresnet100_supplementary_sota_lfw}
\end{figure*}

%% file: figures/fig_iresnet100_supplementary_sota_calfw.tex
\begin{figure*}[h!]
\centering
	\begin{subfigure}[b]{0.95\textwidth}
		\centering
		\includegraphics[width=\textwidth]{figures/iresnet100_bn_sota/legend.pdf}
	\end{subfigure}
\\
	\begin{subfigure}[b]{0.48\textwidth}
		 \centering
		 \includegraphics[width=0.95\textwidth]{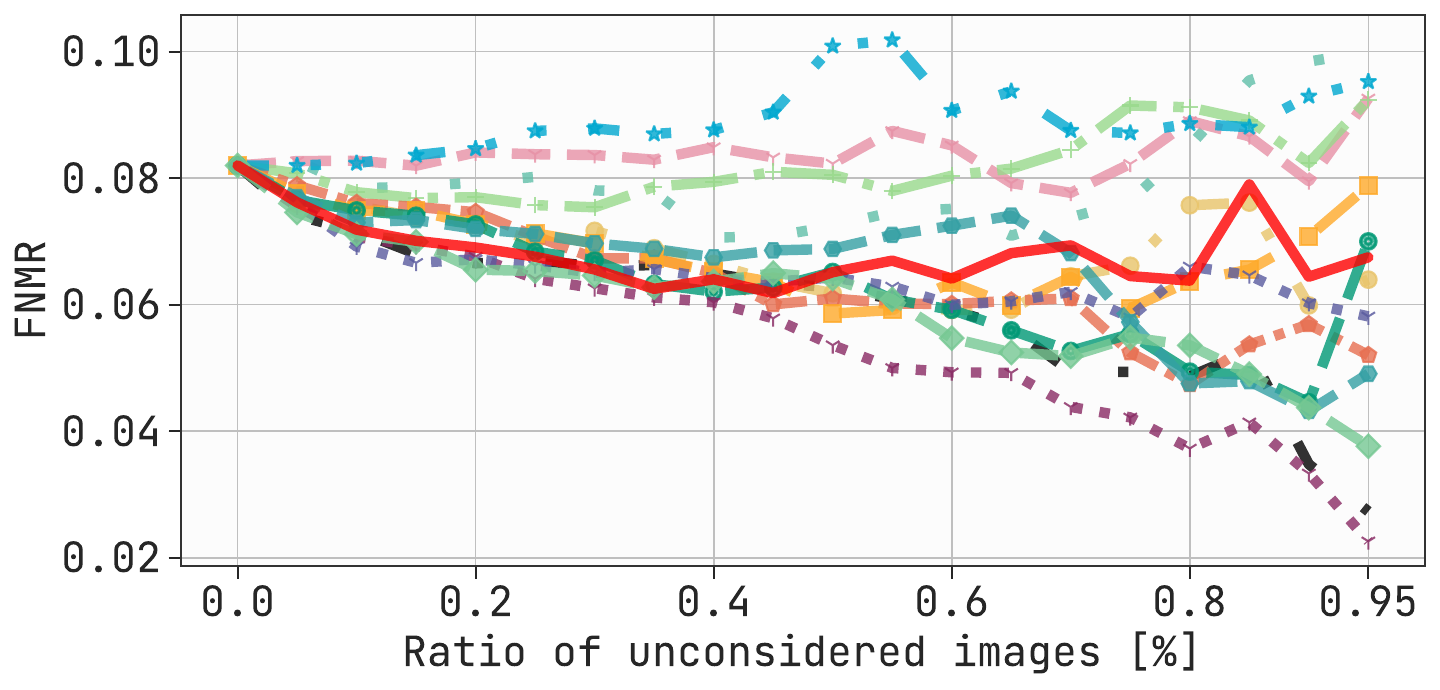}
		 \caption{ArcFace \cite{deng2019arcface} Model, CALFW \cite{CALFW} Dataset \\ \grafiqs with ResNet100, FMR$=1e-3$}
	\end{subfigure}
\hfill
	\begin{subfigure}[b]{0.48\textwidth}
		 \centering
		 \includegraphics[width=0.95\textwidth]{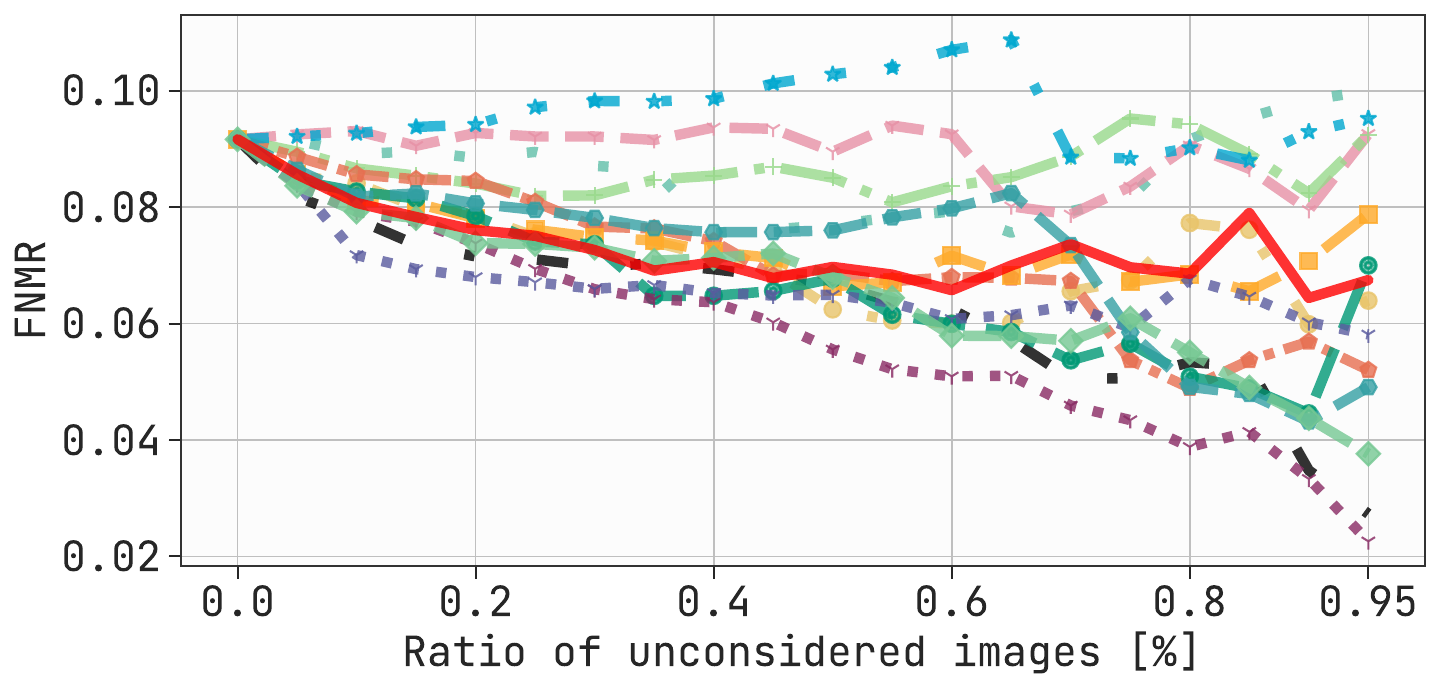}
		 \caption{ArcFace \cite{deng2019arcface} Model, CALFW \cite{CALFW} Dataset \\ \grafiqs with ResNet100, FMR$=1e-4$}
	\end{subfigure}
\\
	\begin{subfigure}[b]{0.48\textwidth}
		 \centering
		 \includegraphics[width=0.95\textwidth]{figures/iresnet100_bn_sota/calfw/ElasticFaceModel/model_ElasticFaceModel_calfw_fnmr3_sota.pdf}
		 \caption{ElasticFace \cite{elasticface} Model, CALFW \cite{CALFW} Dataset \\ \grafiqs with ResNet100, FMR$=1e-3$}
	\end{subfigure}
\hfill
	\begin{subfigure}[b]{0.48\textwidth}
		 \centering
		 \includegraphics[width=0.95\textwidth]{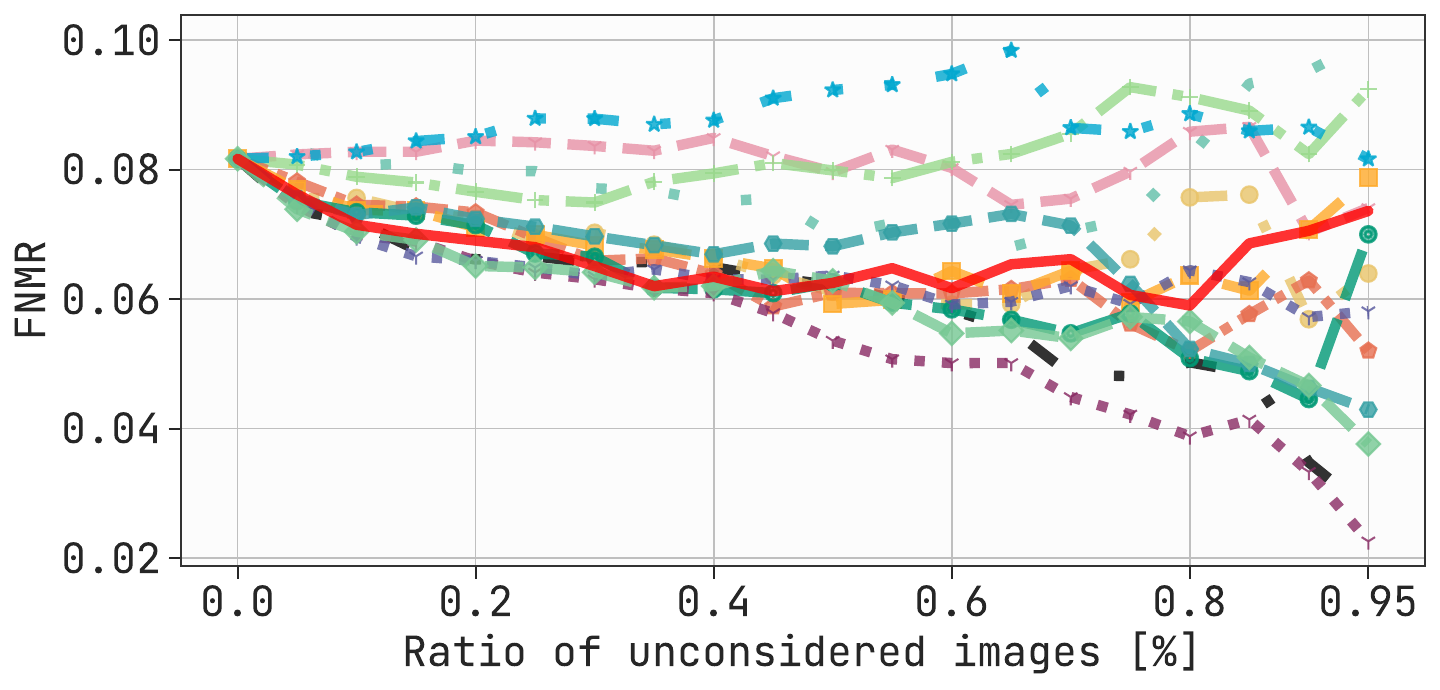}
		 \caption{ElasticFace \cite{elasticface} Model, CALFW \cite{CALFW} Dataset \\ \grafiqs with ResNet100, FMR$=1e-4$}
	\end{subfigure}
\\
	\begin{subfigure}[b]{0.48\textwidth}
		 \centering
		 \includegraphics[width=0.95\textwidth]{figures/iresnet100_bn_sota/calfw/MagFaceModel/model_MagFaceModel_calfw_fnmr3_sota.pdf}
		 \caption{MagFace \cite{meng_2021_magface} Model, CALFW \cite{CALFW} Dataset \\ \grafiqs with ResNet100, FMR$=1e-3$}
	\end{subfigure}
\hfill
	\begin{subfigure}[b]{0.48\textwidth}
		 \centering
		 \includegraphics[width=0.95\textwidth]{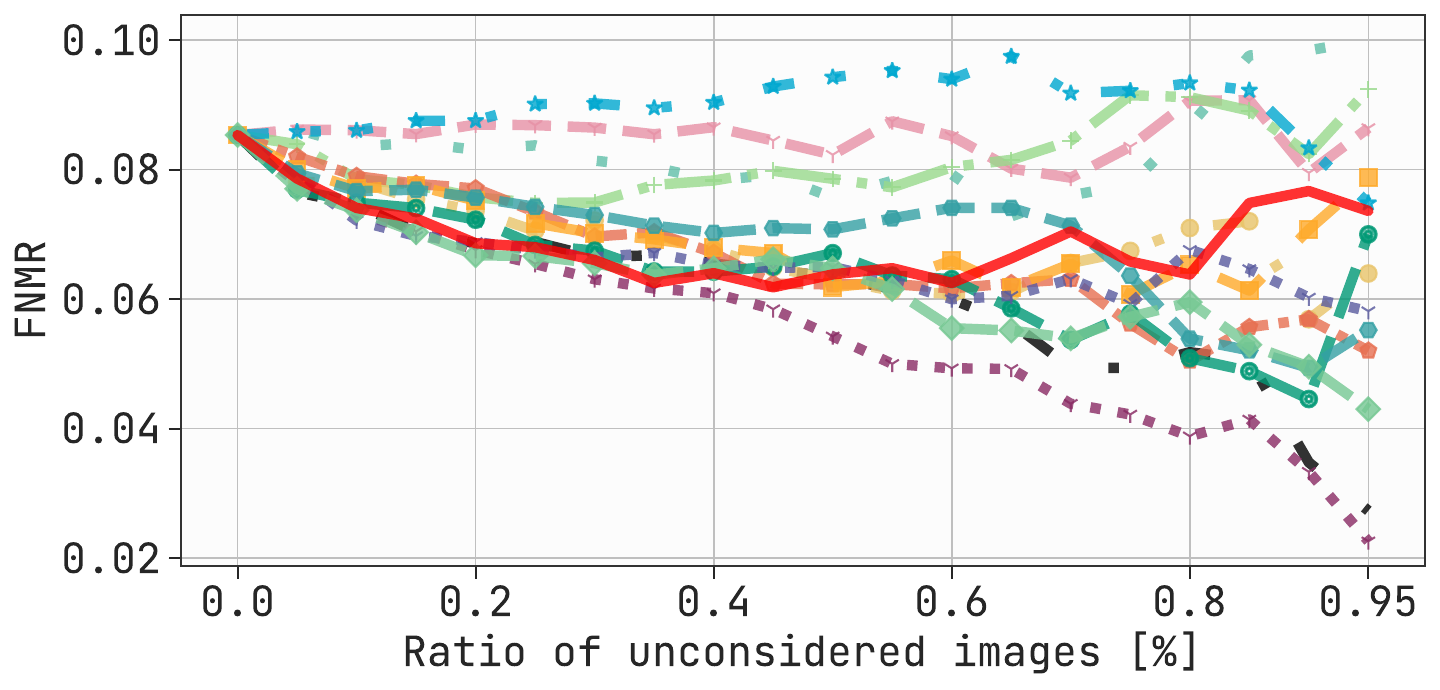}
		 \caption{MagFace \cite{meng_2021_magface} Model, CALFW \cite{CALFW} Dataset \\ \grafiqs with ResNet100, FMR$=1e-4$}
	\end{subfigure}
\\
	\begin{subfigure}[b]{0.48\textwidth}
		 \centering
		 \includegraphics[width=0.95\textwidth]{figures/iresnet100_bn_sota/calfw/CurricularFaceModel/model_CurricularFaceModel_calfw_fnmr3_sota.pdf}
		 \caption{CurricularFace \cite{curricularFace} Model, CALFW \cite{CALFW} Dataset \\ \grafiqs with ResNet100, FMR$=1e-3$}
	\end{subfigure}
\hfill
	\begin{subfigure}[b]{0.48\textwidth}
		 \centering
		 \includegraphics[width=0.95\textwidth]{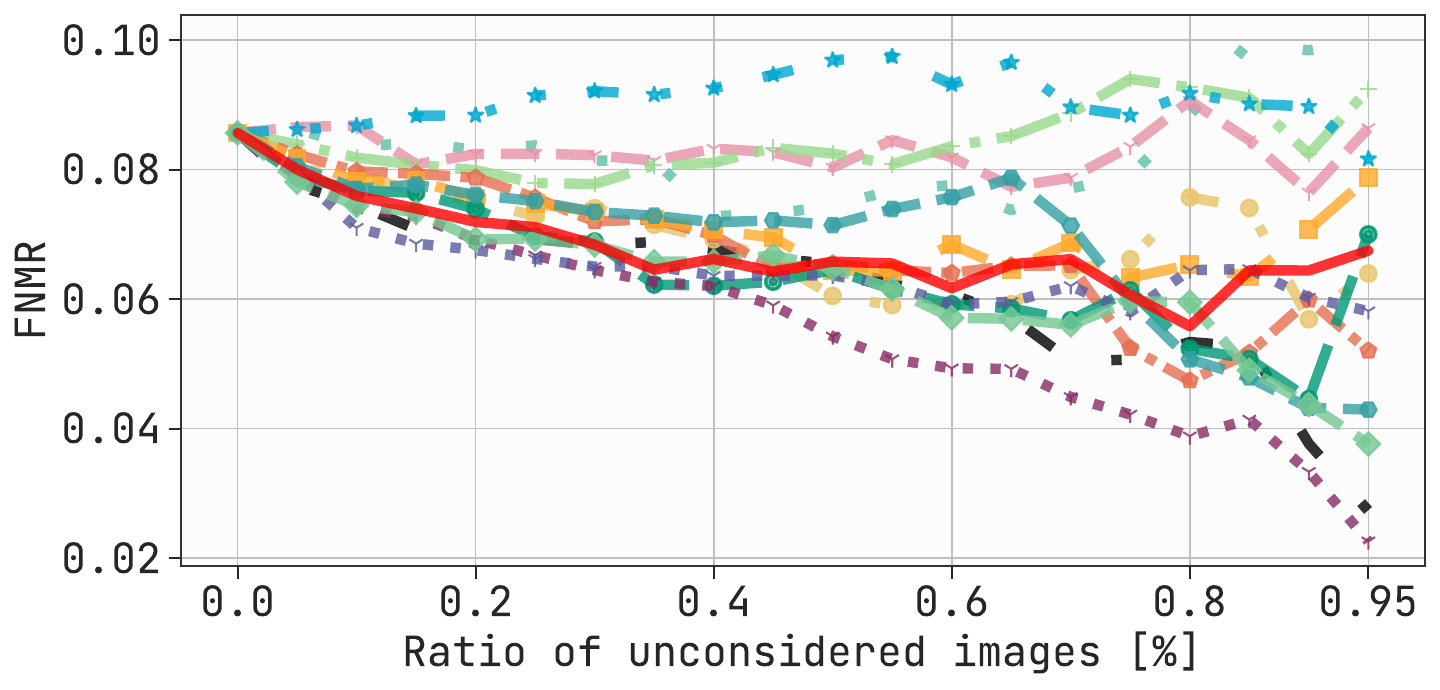}
		 \caption{CurricularFace \cite{curricularFace} Model, CALFW \cite{CALFW} Dataset \\ \grafiqs with ResNet100, FMR$=1e-4$}
	\end{subfigure}
\\
\caption{EDC curves for FNMR@FMR=$1e-3$ and FNMR@FMR=$1e-4$ on dataset CALFW \cite{CALFW} using ArcFace, ElasticFace, MagFace, and CurricularFace FR models. The proposed \grafiqs method, shown in \textcolor{red}{solid red}, utilizes gradient magnitudes and it is reported using the best setting from Table 2 in the main paper.}
\vspace{-4mm}
\label{fig:iresnet100_supplementary_sota_calfw}
\end{figure*}

%% file: figures/fig_iresnet100_supplementary_sota_cplfw.tex
\begin{figure*}[h!]
\centering
	\begin{subfigure}[b]{0.95\textwidth}
		\centering
		\includegraphics[width=\textwidth]{figures/iresnet100_bn_sota/legend.pdf}
	\end{subfigure}
\\
	\begin{subfigure}[b]{0.48\textwidth}
		 \centering
		 \includegraphics[width=0.95\textwidth]{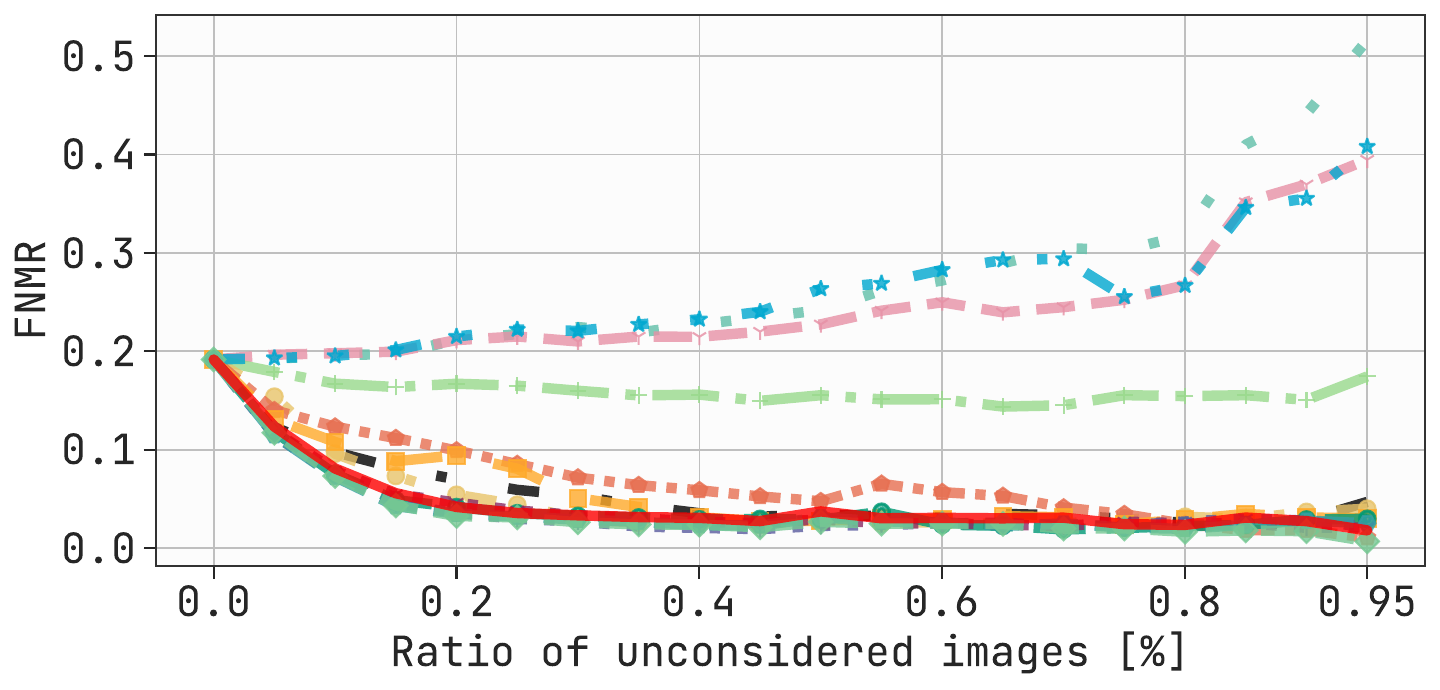}
		 \caption{ArcFace \cite{deng2019arcface} Model, CPLFW \cite{CPLFWTech} Dataset \\ \grafiqs with ResNet100, FMR$=1e-3$}
	\end{subfigure}
\hfill
	\begin{subfigure}[b]{0.48\textwidth}
		 \centering
		 \includegraphics[width=0.95\textwidth]{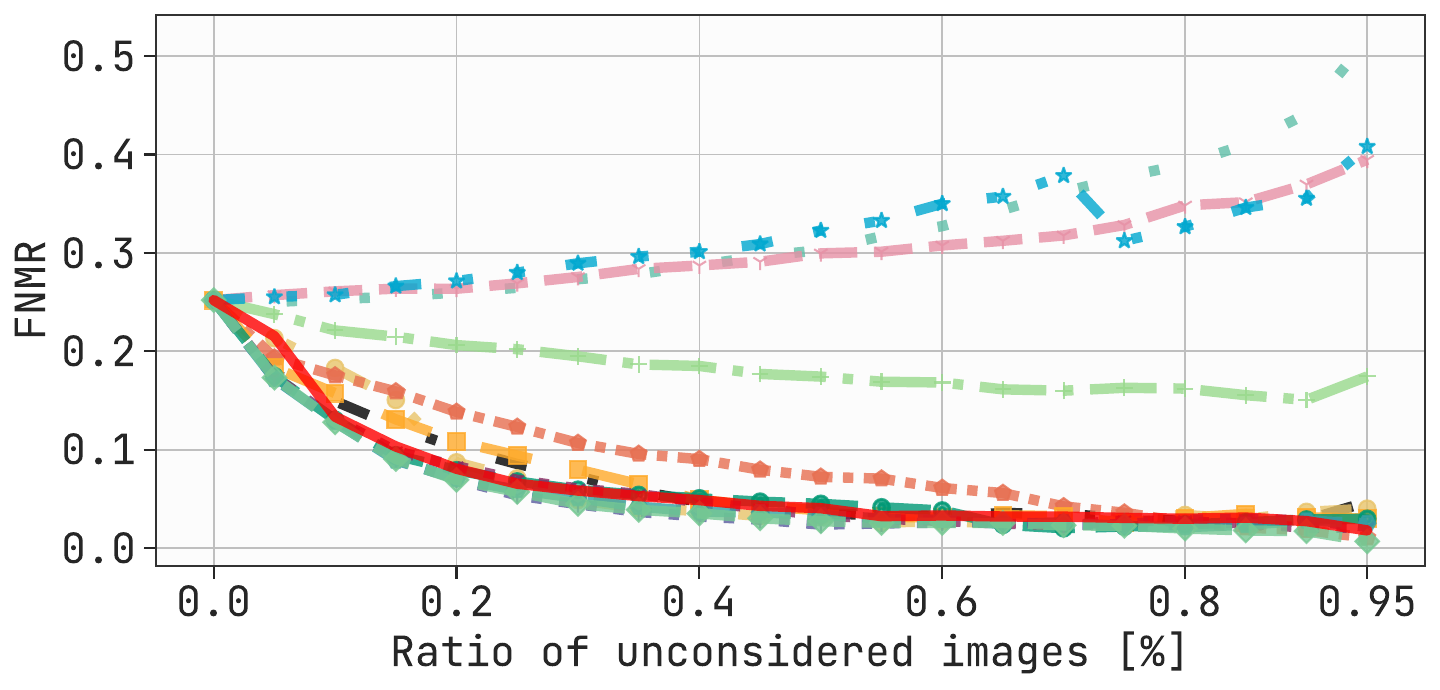}
		 \caption{ArcFace \cite{deng2019arcface} Model, CPLFW \cite{CPLFWTech} Dataset \\ \grafiqs with ResNet100, FMR$=1e-4$}
	\end{subfigure}
\\
	\begin{subfigure}[b]{0.48\textwidth}
		 \centering
		 \includegraphics[width=0.95\textwidth]{figures/iresnet100_bn_sota/cplfw/ElasticFaceModel/model_ElasticFaceModel_cplfw_fnmr3_sota.pdf}
		 \caption{ElasticFace \cite{elasticface} Model, CPLFW \cite{CPLFWTech} Dataset \\ \grafiqs with ResNet100, FMR$=1e-3$}
	\end{subfigure}
\hfill
	\begin{subfigure}[b]{0.48\textwidth}
		 \centering
		 \includegraphics[width=0.95\textwidth]{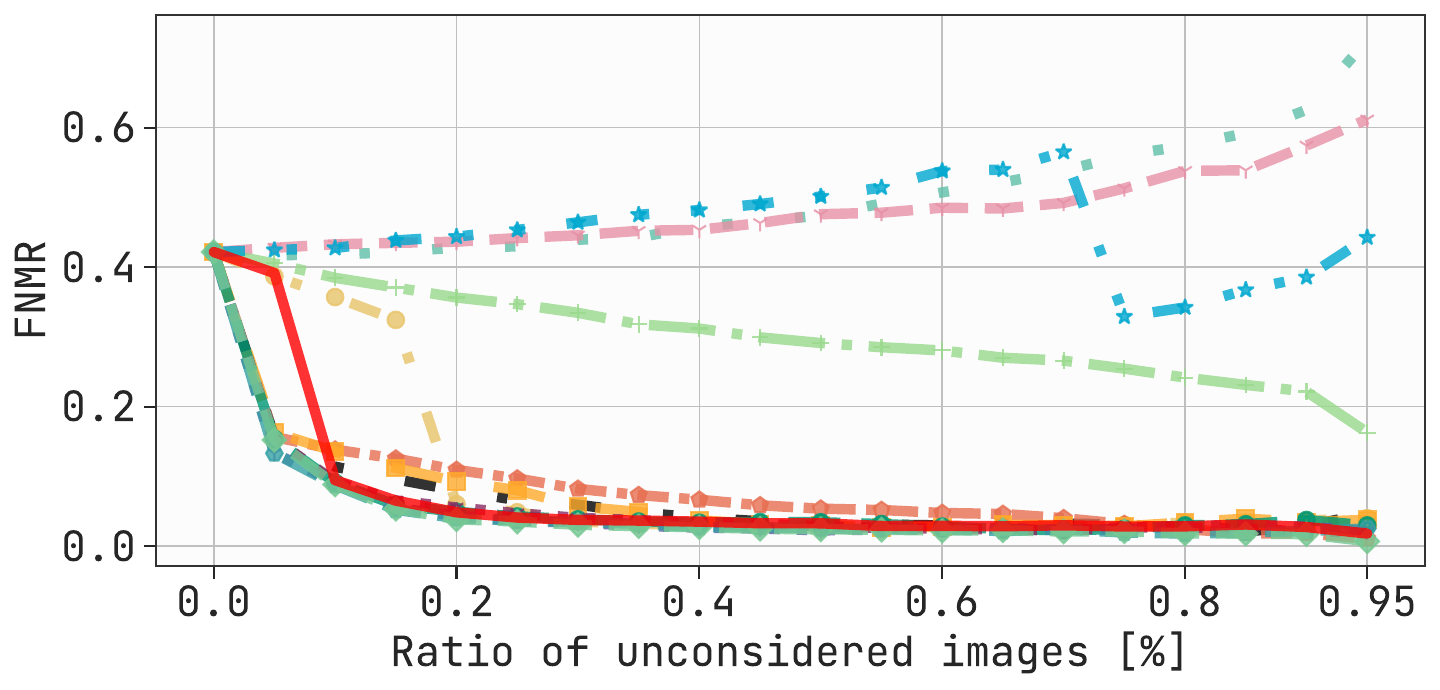}
		 \caption{ElasticFace \cite{elasticface} Model, CPLFW \cite{CPLFWTech} Dataset \\ \grafiqs with ResNet100, FMR$=1e-4$}
	\end{subfigure}
\\
	\begin{subfigure}[b]{0.48\textwidth}
		 \centering
		 \includegraphics[width=0.95\textwidth]{figures/iresnet100_bn_sota/cplfw/MagFaceModel/model_MagFaceModel_cplfw_fnmr3_sota.pdf}
		 \caption{MagFace \cite{meng_2021_magface} Model, CPLFW \cite{CPLFWTech} Dataset \\ \grafiqs with ResNet100, FMR$=1e-3$}
	\end{subfigure}
\hfill
	\begin{subfigure}[b]{0.48\textwidth}
		 \centering
		 \includegraphics[width=0.95\textwidth]{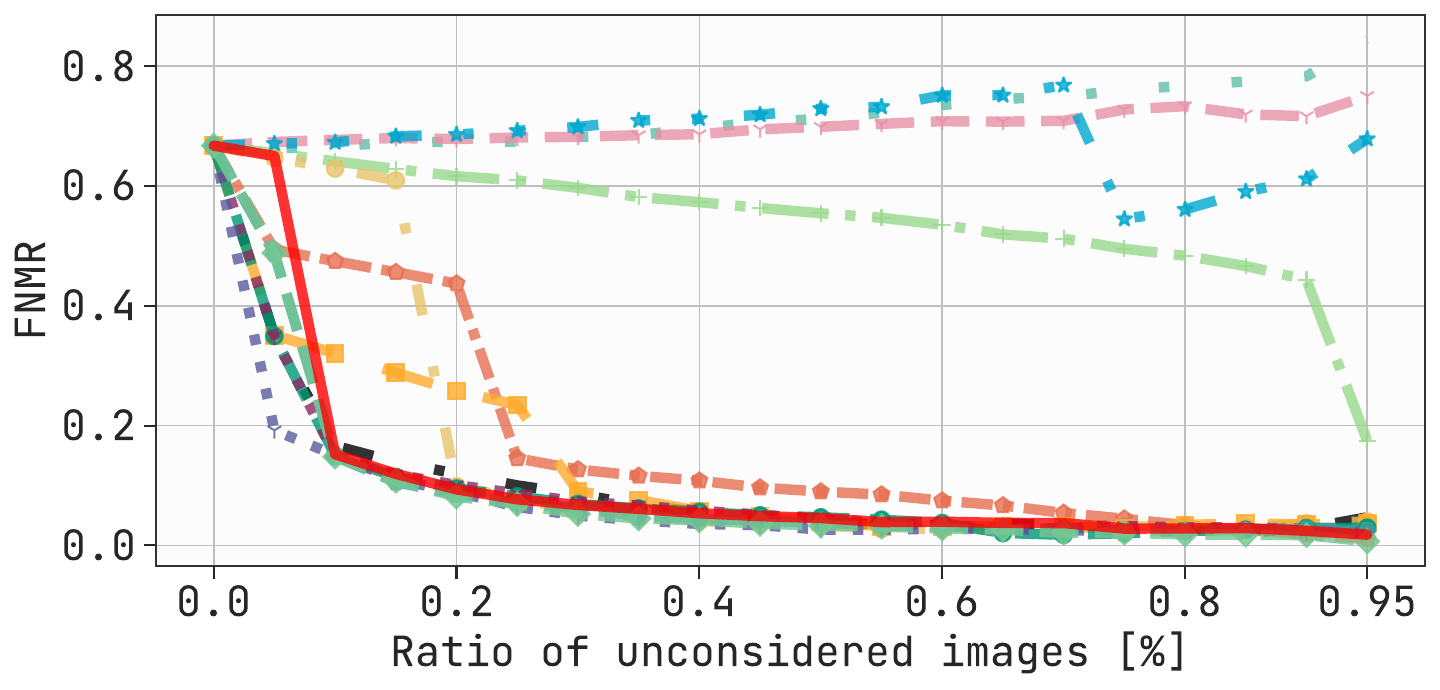}
		 \caption{MagFace \cite{meng_2021_magface} Model, CPLFW \cite{CPLFWTech} Dataset \\ \grafiqs with ResNet100, FMR$=1e-4$}
	\end{subfigure}
\\
	\begin{subfigure}[b]{0.48\textwidth}
		 \centering
		 \includegraphics[width=0.95\textwidth]{figures/iresnet100_bn_sota/cplfw/CurricularFaceModel/model_CurricularFaceModel_cplfw_fnmr3_sota.pdf}
		 \caption{CurricularFace \cite{curricularFace} Model, CPLFW \cite{CPLFWTech} Dataset \\ \grafiqs with ResNet100, FMR$=1e-3$}
	\end{subfigure}
\hfill
	\begin{subfigure}[b]{0.48\textwidth}
		 \centering
		 \includegraphics[width=0.95\textwidth]{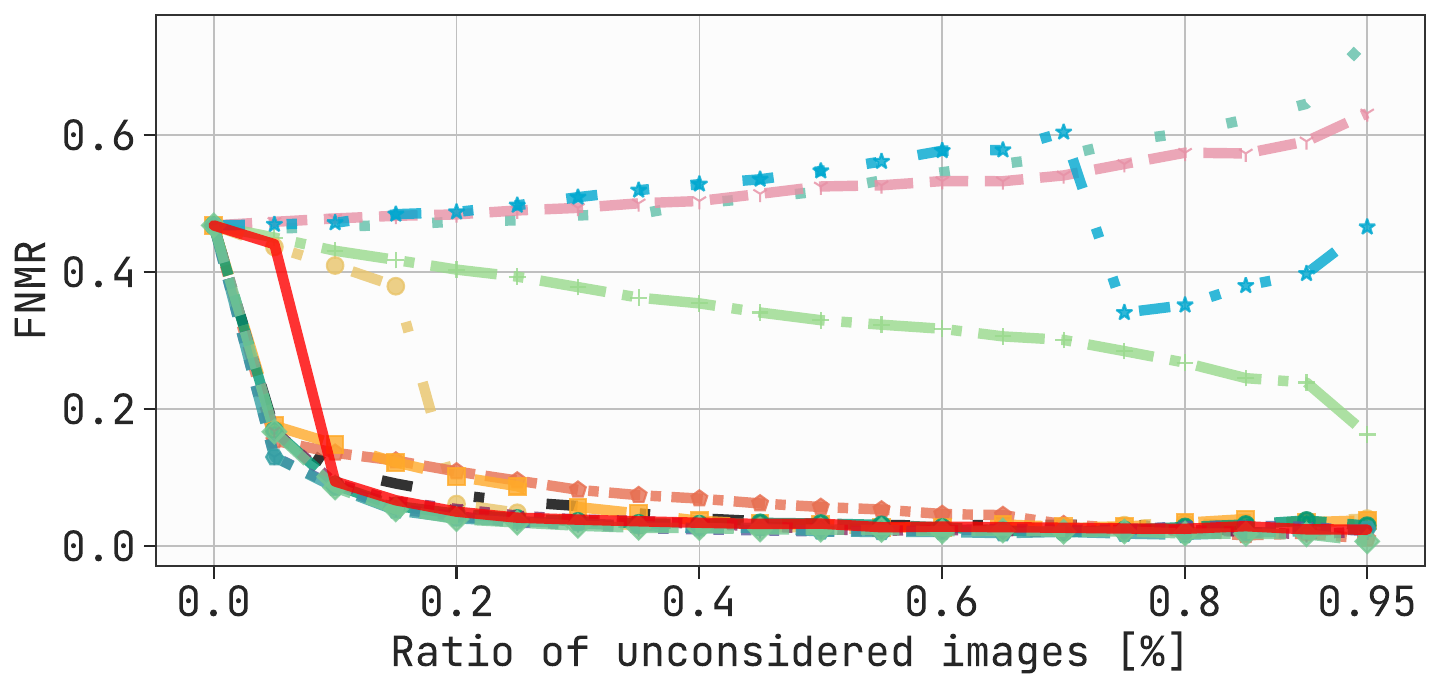}
		 \caption{CurricularFace \cite{curricularFace} Model, CPLFW \cite{CPLFWTech} Dataset \\ \grafiqs with ResNet100, FMR$=1e-4$}
	\end{subfigure}
\\
\caption{EDC curves for FNMR@FMR=$1e-3$ and FNMR@FMR=$1e-4$ on dataset CPLFW \cite{CPLFWTech} using ArcFace, ElasticFace, MagFace, and CurricularFace FR models. The proposed \grafiqs method, shown in \textcolor{red}{solid red}, utilizes gradient magnitudes and it is reported using the best setting from Table 2 in the main paper.}
\vspace{-4mm}
\label{fig:iresnet100_supplementary_sota_cplfw}
\end{figure*}

%% file: figures/fig_iresnet100_supplementary_sota_XQLFW.tex
\begin{figure*}[h!]
\centering
	\begin{subfigure}[b]{0.95\textwidth}
		\centering
		\includegraphics[width=\textwidth]{figures/iresnet100_bn_sota/legend.pdf}
	\end{subfigure}
\\
	\begin{subfigure}[b]{0.48\textwidth}
		 \centering
		 \includegraphics[width=0.95\textwidth]{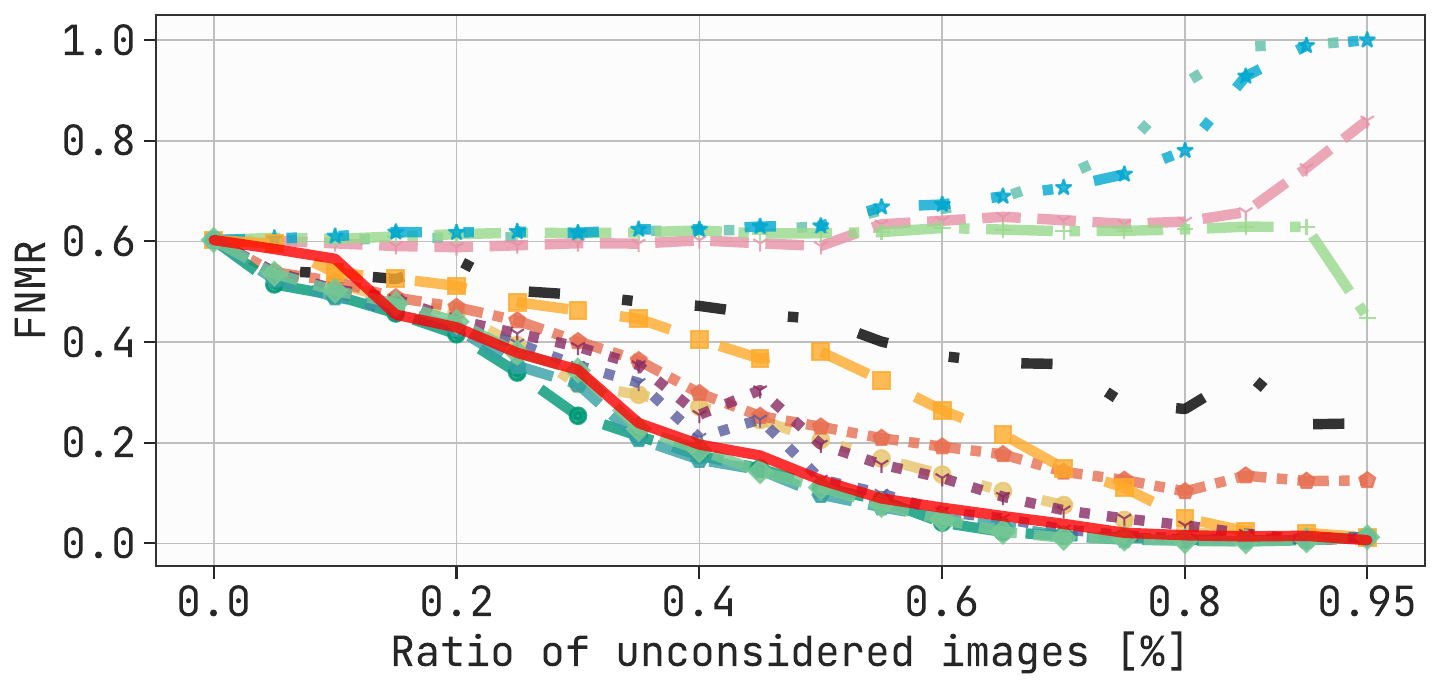}
		 \caption{ArcFace \cite{deng2019arcface} Model, XQLFW \cite{XQLFW} Dataset \\ \grafiqs with ResNet100, FMR$=1e-3$}
	\end{subfigure}
\hfill
	\begin{subfigure}[b]{0.48\textwidth}
		 \centering
		 \includegraphics[width=0.95\textwidth]{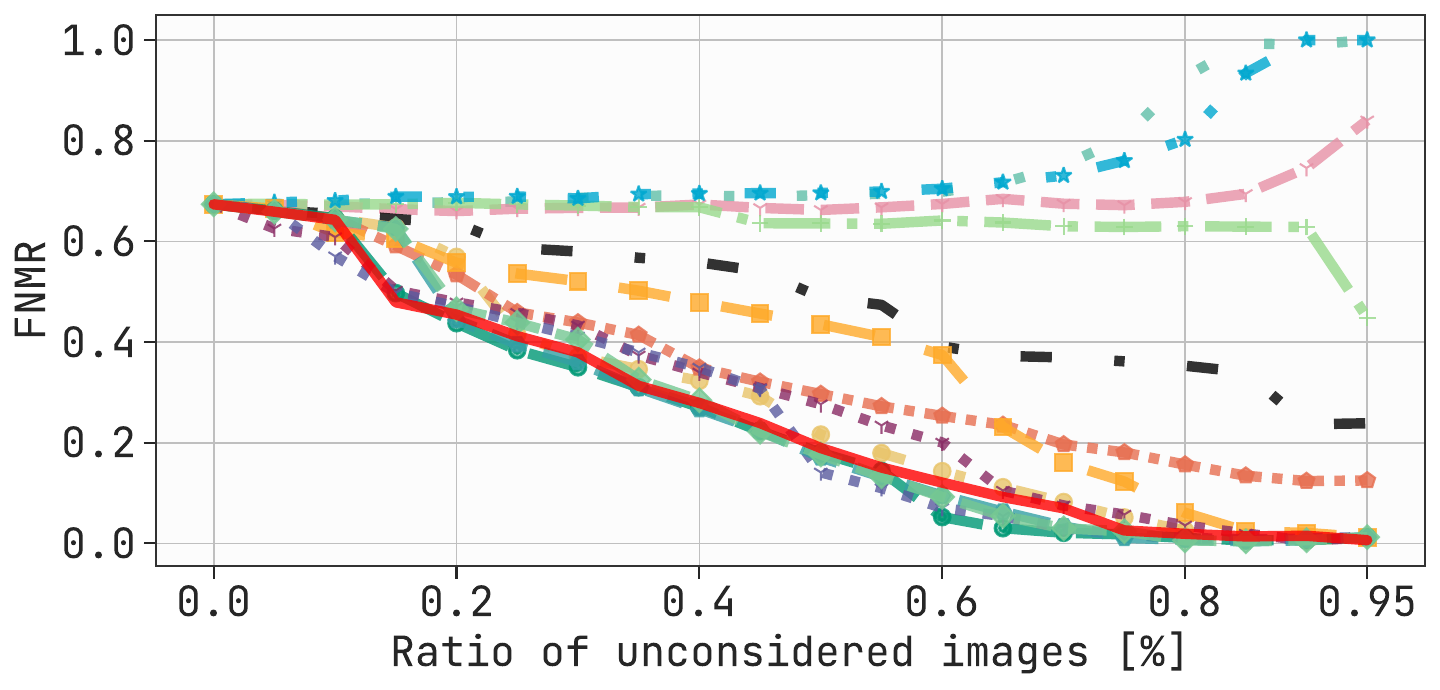}
		 \caption{ArcFace \cite{deng2019arcface} Model, XQLFW \cite{XQLFW} Dataset \\ \grafiqs with ResNet100, FMR$=1e-4$}
	\end{subfigure}
\\
	\begin{subfigure}[b]{0.48\textwidth}
		 \centering
		 \includegraphics[width=0.95\textwidth]{figures/iresnet100_bn_sota/XQLFW/ElasticFaceModel/model_ElasticFaceModel_XQLFW_fnmr3_sota.pdf}
		 \caption{ElasticFace \cite{elasticface} Model, XQLFW \cite{XQLFW} Dataset \\ \grafiqs with ResNet100, FMR$=1e-3$}
	\end{subfigure}
\hfill
	\begin{subfigure}[b]{0.48\textwidth}
		 \centering
		 \includegraphics[width=0.95\textwidth]{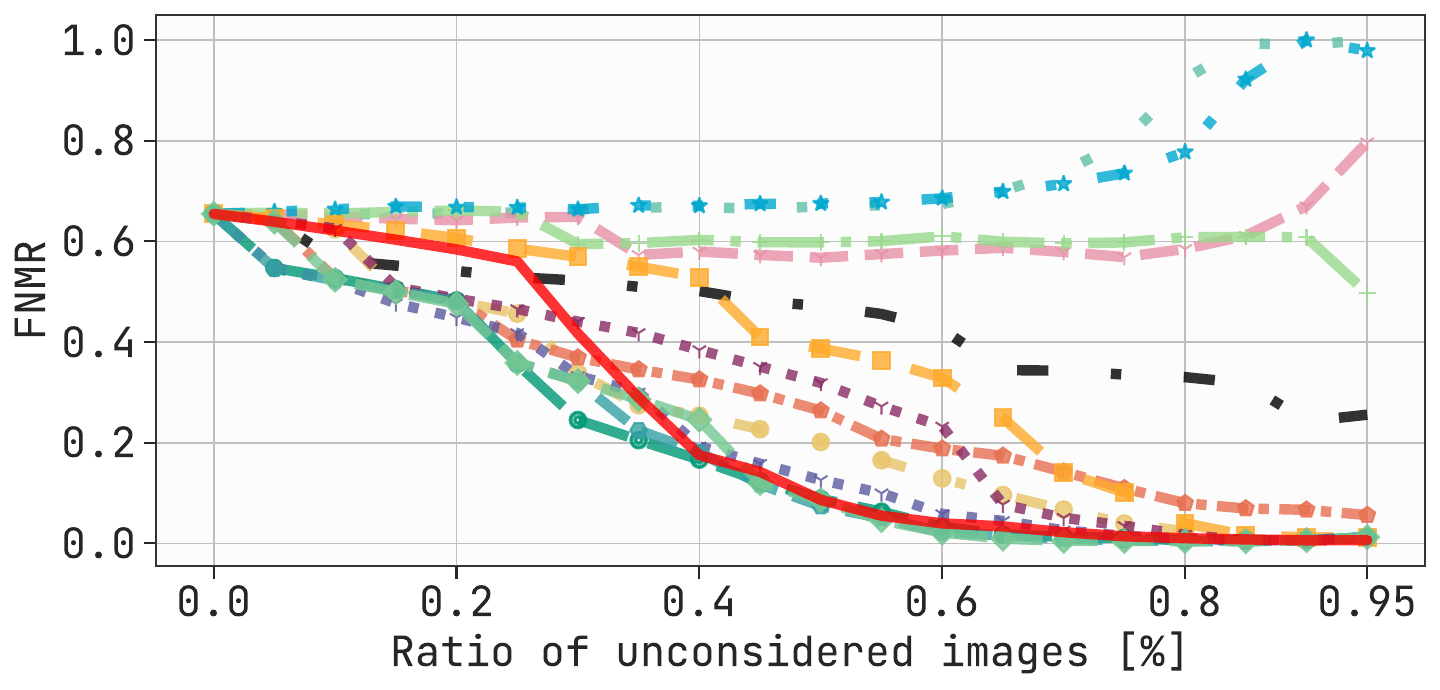}
		 \caption{ElasticFace \cite{elasticface} Model, XQLFW \cite{XQLFW} Dataset \\ \grafiqs with ResNet100, FMR$=1e-4$}
	\end{subfigure}
\\
	\begin{subfigure}[b]{0.48\textwidth}
		 \centering
		 \includegraphics[width=0.95\textwidth]{figures/iresnet100_bn_sota/XQLFW/MagFaceModel/model_MagFaceModel_XQLFW_fnmr3_sota.pdf}
		 \caption{MagFace \cite{meng_2021_magface} Model, XQLFW \cite{XQLFW} Dataset \\ \grafiqs with ResNet100, FMR$=1e-3$}
	\end{subfigure}
\hfill
	\begin{subfigure}[b]{0.48\textwidth}
		 \centering
		 \includegraphics[width=0.95\textwidth]{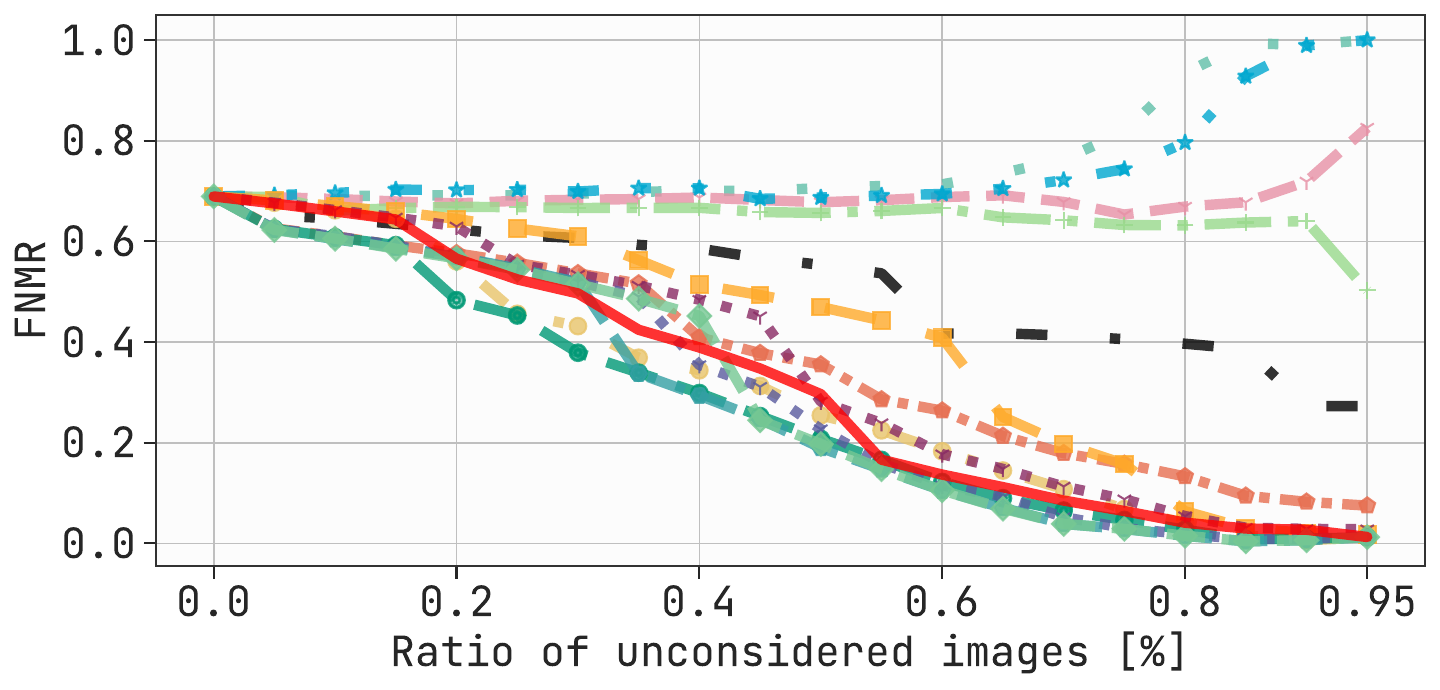}
		 \caption{MagFace \cite{meng_2021_magface} Model, XQLFW \cite{XQLFW} Dataset \\ \grafiqs with ResNet100, FMR$=1e-4$}
	\end{subfigure}
\\
	\begin{subfigure}[b]{0.48\textwidth}
		 \centering
		 \includegraphics[width=0.95\textwidth]{figures/iresnet100_bn_sota/XQLFW/CurricularFaceModel/model_CurricularFaceModel_XQLFW_fnmr3_sota.pdf}
		 \caption{CurricularFace \cite{curricularFace} Model, XQLFW \cite{XQLFW} Dataset \\ \grafiqs with ResNet100, FMR$=1e-3$}
	\end{subfigure}
\hfill
	\begin{subfigure}[b]{0.48\textwidth}
		 \centering
		 \includegraphics[width=0.95\textwidth]{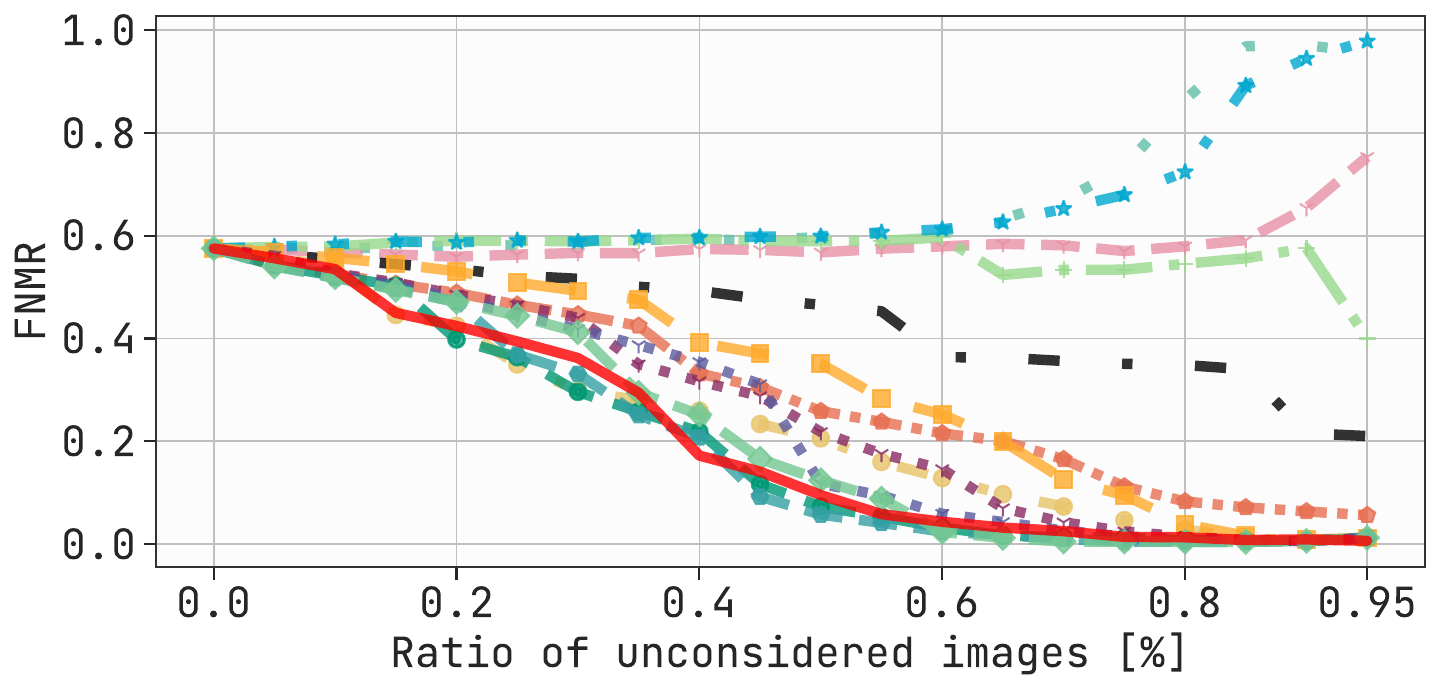}
		 \caption{CurricularFace \cite{curricularFace} Model, XQLFW \cite{XQLFW} Dataset \\ \grafiqs with ResNet100, FMR$=1e-4$}
	\end{subfigure}
\\
\caption{EDC curves for FNMR@FMR=$1e-3$ and FNMR@FMR=$1e-4$ on dataset XQLFW \cite{XQLFW} using ArcFace, ElasticFace, MagFace, and CurricularFace FR models. The proposed \grafiqs method, shown in \textcolor{red}{solid red}, utilizes gradient magnitudes and it is reported using the best setting from Table 2 in the main paper.}
\vspace{-4mm}
\label{fig:iresnet100_supplementary_sota_XQLFW}
\end{figure*}

%% file: main.bbl
\begin{thebibliography}{10}\itemsep=-1pt

\bibitem{Quality_ISO}
{ ISO/IEC JTC 1/SC 37 Biometrics}.
\newblock {ISO/IEC 29794-1 Information technology Biometric sample quality Part 1: Framework}.
\newblock International Organization for Standardization, 2024.

\bibitem{aggarwal_2020_ml_optim}
Charu~C. Aggarwal.
\newblock {\em Linear Algebra and Optimization for Machine Learning - {A} Textbook}.
\newblock Springer, 2020.

\bibitem{DBLP:conf/iwbf/BabnikDS23}
{\v{Z}}iga Babnik, Naser Damer, and Vitomir {\v{S}}truc.
\newblock Optimization-based improvement of face image quality assessment techniques.
\newblock In {\em 11th International Workshop on Biometrics and Forensics, {IWBF} 2023, Barcelona, Spain, April 19-20, 2023}, pages 1--6. {IEEE}, 2023.

\bibitem{10449044}
{\v{Z}}iga Babnik, Peter Peer, and Vitomir {\v{S}}truc.
\newblock Diffiqa: Face image quality assessment using denoising diffusion probabilistic models.
\newblock In {\em 2023 IEEE International Joint Conference on Biometrics (IJCB)}, pages 1--10, 2023.

\bibitem{babnikTBIOM2024}
{\v{Z}}iga Babnik, Peter Peer, and Vitomir {\v{S}}truc.
\newblock {eDifFIQA: Towards Efficient Face Image Quality Assessment based on Denoising Diffusion Probabilistic Models}.
\newblock {\em IEEE Transactions on Biometrics, Behavior, and Identity Science (TBIOM)}, 2024.

\bibitem{best2018learning}
Lacey Best{-}Rowden and Anil~K. Jain.
\newblock Learning face image quality from human assessments.
\newblock {\em {IEEE} Trans. Inf. Forensics Secur.}, 13(12):3064--3077, 2018.

\bibitem{bishop_2023_deeplearning}
Christopher~Michael Bishop and Hugh Bishop.
\newblock {\em Deep Learning - Foundations and Concepts}.
\newblock Springer, 1 edition, 2023.

\bibitem{DEEPIQ_IQA}
Sebastian Bosse, Dominique Maniry, Klaus{-}Robert M{\"{u}}ller, Thomas Wiegand, and Wojciech Samek.
\newblock Deep neural networks for no-reference and full-reference image quality assessment.
\newblock {\em {IEEE} Trans. Image Process.}, 27(1):206--219, 2018.

\bibitem{elasticface}
Fadi Boutros, Naser Damer, Florian Kirchbuchner, and Arjan Kuijper.
\newblock Elasticface: Elastic margin loss for deep face recognition.
\newblock In {\em {IEEE/CVF} Conference on Computer Vision and Pattern Recognition Workshops, {CVPR} Workshops 2022, New Orleans, LA, USA, June 19-20, 2022}, pages 1577--1586. {IEEE}, 2022.

\bibitem{boutros_2023_crfiqa}
Fadi Boutros, Meiling Fang, Marcel Klemt, Biying Fu, and Naser Damer.
\newblock {CR-FIQA:} face image quality assessment by learning sample relative classifiability.
\newblock In {\em {IEEE/CVF} Conference on Computer Vision and Pattern Recognition, {CVPR} 2023, Vancouver, BC, Canada, June 17-24, 2023}, pages 5836--5845. {IEEE}, 2023.

\bibitem{RANKIQ_FIQA}
Jiansheng Chen, Yu Deng, Gaocheng Bai, and Guangda Su.
\newblock Face image quality assessment based on learning to rank.
\newblock {\em {IEEE} Signal Process. Lett.}, 22(1):90--94, 2015.

\bibitem{deng2019arcface}
Jiankang Deng, Jia Guo, Niannan Xue, and Stefanos Zafeiriou.
\newblock Arcface: Additive angular margin loss for deep face recognition.
\newblock In {\em {IEEE} Conference on Computer Vision and Pattern Recognition, {CVPR} 2019, Long Beach, CA, USA, June 16-20, 2019}, pages 4690--4699. Computer Vision Foundation / {IEEE}, 2019.

\bibitem{Adience}
Eran Eidinger, Roee Enbar, and Tal Hassner.
\newblock Age and gender estimation of unfiltered faces.
\newblock {\em {IEEE} Trans. Inf. Forensics Secur.}, 9(12):2170--2179, 2014.

\bibitem{BiyingWACV}
Biying Fu, Cong Chen, Olaf Henniger, and Naser Damer.
\newblock A deep insight into measuring face image utility with general and face-specific image quality metrics.
\newblock In {\em {IEEE/CVF} Winter Conference on Applications of Computer Vision, {WACV} 2022, Waikoloa, HI, USA, January 3-8, 2022}, pages 1121--1130. {IEEE}, 2022.

\bibitem{NISTQuaity}
P. Grother, M.~Ngan A.~Hom, and K. Hanaoka.
\newblock Ongoing face recognition vendor test (frvt) part 5: Face image quality assessment (4th draft).
\newblock In {\em National Institute of Standards and Technology}. Tech. Rep., Sep. 2021.

\bibitem{GT07}
P. Grother and E. Tabassi.
\newblock Performance of biometric quality measures.
\newblock {\em IEEE Trans.~on Pattern Analysis and Machine Intelligence}, 29(4):531--543, Apr. 2007.

\bibitem{guo_2016_ms1m}
Yandong Guo, Lei Zhang, Yuxiao Hu, Xiaodong He, and Jianfeng Gao.
\newblock Ms-celeb-1m: {A} dataset and benchmark for large-scale face recognition.
\newblock In Bastian Leibe, Jiri Matas, Nicu Sebe, and Max Welling, editors, {\em Computer Vision - {ECCV} 2016 - 14th European Conference, Amsterdam, The Netherlands, October 11-14, 2016, Proceedings, Part {III}}, volume 9907 of {\em Lecture Notes in Computer Science}, pages 87--102. Springer, 2016.

\bibitem{he_2016_resnet}
Kaiming He, Xiangyu Zhang, Shaoqing Ren, and Jian Sun.
\newblock Deep residual learning for image recognition.
\newblock In {\em 2016 {IEEE} Conference on Computer Vision and Pattern Recognition, {CVPR} 2016, Las Vegas, NV, USA, June 27-30, 2016}, pages 770--778. {IEEE} Computer Society, 2016.

\bibitem{faceqnetv1}
Javier Hernandez{-}Ortega, Javier Galbally, Julian Fi{\'{e}}rrez, and Laurent Beslay.
\newblock Biometric quality: Review and application to face recognition with faceqnet.
\newblock {\em CoRR}, abs/2006.03298, 2020.

\bibitem{hernandez2019faceqnet}
Javier Hernandez{-}Ortega, Javier Galbally, Julian Fi{\'{e}}rrez, Rudolf Haraksim, and Laurent Beslay.
\newblock Faceqnet: Quality assessment for face recognition based on deep learning.
\newblock In {\em 2019 International Conference on Biometrics, {ICB} 2019, Crete, Greece, June 4-7, 2019}, pages 1--8. {IEEE}, 2019.

\bibitem{LFWTech}
Gary~B. Huang, Manu Ramesh, Tamara Berg, and Erik Learned-Miller.
\newblock Labeled faces in the wild: A database for studying face recognition in unconstrained environments.
\newblock Technical Report 07-49, University of Massachusetts, Amherst, October 2007.

\bibitem{curricularFace}
Yuge Huang, Yuhan Wang, Ying Tai, Xiaoming Liu, Pengcheng Shen, Shaoxin Li, Jilin Li, and Feiyue Huang.
\newblock Curricularface: Adaptive curriculum learning loss for deep face recognition.
\newblock In {\em 2020 {IEEE/CVF} Conference on Computer Vision and Pattern Recognition, {CVPR} 2020, Seattle, WA, USA, June 13-19, 2020}, pages 5900--5909. Computer Vision Foundation / {IEEE}, 2020.

\bibitem{ioffe_2015_batchnormalization}
Sergey Ioffe and Christian Szegedy.
\newblock Batch normalization: Accelerating deep network training by reducing internal covariate shift.
\newblock In Francis Bach and David Blei, editors, {\em Proceedings of the 32nd International Conference on Machine Learning}, volume~37 of {\em Proceedings of Machine Learning Research}, pages 448--456, Lille, France, 07--09 Jul 2015. PMLR.

\bibitem{iso_metric}
{ISO/IEC JTC1 SC37 Biometrics}.
\newblock {ISO/IEC 19795-1:2021 Information technology — Biometric performance testing and reporting — Part 1: Principles and framework}.
\newblock International Organization for Standardization, 2021.

\bibitem{XQLFW}
Martin Knoche, Stefan H{\"{o}}rmann, and Gerhard Rigoll.
\newblock Cross-quality {LFW:} {A} database for analyzing cross- resolution image face recognition in unconstrained environments.
\newblock In {\em 16th {IEEE} International Conference on Automatic Face and Gesture Recognition, {FG} 2021, Jodhpur, India, December 15-18, 2021}, pages 1--5. {IEEE}, 2021.

\bibitem{kolf_2023_idnet}
Jan~Niklas Kolf, Tim Rieber, Jurek Elliesen, Fadi Boutros, Arjan Kuijper, and Naser Damer.
\newblock Identity-driven three-player generative adversarial network for synthetic-based face recognition.
\newblock In {\em {IEEE/CVF} Conference on Computer Vision and Pattern Recognition, {CVPR} 2023 - Workshops, Vancouver, BC, Canada, June 17-24, 2023}, pages 806--816. {IEEE}, 2023.

\bibitem{liu2017rankiqa}
Xialei Liu, Joost van~de Weijer, and Andrew~D. Bagdanov.
\newblock Rankiqa: Learning from rankings for no-reference image quality assessment.
\newblock In {\em {IEEE} International Conference on Computer Vision, {ICCV} 2017, Venice, Italy, October 22-29, 2017}, pages 1040--1049. {IEEE} Computer Society, 2017.

\bibitem{MagFace}
Qiang Meng, Shichao Zhao, Zhida Huang, and Feng Zhou.
\newblock Magface: {A} universal representation for face recognition and quality assessment.
\newblock In {\em {IEEE} Conference on Computer Vision and Pattern Recognition, {CVPR} 2021, virtual, June 19-25, 2021}, pages 14225--14234. Computer Vision Foundation / {IEEE}, 2021.

\bibitem{meng_2021_magface}
Qiang Meng, Shichao Zhao, Zhida Huang, and Feng Zhou.
\newblock Magface: {A} universal representation for face recognition and quality assessment.
\newblock In {\em {IEEE} Conference on Computer Vision and Pattern Recognition, {CVPR} 2021, virtual, June 19-25, 2021}, pages 14225--14234. Computer Vision Foundation / {IEEE}, 2021.

\bibitem{BRISQE_IQA}
Anish Mittal, Anush~Krishna Moorthy, and Alan~Conrad Bovik.
\newblock No-reference image quality assessment in the spatial domain.
\newblock {\em {IEEE} Trans. Image Process.}, 21(12):4695--4708, 2012.

\bibitem{nique}
Anish Mittal, Rajiv Soundararajan, and Alan~C. Bovik.
\newblock Making a "completely blind" image quality analyzer.
\newblock {\em {IEEE} Signal Process. Lett.}, 20(3):209--212, 2013.

\bibitem{agedb}
Stylianos Moschoglou, Athanasios Papaioannou, Christos Sagonas, Jiankang Deng, Irene Kotsia, and Stefanos Zafeiriou.
\newblock Agedb: The first manually collected, in-the-wild age database.
\newblock In {\em 2017 {IEEE} CVPRW, {CVPR} Workshops 2017, Honolulu, HI, USA, July 21-26, 2017}, pages 1997--2005. {IEEE} Computer Society, 2017.

\bibitem{SDDFIQA}
Fu{-}Zhao Ou, Xingyu Chen, Ruixin Zhang, Yuge Huang, Shaoxin Li, Jilin Li, Yong Li, Liujuan Cao, and Yuan{-}Gen Wang.
\newblock {SDD-FIQA:} unsupervised face image quality assessment with similarity distribution distance.
\newblock In {\em {IEEE} Conference on Computer Vision and Pattern Recognition, {CVPR} 2021, virtual, June 19-25, 2021}, pages 7670--7679. Computer Vision Foundation / {IEEE}, 2021.

\bibitem{prince_2023_understanding}
Simon~J.D. Prince.
\newblock {\em Understanding Deep Learning}.
\newblock MIT Press, 2023.

\bibitem{santurkar_2018_batchnormalization}
Shibani Santurkar, Dimitris Tsipras, Andrew Ilyas, and Aleksander Madry.
\newblock How does batch normalization help optimization?
\newblock In {\em Advances in Neural Information Processing Systems 31: Annual Conference on Neural Information Processing Systems 2018, NeurIPS 2018, December 3-8, 2018, Montr{\'{e}}al, Canada}, pages 2488--2498, 2018.

\bibitem{DBLP:journals/csur/SchlettRHGFB22}
Torsten Schlett, Christian Rathgeb, Olaf Henniger, Javier Galbally, Julian Fi{\'{e}}rrez, and Christoph Busch.
\newblock Face image quality assessment: {A} literature survey.
\newblock {\em {ACM} Comput. Surv.}, 54(10s):210:1--210:49, 2022.

\bibitem{cfp-fp}
Soumyadip Sengupta, Jun{-}Cheng Chen, Carlos~Domingo Castillo, Vishal~M. Patel, Rama Chellappa, and David~W. Jacobs.
\newblock Frontal to profile face verification in the wild.
\newblock In {\em 2016 {IEEE} Winter Conference on Applications of Computer Vision, {WACV} 2016, Lake Placid, NY, USA, March 7-10, 2016}, pages 1--9. {IEEE} Computer Society, 2016.

\bibitem{PFE_FIQA}
Yichun Shi and Anil~K. Jain.
\newblock Probabilistic face embeddings.
\newblock In {\em 2019 {IEEE/CVF} International Conference on Computer Vision, {ICCV} 2019, Seoul, Korea (South), October 27 - November 2, 2019}, pages 6901--6910. {IEEE}, 2019.

\bibitem{SERFIQ}
Philipp Terh{\"{o}}rst, Jan~Niklas Kolf, Naser Damer, Florian Kirchbuchner, and Arjan Kuijper.
\newblock {SER-FIQ:} unsupervised estimation of face image quality based on stochastic embedding robustness.
\newblock In {\em 2020 {IEEE/CVF} Conference on Computer Vision and Pattern Recognition, {CVPR} 2020, Seattle, WA, USA, June 13-19, 2020}, pages 5650--5659. Computer Vision Foundation / {IEEE}, 2020.

\bibitem{xu_2020_generativequant}
Shoukai Xu, Haokun Li, Bohan Zhuang, Jing Liu, Jiezhang Cao, Chuangrun Liang, and Mingkui Tan.
\newblock Generative low-bitwidth data free quantization.
\newblock In Andrea Vedaldi, Horst Bischof, Thomas Brox, and Jan{-}Michael Frahm, editors, {\em Computer Vision - {ECCV} 2020 - 16th European Conference, Glasgow, UK, August 23-28, 2020, Proceedings, Part {XII}}, volume 12357 of {\em Lecture Notes in Computer Science}, pages 1--17. Springer, 2020.

\bibitem{yi_2014_casiawebface}
Dong Yi, Zhen Lei, Shengcai Liao, and Stan~Z. Li.
\newblock Learning face representation from scratch.
\newblock {\em CoRR}, abs/1411.7923, 2014.

\bibitem{CPLFWTech}
T. Zheng and W. Deng.
\newblock Cross-pose lfw: A database for studying cross-pose face recognition in unconstrained environments.
\newblock Technical Report 18-01, Beijing University of Posts and Telecommunications, February 2018.

\bibitem{CALFW}
Tianyue Zheng, Weihong Deng, and Jiani Hu.
\newblock Cross-age {LFW:} {A} database for studying cross-age face recognition in unconstrained environments.
\newblock {\em CoRR}, abs/1708.08197, 2017.

\end{thebibliography}
